\documentclass[11pt]{article} %
\usepackage{setspace} %
\usepackage[opt-arxiv]{optional}
\opt{opt-neurips}{\usepackage[preprint]{neurips_2024}}

\usepackage{amsmath,amsfonts,bm}

\def\1{\bm{1}}

\def\eps{{\epsilon}}

\DeclareMathAlphabet{\mathsfit}{\encodingdefault}{\sfdefault}{m}{sl}
\SetMathAlphabet{\mathsfit}{bold}{\encodingdefault}{\sfdefault}{bx}{n}

\def\gC{{\mathcal{C}}}
\def\gD{{\mathcal{D}}}

\def\gF{{\mathcal{F}}}

\def\gH{{\mathcal{H}}}

\def\gM{{\mathcal{M}}}

\def\gO{{\mathcal{O}}}
\def\gP{{\mathcal{P}}}
\def\gQ{{\mathcal{Q}}}

\def\gW{{\mathcal{W}}}
\def\gX{{\mathcal{X}}}
\def\gY{{\mathcal{Y}}}
\def\gZ{{\mathcal{Z}}}

\newcommand{\E}{\mathbb{E}}

\newcommand{\R}{\mathbb{R}}

\usepackage[utf8]{inputenc}
\usepackage[T1]{fontenc}
\usepackage[usenames,dvipsnames]{xcolor}
\usepackage{svg}
\usepackage{tikz}
\usepackage{tikz-3dplot}
\usepackage{upgreek}

\usepackage{amssymb,enumerate,graphicx,verbatim,natbib,amsthm,bbm}
\usepackage{silence}
\usepackage[inline, shortlabels]{enumitem}
\usepackage{thmtools}
\usepackage{thm-restate}
\usepackage[hypertexnames=false]{hyperref}
\usepackage{subcaption}
\usepackage{mwe}
\usepackage[toc,page,header]{appendix}
\usepackage[nohints]{minitoc}
\usepackage[linesnumbered,ruled,vlined]{algorithm2e}
\definecolor{amethyst}{rgb}{0.6, 0.4, 0.8}
\hypersetup{
    colorlinks=true,
    linkcolor=purple,
    filecolor=magenta,      
    urlcolor=cyan,
    citecolor=amethyst,
    bookmarksdepth=2,
}
\usepackage{url}
\usepackage[nameinlink,noabbrev,capitalise]{cleveref}
\Crefname{table}{Table}{Tables}
\Crefname{assumption}{Assumption}{Assumptions}
\crefname{equation}{}{}
\newtheorem{theorem}{Theorem}[section]
\newtheorem{lemma}{Lemma}[section]
\newtheorem{assumption}{Assumption}[section]
\newtheorem{remark}{Remark}[section]
\newtheorem{proposition}{Proposition}[section]

\newtheorem{definition}{Definition}[section]

\newtheorem{corollary}{Corollary}[section]
\newtheorem{example}{Example}[section]

\DeclareFontShape{OT1}{cmr}{m}{scit}{<->ssub * cmr/m/sc}{}
\DeclareFontShape{OT1}{cmr}{bx}{sc}{<->ssub * cmr/bx/n}{}

\makeatletter
\AtBeginDocument{%
  \@ifundefined{r@def:perturbation}{%
    \@ifundefined{r@def:joint-perturbation}{}{%
      \expandafter\global\expandafter\edef\csname r@def:perturbation\endcsname{\csname r@def:joint-perturbation\endcsname}%
    }%
  }{}%
}
\makeatother

\IfFileExists{symbol.tex}{

}{}
\usepackage{color}

\newenvironment{proof-sketch}{\noindent\textbf{\textit{Proof sketch}:}}{\hfill$\square$}
\renewenvironment{proof}{\noindent\textbf{\textit{Proof }:}}{\hfill$\square$}

\newcommand\normx[1]{\left\Vert#1\right\Vert}

\newcommand{\ate}{\textsc{ATE}}
\newcommand{\att}{\textsc{ATT}}
\newcommand{\dis}{\textsc{DS}}
\newcommand{\ape}{\textsc{APE}}
\newcommand{\eqd}{\textsc{EQD}}

\newcommand{\ecc}{\textsc{ECC}}

\newcommand{\wad}{\textsc{WAD}}
\newcommand{\argmin}{\operatornamewithlimits{argmin}}

\newcommand\ci{\protect\mathpalette{\protect\independenT}{\perp}}
\def\independenT#1#2{\mathrel{\rlap{$#1#2$}\mkern2mu{#1#2}}}
\newcommand{\dd}{\textup{d}}
\providecommand{\HH}{H}

\newcommand{\cX}{\ensuremath{{\cal X}}}
\def\independenT#1#2{\mathrel{\rlap{$#1#2$}\mkern2mu{#1#2}}}

\usepackage{color-edits}
\addauthor{vs}{red}
\addauthor{jj}{blue}

\title{Sharp Structure-Agnostic Lower Bounds for General Linear Functional
Estimation}

\opt{opt-arxiv}{
\author{Jikai Jin \thanks{Stanford University. Email: \texttt{jkjin@stanford.edu}. Jikai Jin was supported by NSF Award IIS-2337916.}
~~~~Vasilis Syrgkanis \thanks{Stanford University. Email: \texttt{vsyrgk@stanford.edu}. Vasilis Syrgkanis was supported by NSF Award IIS-2337916.}}}

\opt{opt-arxiv}{
\usepackage[left=2.0cm, right=2.0cm, top=2.3cm]{geometry}}

\begin{document}
\maketitle

\begin{abstract}
    We establish a general statistical optimality theory for estimation problems where the target parameter is a linear functional of an unknown nuisance component that must be estimated from data. This formulation covers many causal and predictive parameters and has applications to numerous disciplines. We adopt the structure-agnostic framework introduced by \citet{balakrishnan2023fundamental}, which poses no structural properties on the nuisance functions other than access to black-box estimators that achieve some statistical estimation rate. This framework is particularly appealing when one is only willing to consider estimation strategies that use non-parametric regression and classification oracles as black-box sub-processes. Within this framework, we first prove the statistical optimality of the celebrated and widely used doubly robust estimators for the Average Treatment Effect (ATE), the most central parameter in causal inference. We then characterize the minimax optimal rate under the general formulation. Notably, we differentiate between two regimes in which double robustness can and cannot be achieved and in which first-order debiasing yields different error rates. Our result implies that first-order debiasing is simultaneously optimal in both regimes. We instantiate our theory by deriving optimal error rates that recover existing results and extend to various settings of interest, including the case when the nuisance is defined by generalized regressions and when covariate shift exists for training and test distribution\footnote{This paper generalizes and subsumes \citet{jin2024structure} by the same authors.}.

\end{abstract}
\onehalfspacing

\section{Introduction}
\label{sec:intro}

Let $\{O_t\}_{t=1}^N$ be i.i.d.\ training samples from an unknown distribution $P_0$ on $\gO=\gZ\times\gW$, where each
$O=(Z,W)$ contains covariates $Z$ and an outcome $W$, and let $\{Z_i\}_{i=1}^N$ be i.i.d.\ target covariate samples from
an unknown distribution $Q_0$ on $\gZ$.
We consider the problem of estimating a linear functional of a regression-type nuisance learned under the training law
$P_0$ but evaluated under the target law $Q_0$:
\begin{equation}
    \label{eq:parameter-of-interest}
    \chi(P_0,Q_0) = \E_{Q_0}\big[ m_1(Z,\gamma(\cdot;P_0)) \big],
\end{equation}
where for any fixed $z$ the mapping $\gamma\mapsto m_1(z,\gamma)$ is linear, and the nuisance function $\gamma(z;P)$ is the solution to a generalized regression problem under the training law:
\begin{equation}
    \label{eq:generalized-regression-formula}
    \gamma(z;P) = \argmin_{\gamma\in L^2(\mu_Z)} \E_{P}\big[ \ell(O,\gamma) \big],
\end{equation}
where $\mu_Z$ is some known measure on $\gZ=\mathrm{supp}(Z)$.

This formulation covers many causal and predictive estimation tasks and has found important applications in numerous
disciplines such as economics \citep{hirano2003efficient,imbens2004nonparametric}, education \citep{oreopoulos2006estimating},
epidemiology \citep{little2000causal,wood2008empirical}, and political science \citep{mayer2011does}.

\begin{example}[ATE]
    \label{example:ate}
    In the standard treatment-effect setup with $O=(X,D,Y)$, squared-loss regression yields the outcome model
    \[
    \gamma(d,x;P_0)=\E[Y\mid D=d,X=x].
    \]
    Under conditional ignorability and overlap, the average treatment effect is
    \[
    \theta^{\ate}=\E[\gamma(1,X;P_0)-\gamma(0,X;P_0)].
    \]
    This fits \eqref{eq:parameter-of-interest} by taking
    $Z=(D,X)$, $W=Y$, and
    \[
    m_1\big((d,x),\gamma\big)=\gamma(1,x)-\gamma(0,x),
    \qquad
    Q_0=P_{0,Z},
    \]
    where $P_{0,Z}$ denotes the marginal distribution of $P_0$ on $\gZ$.
\end{example}

\begin{example}[Average treatment effect on the treated (ATT)]
    \label{example:att}
    The average treatment effect on the treated is
    $\theta^{\att}=\E[\gamma(1,X;P_0)-\gamma(0,X;P_0)\mid D=1]$ and corresponds to the same choice of $\gamma$ and $m_1$,
    but with a \emph{selection} target law $Q_0=P_{0,Z\mid D=1}$.
\end{example}

\begin{example}[Log-odds difference (LOD)]
    \label{example:lod}
    In the same setup with binary outcome $Y\in\{0,1\}$, consider the log-odds regression
    \[
    \gamma(d,x;P_0)=\log\left(\frac{\E[Y\mid D=d,X=x]}{1-\E[Y\mid D=d,X=x]}\right).
    \]
    The log-odds-difference estimand is
    \[
    \chi_{\mathrm{LOD}}(P_0)=\E\big[\gamma(1,X;P_0)-\gamma(0,X;P_0)\big].
    \]
    This fits \eqref{eq:parameter-of-interest} by taking $Z=(D,X)$, $W=Y$, the same
    $m_1\big((d,x),\gamma\big)=\gamma(1,x)-\gamma(0,x)$, and $Q_0=P_{0,Z}$.
\end{example}

Estimating the ATE is one of the central problems in causal inference. In view of its practical importance, a large body
of work is devoted to developing statistically efficient estimators for the ATE based on regression
\citep{robins1994estimation,robins1995analysis,imbens2003mean}, matching \citep{heckman1998matching,rosenbaum1989optimal,abadie2006large},
and propensity scores \citep{rosenbaum1983central,hirano2003efficient}, as well as their combinations.
Beyond ATE, influence-function-based methods have been developed for a range of related estimands, including selection/conditioning targets such as ATT, policy learning objectives, weighted average derivatives, and covariate-shift/data-fusion targets; see, for example, \citet{athey2021policy,newey1993efficiency,powell1989semiparametric,sugiyama2007covariate,reddi2015doubly} and references therein.

Statistical limits for estimating treatment-effect-type parameters are studied in
\citet{robins2009semiparametric,balakrishnan2019hypothesis,kennedy2022minimax,robins2008higher}, typically under
H\"older-smoothness assumptions on the nuisances. When the nonparametric components of the data-generating process are
estimable at a fast enough rate (typically $n^{-1/4}$), semiparametric efficiency \citep{newey1994asymptotic} provides
optimal variance constants multiplying the leading rate. Finally, \citet{bradic2019minimax} characterizes minimax
conditions for root-$n$ estimability, albeit under strong linearity restrictions and constant effects. These works
crucially rely on structural assumptions on the underlying function classes, which enables tight rates but can be
cumbersome to deploy in practice when the relevant structure is unknown or violated.

Since the nuisance function $\gamma$ in \eqref{eq:parameter-of-interest} is unknown and may have complex structures, and since the dimension $K$ of the covariates $X$ can be large relative to the number of data $n$ in many applications, it is extremely suitable to apply modern machine learning (ML) methods for the non-parametric, flexible and adaptive estimation of these nuisance functions, including penalized linear regression methods \citep{belloni2014pivotal,van2014asymptotically,chernozhukov2022automatic,zou2005regularization}, random forest methods \citep{breiman2001random,hastie2009random,biau2008consistency,wager2015adaptive,syrgkanis2020estimation}, gradient boosted forests \citep{friedman2001greedy,buhlmann2003boosting,zhang2005boosting} and neural networks \citep{schmidt2020nonparametric,farrell2021deep}, as well as ensemble and model selection approaches that combine all the above using out-of-sample cross-validation metrics \citep{wolpert1992stacked,zhang1993model,freund1997decision,van2007super,dvzeroski2004combining,sill2009feature,wegkamp2003model,Arlot2010,chetverikov2021cross}. However, ML methods typically require some forms of regularization to avoid overfitting, which can potentially make the resulting estimator severely biased.

A principled way to combine flexible nuisance estimation with accurate estimation of a low-dimensional target is to use
\emph{orthogonal (Neyman-orthogonal) estimating equations} derived from influence-function theory
\citep{robins1995analysis,robins1995semiparametric}. Double/debiased machine learning (DML) is a prominent and widely used
implementation of this idea \citep{chernozhukov2017double,chernozhukov2018double,athey2021policy,chernozhukov2022automatic}:
one estimates the nuisances (often via cross-fitting) and then evaluates an orthogonal score whose first-order sensitivity
to nuisance estimation errors vanishes at the truth. As a consequence, the estimation error admits a decomposition of the
schematic form
\[
\text{(parametric noise)}\;+\;\text{(higher-order remainder depending on nuisance errors)},
\]
where the leading remainder is typically proportional to a product of nuisance estimation errors (and, for some generalized
regression targets, may also include a squared term); see \citet{chernozhukov2018double} and the references therein.

In the special case with \emph{no covariate shift} ($Q_0=P_{0,Z}$) and ordinary least-squares regression
($\ell(o,\gamma)=(w-\gamma)^2/2$, so the score $\rho(o,\gamma)=\gamma-w$ is affine), write $\gamma_0(z):=\gamma(z;P_0)$
(which equals $\E[W\mid Z=z]$ for squared loss). Assume that the linear functional
$h\mapsto \E_{Q_0}[m_1(Z,h)]$ is continuous on $L^2(P_{0,Z})$. Equivalently, there exists a weight
$\nu_m(\cdot;P_0,Q_0)\in L^2(P_{0,Z})$ such that
\[
\E_{Q_0}\big[m_1(Z,h)\big]=\E_{P_{0,Z}}\big[h(Z)\,\nu_m(Z;P_0,Q_0)\big]\qquad\text{for all }h\in L^2(P_{0,Z}).
\]
Since $\rho(o,\gamma)=\gamma-w$ has derivative $1$ in its regression argument, we define the orthogonal weight
\[
\alpha_0(z):=\alpha(z;P_0,Q_0):=-\nu_m(z;P_0,Q_0).
\]
The corresponding orthogonal score is $\psi(O;\gamma,\alpha):=m_1(Z,\gamma)+\alpha(Z)\rho(O,\gamma(Z))$, and the
cross-fitted estimator can be written in the augmented plug-in form
\[
\hat\chi
=
\frac{1}{n}\sum_{i=1}^n m_1\big(Z_i,\hat\gamma\big)
\;+\;
\frac{1}{n}\sum_{i=1}^n \hat\alpha(Z_i)\,\big(\hat\gamma(Z_i)-W_i\big),
\]
with sample splitting/cross-fitting to ensure independence between the evaluation sample and the first-stage fits.
Moreover, letting $\mathbb{P}_n$ denote the empirical measure of $\{O_i\}_{i=1}^n$, a standard orthogonality expansion
gives (up to negligible empirical-process terms controlled by cross-fitting)
\[
\hat\chi-\chi(P_0,Q_0)
=
(\mathbb{P}_n-P_0)\psi(O;\gamma_0,\alpha_0)
\;+\;
\E_{P_0}\Big[\big(\hat\alpha(Z)-\alpha_0(Z)\big)\big(\hat\gamma(Z)-\gamma_0(Z)\big)\Big],
\]
so $\hat\chi-\chi(P_0,Q_0)=\gO_P(n^{-1/2}+\eps_{n,\gamma}\eps_{n,\alpha})$ under the mean-squared error constraints
imposed below. This approach also generalizes to the setting with covariate shift and generalized regression, as shown in \citet{chernozhukov2023automatic} and \citet{chernozhukov2021automatic} respectively. The generalized approach will be discussed in details in Section \ref{sec:first-order-debiasing-general}.

Motivated by the wide adoption and use of black-box adaptive estimation methods \citep{polley2019package,ledell2020h2o,wang2021flaml,karmaker2021automl} for these non-parametric components of the data generating process, as well as their superior empirical performance \citep{bach2024hyperparameter}, we will examine the statistical optimality of the aforementioned procedure within the \emph{structure-agnostic} minimax framework that was recently introduced in \citet{balakrishnan2023fundamental}. In particular, the only assumption that we will be making about our data generating process is that we have access to estimates $\hat{\gamma}$ and $\hat{\alpha}$ that achieve some statistical error rate, as measured by the mean-squared error, i.e.
$\|\hat{\gamma}(Z)-\gamma(Z;P_0)\|_{P_{0,Z},2}\leq \eps_{n,\gamma}$ and
$\|\hat{\alpha}(Z)-\alpha(Z;P_0,Q_0)\|_{P_{0,Z},2}\leq \eps_{n,\alpha}$, where for any function $v:\cX \to \R$, we denote $\|v(X)\|_{P_X,2}:=\sqrt{\E[v(X)^2]}$. Having access to such estimates for these two non-parametric components and imposing the aforementioned estimation error constraints on the data generating process, \emph{we resolve the optimal statistical rate achievable by any estimation algorithm for the parameters of interest.}

The structure-agnostic framework is particularly appealing as it essentially restricts any estimation approach to only use non-parametric regression estimates as a black-box and not tailor the estimation strategy to particular structural assumptions about the regression function or the propensity. These further structural assumptions can many times be brittle and violated in practice, rendering the tailored estimation strategy invalid or low-performing. Hence, the structure-agnostic statistical lower bound framework has the benefit that it yields lower bounds that can be matched by estimation procedures that are easy to deploy and robust.

\paragraph{Contributions and main message.}
Our main contribution is a general, sharp structure-agnostic lower bound theory for a broad class of functionals of the form \eqref{eq:parameter-of-interest}, where the nuisance $\gamma(\cdot;P)$ is defined as the solution to a (generalized) regression problem and the functional is linear in $\gamma$.  The class includes the average treatment effect and a range of causal and policy estimands that admit influence-function-based orthogonal scores.  Under assumptions that we verify for a collection of examples, our results identify two regimes:
\begin{itemize}
    \item In a \emph{mixed-bias} regime (covering standard regression residuals that are affine in $\gamma$), the minimax structure-agnostic error is lower bounded by
    \[
        \Omega\big(\eps_{n,\gamma}\eps_{n,\alpha}+n^{-1/2}\big).
    \]
    \item In a more general regime (covering generalized-regression targets), the minimax error is lower bounded by
    \[
        \Omega\big(\eps_{n,\gamma}\eps_{n,\alpha}+\eps_{n,\gamma}^2+n^{-1/2}\big).
    \]
\end{itemize}
In both regimes we provide matching upper bounds via first-order debiasing/DML estimators. Consequently, without additional structural information beyond mean-squared error guarantees for nuisance estimation, one cannot improve the dependence on nuisance errors beyond what is achieved by DML.  

For general non-parametric functional estimation, it has been known for decades that if the function possesses certain smoothness properties, then higher-order debiasing schemes can be designed that lead to improved error rates \citep{bickel1988estimating,birge1995estimation}. Specifically, first-order debiasing methods are suboptimal even when the nuisance function estimators are minimax optimal. Estimators based on higher-order debiasing have also been proposed and analyzed for functionals that arise in causal inference problems \citep{robins2008higher,van2014higher,robins2017minimax,liu2017semiparametric,kennedy2022minimax}. However, none of these approaches enjoy the structure-agnostic property that we explicitly impose in our minimax framework.

We then instantiate the general theory for a broad range of estimands, including the average treatment effect (ATE), average treatment effect on the treated (ATT), expected conditional covariance (ECC), weighted average derivative (WAD), distribution shift (DS), average policy effect (APE), log-odds-difference (LOD) and expected derivatives of conditional quantiles (EQD).

Prior to this work, optimal structure-agnostic error rates are not well understood, except for a few specific problem instances.  \citet{balakrishnan2023fundamental} was the first to establish sharp structure-agnostic lower bounds, but their proof techniques only apply to inner product functionals like ECC (see additional discussions in Section \ref{subsec:previous-proof-limitation}). Later, \citep{jin2024structure} established similar results separately for ATE and ATT. This paper is a generalized version of \citet{jin2024structure} and subsumes the results therein. Another recent work \citep{jin2025its} considered structure-agnostic estimation in a partial linear outcome model and an in-depth discussion of their results can be found in Remark~\ref{rem:plm-jin2025its-connection}.

\paragraph{Technical contributions.} The main technical contribution is a general lower-bound principle that applies uniformly across a broad class of statistical estimands, including targets that involve generalized regression that fall outside the mixed-bias regime. Our proof relies on a number of novel technical ideas, as we explain next.

Our lower bounds are proved via the method of fuzzy hypotheses, reducing estimation to testing between carefully constructed mixtures.  The core difficulty is to build composite null and alternative hypotheses that (i) remain within the prescribed structure-agnostic nuisance neighborhood and (ii) induce separation in the target functional of the desired order, while keeping the two mixtures close in Hellinger distance.  To achieve this we introduce a \emph{two-step sequential perturbation} construction that decouples feasibility (staying inside the nuisance neighborhood) from separation (moving the target functional).  A key ingredient is a geometric partitioning/``pairing'' argument (based on ham-sandwich-type results) that lets us place localized perturbations while enforcing the exact invariances required by our lower-bound theorem. A complete overview of our proofs can be found in Section \ref{subsec:proof-sketch}.

\subsection{Notation}
\label{subsec:notations}

We write $O=(Z,W)\in\gO=\gZ\times\gW$ for a generic observation. We observe i.i.d.\ training samples
$\{O_t\}_{t=1}^N\sim P_0$ on $\gO$ and i.i.d.\ target covariates $\{Z_i\}_{i=1}^N\sim Q_0$ on $\gZ$; more generally we
write $P$ and $Q$ for candidate training and target laws. The nuisance $\gamma(\cdot;P)$ is a function on $\gZ$ (typically
in $L^2(\mu_Z)$), and the target functional has the form $\chi(P,Q)=\E_Q[m_1(Z,\gamma(\cdot;P))]$ with $m_1(z,\cdot)$
linear, as in \eqref{eq:parameter-of-interest}--\eqref{eq:generalized-regression-formula}. We use subscripts to denote
marginals: if $P$ is a distribution on a product space,
we write $P_Z$ (resp.\ $P_X$) for the marginal law of $Z$ (resp.\ $X$). We write $\E_P[\cdot]$ for expectation under $P$
(and similarly $\E_Q[\cdot]$), and we use $\mathbb{P}_n$ for the empirical measure of an i.i.d.\ sample of size $n$ when
this is convenient.

For any function $f:\R^n\mapsto\R^k$ and distribution $P$ over $\R^n$, we define its $L^r(P)$ norm as
$\normx{f}_{P,r}=\big(\int \normx{f(x)}^r\,\dd P(x)\big)^{1/r}$ for $r\in(0,\infty)$, and
$\normx{f}_{P,\infty}=\text{ess sup}\{\normx{f(X)}:X\sim P\}$. When the distribution is clear from context we also write
$\normx{f}_r$. For deterministic sequences $(a_n)_{n\ge 1}$ and $(b_n)_{n\ge 1}$ we write $a_n=\gO(b_n)$ if there exists
$C>0$ such that $|a_n|\le C|b_n|$ for all $n$, and $a_n=\Omega(b_n)$ if there exists $c>0$ such that $|a_n|\ge c|b_n|$
for all $n$. For random variables, $X_n=\gO_P(b_n)$ means $X_n/b_n$ is bounded in probability.
We write $L^r(P)$ for the corresponding function space $\{f:\normx{f}_{P,r}<\infty\}$.

Throughout this paper we fix $\sigma$-finite reference measures $\mu_Z$ on $\gZ$ and $\mu_W$ on $\gW$, and write
$\mu=\mu_Z\otimes\mu_W$ on $\gO=\gZ\times\gW$ (often the uniform measure on its support). Our theory applies to
probability measures that are absolutely continuous with respect to these reference measures. For $P\ll\mu$ we write
$p=\dd P/\dd\mu$ for the density and $p_Z(z):=\int p(z,w)\,\dd\mu_W(w)$ (also denoted $p(z,\cdot)$) for the
$Z$-marginal density. For $Q\ll\mu_Z$ we write $q=\dd Q/\dd\mu_Z$ for its density. For any two distributions $P_1,P_2\ll\mu$
with densities $p_1,p_2$ and common support $\gO$, we define their $L_\infty$ distance by
$d_{\mu,\infty}(P_1,P_2)=\text{ess sup}_{o\in\gO}|p_1(o)-p_2(o)|$.

We define the directional derivative of a functional $\chi(P,Q)$ at $(P,Q)$ in the direction of a joint perturbation pair
$(H,K)$ (when it exists) as
\[
\chi_{(P,Q)}'(P,Q)[H,K]
:=
\left.\frac{\dd}{\dd t}\right|_{t=0}\chi(P+tH,Q+tK),
\]
where $H$ is a finite signed measure on $\gO$ and $K$ is a finite signed measure on $\gZ$. Similarly, we define the mixed
second derivative in directions $(H_0,K_0)$ and $(H_1,K_1)$ by
\[
\chi''(P,Q)[(H_0,K_0),(H_1,K_1)]
:=
\left.\frac{\dd^2}{\dd t\,\dd s}\right|_{t=s=0}\chi(P+tH_0+sH_1,Q+tK_0+sK_1),
\]
and
\[
\chi''(P,Q)[(H,K)]:=\chi''(P,Q)[(H,K),(H,K)].
\]
When the functional of interest is of the form $\Psi(U,P,Q)$ where $U$ is an additional parameter, we use
$\Psi_P'(U_0,P_0,Q_0)[H]$, $\Psi_Q'(U_0,P_0,Q_0)[K]$ and their second-order analogues to denote partial distributional
derivatives.

\section{Overview of our contributions}

\subsection{Our main results in a nutshell}
The main contribution of this paper is general lower bounds on the estimation error for functionals of the form \eqref{eq:parameter-of-interest}, in the case where no structural priors are available. In this subsection, we provide an overview of these results before stating the formal results and assumptions. 

Given estimates $\hat{\gamma}, \hat{\alpha}$ of $\gamma$ and $\alpha$ (defined later in Section \ref{sec:first-order-debiasing-general}, where $\alpha$ is some transformation of $m$ for ATE) and some specified error bounds $\eps_{n,\gamma}$ and $\eps_{n,\alpha}$, the set of all plausible ground-truth data distributions $P$ consists of those with nuisance functions $\gamma(Z;P)$ and $\alpha(Z;P)$ satisfying
\begin{equation}
    \label{eq:nuisance-constraint}
    \normx{\hat{\gamma}(Z)-\gamma(Z;P)}_{P_Z,2} \leq \eps_{n,\gamma},\quad \normx{\hat{\alpha}(Z)-\alpha(Z;P)}_{P_Z,2} \leq \eps_{n,\alpha}.
\end{equation}
Any estimator $\hat{\theta}$ can be viewed as a (possibly random) mapping from the observed data $\{O_i\}_{i=1}^n$ to $\mathbb{R}$. For any distribution $P$, when $\{O_i\}_{i=1}^n$ are i.i.d. samples from $P$, the estimator $\hat{\theta}$ induces a distribution of estimates on $\mathbb{R}$. Let $\xi\in(0,1)$ be a pre-specified tolerance probability. By comparing this distribution with the true parameter $\theta(P)$, we can measure the quality of the estimator $\hat{\theta}$ via the $(1-\xi)$-quantile of $|\hat{\theta}-\theta(P)|$. The worst-case error of $\hat{\theta}$ is then naturally defined as the supremum of this quantile over all possible $P$ satisfying the nuisance constraint in \eqref{eq:nuisance-constraint}. Our main result can be summarized as follows:
\begin{theorem}[Informal minimax structure-agnostic rates]
    Under certain assumptions that we verify for a broad class of functionals, the optimal worst-case error for estimating $\theta$ in \eqref{eq:parameter-of-interest} is either $\Omega(\eps_{n,\gamma}\eps_{n,\alpha}+n^{-1/2})$ or $\Omega(\eps_{n,\gamma}\eps_{n,\alpha}+\eps_{n,\gamma}^2+n^{-1/2})$. Both rates are attainable by DML.
    This is an informal consequence of Theorems~\ref{thm:main-mixed-bias} and \ref{thm:main}; see Section~\ref{sec:proof-main-thms}.
\end{theorem}

\subsection{Our main technical contribution}
\label{subsec:previous-proof-limitation}

Our proof of the lower bounds uses the method of fuzzy hypotheses, which reduces our estimation problem to testing between a pair of \textit{mixtures} of hypotheses. While such methods are widely adopted in establishing lower bounds for non-parametric functional estimation problems \cite{tsybakov2008introduction,robins2009semiparametric,kennedy2022minimax,balakrishnan2023fundamental}, we introduce a novel \emph{two-step sequential perturbation} technique to construct the null and alternative hypotheses with the desired properties. The two perturbation steps are asymmetric in general, and interchanging them would lead to two different types of optimal rates. We elaborate on this technique in Section \ref{subsec:proof-sketch}. Due to the more complicated relationship between the estimand and the data distribution, existing constructions of composite hypotheses \cite{robins2009semiparametric,kennedy2022minimax,balakrishnan2023fundamental} do not apply to our setting, as we explain next.

In \cite{balakrishnan2023fundamental}, the authors investigate the estimation problem of three functionals: quadratic functionals in Gaussian sequence models, quadratic integral functionals, and the expected conditional covariance. They establish their lower bound by reducing it to a related hypothesis testing problem. The testing error is then lower-bounded by constructing priors (mixtures) over the composite null and alternative hypotheses. The priors they construct are based on adding or subtracting ``bumps'' on top of a fixed hypothesis in a symmetric manner, which is a standard proof strategy for functional estimation problems \cite{ingster1994minimax,robins2009semiparametric,arias2018remember,balakrishnan2019hypothesis}. The reason why the proof strategy of \cite{balakrishnan2023fundamental} fails for \textsc{ATE} and most other functionals is that the functional relationships between the nuisance parameters and these target parameters take significantly different forms. Specifically, the target parameters that \cite{balakrishnan2023fundamental} investigates are all of the form
\begin{equation}
    \label{eq:form}
    T(f,g)=\left\langle f,g\right\rangle_{\gH},
\end{equation}
where $f,g$ are unknown nuisance parameters that lie in some Hilbert space $\gH$. To be concrete, consider the example of the expected conditional covariance $\theta^{\textsc{Cov}}$. Let $\mu_0(x) = \E\left[ Y \mid X=x\right]$, then we have that
\begin{equation}
    \theta^{\textsc{Cov}} = \E[DY] - \int m_0(x)\mu_0(x)\text{d} p_X(x)
\end{equation}
where $p_X$ is the marginal density of $X$. The first term, $\E[DY]$, can be estimated at the standard $\gO(n^{-1/2})$ rate, so it suffices to estimate the second term, which is exactly in the form of \eqref{eq:form}. However, the \ate~ functional does not take this inner product form. Instead, it is of the form
\opt{opt-arxiv}{\begin{align*}
    \notag
    T_1 (m_0,g_0) :=~& \E_{X} \left[g_0(1,X)-g_0(0,X) \right]= \E_{D,X} \left[\frac{D-m_0(X)}{m_0(X) (1-m_0(X))}g_0(D,X) \right].
\end{align*}}
\opt{opt-or}{
\begin{equation}
    \begin{aligned}
        &\quad T_1 (m_0,g_0) := \E_{X} \left[g_0(1,X)-g_0(0,X) \right]\\
        &= \E_{D,X} \left[\frac{D-m_0(X)}{m_0(X) (1-m_0(X))}g_0(D,X) \right].
    \end{aligned}
\end{equation}
}
Stepping outside of the realm of inner product functionals is the major challenge in extending existing approaches of establishing lower bounds to \ate~ and other relevant functionals, and is our main technical innovation.

\section{Optimality of first-order debiasing: the ATE case}
\label{sec:main-result}

Before going into full generality, we first revisit the ATE example to build intuition for first-order debiasing and the structure-agnostic viewpoint. In the standard setting, we observe $O=(X,D,Y)$, where $X$ is a
high-dimensional covariate vector, $D\in \{0,1\}$ is a binary treatment, and $Y\in\R$ is an outcome. Let $Y(1)$ and
$Y(0)$ denote the potential outcomes under each treatment level. The \emph{average treatment effect} (\ate) is defined as
\begin{align}
	    \theta^{\ate} :=~& \E\left[Y(1) - Y(0)\right]
\end{align}

We consider the case when all potential confounders $X\in \cX \subseteq \R^K$ of the treatment and outcome are observed,
a setting that has received substantial attention in the causal inference literature. In particular, we will make the
widely used assumption of \emph{conditional ignorability}:
\begin{align}
	    Y(1), Y(0) \ci D \mid X.
\end{align}

We assume that we are given data that consist of samples of the tuple of random variables $(X, D, Y)$, that satisfy the basic \emph{consistency} property 
\begin{align}
    Y = Y(D).
\end{align} 

Without loss of generality, the data generating process obeys the regression equations:
\begin{equation}
    \label{model}
    \begin{aligned}
        Y =~& g_0(D,X) + U, & \E\left[ U\mid D,X\right]=~& 0\\
        D =~& m_0(X) + V, & \E\left[V\mid X\right]=~& 0
    \end{aligned}
\end{equation}
where $U,V$ are noise variables. Note that when the outcome $Y$ is also binary, then the non-parametric functions $g_0$ and $m_0$, as well as the marginal probability law of the covariates $X$, fully determine the likelihood of the observed data.

Under conditional ignorability, consistency and the \emph{overlap assumption} that both treatment values are probable conditional on $X$, i.e., $m_0(X)\in [c, 1-c]$ almost surely, for some $c>0$, it is well known that the \ate~ is identified by the statistical estimands:
\begin{align}
\theta^{\ate} =~& \E[g_0(1, X) - g_0(0, X)].
\end{align}
This is the no-shift specialization ($Q_0=P_{0,Z}$) of \eqref{eq:parameter-of-interest}, with squared-loss regression for $\gamma$.

If we have access to a nuisance estimate $\hat{g}$, a straightforward approach is to plug it into \eqref{eq:parameter-of-interest} and replace the expectation with a sample average. However, this approach makes the estimation accuracy of the target parameter highly susceptible to errors in the outcome regression nuisance function, which can be large due to high dimensionality, regularization, and model selection. Moreover, the function spaces over which these estimators operate might be complex and do not necessarily satisfy commonly invoked Donsker conditions \citep{dudley2014uniform}.

To mitigate this dependence on the outcome regression model and to lift restrictions on the nuisance estimation algorithm
beyond root-mean-squared-error (RMSE) accuracy, first-order debiasing/DML uses sample splitting and an orthogonal score.
For the \ate, this yields a sample-splitting variant of the well-known doubly robust estimator; see, for example,
\citet{robins1995analysis,chernozhukov2018double,foster2023orthogonal}:
\opt{opt-arxiv}{
\begin{align}
        \label{eq:ate-debias-estimator}
        \hat{\theta}^{\ate} = \frac{1}{n}\sum_{i=1}^n \left[ \hat{g}(1,X_i) - \hat{g}(0,X_i) + \frac{D_i - \hat{m}(X_i)}{\hat{m}(X_i)\,(1-\hat{m}(X_i))}\left( Y_i - \hat{g}(D_i,X_i)\right) \right],
\end{align}}
\opt{opt-or}{
\begin{equation}
    \label{eq:ate-debias-estimator}
    \begin{aligned}
        &\hat{\theta}^{\ate} = \frac{1}{n}\sum_{i=1}^n \bigg[ \hat{g}(1,X_i) - \hat{g}(0,X_i) \\
        &+ \frac{D_i - \hat{m}(X_i)}{\hat{m}(X_i)\,(1-\hat{m}(X_i))}\left( Y_i - \hat{g}(D_i,X_i)\right) \bigg],
    \end{aligned}
\end{equation}}
where $\hat{g}, \hat{m}$ are estimates of $g_0$ and $m_0$ respectively. 

Even though the $n^{1/4}$ requirement can be achieved by a broad range of machine learning methods \citep{bickel2009simultaneous,belloni2011l1,belloni2013least,chen1999improved,wager2018estimation,athey2019generalized} (under assumptions), it can often be violated in practice. Even when this requirement is violated, a small modification of the arguments in \cite{chernozhukov2018double,foster2023orthogonal} can be used to show that $\hat{\theta}^{\ate} - \theta^{\ate} = \gO_P\left( \eps_{n,m}\eps_{n,g} + n^{-1/2}\right)$, under the assumption that
\opt{opt-arxiv}{\begin{equation}
\label{eq:nuisance-estimation-error}
    \|\hat{g}(d, X) - g_0(d, X)\|_{P_X, 2} \leq \eps_{n,g},  d\in\{0,1\} \quad\text{and}\quad
    \|\hat{m}(X) - m_0(X)\|_{P_X, 2} \leq \eps_{n,m}.
\end{equation}}
\opt{opt-or}{\begin{equation}
\label{eq:nuisance-estimation-error}
\begin{aligned}
    & \|\hat{g}(d, X) - g_0(d, X)\|_{P_X, 2} \leq \eps_{n,g},  d\in\{0,1\} \quad\text{and}\\
    & 
    \|\hat{m}(X) - m_0(X)\|_{P_X, 2} \leq \eps_{n,m}.
\end{aligned}
\end{equation}}

The formal statement of this result is presented below. The proof is in Section~\ref{sec:proof:ate-upper-bound}.
\begin{theorem}[Doubly robust ATE upper bound]
\label{thm:ate-upper-bound}
    Suppose that there exists a constant $c\in(0,1)$ such that $c\leq \hat{m}(x)\leq 1-c, \forall x\in\mathrm{supp}(X)$ and $|Y|\leq G$ \emph{a.s.}, for some constant $G$. Then for any $\delta>0$, there exists a constant $C_{\delta}$ such that the doubly robust estimator of the \ate~ (defined in \eqref{eq:ate-debias-estimator})
    achieves estimation error
    \begin{equation}
        \notag
        |\hat{\theta}^{\ate}-\theta^{\ate}| \leq C_{\delta}\left( \eps_{n,m}\eps_{n,g} + n^{-1/2} \right).
    \end{equation}
    with probability $\geq 1-\delta$.
\end{theorem}

Theorem \ref{thm:ate-upper-bound} highlights an important practical benefit of the doubly robust estimator: its accuracy depends only on the root-mean-squared error (RMSE) rates of the nuisance estimators, with no explicit structural assumptions on the nuisance classes. This is in stark contrast with higher-order debiasing schemes, which can lead to improved error rates \citep{bickel1988estimating,birge1995estimation,robins2008higher,van2014higher,robins2017minimax,liu2017semiparametric,kennedy2022minimax} under smoothness assumptions but no longer apply when these assumptions are violated.

To establish the matching lower bound, we restrict ourselves to the case of binary outcomes:

\begin{assumption}
    \label{asmp:outcome-binary}
    The outcome variable $Y$ is binary, \emph{i.e.}, $Y\in\{0,1\}$.
\end{assumption}

Given that the black-box nuisance function estimators satisfy \eqref{eq:nuisance-estimation-error}, we define the following constraint set
\opt{opt-arxiv}{\begin{equation}
    \label{local-set}
    \begin{aligned}
        \gM_1(\hat{P};\eps_{n,m},\eps_{n,g}) = \Bigl\{ & (m,g) \mid \mathrm{supp}(X)=[0,1]^K, P_X=\mathrm{Uniform}\bigl([0,1]^K\bigr),  \\
        & \Vert g(d,X)-\hat{g}(d,X)\Vert_{P_X,2} \leq \eps_{n,g},\ d\in\{0,1\}, \Vert m(X)-\hat{m}(X)\Vert _{P_X,2} \leq \eps_{n,m}, \\
        & 0\leq m(x),g(d,x)\leq 1,\forall x\in[0,1]^K\Bigr\}
    \end{aligned}
\end{equation}}
\opt{opt-or}{\begin{equation}
    \label{local-set}
    \begin{aligned}
        & \gM_1(\hat{P};\eps_{n,m},\eps_{n,g}) = \Bigl\{  (m,g) \mid \mathrm{supp}(X)=[0,1]^K, \\
        & P_X=\mathrm{Uniform}\bigl([0,1]^K\bigr),  \Vert g(d,X)-\hat{g}(d,X)\Vert_{P_X,2} \leq \eps_{n,g},\ d\in\{0,1\}, \\
        & \Vert m(X)-\hat{m}(X)\Vert _{P_X,2} \leq \eps_{n,m}, 0\leq m(x),\\
        & 0\leq g(d,x)\leq 1,\forall x\in[0,1]^K\Bigr\}
    \end{aligned}
\end{equation}}
where $\eps_{n,m},\eps_{n,g} = o(1)\quad (n\to +\infty).$ Note that introducing Assumption \ref{asmp:estimator-bounded} and constraints on $P_X$ in \eqref{local-set} only strengthens the lower bound that we are going to prove, since they provide additional information on the ground-truth model. Moreover, the constraints $0\leq m(x), g(d,x)\leq 1$ naturally holds due to the fact that both the treatment and outcome variables are binary.

We then define the minimax $(1-\xi)$-quantile risk of estimating $\theta^{\ate}$ over a function space $\gF$ as $$\mathfrak{M}_{n,\xi}^{\ate}\left(\gF\right) = \inf_{\hat{\theta}:\left(\gX\times\gD\times\gY\right)^n\mapsto\R} \sup_{(m^*,g^*)\in\gF} \gQ_{P_{m^*,g^*},1-\xi}\left( |\hat{\theta}-\theta^{\ate}| \right),$$
where $\gQ_{P,\gamma}(X)=\inf\left\{x\in\R: P[X\leq x]\geq\gamma\right\}$ denotes the quantile function of a random variable $X$ with distribution $P$, and $P_{m^*,g^*}$ is the joint distribution of $(X,D,Y)$ which is uniquely determined by the functions $m^*$ and $g^*$. Specifically, let $\mu$ be the uniform distribution on $\gX\times\gD\times\gY=[0,1]^K\times\{0,1\}\times\{0,1\}$, then the density $p_{m^*,g^*}=\dd P_{m^*,g^*}/{\dd\mu}$ can be expressed as $p_{m^*,g^*}(x,d,y) = m^*(x)^d(1-m^*(x))^{1-d}g^*(d,x)^y(1-g^*(d,x))^{1-y}.$

By definition, $\mathfrak{M}_{n,\xi}^{\ate}\left(\gF\right) \geq \rho$ would imply that for any estimator $\hat{\theta}$ of \textsc{ATE}, there must exist some $(m^*,g^*)\in\gF$, such that under the induced data distribution, the probability of $\hat{\theta}$ having estimation error $\geq\rho$ is at least $1-\xi$. This provides a stronger form of lower bound compared with the minimax \emph{expected} risk defined in \cite{balakrishnan2023fundamental}, in the sense that the lower bound $\mathfrak{M}_{n,\xi}^{\ate}\left(\gF\right) \geq \rho$ implies a lower bound $(1-\xi)\rho$ of the minimax expected risk, but the converse does not necessarily hold.

The main objective of this section is to derive lower bounds for $\mathfrak{M}_{n,\xi}^{\ate}\left(\gM_1(\hat{P};\eps_{n,m},\eps_{n,g})\right)$ in terms of $\eps_{n,m},\eps_{n,g}$ and $n$. To derive our lower bound, we also need to assume that the estimators $\hat{m}(x): [0,1]^K\mapsto[0,1]$ and $\hat{g}(d,x):\{0,1\}\times[0,1]^K\mapsto[0,1]$ are bounded away from $0$ and $1$.

\begin{assumption}
\label{asmp:estimator-bounded}
    There exists a constant $c$ such that $c \leq \hat{m}(x), \hat{g}(d,x) \leq 1-c$ for all $d\in\{0,1\}$ and $x\in[0,1]^K$.
\end{assumption}

The assumption that $c \leq \hat{m}(x)\leq 1-c$ is common in deriving upper bounds for the error induced by debiased estimators. On the other hand, the assumption that $c\leq \hat{g}(d,x)\leq 1-c$ is typically not needed for deriving upper bounds, but it is also made in prior works for proving \textit{lower bounds} of estimating the expected conditional covariance $\E\left[\mathrm{Cov}(D,Y)\mid X \right]$ \citep{robins2009semiparametric,balakrishnan2023fundamental}.

Now we are ready to state our main results for \ate.

\begin{theorem}[Minimax lower bound for ATE]
    \label{thm:ate-only}
    For any constant $\xi\in\left(\frac{1}{2},1\right)$ and estimators $\hat{m}(x)$ and $\hat{g}(d,x)$ that satisfy Assumption \ref{asmp:estimator-bounded}, the minimax risk of estimating the \ate~ is
    \begin{equation}
        \notag
        \mathfrak{M}_{n,\xi}^{\ate}\left(\gM_1(\hat{P};\eps_{n,m},\eps_{n,g})\right) = \Omega\left( \eps_{n,m}\eps_{n,g} + \min\{\eps_{n,g},n^{-1/2}\} \right)
    \end{equation}
\end{theorem}

The proof can be found in Section~\ref{sec:proof:ate-only}. As discussed in Section~\ref{subsec:previous-proof-limitation}, it relies on a fundamentally different construction of fuzzy hypotheses compared with the lower bound proof of ECC in \citet{balakrishnan2023fundamental}.

\begin{remark}
\label{remark:known-baseline}
    If we only assume that $c\leq\hat{m}(x), \hat{g}(1,x) \leq 1-c$ in Assumption \ref{asmp:estimator-bounded}, then we would still have the same lower bound.
    Furthermore, this lower bound still holds in the case where we know the baseline response, \emph{i.e.,} $\hat{g}(0,x)=g_0(0,x)=0$.
\end{remark}

\section{General error rates of first-order debiasing estimators}
\label{sec:first-order-debiasing-general}

In this section, we present the generic debiased estimator and its error bound (Theorem~\ref{thm:upper-bound-general})
in the general setting described in Section \ref{sec:intro}.
We then isolate the special ``affine-score'' case, in which the quadratic term $\eps_{N,\gamma}^2$ disappears and
the error becomes purely doubly robust. This affine case coincides with the \emph{mixed bias property} discussed in
Section~\ref{subsec:mixed-bias} and eventually leads to a different optimal error rate, as we will see in
Theorem~\ref{thm:main-mixed-bias}.

Assume that the linear functional $\gamma\mapsto \E_{Q}[m_1(Z,\gamma)]$ is continuous on $L^2(P_Z)$. Equivalently,
there exists a function $\nu_m(\cdot;P,Q)\in L^2(P_Z)$ such that for any $\gamma\in L^2(P_Z)$,
\begin{equation}
    \label{eq:riesz-representer}
    \E_{Q}\big[ m_1\big(Z,\gamma\big) \big] = \E_P \big[ \gamma(Z)\,\nu_m(Z;P,Q) \big].
\end{equation}

We think of $\nu_m(\cdot;P,Q)$ as the ``cross-population Riesz weight'' representing the linear functional
$\gamma\mapsto \E_{Q}[m_1(Z,\gamma)]$ under the $L^2(P_Z)$ inner product.
This identity is the direct analogue of the key representer condition used in the DML literature \citep{chernozhukov2018double} and still works in the presence of covariate shift \citep{chernozhukov2023automatic},

\paragraph{Generalized regression for $\gamma$ and the score $\rho$.}
The nuisance $\gamma(\cdot;P)$ is defined via generalized regression, i.e.\ as a pointwise minimizer of an expected
loss \eqref{eq:generalized-regression-formula} under the \emph{training} law $P$. By first-order optimality,
\begin{equation}
\label{eq:first-order-optimality-cond}
    \E_P\big[\rho\big(O,\gamma(Z;P)\big)\mid Z \big] = 0,
\end{equation}
where the score $\rho$ is the derivative of the loss in the regression direction,
$\rho(o,\gamma)=\frac{\dd}{\dd a}\ell(o,\gamma+a)\big|_{a=0}$.

\paragraph{The weighted Riesz identity and the auxiliary nuisance $\alpha$.}
Assuming the derivative $\nu_\rho(z;P)$ defined below exists and is nonzero, we define the auxiliary nuisance
\opt{opt-arxiv}{\begin{equation}
    \label{eq:weighted-riesz}
    \nu_{\rho}(z;P) := \frac{\dd}{\dd a} \E_P\big[\rho\big(O,\gamma(Z;P)+a\big)\mid Z=z \big]\Big|_{a=0}\quad \text{and}\quad
    \alpha(z;P,Q) = -\frac{\nu_m(z;P,Q)}{\nu_{\rho}(z;P)}.
\end{equation}}
\opt{opt-or}{
\begin{equation}
    \label{eq:weighted-riesz}
    \begin{aligned}
        \nu_{\rho}(z;P) &:= \frac{\dd}{\dd a} \E_P\big[\rho\big(O,\gamma(Z;P)+a\big)\mid Z=z \big]\Big|_{a=0},\\
        \alpha(z;P,Q) &:= -\frac{\nu_m(z;P,Q)}{\nu_{\rho}(z;P)}.
    \end{aligned}
\end{equation}
}
The definition of $\alpha$ is chosen so that the first-order sensitivity of the debiased estimator to
$\gamma$-perturbations cancels. Note that, unlike in the single-distribution setting, $\alpha(\cdot;P,Q)$ depends on
both the training law $P$ (through $\nu_\rho$) and the target law $Q$ (through $\nu_m$).

\paragraph{The first-order debiased estimator.}
Given black-box estimators $\hat{\gamma},\hat{\alpha}$ for $\gamma(\cdot;P_0)$ and $\alpha(\cdot;P_0,Q_0)$, the
(debiased / orthogonal) estimator is
\begin{equation}
    \label{eq:dml-estimator}
    \hat{\chi}
    =
    \frac{1}{N}\sum_{i=1}^N m_1\big(Z_i,\hat{\gamma}\big)
    +
    \frac{1}{N}\sum_{t=1}^N \hat{\alpha}(Z_t)\,\rho\big(O_t,\hat{\gamma}(Z_t)\big).
\end{equation}
In practice $\hat\gamma,\hat\alpha$ are obtained by sample splitting / cross-fitting; we suppress these details
since our focus is on the error scaling in $(\eps_{N,\gamma},\eps_{N,\alpha})$. The following theorem provides an upper bound on this scaling and the proof can be found in Section \ref{sec:proof:upper-bound-general}.

\begin{theorem}[Generic first-order debiasing upper bound under covariate shift]
\label{thm:upper-bound-general}
Suppose that $|\hat{\alpha}(z)|\leq A$, $|\alpha(z;P_0,Q_0)|\leq A$ and $|m_1(z,\hat{\gamma})|\leq C_m$ are uniformly bounded for $z\in\gZ$.
Assume also that the score is uniformly bounded at the truth and uniformly Lipschitz in its regression argument under the training law $P_0$:
$|\rho(O,\gamma(Z;P_0))|\le C_{\rho,0}$ almost surely and
\[
|\rho(o,\gamma)-\rho(o,\gamma')|\le C_{\rho,1}\,|\gamma-\gamma'|
\qquad\text{for all }o\in\gO\text{ and all }\gamma,\gamma'\in\R.
\]
Finally, assume there exist constants $C_{\rho,2}, r_{\rho,2} >0$ such that for any $\tilde{\gamma}\in L^2(P_{0,Z})$
satisfying $\|\tilde{\gamma}(Z)-\gamma(Z;P_0)\|_{P_{0,Z},2} \leq r_{\rho,2}$,
\opt{opt-arxiv}{
\begin{equation}
    \label{eq:def-2nd-order-effect-rho-equiv}
    \begin{aligned}
        \E_{P_0}\Bigg[
        \Bigg|
        \E_{P_0}\Big[
        &\rho\big(O,\tilde{\gamma}(Z)\big)-\rho\big(O,\gamma(Z;P_0)\big)\\
        &\quad-\nu_{\rho}(Z;P_0)\big(\tilde{\gamma}(Z)-\gamma(Z;P_0)\big)
        \,\Big|\, Z
        \Big]
        \Bigg|
        \Bigg]
        &\leq
        C_{\rho,2} \big\|\tilde{\gamma}(Z)-\gamma(Z;P_0)\big\|_{P_{0,Z},2}^2.
    \end{aligned}
\end{equation}
}
\opt{opt-or}{
\begin{equation}
    \label{eq:def-2nd-order-effect-rho-equiv}
    \begin{aligned}
        \E_{P_0}\Bigg[
        \Bigg|
        \E_{P_0}\Big[
        \rho\big(O,\tilde{\gamma}(Z)\big)-\rho\big(O,\gamma(Z;P_0)\big)
        -\nu_{\rho}(Z;P_0)\big(\tilde{\gamma}(Z)-\gamma(Z;P_0)\big)
        \,\Big|\, Z
        \Big]
        \Bigg|
        \Bigg]
        \leq
        C_{\rho,2} \big\|\tilde{\gamma}(Z)-\gamma(Z;P_0)\big\|_{P_{0,Z},2}^2.
    \end{aligned}
\end{equation}
}
If the nuisance estimators satisfy
$\|\hat{\gamma}(Z)-\gamma(Z;P_0)\|_{P_{0,Z},2}\leq \eps_{N,\gamma}$ and
$\|\hat{\alpha}(Z)-\alpha(Z;P_0,Q_0)\|_{P_{0,Z},2}\leq \eps_{N,\alpha}$,
and are constructed in a way such that the evaluation samples used in \eqref{eq:dml-estimator} are independent of $(\hat{\gamma},\hat{\alpha})$ (e.g., by sample splitting/cross-fitting), then for any $\delta>0$ there exists $C_{\delta}>0$ such that
\opt{opt-arxiv}{
\begin{equation}
    \label{eq:upper-bound-general}
    |\hat{\chi}-\chi(P_0,Q_0)|
    \leq
    C_{\delta}\Big(
    C_{\rho,1}\eps_{N,\gamma}\eps_{N,\alpha}
    + A\,C_{\rho,2}\eps_{N,\gamma}^2
    + (C_m + A\,C_{\rho,0})N^{-1/2}
    \Big)
\end{equation}
}
\opt{opt-or}{
\begin{equation}
    \label{eq:upper-bound-general}
    \begin{aligned}
    |\hat{\chi}-\chi(P_0,Q_0)|
    \leq\;
    C_{\delta}\Big(
    C_{\rho,1}\eps_{N,\gamma}\eps_{N,\alpha}
    + A\,C_{\rho,2}\eps_{N,\gamma}^2
    + (C_m + A\,C_{\rho,0})N^{-1/2}
    \Big).
    \end{aligned}
\end{equation}
}
with probability at least $1-\delta$.
In particular, if $\rho(o,\gamma)$ is affine in $\gamma$, then the conditional remainder in
\eqref{eq:def-2nd-order-effect-rho-equiv} vanishes (so one may take $C_{\rho,2}=0$) and
\eqref{eq:upper-bound-general} reduces to
\opt{opt-arxiv}{
\begin{equation}
    \label{upper-bound-affine}
    |\hat{\chi}-\chi(P_0,Q_0)|
    \leq
    C_{\delta}\Big(
    C_{\rho,1}\eps_{N,\gamma}\eps_{N,\alpha}
    + (C_m + A\,C_{\rho,0})N^{-1/2}
    \Big).
\end{equation}
}
\opt{opt-or}{
\begin{equation}
    \label{upper-bound-affine}
    \begin{aligned}
    |\hat{\chi}-\chi(P_0,Q_0)|
    \leq\;
    C_{\delta}\Big(
    C_{\rho,1}\eps_{N,\gamma}\eps_{N,\alpha}
    + (C_m + A\,C_{\rho,0})N^{-1/2}
    \Big).
    \end{aligned}
\end{equation}
}
\end{theorem}

Theorem~\ref{thm:upper-bound-general} shows that first-order debiasing yields a \emph{structure-agnostic} error bound:
up to the sampling term $N^{-1/2}$, the dominant contribution is either the product
$\eps_{N,\gamma}\eps_{N,\alpha}$ (the doubly robust rate) or, in the presence of curvature in the score, the additional
$\eps_{N,\gamma}^2$ term. Our main theorems show that these rates are not artifacts of the proof technique. Rather, they are actually
minimax optimal in terms of the nuisance estimation errors.

\subsection{The doubly robust regime and the mixed bias property}
\label{subsec:mixed-bias}

Theorem~\ref{thm:upper-bound-general} distinguishes two regimes: an \emph{affine-score} regime with doubly robust rate
$\eps_{N,\gamma}\eps_{N,\alpha}$, and a general regime with an extra $\eps_{N,\gamma}^2$ term.
We now explain the structural reason behind the affine-score regime. In this case the target functional admits a
\emph{second} linear representation in terms of the auxiliary nuisance $\alpha$. This is the mixed bias property of
\cite{rotnitzky2021characterization}, and it is exactly the condition under which the quadratic term disappears in both
upper and lower bounds.

\medskip

Suppose that $\rho(o,\gamma)$ is affine in $\gamma$, i.e.\ there exist measurable functions $\rho_0,\rho_1:\gO\to\R$ such that
\begin{equation}
    \label{eq:mixed-bias}
    \rho(o,\gamma)=\rho_0(o)+\rho_1(o)\gamma.
\end{equation}
Then the conditional first-order condition \eqref{eq:first-order-optimality-cond} becomes
$\E_P[\rho_0(O)\mid Z] + \E_P[\rho_1(O)\mid Z]\,\gamma(Z;P)=0$, hence
\[
\gamma(z;P)=-\frac{\E_P[\rho_0(O)\mid Z=z]}{\E_P[\rho_1(O)\mid Z=z]}
\qquad\text{and}\qquad
\nu_{\rho}(z;P)=\E_P[\rho_1(O)\mid Z=z],
\]
provided the denominator is nonzero.

Since $\alpha(z;P,Q)=-\nu_m(z;P,Q)/\nu_{\rho}(z;P)$, we have $\nu_m(z;P,Q)=-\alpha(z;P,Q)\nu_{\rho}(z;P)$ and therefore
\begin{align*}
\chi(P,Q)
&=\E_Q\big[m_1(Z,\gamma(\cdot;P))\big]
 =\E_P\big[\gamma(Z;P)\nu_m(Z;P,Q)\big]
 = -\E_P\big[\gamma(Z;P)\alpha(Z;P,Q)\nu_{\rho}(Z;P)\big] \\
&= \E_P\Big[\alpha(Z;P,Q)\,\E_P[\rho_0(O)\mid Z]\Big]
 = \E_P\big[\rho_0(O)\,\alpha(Z;P,Q)\big],
\end{align*}
where the last step uses that $\alpha(Z;P,Q)$ is $Z$-measurable.
Thus $\chi(P,Q)$ can also be written as the expectation of a linear functional of $\alpha$ (under the \emph{training} law):
\begin{equation}
    \label{eq:expectation-m2}
    \chi(P,Q) = \E_{P} \Big[ m_2\big(O,\alpha(Z;P,Q)\big)  \Big],
    \qquad
    m_2(o,h):=\rho_0(o)\,h.
\end{equation}
This is the \emph{mixed bias property}. In this regime the score has zero curvature in the $\gamma$ direction,
which is why the $\eps_{N,\gamma}^2$ term vanishes in Theorem~\ref{thm:upper-bound-general}.
Our lower-bound Theorem~\ref{thm:main-mixed-bias} shows that the remaining doubly robust rate
$\eps_{N,\gamma}\eps_{N,\alpha}$ is minimax optimal.

\section{A general framework for structure-agnostic estimation}
\label{sec:structure-agnostic-general}

\subsection{Problem set-up and main assumptions}
\label{subsec:main-asmp}

Our goal is to understand the best possible (minimax) accuracy for estimating a semiparametric functional \eqref{eq:parameter-of-interest} when the underlying data-generating mechanisms are only partially
identified through first-stage nuisance estimates. As in the ATE analysis, we work in a deliberately
\emph{structure-agnostic} fashion: we treat the first-stage learners as black boxes, and we quantify their quality only
through $L^2$-type error tolerances.

In this section we formalize the general lower bound framework by specifying
(i) the risk criterion and the data-generating experiment,
(ii) the uncertainty set of distribution pairs compatible with given nuisance-error tolerances, and
(iii) the regularity and curvature conditions under which our lower-bound constructions operate.

\begin{figure}[!htbp]
\vspace{-140pt}
\centering
\begin{tikzpicture}[scale=1]
\def\R{3}        %
\def\yc{1.4}     %
\def\f{0.35}     %
\def\W{7}                     %
\pgfmathsetmacro{\H}{\W*\f}   %
\pgfmathsetmacro{\eps}{0.03}  %
\pgfmathsetmacro{\aExact}{sqrt(\R*\R-\yc*\yc)}   %
\pgfmathsetmacro{\a}{\aExact-\eps}               %
\pgfmathsetmacro{\b}{\f*\a}                      %
\def\Diamond{(-\W,0) -- (0,\H) -- (\W,0) -- (0,-\H) -- cycle}
\begin{scope}
  \clip (-\W,0) rectangle (\W,\H);
  \fill[gray!25,opacity=0.30] \Diamond;
\end{scope}
\shade[ball color=white!90!black,opacity=0.25] (0,\yc) circle (\R);
\begin{scope}
  \clip (-\W,-\H) rectangle (\W,0);
  \fill[gray!55,opacity=0.30] \Diamond;
\end{scope}
\fill[cyan!60,opacity=0.60] (0,0) ellipse ({\a} and {\b});
\draw[thick]         (-\a,0) arc (180:360:{\a} and {\b}); %
\draw[thick,dashed]   (\a,0) arc (  0:180:{\a} and {\b}); %
\draw[thin,gray] node[above left] {uncertainty set};
\begin{scope}
  \clip (-\W,0) rectangle (\W,4*\H);   %
  \draw[thick] (0,\yc) circle (\R);
\end{scope}
\draw[thin,gray] \Diamond node[above right] {feasible nuisance space};
\draw[dashed] (0,\yc) -- (0,0);
\fill (0,\yc) circle (2pt) node[above left=-1pt] {$\hat{h}$};
\fill (0,0)   circle (2pt) node[right=4pt]       {$h(\cdot; \hat{P},\hat{Q})$};
\end{tikzpicture}
\caption{Schematic view of the \emph{anchored} analysis in the covariate-shift setting. The blue intersection represents the
uncertainty set of feasible pairs compatible with the nuisance-error constraints.}
\label{fig:anchoring-geometry}
\end{figure}
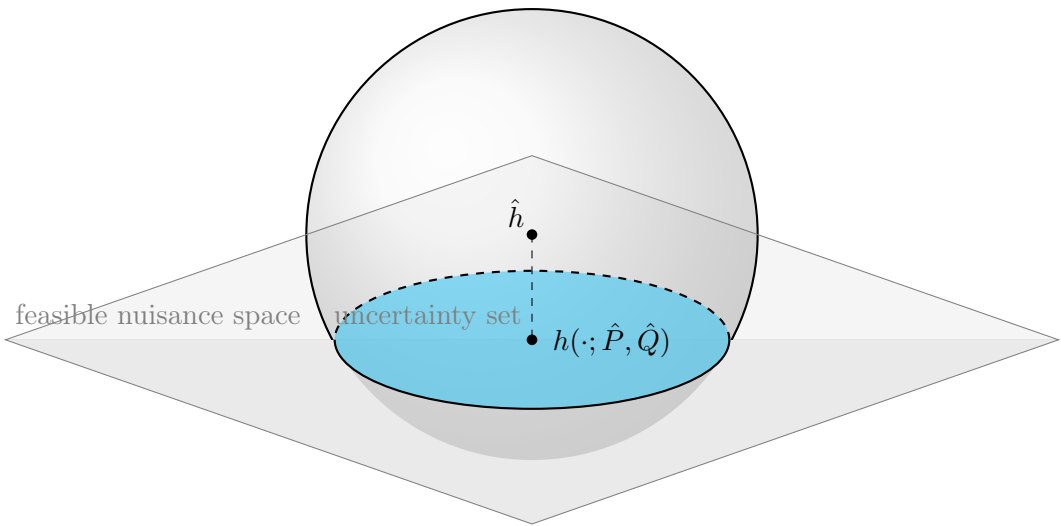

\paragraph{Target estimand.}
Recall that we consider semiparametric functionals of the form
\begin{equation}
\label{eq:cs-target-functional}
\chi(P,Q)
~:=~
\E_{Z\sim Q}\left[m_1\left(Z,\gamma(Z;P)\right)\right],
\end{equation}
where $P$ denotes a \emph{training} distribution for a generic observation $O=(Z,W)\in\gO$ and $Q$ denotes a
(possibly different) \emph{target} distribution for covariates $Z\in\gZ$.\footnote{Throughout this section we
take $\gZ=\gZ_1\times\gZ_2$ and $\gO=\gZ\times\gW$ as in Assumption~\ref{asmp:cond-prob-functional}.}
We emphasize that while the experiment provides \emph{separate samples} from $P$ and $Q$, the underlying model
class $\gP_0$ may impose \emph{coupling constraints} between them. Important special cases include:
(i) \emph{no covariate shift}, where $Q$ equals the $Z$-marginal of $P$ (often written informally as ``$Q=P$''), and
(ii) selection/conditioning operators such as ATT, where $Q$ is a conditional distribution derived from $P$, as shown in Example \ref{example:att}.
Throughout, we work under the simplifying assumption that the training and target sample sizes are equal, and we write
$\eps_{N,\gamma}$ and $\eps_{N,\alpha}$ for the nuisance-error tolerances of $\gamma$ and $\alpha$, respectively.

\paragraph{Risk criterion.}
For $\xi\in(0,1)$ and any collection $\mathcal{P}$ of candidate pairs $(P,Q)$, we define the minimax
$(1-\xi)$-quantile risk as
\begin{equation}
\label{eq:cs-minimax-risk}
\mathfrak{M}^{\chi}_{N,\xi}(\mathcal{P})
~:=~
\inf_{\hat\chi:\,\gO^N\times\gZ^N\to\mathbb{R}}
\sup_{(P,Q)\in\mathcal{P}}
Q_{P^{\otimes N}\otimes Q^{\otimes N},\,1-\xi}\left(\left|\hat\chi-\chi(P,Q)\right|\right),
\end{equation}
where $Q_{R,1-\xi}(\cdot)$ denotes the $(1-\xi)$-quantile under $R$. Quantile risk avoids imposing tail assumptions on
$\hat\chi-\chi(P,Q)$ and is convenient for the fuzzy-hypothesis arguments used in the proofs.
We use the same letter $N$ for both the training and the target sample sizes.

\subsubsection{Structure-agnostic uncertainty sets}

As in our upper bound analysis, we treat first-stage nuisance estimators as
black boxes and quantify their quality only through $L^2$-type error bounds. Let $\hat\gamma:\gZ\to\mathbb{R}$ and
$\hat\alpha:\gZ\to\mathbb{R}$ denote (possibly data-dependent) estimators of the nuisance functions
$\gamma(\cdot;P)$ and $\alpha(\cdot;P,Q)$.
Given error tolerances $\eps_{N,\gamma}$ and $\eps_{N,\alpha}$, define the (data-dependent) collection of admissible
distributional pairs by
\begin{equation}
\label{eq:cs-uncertainty-set-data}
\gP\left(\hat\gamma,\hat\alpha;\,\eps_{N,\gamma},\eps_{N,\alpha}\right)
:=
\left\{
(P,Q)\in\gP_0:
\left\|\hat\gamma(Z)-\gamma(Z;P)\right\|_{P_Z,2}\le \eps_{N,\gamma},~
\left\|\hat\alpha(Z)-\alpha(Z;P,Q)\right\|_{P_Z,2}\le \eps_{N,\alpha}
\right\},
\end{equation}
where $P_Z$ is the $Z$-marginal of the training law $P$.

As before, we lower bound the minimax risk by anchoring at a single feasible pair. In general, the first-stage
nuisance estimates $\hat h$ (think $(\hat\gamma,\hat\alpha)$) need not be exactly induced by any feasible model pair.
Lemma~\ref{lemma:anchoring} shows that for lower bounds it is enough to work in an anchored neighborhood around a feasible
pair $(\hat P,\hat Q)$ whose induced nuisances are within the same error tolerances. This reduction is illustrated in Figure \ref{fig:anchoring-geometry}.

\begin{lemma}[Anchoring to a feasible nuisance pair]
\label{lemma:anchoring}
Let $\gP_1\subseteq\gP_0$.
Suppose there exists $(\hat{P},\hat{Q})\in\gP_0$ such that
$\|\hat{\gamma}(Z)-\gamma(Z;\hat{P})\|_{\hat{P}_Z,2} \leq \eps_{n,\gamma}/2$ and
$\|\hat{\alpha}(Z)-\alpha(Z;\hat{P},\hat{Q})\|_{\hat{P}_Z,2} \leq \eps_{n,\alpha}/2$.
Then
\begin{equation}
    \notag
    \mathfrak{M}_{N,\xi}^{\chi}\!\left(\gP_1\cap\gP(\hat{\gamma},\hat{\alpha};\eps_{n,\gamma},\eps_{n,\alpha})\right)
    \geq
    \mathfrak{M}_{N,\xi}^{\chi}\!\left(\gP_1\cap\gP\big(\gamma(\cdot;\hat{P}),\alpha(\cdot;\hat{P},\hat{Q});\eps_{n,\gamma}/2,\eps_{n,\alpha}/2\big)\right).
\end{equation}
\end{lemma}

\begin{proof}
By the triangle inequality, if a distribution pair $(P,Q)\in\gP_0$ satisfies
$\|\gamma(Z;\hat{P})-\gamma(Z;P)\|_{P_Z,2} \leq \eps_{N,\gamma}/2$ and
$\|\alpha(Z;\hat{P},\hat{Q})-\alpha(Z;P,Q)\|_{P_Z,2} \leq \eps_{N,\alpha}/2$,
then we necessarily have. 
\[
\|\gamma(Z;P)-\gamma(Z;\hat P)\|_{P_Z,2}
~\le~
\|\gamma(Z;P)-\hat\gamma(Z)\|_{P_Z,2}+\|\hat\gamma(Z)-\gamma(Z;\hat P)\|_{P_Z,2}
~\le~\eps_{N,\gamma},
\]
and likewise $\|\alpha(Z;P,Q)-\alpha(Z;\hat P,\hat Q)\|_{P_Z,2}\le \eps_{N,\alpha}$.
Hence
\[
\gP\big(\gamma(\cdot;\hat{P}),\alpha(\cdot;\hat{P}, \hat{Q});\eps_{N,\gamma}/2,\eps_{N,\alpha}/2\big)
\subseteq
\gP(\hat{\gamma},\hat{\alpha};\eps_{N,\gamma},\eps_{N,\alpha}),
\]
and the claimed lower bound follows by monotonicity in the distribution class.
\end{proof}

In what follows we fix such an anchor pair $(\hat P,\hat Q)$ and focus on the risk over the anchored class
$\gM((\hat P,\hat Q);\eps_{N,\gamma},\eps_{N,\alpha})$.
For notational simplicity we also set
\[
\hat\gamma(\cdot):=\gamma(\cdot;\hat P),
\qquad
\hat\alpha(\cdot):=\alpha(\cdot;\hat P,\hat Q),
\]
so that the nuisance constraints are centered at the anchor.

\subsubsection{Technical assumptions}

We now state the main conditions needed for establishing our general lower bounds.

\begin{assumption}[Bounded densities]
\label{asmp:density-bounded}
The anchor pair $(\hat P,\hat Q)$ is absolutely continuous with respect to dominating measures $\mu$ on $\gO$
and $\mu_Z$ on $\gZ$, with densities $\hat p:=d\hat P/d\mu$ and $\hat q:=d\hat Q/d\mu_Z$ satisfying, for some
constants $0<b_0\le b_1<\infty$,
\[
b_0\le \hat p(o)\le b_1 \ \ \text{for $\mu$-a.e.\ $o\in\gO$},
\qquad
b_0\le \hat q(z)\le b_1 \ \ \text{for $\mu_Z$-a.e.\ $z\in\gZ$}.
\]
\end{assumption}

Assumption~\ref{asmp:density-bounded} requires that the anchor training and target laws admit densities (with respect to
the chosen dominating measures) that are uniformly bounded above and away from zero.

\begin{definition}[Nondegenerate measure space]
\label{def:nondegenerate}
We say that a measure space $(\mathcal{Z},\mu)$ is $K$-nondegenerate if there exist bounded
$\mu$-measurable functions $f_1,\ldots,f_K$ on $\mathcal{Z}$ such that for any
$(\lambda_1,\ldots,\lambda_K)\neq 0$,
\[
  \mu\Big(\Big\{ z\in \mathcal{Z} : \sum_{k=1}^K \lambda_k f_k(z)=0\Big\}\Big)=0.
\]
\end{definition}

\begin{assumption}[Nondegenerate slicing covariate]
\label{asmp:cond-space-nontrivial}
Let $K^*:=10$. The measure space $(\gZ_1,\mu_{Z_1})$ is $K^*$-nondegenerate
(Definition~\ref{def:nondegenerate}).
\end{assumption}

Assumption~\ref{asmp:cond-space-nontrivial} requires that $(\gZ_1,\mu_{Z_1})$ admits enough ``degrees of freedom''. A sufficient condition in our applications
is that $\gZ_1\subset\mathbb{R}^d$ and $\mu_{Z_1}$ has a density on a set with non-empty interior (e.g., one may take
$f_k$ as coordinate monomials up to the required order).

\begin{assumption}[Conditional-density factorization and orthogonal-score objects]
\label{asmp:cond-prob-functional}
Let $\gO=\gZ_1\times\gZ_2\times \gW$ and write $O=(Z_1,Z_2,W)$, $Z=(Z_1,Z_2)$.
Let $\mu=\mu_{Z_1}\otimes\mu_{Z_2}\otimes\mu_W$ and $\mu_Z=\mu_{Z_1}\otimes\mu_{Z_2}$.
Let $\gM$ be a convex set of $\mu_{Z_2}\otimes\mu_W$-integrable functions.

\begin{enumerate}
\item Local dependence on conditional slices.
For each training law $P$ with density $p$ and each $z_1\in\gZ_1$, define the slice
$p_{z_1}(z_2,w):=p(z_1,z_2,w)$.
For each target law $Q$ with density $q$ and each $z_1\in\gZ_1$, define the slice
$q_{z_1}(z_2):=q(z_1,z_2)$.
There exist mappings
\[
\Gamma_\gamma:\gM\to L^2(\mu_{Z_2}),\qquad
\Gamma_\alpha:\gM\times L^1(\mu_{Z_2})\to L^2(\mu_{Z_2})
\]
such that, whenever $p_{z_1}\in\gM$ for $\mu_{Z_1}$-a.e.\ $z_1$ and $q_{z_1}\in L^1(\mu_{Z_2})$,
\[
\gamma(z_1,z_2;P)=\Gamma_\gamma(p_{z_1})(z_2),
\qquad
\alpha(z_1,z_2;P,Q)=\Gamma_\alpha(p_{z_1},q_{z_1})(z_2).
\]

\item Linear functional and (cross-)Riesz representers.
There exists a mapping $m_1:\gZ\times L^2(\mu_Z)\to \mathbb{R}$ which is linear in its second argument and a
measurable function $\nu_\rho(\cdot;P)$ such that the (cross-population) Riesz identity
\eqref{eq:riesz-representer} and weighted Riesz identity \eqref{eq:weighted-riesz} hold, with a representer
$\nu_m(\cdot;P,Q)$ that may depend on both $P$ and $Q$, and with
\[
\alpha(z;P,Q)=-\frac{\nu_m(z;P,Q)}{\nu_\rho(z;P)}.
\]
In particular, the target functional \eqref{eq:cs-target-functional} can be written as
$\chi(P,Q)=\E_{Z\sim Q}[m_1(Z,\gamma(Z;P))]$.
\end{enumerate}
\end{assumption}

Assumption~\ref{asmp:cond-prob-functional}(1) formalizes that the training nuisance $\gamma$ is a ``conditional''
functional: for each fixed $Z_1=z_1$, the function $z_2\mapsto \gamma(z_1,z_2;P)$ depends on $P$ only through the slice
$p_{z_1}(z_2,w)=p(z_1,z_2,w)$. The same is true for $\alpha$, except that in the covariate-shift setting $\alpha$ may
also depend on the corresponding target slice $q_{z_1}(z_2)=q(z_1,z_2)$. This setup covers many familiar nuisances (e.g., outcome
regressions and propensities) that are computed by conditioning on part of the covariates.

\begin{definition}[Feasible training laws and feasible covariate-shift pairs]
\label{def:feasible-pair}
In the setup of Assumption~\ref{asmp:cond-prob-functional}, we call a training distribution $P$ on $\gO$ \emph{$\gM$-feasible}
if $P\ll\mu$ and $p_{z_1}\in\gM$ for $\mu_{Z_1}$-a.e.\ $z_1\in\gZ_1$. We call a pair $(P,Q)$ \emph{$\gM$-feasible} if $P$ is
$\gM$-feasible and $Q\ll\mu_Z$.
\end{definition}

The role of $\gM$ is to encode any regularity restrictions needed to make the nuisances well-defined (e.g., overlap or
boundedness of denominators). Importantly, $\gM$ is a \emph{local} constraint: it must hold slice-by-slice in $z_1$.

We now formally define distributional perturbations that are allowed in our setting, where the model class $\gP_0$ may impose
\emph{coupling constraints} between the training law $P$ and the target law $Q$ (e.g.\ $Q=P_Z$ in the no-shift case, or
conditioning/selection operators such as the ATT in Example \ref{example:att}).
A \emph{training perturbation} $G$ is a finite signed measure on $\gO$ with $G(\gO)=0$ and $G\ll \mu$, and we write
$P+tG$ for the signed measure with density $p+t g$ with respect to $\mu$ whenever $g=\dd G/\dd\mu$ exists.
A \emph{target perturbation} $K$ is a finite signed measure on $\gZ$ with $K(\gZ)=0$ and $K\ll \mu_Z$, and we write
$Q+tK$ analogously.

\begin{definition}[Feasible joint perturbations]
\label{def:joint-perturbation}
Fix an anchor pair $(\hat P,\hat Q)\in\gP_0$.
A pair $(G,K)$ of signed measures, with $G$ on $\gO$ and $K$ on $\gZ$, is called a \emph{feasible joint perturbation}
(at $(\hat P,\hat Q)$, relative to $\gP_0$) if:
\begin{enumerate}
\item $G(\gO)=0$, $K(\gZ)=0$, and $G\ll\mu$, $K\ll\mu_Z$ with essentially bounded densities; and
\item there exists $r_{G,K}>0$ such that for all $t\in[-r_{G,K},r_{G,K}]$,
\[
(\hat P+tG,\hat Q+tK)\in \gP_0,
\qquad
d_{\mu,\infty}(\hat P+tG,\hat P)<\infty,
\qquad
d_{\mu_Z,\infty}(\hat Q+tK,\hat Q)<\infty,
\]
so that $(\hat P+tG,\hat Q+tK)$ remains absolutely continuous with respect to $(\mu,\mu_Z)$ and stays within the local
neighborhood on which Assumption~\ref{asmp:regularity} applies.
\end{enumerate}
\end{definition}

\begin{definition}[$Z_1$-modulation closure of a feasible joint perturbation]
\label{def:z1-modulation-closure}
Fix an anchor pair $(\hat P,\hat Q)\in\gP_0$ and let $(G,K)$ be a feasible joint perturbation at $(\hat P,\hat Q)$ in
the sense of Definition~\ref{def:joint-perturbation}.
For any measurable $\psi:\gZ_1\to\R$ with $\|\psi\|_{\infty}<\infty$, define the \emph{$Z_1$-modulated} signed measures
$(G^\psi,K^\psi)$ by
\[
\frac{\dd G^\psi}{\dd\mu}(o):=\psi(z_1)\,\frac{\dd G}{\dd\mu}(o),
\qquad
\frac{\dd K^\psi}{\dd\mu_Z}(z):=\psi(z_1)\,\frac{\dd K}{\dd\mu_Z}(z).
\]
We say that $(G,K)$ is \emph{$Z_1$-modulation closed at $(\hat P,\hat Q)$} if there exists
$r_{G,K}^{\mathrm{mod}}>0$ such that for every $\psi$ with $\|\psi\|_{\infty}\le 1$ satisfying the centering
conditions $G^\psi(\gO)=0$ and $K^\psi(\gZ)=0$, we have
\[
(\hat P+tG^\psi,\hat Q+tK^\psi)\in \gP_0
\qquad\text{for all }t\in[-r_{G,K}^{\mathrm{mod}},r_{G,K}^{\mathrm{mod}}].
\]
\end{definition}

\begin{remark}[No-shift coupling]
In the no-shift model class where $Q=P_Z$, feasible joint perturbations necessarily satisfy $K=G_Z$.
For any $\psi$ as above, the marginalization commutes with $Z_1$-modulation, i.e.\ $(G^\psi)_Z=K^\psi$.
Thus, once the centering condition $G^\psi(\gO)=0$ holds (equivalently $K^\psi(\gZ)=0$), the pair
$(\hat P+tG^\psi,\hat Q+tK^\psi)$ automatically satisfies the coupling constraint $Q_t=(P_t)_Z$ for sufficiently
small $|t|$.
\end{remark}

In coupled models, it is generally \emph{not} possible to perturb $P$ while keeping $Q$ fixed. Accordingly, all
directional derivatives in the lower bound will be taken along feasible \emph{joint} perturbations $(G,K)$.

\begin{assumption}[Uniform smoothness on a local neighborhood]
\label{asmp:regularity}
There exist finite constants $r$, $c_t$, $L_{1}$, $L_{2}$, and $L_{\chi,2}$ such that the following hold uniformly for every pair
$(P,Q)\in\gP_0$ satisfying
\[
d_{\mu,\infty}(P,\hat P)\le r,
\qquad
d_{\mu_Z,\infty}(Q,\hat Q)\le r.
\]
In the statements below, all $L^2$ norms are taken with respect to $\hat P_Z$, and ``feasible joint perturbation at
$(P,Q)$'' is understood in the sense of Definition~\ref{def:joint-perturbation} with anchor $(P,Q)$.

\begin{enumerate}
\item (Second-order directional (G\^ateaux) differentiability.)
The map $P\mapsto \gamma(\cdot;P)$ is twice directionally (G\^ateaux) differentiable at $P$, and the bivariate map
$(P,Q)\mapsto \alpha(\cdot;P,Q)$ is twice directionally (G\^ateaux) differentiable at $(P,Q)$.
We denote the first and second derivatives by
\[
\gamma_P'(\cdot;P)[G],\quad \gamma_P''(\cdot;P)[G_0,G_1],
\qquad
\alpha_{(P,Q)}'(\cdot;P,Q)[G,K],\quad
\alpha_{(P,Q)}''(\cdot;P,Q)[G_0,K_0;G_1,K_1].
\]

\item (Second-order remainder bounds.)
For all feasible joint perturbations $(G,K)$ at $(P,Q)$ and all $|t|\le c_t$,
\begin{align*}
\left\|\gamma(\cdot;P+tG)-\gamma(\cdot;P)-t\,\gamma_P'(\cdot;P)[G]\right\|_{\hat P_Z,2}
&\le L_{2}\,t^2\|G\|_{\mathrm{TV}}^2,\\
\left\|\alpha(\cdot;P+tG,Q+tK)-\alpha(\cdot;P,Q)-t\,\alpha_{(P,Q)}'(\cdot;P,Q)[G,K]\right\|_{\hat P_Z,2}
&\le L_{2}\,t^2\big(\|G\|_{\mathrm{TV}}+\|K\|_{\mathrm{TV}}\big)^2.
\end{align*}

			\item (Local Lipschitz bounds.)
			For all feasible joint perturbations $(G_0,K_0)$, $(G_1,K_1)$ at $(P,Q)$ and all $|t|\le c_t$,
			\begin{align*}
			\left\|\gamma_P'(\cdot;P+tG_0)[G_1]-\gamma_P'(\cdot;P)[G_1]\right\|_{\hat P_Z,2}
			&\le L_{1}\,|t|\,\|G_0\|_{\mathrm{TV}}\|G_1\|_{\mathrm{TV}},\\
			\left\|\alpha_{(P,Q)}'(\cdot;P+tG_0,Q+tK_0)[G_1,K_1]-\alpha_{(P,Q)}'(\cdot;P,Q)[G_1,K_1]\right\|_{\hat P_Z,2}
			&\le L_{1}\,|t|\,\big(\|G_0\|_{\mathrm{TV}}+\|K_0\|_{\mathrm{TV}}\big)\\
				&\quad\times \big(\|G_1\|_{\mathrm{TV}}+\|K_1\|_{\mathrm{TV}}\big).
				\end{align*}

		\item (Uniform second-order remainder for $\chi$.)
		For each $(P,Q)$ and direction $(G,K)$ as above, define
		\[
		\chi_{(P,Q)}'(P,Q)[G,K]
		:=
	\left.\frac{\partial}{\partial t}\right|_{t=0}\chi(P+tG,Q+tK).
	\]
	Then for all feasible joint perturbations $(G,K)$ at $(P,Q)$ and all $|t|\le c_t$,
	\[
	\left|\chi(P+tG,Q+tK)-\chi(P,Q)-t\,\chi_{(P,Q)}'(P,Q)[G,K]\right|
	\le L_{\chi,2}\,t^2\big(\|G\|_{\mathrm{TV}}+\|K\|_{\mathrm{TV}}\big)^2.
	\]
			\end{enumerate}

		\noindent Moreover, for all such $(P,Q)$ and all feasible joint perturbations $(G,K)$ and $(G_1,K_1)$ at $(P,Q)$ with
	$\|G\|_{\mathrm{TV}}+\|K\|_{\mathrm{TV}}\le 1$ and $\|G_1\|_{\mathrm{TV}}+\|K_1\|_{\mathrm{TV}}\le 1$, we have
\begin{equation}
\label{eq:asmp-regularity-derivative-bounded}
\begin{aligned}
\max\Big\{\big|\gamma_P'(z;P)[G]\big|, \big|\alpha_{(P,Q)}'(z;P,Q)[G,K]\big|\Big\} &\le L_{1},\\
\max\Big\{\big|\gamma_P''(z;P)[G,G_1]\big|, \big|\alpha_{(P,Q)}''(z;P,Q)[G,K;G_1,K_1]\big|\Big\} &\le L_{2},
\qquad\forall z\in\gZ.
\end{aligned}
\end{equation}
\end{assumption}

Assumption~\ref{asmp:regularity} is a local smoothness condition: in a neighborhood of the anchor pair, the maps
$P\mapsto \gamma(\cdot;P)$ and $(P,Q)\mapsto \alpha(\cdot;P,Q)$ admit second-order expansions along feasible perturbation
paths, with remainders that are uniformly controlled. The same type of control is imposed directly on the target
functional $\chi(P,Q)$. Intuitively, this means that small distributional changes lead to small and predictable changes in
the nuisances and the estimand (up to quadratic error), rather than producing discontinuous jumps. Note that $\gamma$
depends only on the training law, while $\alpha$ may depend on both the training and target laws through the
cross-population Riesz object.

We finally state the key assumption needed for our main results.

\begin{assumption}[Invariant directions, non-degenerate curvature, and a feasible parametric direction]
\label{asmp:main}
We write
\[
\chi''(\hat P,\hat Q)[(G_0,K_0),(G_1,K_1)]
\]
for the mixed second derivative of the bivariate map $(P,Q)\mapsto \chi(P,Q)$ at $(\hat P,\hat Q)$ in directions
$(G_0,K_0)$ and $(G_1,K_1)$.
There exist feasible joint perturbations $(G_0,K_0)$, $(G_1,K_1)$, $(H_0,L_0)$, $(H_1,L_1)$ at $(\hat P,\hat Q)$
and a constant $c_t>0$ such that:

\begin{enumerate}
\item (Invariant directions.)
For all $|t|\le c_t$,
\[
\gamma(\cdot;\hat P+tG_0)=\gamma(\cdot;\hat P),
\qquad
\alpha(\cdot;\hat P+tH_0,\hat Q+tL_0)=\alpha(\cdot;\hat P,\hat Q).
\]

\item (Two-step feasibility along the lower-bound directions.)
For every measurable $\psi:\gZ_1\to\R$ with $\|\psi\|_{\infty}\le 1$ such that the $Z_1$-modulated pairs
$(G_k^\psi,K_k^\psi)$, $k\in\{0,1\}$, satisfy the centering conditions in Definition~\ref{def:z1-modulation-closure}, we have
\[
(\hat P+sG_0^\psi+tG_1^\psi,\ \ \hat Q+sK_0^\psi+tK_1^\psi)\in\gP_0
\qquad\text{for all }|s|\le c_t,\ |t|\le c_t,
\]
where $(G_k^\psi,K_k^\psi)$ denotes the $Z_1$-modulation of $(G_k,K_k)$ by $\psi$.
The same condition holds with $(H_k,L_k)$ in place of $(G_k,K_k)$.

\item (Non-degenerate mixed curvature.)
The mixed second derivatives satisfy
\[
\chi''(\hat P,\hat Q)[(G_0,K_0),(G_1,K_1)]\neq 0,
\qquad
\chi''(\hat P,\hat Q)[(H_0,L_0),(H_1,L_1)]\neq 0.
\]

\item (A feasible parametric direction for $\chi$.)
There exists a feasible joint perturbation $(G_{\mathrm{LC}},K_{\mathrm{LC}})$ at $(\hat P,\hat Q)$ such that the
first-order directional derivative of $(P,Q)\mapsto\chi(P,Q)$ along this direction is nonzero:
\[
\chi_{(P,Q)}'(\hat P,\hat Q)[G_{\mathrm{LC}},K_{\mathrm{LC}}]\neq 0,
\]
where $\chi_{(P,Q)}'(\hat P,\hat Q)[\cdot,\cdot]$ is defined in Assumption~\ref{asmp:regularity}(4).
\end{enumerate}
\end{assumption}

Assumption~\ref{asmp:main} requires the existence, at the anchor, of feasible perturbation directions that keep the key
nuisances fixed along small paths (invariance), while the target functional still exhibits a nonzero mixed second-order
response when combining these directions (non-degenerate curvature). In addition, it posits the existence of a feasible
direction along which the estimand varies to first order. 

The following proposition shows that this assumption is naturally satisfied in a canonical non-shift setting. The proof can be found in Section~\ref{sec:proof:main-asmp-suff-cond}.

\begin{proposition}[Sufficient condition for invariant perturbations]
\label{prop:main-asmp-suff-cond}
Consider the no-shift case where $Q=P_{Z}$ and assume that $\alpha(z;P)=\E_P[F_0(O)\mid Z=z]/\E_P[F_1(O)\mid Z=z]$ is uniformly bounded, where $F_i\in L^2(\gO)$ are bounded functions
and the conditional expectations are well-defined. Suppose further that for all $z\in\gZ$,
\begin{equation}
    \label{eq:non-affine}
    \min_{a,b\in\R}\E_{\mu} \big[ (F_0(Z, W)-aF_1(Z, W)-b)^2 \mid Z = z \big] > c,
\end{equation}
for some constant $c>0$ (i.e., $F_0$ is not nearly affine in $F_1$ conditional on $Z$).
If $(\gW,\mu_{\gW})$ is $3$-nondegenerate (cf.~Definition~\ref{def:nondegenerate}) and $\hat{P}$ is in the interior of $\gM$
under the distance $d_{\mathsf{c},\infty}(\hat{P},P):=\sup_{o=(z,w)\in\gO}|\hat{p}(w\mid z)-p(w\mid z)|$,
then there exists a $\gM$-feasible perturbation $H_0$ (with $L_0 = H_{0,Z}$) that satisfies Assumption~\ref{asmp:main}(1) for $\alpha$.
Moreover, if $\alpha$ is not locally constant in a $d_{\mathsf{c},\infty}$-neighborhood of $\hat{P}$, then there exists a $\gM$-feasible perturbation
$H_1$ such that $\chi''(\hat{P})[H_0,H_1] \neq 0$. The same conclusion holds with $\alpha$ replaced by $\gamma$.
\end{proposition}

Finally, we define the anchored candidate class we will lower bound.
\begin{equation}
\label{eq:candidate-distribution}
\gM\left((\hat P,\hat Q);\eps_{N,\gamma},\eps_{N,\alpha}\right)
:=
\left\{
(P,Q) \text{ is } \gM\text{-feasible} \mid \|\gamma(\cdot;P)-\hat\gamma(\cdot)\|_{P_Z,2}\le \eps_{N,\gamma}, \|\alpha(\cdot;P,Q)-\hat\alpha(\cdot)\|_{P_Z,2}\le \eps_{N,\alpha}
\right\}.
\end{equation}

\section{Optimality of first-order debiasing: the general case}

\subsection{Main lower bound results}
\label{subsec:main-results-lowerbound}

We now state the minimax lower bounds for the covariate shift functional \eqref{eq:cs-target-functional} over the
anchored, structure-agnostic uncertainty set \eqref{eq:candidate-distribution}. Recall that we observe an i.i.d.\
training sample of size $N$ from $P$ and an independent i.i.d.\ target sample of size $N$ from $Q$.

\begin{theorem}[Mixed-bias lower bound under covariate shift]
\label{thm:main-mixed-bias}
Assume \ref{asmp:density-bounded}, \ref{asmp:cond-space-nontrivial}, \ref{asmp:cond-prob-functional},
\ref{asmp:regularity}, and \ref{asmp:main} hold for some joint perturbations $(G_0,K_0)$, $(G_1,K_1)$, $(H_0,L_0)$,
$(H_1,L_1)$, and $(G_{\mathrm{LC}},K_{\mathrm{LC}})$.
Assume in addition that each of these perturbation pairs is $Z_1$-modulation closed at $(\hat P,\hat Q)$ in the sense
of Definition~\ref{def:z1-modulation-closure}.
In addition, assume that the mapping $\gamma\mapsto\rho(o,\gamma)$ is affine in $\gamma$.
Assume further that the nuisance-error tolerances are not smaller than the parametric scale: there exists a constant
$c_{\min}>0$ such that, for all sufficiently large $N$,
\[
\eps_{N,\gamma}\ge c_{\min}N^{-1/2}
\qquad\text{and}\qquad
\eps_{N,\alpha}\ge c_{\min}N^{-1/2}.
\]

Then for any $\xi\in(1/2,1)$, there exists a constant $\delta>0$ (depending only on the constants in the
assumptions) such that, for all sufficiently large $N$,
\[
\mathfrak{M}^{\chi}_{N,\xi}\left(\gM\left((\hat P,\hat Q);\eps_{N,\gamma},\eps_{N,\alpha}\right)\right)
~=~
\Omega\left(\eps_{N,\gamma}\,\eps_{N,\alpha}+N^{-1/2}\right).
\]
\end{theorem}

\begin{theorem}[Non-affine lower bound under covariate shift]
\label{thm:main}
Assume \ref{asmp:density-bounded}, \ref{asmp:cond-space-nontrivial}, \ref{asmp:cond-prob-functional},
\ref{asmp:regularity}, and \ref{asmp:main} hold for some joint perturbations $(G_0,K_0)$, $(G_1,K_1)$, $(H_0,L_0)$,
$(H_1,L_1)$, and $(G_{\mathrm{LC}},K_{\mathrm{LC}})$.
Assume in addition that each of these perturbation pairs is $Z_1$-modulation closed at $(\hat P,\hat Q)$ in the sense
of Definition~\ref{def:z1-modulation-closure}.
Assume further that the nuisance-error tolerances are not smaller than the parametric scale: there exists a constant
$c_{\min}>0$ such that, for all sufficiently large $N$,
\[
\eps_{N,\gamma}\ge c_{\min}N^{-1/2}
\qquad\text{and}\qquad
\eps_{N,\alpha}\ge c_{\min}N^{-1/2}.
\]
Assume further that 
\begin{equation}
    \label{eq:nonaffine-extra-asmp}
    \chi''(\hat P,\hat Q)[(H_0,L_0),(H_0,L_0)]\neq 0.
\end{equation}
Then for any $\xi\in(1/2,1)$, there exists a constant $\delta>0$ such that, for all sufficiently large $N$,
\[
\mathfrak{M}^{\chi}_{N,\xi}\left(\gM\left((\hat P,\hat Q);\eps_{N,\gamma},\eps_{N,\alpha}\right)\right)
~=~
\Omega\left(\eps_{N,\gamma}\,\eps_{N,\alpha}+\eps_{N,\gamma}^2+N^{-1/2}\right).
\]
\end{theorem}

Compared with Theorem~\ref{thm:main-mixed-bias}, the lower bound in Theorem~\ref{thm:main} contains the additional
term $\eps_{N,\gamma}^2$. This term is the price of curvature: it is driven by the second-order response of the
functional along $\gamma$-directions that are compatible with the nuisance-error constraints.
When $\rho$ is affine in $\gamma$, this curvature vanishes and one necessarily has $\chi''(\hat{P})[(H_0,L_0),(H_0,L_0)]=0$
(Proposition~\ref{prop:compute-2nd-order-differential}), which is why the affine-score regime is covered separately by
Theorem~\ref{thm:main-mixed-bias}.
Theorem~\ref{thm:main} matches the generic upper bound in Theorem~\ref{thm:upper-bound-general} and shows that, in the
absence of the mixed-bias structure, achieving pure double robustness is fundamentally impossible in a
structure-agnostic sense.

\subsection{Proof overview}
\label{subsec:proof-sketch}

\paragraph{Overview.}
The proof has two main components:
(i) a fuzzy-hypothesis lower bound that reduces minimax risk to constructing a family of local alternatives with
small pairwise divergences, and
(ii) a second-order Taylor analysis showing that the parameter separation between null and alternatives scales as
$\eps_{N,\gamma}\eps_{N,\alpha}$ (and, in the non-affine case, $\eps_{N,\gamma}^2$) while staying inside the uncertainty set.

\medskip

\paragraph{Method of fuzzy hypotheses.}
We use the method of fuzzy hypotheses \citep{robins2009semiparametric,kennedy2022minimax,balakrishnan2023fundamental}.
One constructs a null distribution (the anchor $\hat P$) and a carefully designed mixture of alternatives
$\{P_{\lambda}\}$ such that:
\begin{itemize}
\item (\emph{Indistinguishability}) the average $\chi^2$ or KL divergence between the null and the mixture is bounded,
so that no estimator can reliably identify which hypothesis generated the data; and
\item (\emph{Separation}) the parameter values $\chi(P_{\lambda})$ differ from $\chi(\hat P)$ by at least a target amount.
\end{itemize}

\paragraph{Two-step perturbations and second-order separation.}
The alternatives $P_{\lambda}$ are built as \emph{two-step} local perturbations of $\hat P$.
The first step moves along an invariant direction for either $\gamma$ or $\alpha$ (Assumption~\ref{asmp:main}(1)),
so that the corresponding nuisance remains exactly unchanged and the alternative stays inside the uncertainty set.
The second step is chosen to (i) satisfy the remaining nuisance-error constraint and (ii) create a nontrivial
second-order change in $\chi$ (Assumption~\ref{asmp:main}(2)).
The sizes of the two steps are calibrated as
\[
\text{(first step)}\asymp \max\{\eps_{N,\gamma},\eps_{N,\alpha}\},
\qquad
\text{(second step)}\asymp \min\{\eps_{N,\gamma},\eps_{N,\alpha}\},
\]
so that the resulting parameter shift is of order $\eps_{N,\gamma}\eps_{N,\alpha}$ when the score is affine.

\paragraph{Why the $\eps_{N,\gamma}^2$ term appears in the non-affine case?}
If $\rho$ is not affine, the conditional score has curvature captured by $\upsilon_{\rho}$ in
Assumption~\ref{asmp:regularity}. In this case, even when we use a $\gamma$-invariant first step, the Taylor expansion of
$\chi(P_{\lambda})$ can contain a pure $\eps_{N,\gamma}^2$ term (formalized via the condition $\chi''(\hat P)[H_0,H_0]\neq 0$).
This is precisely the mechanism behind Theorem~\ref{thm:main}.

\paragraph{The role of the mixed-bias property.}
When $\rho$ is affine, the curvature term disappears and one can symmetrically control the second-order expansion so that
the leading term is always $\eps_{N,\gamma}\eps_{N,\alpha}$.
This corresponds to the mixed-bias representation discussed in Section~\ref{subsec:mixed-bias} and yields
Theorem~\ref{thm:main-mixed-bias}.

\section{Instantiating the lower bounds}
\label{sec:examples}

In this section, we revisit several widely studied structural parameter estimation problems in the literature, and deduce structure-agnostic lower bounds as corollaries of Theorem \ref{thm:main-mixed-bias} and Theorem \ref{thm:main}.

\subsection{Average treatment effect}

We first show how Theorem \ref{thm:ate-only} can be derived as a special case of Theorem \ref{thm:main-mixed-bias}.
Recall that in the ATE case, we are given observational data $\{(X_i,D_i,Y_i)\}_{i=1}^n$, where $X_i\in\R^K$ is the covariate, $D_i$ is a binary treatment variable and $Y_i=Y_i(D_i)$ is the corresponding binary outcome. The covariate space $\gX = \mathrm{supp}(X)$ can be either continuous or discrete. Let $P_0$ be the ground-truth observation distribution, recall that under the conditional ignorability assumption, the ATE is identified as
\begin{equation}
    \notag
    \chi_{\ate}(P_0) = \E_{P_0} \big[ g_0(1,X) - g_0(0,X)\big].
\end{equation}

Now we show how Theorem \ref{thm:main} directly implies a lower bound for structure-agnostic estimation of ATE. Let the base measure $\mu$ be the uniform distribution on $\gX\times\gD\times\gY$, $\gZ = \gZ_1 = \gX\times\gD$, $\gW = \gY$. For any distribution $P\ll \mu$, its density can be written as $p(x,d,y) := p_X(x) \pi(x;P)^d \big(1-\pi(x;P)\big)^{1-d} g(d,x;P)^y \big(1-g(d,x;P)\big)^{1-y}, $
where $\pi(x;P)=\E_P[D\mid X=x]$ and $g(d,x;P)=\E_P[Y\mid X=x, D=d]$ are the nuisance functions under the distribution $P$. We then have the following theorem:

\begin{theorem}[Average treatment effect (mixed-bias) lower bound]
\label{thm:ate}
    Let $c\in(0,\frac{1}{2})$ be some constant. Suppose that $\hat{P}$ satisfies:
    \begin{enumerate}[(1).]
        \item $c \leq \pi(x;\hat{P}), g(d,x;\hat{P}) \leq 1-c, \forall x\in\gX$, and
        \item The marginal density of $\hat{P}$ on $\gX$, which we denote by $\hat{p}_{\gX}(\cdot)$, satisfies $l_{\hat{P}}\leq \hat{p}_{\gX} (x)\leq u_{\hat{P}}, \forall x\in\gX$ for some constants $l_{\hat{P}},u_{\hat{P}}>0$,
    \end{enumerate}
    and that $(\gX,\mu_{\gX})$ satisfies Assumption \ref{asmp:cond-space-nontrivial} with $\mu_{\gX}$ being the uniform distribution on $\gX$.
    Let $Z_1=X\in\gX, Z_2=D\in\gD=\{0,1\}, W=Y\in\gY=\{0,1\}$, $\gM$ be the set of all functions from $\{0,1\}^2$ to $[0,1]$, $\gamma(x,d;P) = g(d,x;P)$, $\alpha(x,d;P) = (2d-1)/{\big[\pi(x;P)^d (1-\pi(x;P))^{1-d}\big]}, m_1(o,h)= h(x,1)-h(x,0), \rho(o,\gamma)=y-\gamma(x,d)$ for all $ o=(x,d,y)\in\gO$. Then Assumptions \ref{asmp:cond-prob-functional} and \ref{asmp:regularity} hold, and there exists perturbations $G_i,H_i,i\in\{0,1\}$ that satisfy Assumption \ref{asmp:main} and that $\chi''(\hat{P})[G_0,G_1], \chi''(\hat{P})[H_0,H_1]\neq 0$. Since $\rho$ is affine in $\gamma$, we can deduce from Theorem \ref{thm:main-mixed-bias} that
    \begin{equation}
        \notag
        \mathfrak{M}_{n,\xi}^{\chi_{\ate}}\left(\gM(\hat{P}; \eps_{n,\gamma},\eps_{n,\alpha})\right) = \Omega\big( \eps_{n,\gamma}\eps_{n,\alpha}+{1}/{\sqrt{n}}\big),
    \end{equation}
    where by definition, we have $\gM(\hat{P}; \eps_{n,\gamma},\eps_{n,\alpha}) = \big\{ P\ll\mu : \|\gamma(Z;\hat{P})-\gamma(Z;P)\|_{P,2} \leq \eps_{n,\gamma}, \|\alpha(Z;\hat{P})-\alpha(Z;P)\|_{P,2} \leq \eps_{n,\alpha} \big\}.$
\end{theorem}
\noindent The proof is in Section~\ref{sec:proof:ate}.

Theorem \ref{thm:ate} states that the minimax structure-agnostic rate for such type of distributions is lower-bounded by $\eps_{n,\gamma}\eps_{n,\alpha}+1/{\sqrt{n}}$ up to a constant factor. Compared with Theorem \ref{thm:ate-only}, the only difference is that here, the nuisances are chosen as $\gamma$ and $\alpha$ rather than $\gamma$ and $\pi$. However, they are equivalent up to a constant factor depending only on $c$ since
\opt{opt-arxiv}{\begin{equation}
    \begin{aligned}
        &\quad \|\alpha(D,X;\hat{P})-\alpha(D,X;P)\|_{P,2}^2 \\
        &= \E\bigg[ \pi(X;P)\bigg(\frac{1}{\pi(X;P)}-\frac{1}{\pi(X;\hat{P})}\bigg)^2 + (1-\pi(X;P))\bigg(\frac{1}{1-\pi(X;P)}-\frac{1}{1-\pi(X;\hat{P})}\bigg)^2 \bigg] \\
        &= \E\bigg[ \bigg(\frac{1}{\pi(X;P)\pi(X;\hat{P})^2}+\frac{1}{(1-\pi(X;P))(1-\pi(X;\hat{P}))^2}\bigg) \big(\pi(X;P)-\pi(X;\hat{P})\big)^2 \bigg],
    \end{aligned}
\end{equation}}
\opt{opt-or}{$\|\alpha(D,X;\hat{P})-\alpha(D,X;P)\|_{P,2}^2 = \E\big[ \big(\frac{1}{\pi(X;P)\pi(X;\hat{P})^2}+\frac{1}{(1-\pi(X;P))(1-\pi(X;\hat{P}))^2}\big) \big(\pi(X;P)-\pi(X;\hat{P})\big)^2 \big]$}
so that under Assumption (2) in Theorem \ref{thm:ate}, we have $\sqrt{2}\|\pi(X;P)-\pi(X;\hat{P})\|_{P,2}\leq \|\alpha(D,X;\hat{P})-\alpha(D,X;P)\|_{P,2} \leq \sqrt{2/c^{3}}\|\pi(X;P)-\pi(X;\hat{P})\|_{P,2}$. Hence
we reproduce the lower bound for ATE directly by applying Theorem \ref{thm:main-mixed-bias}.

\subsection{Average treatment effect on the treated}

We next derive a structure-agnostic lower bound for the average treatment effect on the treated (ATT).
Let $P_0$ denote the observational distribution of $O=(X,D,Y)$, where $X\in\gX$ is a vector of covariates,
$D\in\{0,1\}$ is a binary treatment indicator, and $Y=Y(D)\in\{0,1\}$ is the observed outcome.
Under conditional ignorability and overlap, the ATT is identified by
\begin{equation}
\label{eq:att-id}
\theta_{\att}(P_0)
~=~
\E_{P_0}[Y\mid D=1]-\E_{P_0}\!\left[g_0(X)\mid D=1\right],
\qquad
g_0(x):=\E_{P_0}[Y\mid X=x,D=0].
\end{equation}

To connect \eqref{eq:att-id} to our general covariate shift functional \eqref{eq:cs-target-functional}, define the
training law $P$ as the joint law of $(X,D,Y)$ and define the target law $Q$ as the distribution of $X$ among
treated units, i.e.\ $Q(\cdot)=P(X\in\cdot\mid D=1)$.\footnote{Equivalently, one may allow $Q$ to be any covariate
distribution of interest and interpret $\E_{P_0}[g_0(X)\mid D=1]$ as $\E_{X\sim Q}[g_0(X)]$; the coupled choice
$Q=P(\cdot\mid D=1)$ is the canonical ATT instance.}  Set $Z=X$ and $W=(D,Y)$ so that $O=(Z,W)$.
Let $\gamma(\cdot;P)$ be the control regression $x\mapsto g_0(x)$, which can be characterized as the unique
solution to the conditional moment restriction
\begin{equation}
\label{eq:att-gamma-moment}
\E_P\!\left[(1-D)\{Y-\gamma(X)\}\mid X\right]=0.
\end{equation}
Define
\begin{equation}
\label{eq:att-cs-functional}
\chi_{\att}(P,Q):=\E_{X\sim Q}\!\left[\gamma(X;P)\right],
\end{equation}
so that $\theta_{\att}(P_0)=\E_{P_0}[Y\mid D=1]-\chi_{\att}(P_0,Q_0)$ with $Q_0=P_0(\cdot\mid D=1)$.
Since $\E_{P_0}[Y\mid D=1]$ is a regular (parametric-rate) functional of $P_0$, the structure-agnostic difficulty of
ATT is governed by $\chi_{\att}$.

\begin{theorem}[Average treatment effect on the treated (mixed-bias) lower bound]
\label{thm:att}
Let $c\in(0,\frac{1}{2})$ be a constant. Suppose $(\hat P,\hat Q)$ is an anchor pair such that:
\begin{enumerate}[(1).]
\item $c\le \pi(x;\hat P)\le 1-c$ and $c\le g(d,x;\hat P)\le 1-c$ for all $x\in\gX$ and $d\in\{0,1\}$, where
$\pi(x;P)=\E_P[D\mid X=x]$ and $g(d,x;P)=\E_P[Y\mid X=x,D=d]$;
\item the $X$-marginal densities of $\hat P$ and $\hat Q$ satisfy $0<l\le \hat p_{\gX}(x)\le u<\infty$ and
$0<l\le \hat q_{\gX}(x)\le u<\infty$ for all $x\in\gX$ for some constants $l,u$; and
\item $(\gX,\mu_{\gX})$ satisfies Assumption~\ref{asmp:cond-space-nontrivial}, where $\mu_{\gX}$ is the uniform
distribution on $\gX$.
\end{enumerate}
Let $Z=X$, $W=(D,Y)$, define $m_1(z,h)=h(z)$ and $\rho(o,\gamma)=(1-d)(y-\gamma(x))$ for $o=(x,d,y)$.
Let $\gamma(x;P)=\E_P[Y\mid X=x,D=0]$ and let
\[
\alpha(x;P,Q):=\frac{\dd Q}{\dd P_X}(x)\cdot\frac{1}{1-\pi(x;P)},
\]
where $P_X$ is the $X$-marginal of $P$. Then Assumptions \ref{asmp:density-bounded},
\ref{asmp:cond-prob-functional}, and \ref{asmp:regularity} hold for the functional $\chi_{\att}$ in
\eqref{eq:att-cs-functional}, and one can construct feasible perturbation pairs satisfying the invariance and
non-degeneracy requirements of Assumption~\ref{asmp:main}. Moreover, $\rho$ is affine in $\gamma$ and
$\chi_{\att}$ satisfies the mixed-bias property. Consequently, Theorem~\ref{thm:main-mixed-bias} implies that for
any $\xi\in(1/2,1)$,
\[
\mathfrak{M}^{\chi_{\att}}_{N,\xi}\!\left(\gM\!\left((\hat P,\hat Q);\eps_{N,\gamma},\eps_{N,\alpha}\right)\right)
~=~
\Omega\!\left(\eps_{N,\gamma}\eps_{N,\alpha}+N^{-1/2}\right).
\]
In particular, the same lower bound holds for estimating $\theta_{\att}$ in \eqref{eq:att-id}, up to addition of the
parametric term $N^{-1/2}$ coming from $\E_P[Y\mid D=1]$.
\end{theorem}
\noindent The proof is in Section~\ref{sec:proof:att}.

\subsection{Weighted average derivative}
\label{subsec:wad}

When the treatment variable is continuous, the weighted average derivative (\wad) is a commonly considered parameter of interest in econometrics with applications in index models and demand analysis \citep{hardle1991empirical,powell1989semiparametric,newey1993efficiency,imbens2009identification}; see also \citet[Example 2]{chernozhukov2021automatic}  and \citet[Example 4]{rotnitzky2021characterization} for formal definitions. WAD can naturally be interpreted as the continuous version of the ATE. Given observational data $\{(X_i,D_i,Y_i)\}_{i=1}^n$ where $X_i\in\R^K$ is the covariate, $D_i$ is a real-valued treatment variable and $Y_i=Y_i(D_i)$ is the corresponding binary outcome. Define $g(x,d;P) = \E_P[Y\mid X=x, D=d]$. Suppose that $g$ is $\gC^1$ in $d$, then we are interested in estimating
\begin{equation}
    \chi_{\wad}(P) = \E_P\bigg[ \int \omega(u)\frac{\partial g(X,u;P)}{\partial u} \dd u \bigg],
\end{equation}
where $\omega$ is a known probability density function (PDF). Assuming that $\omega$ is continuously differentiable and has support on $(0,1)$, integration by parts implies that
\opt{opt-arxiv}{
\begin{equation}
    \chi_{\wad}(P) = \E_P\bigg[ \int_0^1 s(u)g(x,u;P)\omega(u)\dd u \bigg] = \E\big[ s(U)g(X,U;P) \big],
\end{equation}
}
\opt{opt-or}{
\begin{equation}
    \begin{aligned}
        \chi_{\wad}(P) &= \E_P\bigg[ \int_0^1 s(u)g(x,u;P)\omega(u)\dd u \bigg] \\
        &= \E\big[ s(U)g(X,U;P) \big],
    \end{aligned}
\end{equation}
}
where $s(u)=-\omega(u)^{-1}\omega'(u)$ and $U$ is a random variable independent of $O=(X,D,Y)$. The following theorem provides a lower bound for structure-agnostic estimation of \wad.

\begin{theorem}[Weighted average derivative (mixed-bias) lower bound]
\label{thm:wad}
    Suppose that the density $\hat{p}=\dd\hat{P}/\dd\mu$ satisfies: 
    \begin{enumerate}[(1).]
        \item For all $x\in\gX, y\in\gY$, $\hat{p}(x,d,y)$ is continuously differentiable in $d$ on $[0,1]$ with derivative uniformly bounded by $C_d$, and
        \item There exists constants $l_{\hat{p}}, u_{\hat{p}}>0$ such that $l_{\hat{p}} \leq \hat{p}(x,d,y) \leq u_{\hat{p}}$ holds for all $(x,d,y)\in\gX\times\gD\times\gY$ except from a $\mu$-null subset,
    \end{enumerate}
    and that $(\gX,\mu_{\gX})$ satisfies Assumption \ref{asmp:cond-space-nontrivial}.
    Let $Z_1=X\in\gX, Z_2=D\in\gD=[0,1], W=Y\in\gY=\{0,1\}$, $\gM$ be the set of all functions $h:\gD\times\gY\mapsto\R_{\geq 0}$ such that $h(d,y)$ is continuously differentiable in $d$ satisfying $|{\partial h}/{\partial d}|\leq 2C_d$. For all $\gM$-feasible distribution, we define $\gamma(z;P)=g(x,d;P)$ for $z=(x,d)$,
    \[
        m_1(o,h)=\int_0^1 s(u)h(x,u)\omega(u)\dd u,\quad \rho(o,\gamma)=y-\gamma(x,d),
    \]
    and
    \[
        \alpha(z;P)=-\frac{\omega'(d)}{p(d\mid x)}=\frac{s(d)\omega(d)}{p(d\mid x)}.
    \]
    Then Assumptions \ref{asmp:cond-prob-functional} and \ref{asmp:regularity} hold and there exists perturbations $G_i,H_i,i\in\{0,1\}$ that satisfy Assumption \ref{asmp:main} and that $\chi''_{\wad}(\hat{P})[G_0,G_1], \chi''_{\wad}[H_0,H_1]\neq 0$. Hence we can deduce from Theorem \ref{thm:main-mixed-bias} that
    \begin{equation}
        \notag
        \mathfrak{M}_{n,\xi}^{\chi_{\wad}}\left(\gM(\hat{P}; \eps_{n,\gamma},\eps_{n,\alpha})\right) = \Omega\big( \eps_{n,\gamma}\eps_{n,\alpha}+{1}/{\sqrt{n}}\big),
    \end{equation}
    where by definition, we have $\gM(\hat{P}; \eps_{n,\gamma},\eps_{n,\alpha})= \Big\{ P\ll\mu : \|\gamma(Z;\hat{P})-\gamma(Z;P)\|_{P,2} \leq \eps_{n,\gamma}, \|\alpha(Z;\hat{P})-\alpha(Z;P)\|_{P,2} \leq \eps_{n,\alpha}$, and $\sup_{o=(x,d,y)\in\gO}\big|{\partial p(x,d,y)}/{\partial d}\Big| \leq 2C_d \big\}.$
\end{theorem}
\noindent The proof is in Section~\ref{sec:proof:wad}.

\subsection{Average policy effect}
Assume that $D\in[0,1]$. We consider the average policy effect as in \citet{stock1989nonparametric}:
\[
  \chi_{\ape}(P)=\E\big[g\{X,\tau(D);P\}-g(X,D;P)\big],
\]
where $\tau:[0,1]\to[0,1]$ is a known counterfactual transformation. Throughout this example we assume that $\tau$ is a $C^1$-bijection of $[0,1]$ onto itself and that there exist constants $0<\underline\tau\le \overline\tau<\infty$ such that
\[
  \underline\tau\le |\tau'(d)|\le \overline\tau,\qquad \forall d\in[0,1],
\]
so that $\tau^{-1}$ is well-defined and Lipschitz. This estimand fits our framework by taking $Z_1=X$, $Z_2=D$, $W=Y$ and
\[
  m_1(o,h)=h\{x,\tau(d)\}-h(x,d),\qquad \rho(o,\gamma)=y-\gamma(x,d).
\]
The associated Riesz representer depends on the conditional density of $\tau(D)$ given $X$, which can be obtained by a change of variables.

\begin{theorem}[Average policy effect (mixed-bias) lower bound]\label{thm:ape}
Let $Z_1=X$, $Z_2=D$, $W=Y\in\gY=\{0,1\}$, and let $\gM$ be the set of all functions from $\gD\times \gY$ to $[0,1]$. For any $\gM$-feasible distribution $P$ with density $p$ (w.r.t.~$\mu_{\gX}\otimes\mathrm{Leb}\otimes\mu_{\{0,1\}}$), define
\[
  \begin{aligned}
  \gamma(z;P) &= g(x,d;P),\\
  \alpha(z;P) &= \frac{p_{\tau}(d\mid x)}{p(d\mid x)}-1,\\
  m_1(o,h) &= h\{x,\tau(d)\}-h(x,d),\\
  \rho(o,\gamma) &= y-\gamma(x,d),
  \end{aligned}
\]
where $p(d\mid x)=p(x,d,\cdot)/p(x,\cdot,\cdot)$ is the conditional density of $D$ given $X=x$ and $p_{\tau}(\cdot\mid x)$ is the conditional density of $\tau(D)$ given $X=x$ (equivalently,
$p_{\tau}(d\mid x)=p\{x,\tau^{-1}(d),\cdot\}/\{|\tau'|\{\tau^{-1}(d)\}p(x,\cdot,\cdot)\}$).
Assume that Assumptions \ref{asmp:cond-prob-functional} and \ref{asmp:regularity} hold, and there exist perturbations satisfying Assumption \ref{asmp:main}. Then Theorem \ref{thm:main-mixed-bias} implies that
\[
  \mathcal{R}(\hat P,\eps_{n,\gamma},\eps_{n,\alpha})\gtrsim \eps_{n,\gamma}\eps_{n,\alpha}+\frac{1}{\sqrt{n}},
\]
where by definition, we have $\gM(\hat{P}; \eps_{n,\gamma},\eps_{n,\alpha}) = \{ P\ll\mu : \|\gamma(Z;\hat{P})-\gamma(Z;P)\|_{P,2} \leq \eps_{n,\gamma}, \|\alpha(Z;\hat{P})-\alpha(Z;P)\|_{P,2} \leq \eps_{n,\alpha}\}.$
\end{theorem}
\noindent The proof is in Section~\ref{sec:proof:ape}.

\subsection{Expected conditional covariance}
\label{subsec:ecc}

We consider the DGP given in \eqref{model}, and the goal is to estimate the expected conditional covariance (\ecc), which is defined as

\begin{equation}
    \label{def:ecc}
    \chi_{\ecc}(P) = \E_P \big[\mathrm{Cov}(D,Y\mid X)\big].
\end{equation}

In \citet{robins2009semiparametric}, the authors derive minimax rates for estimating ECC under Holder-smoothness assumptions on the nuisance functions $m,g$ and the CATE function. \citet{balakrishnan2023fundamental} considers a structure-agnostic setting as our paper and shows that the minimax rate scales as $\eps_{n,\gamma}\eps_{n,\alpha}+1/{\sqrt{n}}$. It is worth noticing that this rate applies to the fully nonparametric regression model \eqref{model}.
One may wonder, however, if this rate is still optimal in a partial linear outcome model, \emph{i.e.,} if one additionally assumes that the treatment effect is \emph{constant} in $X$, namely
\begin{equation}\label{eq:plm-model}
Y=\theta_0 T+f_0(X)+\varepsilon,
\qquad 
\mathbb E[\varepsilon\mid X,T]=0,
\qquad 
X\in\gX=[0,1]^K,\ T\in\{0,1\},\ Y\in\{0,1\}.
\end{equation}
Let
\[
g(x;P):=\mathbb P_P(T=1\mid X=x),
\qquad 
q(x;P):=\mathbb E_P[Y\mid X=x].
\]
Then the ECC functional can be written as the residual-covariance numerator
\begin{equation}\label{eq:ecc-plm-num}
\chi_{\ecc}(P)=\mathbb E_P\Big[\mathrm{Cov}(T,Y\mid X)\Big]
=\mathbb E_P\Big[(T-g(X;P))(Y-q(X;P))\Big].
\end{equation}
The next theorem shows that \emph{even under the PLM restriction} --- which is a strict submodel of \eqref{model} --- one cannot improve on the doubly robust rate in a structure-agnostic neighborhood.
In this sense, it is strictly stronger than the ECC lower bound of \citet{balakrishnan2023fundamental}, which do not impose the partial linear assumption.

Fix an anchor PLM distribution $\hat P$ such that $X\sim\mathrm{Unif}([0,1]^K)$ and $T,Y\in\{0,1\}$.
For radii $\eps_{n,g},\eps_{n,q}>0$, define the PLM-restricted anchored neighborhood
\[
\gM_{\mathrm{PLM}}(\hat P;\eps_{n,g},\eps_{n,q})
:=
\Big\{P\in\mathcal P_{\mathrm{PLM}}:\ 
\|g(X;P)-g(X;\hat P)\|_{P,2}\le \eps_{n,g},\ 
\|q(X;P)-q(X;\hat P)\|_{P,2}\le \eps_{n,q}\Big\},
\]
where $\mathcal P_{\mathrm{PLM}}$ denotes the set of all distributions satisfying \eqref{eq:plm-model} with $X\sim\mathrm{Unif}([0,1]^K)$.

\begin{theorem}[PLM expected conditional covariance (mixed-bias) lower bound]\label{thm:plm-numerator}
Let $\xi\in(0,1/4)$.
Assume that $\hat P\in\mathcal P_{\mathrm{PLM}}$ satisfies:
\begin{enumerate}[(1).]
    \item (\emph{Overlap}) there exists $c\in(0,1/2)$ such that $c\le g(x;\hat P),q(x;\hat P)\le 1-c$ for all $x\in\gX$;
    \item (\emph{Bounded density}) $\hat P$ satisfies Assumption~\ref{asmp:density-bounded}.
\end{enumerate}
Then
\begin{equation}\label{eq:plm-num-lb}
\mathfrak{M}_{n,\xi}^{\chi_{\ecc}}\Big(\gM_{\mathrm{PLM}}(\hat P;\eps_{n,g},\eps_{n,q})\Big)
=
\Omega\Big(\eps_{n,g}\eps_{n,q}+\frac{1}{\sqrt n}\Big),
\end{equation}
uniformly over $n\in\mathbb N$ and $\eps_{n,g},\eps_{n,q}>0$.
\end{theorem}
\noindent The proof is in Section~\ref{sec:proof:plm-numerator}.
It constructs a PLM-preserving perturbation family directly at the density level (starting from $\hat p(x,t,y)$),
verifies the ``perturbation invariance'' and nondegenerate mixed second derivative conditions of
Assumption~\ref{asmp:main}, and then applies Theorem~\ref{thm:main-mixed-bias}.

\begin{remark}[Connection to \citet{jin2025its} and the role of invariant perturbations]\label{rem:plm-jin2025its-connection}
The lower bound proved above is obtained by constructing a local alternative family
$\{P_\lambda\}$, subject to the
\emph{invariant perturbation constraint} stated in Assumption~\ref{asmp:main}. Recently, \citet[Theorem~B.1]{jin2025its} established the optimal rate for estimating the
PLM coefficient $\theta$ with binary treatment.while that result cannot be derived as a special case of Theorem~\ref{thm:main-mixed-bias}, it is worth noticing that the \emph{same} invariant
perturbation construction idea can be used \emph{directly} to prove that lower bound. Hence, we believe that Assumption~\ref{asmp:main}
could be a common backbone in establishing structure-agnostic minimax rates that may extend beyond the functionals covered by our main theorem.
\end{remark}

\subsection{Distribution shift}
Let $\{O_i\}_{i=1}^n$ be i.i.d. observations, where $O_i=(X_i,Y_i)$, $X_i\in\gX$ and $Y_i\in\gY=\{0,1\}$. (For the lower bound it suffices to consider the binary-outcome case.) Let $F_1,F_2$ be two known distributions on $\gX$ with densities $f_1,f_2$ w.r.t.~$\mu_{\gX}$. Consider the distribution-shift estimand
\[
  \chi_{\dis}(P)=\int_{\gX}\gamma(x;P)\,\dd(F_2-F_1)(x),\qquad 
  \gamma(x;P)=\E_P[Y\mid X=x].
\]
This estimand fits our framework by taking $Z_1=X$, $Z_2=\varnothing$, $W=Y$, with
\[
  m_1(o,h)=\int_{\gX}h(x)\{f_2(x)-f_1(x)\}\,\dd\mu_{\gX}(x),\qquad 
  \rho(o,\gamma)=y-\gamma(x).
\]

\begin{theorem}[Distribution shift (mixed-bias) lower bound]\label{thm:ds}
Let $Z_1=X$, $Z_2=\varnothing$, $W=Y\in\gY=\{0,1\}$, and let $\gM$ be the set of all functions from $\gY$ to $\R_+$. For any $\gM$-feasible distribution $P$ with density $p$ (w.r.t.~$\mu_{\gX}\otimes\mu_{\{0,1\}}$), define
\[
  \gamma(x;P)=\frac{p(x,1)}{p(x,\cdot)},\qquad 
  \alpha(x;P)=\frac{f_2(x)-f_1(x)}{f(x)}
\]
and
\[
    m_1(o,h)=\int_{\gX}h(x)\{f_2(x)-f_1(x)\}\,\dd\mu_{\gX}(x),\qquad 
  \rho(o,\gamma)=y-\gamma(x),
\]
where $p(x,\cdot)=p(x,0)+p(x,1)$ and $f(x)=p(x,\cdot)$ is the marginal density of $X$ under $P$. Assume that $|f_1(x)|,|f_2(x)|\le C_F$ for all $x$ and let $\hat P$ satisfy Assumption \ref{asmp:density-bounded}. If Assumptions \ref{asmp:cond-prob-functional} and \ref{asmp:regularity} hold and there exist perturbations satisfying Assumption \ref{asmp:main}, then Theorem \ref{thm:main-mixed-bias} implies that
\[
  \mathcal{R}(\hat P,\eps_{n,\gamma},\eps_{n,\alpha})\gtrsim \eps_{n,\gamma}\eps_{n,\alpha}+\frac{1}{\sqrt{n}},
\]
where by definition, we have $\gM(\hat{P}; \eps_{n,\gamma},\eps_{n,\alpha}) = \Big\{ P\ll\mu : \|\gamma(Z;\hat{P})-\gamma(Z;P)\|_{P,2} \leq \eps_{n,\gamma}, \|\alpha(Z;\hat{P})-\alpha(Z;P)\|_{P,2} \leq \eps_{n,\alpha} \Big\}.$
\end{theorem}
\noindent The proof is in Section~\ref{sec:proof:ds}.

\subsection{Log-odds-difference}
\label{subsec:lod}

Let $O=(X,D,Y)\in \gO:=\gX\times\{0,1\}\times\{0,1\}$ where $X\in\gX:=[0,1]^K$, $D\in\{0,1\}$ is a binary treatment indicator, and $Y\in\{0,1\}$ is a binary response. We set $Z_1:=X$, $Z_2:=D$, $W:=Y$, and $Z:=(Z_1,Z_2)=(X,D)$. Let
\[
g(d,x;P):=\E_P[Y\mid D=d,X=x],\qquad \pi(x;P):=\E_P[D\mid X=x],
\]
and define the log-odds regression function
\[
\gamma(d,x;P):=\log\left(\frac{g(d,x;P)}{1-g(d,x;P)}\right),\qquad (d,x)\in\{0,1\}\times\gX.
\]
The log-odds-difference estimand is
\[
\chi_{\mathrm{LOD}}(P):=\E_P\big[\gamma(1,X;P)-\gamma(0,X;P)\big].
\]
Let $\Lambda(t):=(1+\exp(-t))^{-1}$ denote the logistic link and consider the generalized regression score
\[
\rho(o,\gamma):=\frac{y-\Lambda(\gamma)}{\Lambda(\gamma)\{1-\Lambda(\gamma)\}},\qquad o=(x,d,y).
\]
Then $\E_P[\rho\{O,\gamma(Z;P)\}\mid Z]=0$ and $\Lambda\{\gamma(d,x;P)\}=g(d,x;P)$. Moreover, one can verify that $\nu_\rho(z;P)\equiv -1$ and $\upsilon_\rho(z;P)=1-2g(z;P)$, where $g(z;P):=\E_P[Y\mid Z=z]$. Finally, the Riesz representer of $h\mapsto\E_P\{h(1,X)-h(0,X)\}$ is
\[
\nu_m(z;P)=\frac{d}{\pi(x;P)}-\frac{1-d}{1-\pi(x;P)},
\]
and hence $\alpha(z;P)=-\nu_m(z;P)/\nu_\rho(z;P)=\nu_m(z;P)$.

\begin{theorem}[Log-odds difference (non-affine) lower bound]
\label{thm:lod}
Fix $K\geq 1$ and let $\mu=\mu_X\otimes\mu_D\otimes\mu_Y$, where $\mu_X$ is Lebesgue measure on $\gX=[0,1]^K$ and $\mu_D,\mu_Y$ are counting measures on $\{0,1\}$. Let $\hat P\ll\mu$ with density $\hat p$ satisfying $0<p_{\mathrm{lb}}\leq \hat p\leq p_{\mathrm{ub}}<\infty$.
Assume overlap: there exists $\eta\in(0,1/2)$ such that for $\hat P$-a.e. $x\in\gX$,
\[
\eta \leq \pi(x;\hat P)\leq 1-\eta,\qquad 
\eta \leq g(d,x;\hat P)\leq 1-\eta,\quad d\in\{0,1\}.
\]
Assume also that $\hat P_X\{x:g(1,x;\hat P)\neq 1/2\}>0$ and that $(X,\mu_X)$ is $(K+1)$-nondegenerate.
Then Assumptions \ref{asmp:density-bounded}, \ref{asmp:cond-prob-functional}, \ref{asmp:cond-space-nontrivial}, \ref{asmp:regularity}, and \ref{asmp:main} hold for the log-odds-difference estimand $\chi_{\mathrm{LOD}}$ defined above and, in addition, there exists an $\gM$-feasible perturbation direction $H_0$ such that $\chi_{\mathrm{LOD}}''(\hat P)[H_0,H_0]\neq 0$.
Consequently,
\[
\inf_{\hat\chi}\sup_{P\in \gM(\hat P;\eps_{n,\gamma},\eps_{n,\alpha})}\E_P\big[|\hat\chi-\chi_{\mathrm{LOD}}(P)|\big]
=\Omega\left(\eps_{n,\gamma}^2+\eps_{n,\gamma}\eps_{n,\alpha}+\frac{1}{\sqrt{n}}\right),
\]
where the infimum ranges over all estimators $\hat\chi$ based on $n$ i.i.d. observations from $P$.
\end{theorem}
\noindent The proof is in Section~\ref{sec:proof:lod}.

\subsection{Expected derivative of conditional quantiles}
\label{subsec:eqd}

Suppose that we have i.i.d. data $\{(X_i,Y_i)\}_{i=1}^n$ from some distribution $P$ where $X_i\in\gX\subseteq\R^K$ and $Y_i\in\gY\subseteq\R$. Suppose that $X_i=(X_{i,-1},X_{i,1})\in\gX_{-1}\times\gX_{1}$ where $X_{i,1}$ is a scalar variable of interest and $X_{i,-1}$ are the remaining control variables. For some fixed $q\in(0,1)$, let $\gamma(x;P)=\gQ_{q}(P(Y\mid X=x))$ be the $q$-th quantile of the conditional distribution of $Y$ given $X=x$, we are interested in estimating a weighted average of the partial derivative of $\gamma(x;P)$ in the direction of $x_1$
\begin{equation}
    \notag
    \chi_{\eqd} = \E_P\bigg[ w(X)\frac{\partial \gamma(X;P)}{\partial x_1} \bigg]
\end{equation}
where $w(x)$ is a known weight function.
First-order debiasing estimators of $\chi_{\eqd}$ have been constructed in previous works \citep{chernozhukov2022automatic,sasaki2022unconditional}. To present our lower bound for estimating $\chi_{\eqd}$, we need a few more regularity assumptions:

\begin{assumption}
    \label{asmp:eqd-weight}
    $|w(x)|\leq W$ and $|{\partial w(x)}/{\partial x_1}|\leq C_{W,1}$ for all $x\in\gX$.
\end{assumption}

\begin{assumption}
    \label{asmp:eqd-density}
    The density function $\hat{p}(x,y)$ of $\hat{P}$ with respect to $\mu$ is differentiable in $y$ and twice differentiable in $x_1$. There exists constants $l_{\hat{P}},u_{\hat{P}}>0$ such that $l_{\hat{P}}\leq \hat{p}(x,y) \leq u_{\hat{P}}$ holds for all $(x,y)\in\gX\times\gY$ except from a $\mu$-null set. Moreover, $C_{X,1}=\sup_{o=(x,y)\in\gO}|{\partial \hat{p}}/{\partial x_1}|<\infty$ and $C_{Y,k}=\sup_{o=(x,y)\in\gO}|{\partial^k \hat{p}}/{\partial y^k}|<\infty, k\in\{1,2\}$.
\end{assumption}

\begin{theorem}[Expected derivative of conditional quantiles (non-affine) lower bound]
    \label{thm:eqd}
    Suppose that $K\geq 2$, Assumptions \ref{asmp:eqd-weight} and \ref{asmp:eqd-density} are satisfied, and $(\gX_{-1}, \mu_{\gX_{-1}})$ satisfies Assumption \ref{asmp:cond-space-nontrivial} with $\mu_{\gX_{-1}}$ be the uniform distribution on $\gX_{-1}$.
    Let $Z_1=X_{-1}\in\gX_{-1}, Z_2=X_1\in\gX_1=[0,1], W=Y\in\gY=[0,1]$, $\gM$ be the set of all $\mu$-integrable functions $h:\gX\times\gY\mapsto\R_+$ such that $|{\partial h}/{\partial x_1}|\leq 2C_{X,1}$ and $|{\partial^k h}/{\partial y^k}|\leq 2C_{Y,k}, k\in\{1,2\}$.
    For all $\gM$-feasible distributions, we define $\gamma(x;P)$ as above, $\alpha(x;P)=p(x,\gamma(x;P))^{-1}{\partial(w(x)p(x,\cdot))}/{\partial x_1}$, where $p$ is the density of $P$ and $p(x,\cdot)$ is the marginal density of $X$ under $P$,
    \[
    m_1(o,h)= w(x)\frac{\partial h(x)}{\partial x_1},\qquad
    \rho(o,\gamma)=\mathbbm{1}\{y<\gamma(x)\}-q,
    \]
    \[
    \nu_m(x;P)=-p(x,\cdot)^{-1}\frac{\partial (w(x)p(x,\cdot))}{\partial x_1},\quad
    \nu_{\rho}(x;P)=\frac{p(x,\gamma(x;P))}{p(x,\cdot)},\quad
    \upsilon_{\rho}(x;P)=\frac{p_y'(x,\gamma(x;P))}{p(x,\cdot)},
    \]
    for all $o=(x,y)$. Suppose that $\alpha(x;\hat{P})$ is not zero $\mu_X$-a.s. and $\mu_X\big(\big\{x:\alpha(x;\hat{P})\upsilon_{\rho}(x;\hat{P})\neq 0\big\}\big)>0$, then Assumptions \ref{asmp:cond-prob-functional} and \ref{asmp:regularity} hold and there exists perturbations $G_i,H_i,i\in\{0,1\}$ that satisfy Assumption \ref{asmp:main} and $\chi_{\eqd}''(\hat{P})[G_0,G_1], \chi_{\eqd}''(\hat{P})[H_0,H_0]\neq 0$. Hence we can deduce from Theorem \ref{thm:main} that
    \[
    \mathfrak{M}_{n,\xi}^{\chi_{\eqd}}\left(\gM(\hat{P}; \eps_{n,\gamma},\eps_{n,\alpha})\right)
    = \Omega\Big( \eps_{n,\gamma}^2+\eps_{n,\gamma}\eps_{n,\alpha}+\frac{1}{\sqrt{n}}\Big),
    \]
    where
    \[
    \gM(\hat{P}; \eps_{n,\gamma},\eps_{n,\alpha})
    = \Big\{ P\ll\mu : \begin{aligned}[t]
    &\|\gamma(Z;\hat{P})-\gamma(Z;P)\|_{P,2} \leq \eps_{n,\gamma},\\
    &\|\alpha(Z;\hat{P})-\alpha(Z;P)\|_{P,2}\leq \eps_{n,\alpha},\\
    &\big|{\partial p(x,y)}/{\partial x_1}\big| \leq 2C_{X,1},\\
    &\big|{\partial^k p(x,y)}/{\partial y^k}\big| \leq 2C_{Y,k},\qquad k=1,2
    \end{aligned}\Big\}.
    \]
\end{theorem}
\noindent The proof is in Section~\ref{sec:proof:eqd}.

\section{Conclusion}

This paper develops sharp \emph{structure-agnostic} minimax lower bounds for a broad class of semiparametric functionals built from (generalized) regression nuisances. Assuming only $L^2$ error rates for black-box nuisance estimates, our main theorems identify two regimes: an \emph{affine-score / mixed-bias} regime in which the optimal error is of order $\eps_{n,\gamma}\eps_{n,\alpha}+n^{-1/2}$ (matching the doubly robust rate), and a more general \emph{non-affine} regime in which an additional term $\eps_{n,\gamma}^2$ is unavoidable.  These lower bounds match the generic first-order debiasing upper bounds and therefore imply that these methods are unimprovable without additional modeling structure.

Technically, the lower bounds are proved via the method of fuzzy hypotheses, reducing estimation to testing between carefully constructed mixtures of local alternatives.  The key new ingredient is a \emph{two-step sequential perturbation} scheme that decouples feasibility (staying inside the anchored nuisance neighborhood) from separation (creating the desired second-order change in the target functional), together with a ham-sandwich-style partitioning argument that enforces exact invariances needed to “hide” perturbations from the nuisance constraints.  We verify the required conditions for a collection of canonical examples, illustrating how the abstract theory specializes to concrete causal, policy, and quantile-based targets.

Several directions are suggested by these results.  First, it would be valuable to broaden the class of functionals for which a unified optimality theory can be established.  Second, our analysis is minimax by design; a natural next step is to disentangle the roles of approximation and stochastic errors as suggested in \citet{pmlr-v291-gu25b}.  Third, extending the framework beyond i.i.d.\ sampling (e.g.\ dependent data, clustering, distribution shift with weak overlap, or heavy-tailed outcomes) and connecting these lower bounds more directly to finite-sample inference remain important open problems.

\bibliographystyle{plainnat}%
\bibliography{example}

\begin{thebibliography}{83}
\providecommand{\natexlab}[1]{#1}
\providecommand{\url}[1]{\texttt{#1}}
\expandafter\ifx\csname urlstyle\endcsname\relax
  \providecommand{\doi}[1]{doi: #1}\else
  \providecommand{\doi}{doi: \begingroup \urlstyle{rm}\Url}\fi

\bibitem[Abadie and Imbens(2006)]{abadie2006large}
Alberto Abadie and Guido~W Imbens.
\newblock Large sample properties of matching estimators for average treatment effects.
\newblock \emph{econometrica}, 74\penalty0 (1):\penalty0 235--267, 2006.

\bibitem[Arias-Castro et~al.(2018)Arias-Castro, Pelletier, and Saligrama]{arias2018remember}
Ery Arias-Castro, Bruno Pelletier, and Venkatesh Saligrama.
\newblock Remember the curse of dimensionality: The case of goodness-of-fit testing in arbitrary dimension.
\newblock \emph{Journal of Nonparametric Statistics}, 30\penalty0 (2):\penalty0 448--471, 2018.

\bibitem[Arlot and Celisse(2010)]{Arlot2010}
Sylvain Arlot and Alain Celisse.
\newblock {A survey of cross-validation procedures for model selection}.
\newblock \emph{Statistics Surveys}, 4\penalty0 (none):\penalty0 40 -- 79, 2010.
\newblock \doi{10.1214/09-SS054}.
\newblock URL \url{https://doi.org/10.1214/09-SS054}.

\bibitem[Athey and Wager(2021)]{athey2021policy}
Susan Athey and Stefan Wager.
\newblock Policy learning with observational data.
\newblock \emph{Econometrica}, 89\penalty0 (1):\penalty0 133--161, 2021.

\bibitem[Athey et~al.(2019)Athey, Tibshirani, and Wager]{athey2019generalized}
Susan Athey, Julie Tibshirani, and Stefan Wager.
\newblock Generalized random forests.
\newblock \emph{The Annals of Statistics}, 47\penalty0 (2):\penalty0 1148, 2019.

\bibitem[Bach et~al.(2024)Bach, Schacht, Chernozhukov, Klaassen, and Spindler]{bach2024hyperparameter}
Philipp Bach, Oliver Schacht, Victor Chernozhukov, Sven Klaassen, and Martin Spindler.
\newblock Hyperparameter tuning for causal inference with double machine learning: A simulation study.
\newblock \emph{arXiv preprint arXiv:2402.04674}, 2024.

\bibitem[Balakrishnan and Wasserman(2019)]{balakrishnan2019hypothesis}
S~Balakrishnan and L~Wasserman.
\newblock Hypothesis testing for densities and high-dimensional multinomials: Sharp local minimax rates.
\newblock \emph{Annals of Statistics}, 47\penalty0 (4):\penalty0 1893--1927, 2019.

\bibitem[Balakrishnan et~al.(2023)Balakrishnan, Kennedy, and Wasserman]{balakrishnan2023fundamental}
Sivaraman Balakrishnan, Edward~H Kennedy, and Larry Wasserman.
\newblock The fundamental limits of structure-agnostic functional estimation.
\newblock \emph{arXiv preprint arXiv:2305.04116}, 2023.

\bibitem[Belloni and Chernozhukov(2011)]{belloni2011l1}
Alexandre Belloni and Victor Chernozhukov.
\newblock l1-penalized quantile regression in high-dimensional sparse models.
\newblock \emph{The Annals of Statistics}, 39\penalty0 (1):\penalty0 82, 2011.

\bibitem[Belloni and Chernozhukov(2013)]{belloni2013least}
Alexandre Belloni and Victor Chernozhukov.
\newblock Least squares after model selection in high-dimensional sparse models.
\newblock \emph{Bernoulli}, 19\penalty0 (2):\penalty0 521--547, 2013.

\bibitem[Belloni et~al.(2014)Belloni, Chernozhukov, and Wang]{belloni2014pivotal}
Alexandre Belloni, Victor Chernozhukov, and Lie Wang.
\newblock Pivotal estimation via square-root lasso in nonparametric regression.
\newblock \emph{The Annals of Statistics}, 42\penalty0 (2):\penalty0 757, 2014.

\bibitem[Biau et~al.(2008)Biau, Devroye, and Lugosi]{biau2008consistency}
G{\'e}rard Biau, Luc Devroye, and G{\"a}bor Lugosi.
\newblock Consistency of random forests and other averaging classifiers.
\newblock \emph{Journal of Machine Learning Research}, 9\penalty0 (9), 2008.

\bibitem[Bickel and Ritov(1988)]{bickel1988estimating}
Peter~J Bickel and Yaacov Ritov.
\newblock Estimating integrated squared density derivatives: sharp best order of convergence estimates.
\newblock \emph{Sankhy{\=a}: The Indian Journal of Statistics, Series A}, pages 381--393, 1988.

\bibitem[Bickel et~al.(2009)Bickel, Ritov, and Tsybakov]{bickel2009simultaneous}
Peter~J Bickel, Ya'acov Ritov, and Alexandre~B Tsybakov.
\newblock Simultaneous analysis of lasso and dantzig selector.
\newblock \emph{The Annals of Statistics}, pages 1705--1732, 2009.

\bibitem[Birg{\'e} and Massart(1995)]{birge1995estimation}
Lucien Birg{\'e} and Pascal Massart.
\newblock Estimation of integral functionals of a density.
\newblock \emph{The Annals of Statistics}, 23\penalty0 (1):\penalty0 11--29, 1995.

\bibitem[Bradic et~al.(2019)Bradic, Chernozhukov, Newey, and Zhu]{bradic2019minimax}
Jelena Bradic, Victor Chernozhukov, Whitney~K Newey, and Yinchu Zhu.
\newblock Minimax semiparametric learning with approximate sparsity.
\newblock \emph{arXiv preprint arXiv:1912.12213}, 2019.

\bibitem[Breiman(2001)]{breiman2001random}
Leo Breiman.
\newblock Random forests.
\newblock \emph{Machine learning}, 45:\penalty0 5--32, 2001.

\bibitem[B{\"u}hlmann and Yu(2003)]{buhlmann2003boosting}
Peter B{\"u}hlmann and Bin Yu.
\newblock Boosting with the l 2 loss: regression and classification.
\newblock \emph{Journal of the American Statistical Association}, 98\penalty0 (462):\penalty0 324--339, 2003.

\bibitem[Chen and White(1999)]{chen1999improved}
Xiaohong Chen and Halbert White.
\newblock Improved rates and asymptotic normality for nonparametric neural network estimators.
\newblock \emph{IEEE Transactions on Information Theory}, 45\penalty0 (2):\penalty0 682--691, 1999.

\bibitem[Chernozhukov et~al.(2017)Chernozhukov, Chetverikov, Demirer, Duflo, Hansen, and Newey]{chernozhukov2017double}
Victor Chernozhukov, Denis Chetverikov, Mert Demirer, Esther Duflo, Christian Hansen, and Whitney Newey.
\newblock Double/debiased/neyman machine learning of treatment effects.
\newblock \emph{American Economic Review}, 107\penalty0 (5):\penalty0 261--265, 2017.

\bibitem[Chernozhukov et~al.(2018)Chernozhukov, Chetverikov, Demirer, Duflo, Hansen, Newey, and Robins]{chernozhukov2018double}
Victor Chernozhukov, Denis Chetverikov, Mert Demirer, Esther Duflo, Christian Hansen, Whitney Newey, and James Robins.
\newblock Double/debiased machine learning for treatment and structural parameters: Double/debiased machine learning.
\newblock \emph{The Econometrics Journal}, 21\penalty0 (1), 2018.

\bibitem[Chernozhukov et~al.(2021)Chernozhukov, Newey, Quintas-Martinez, and Syrgkanis]{chernozhukov2021automatic}
Victor Chernozhukov, Whitney~K Newey, Victor Quintas-Martinez, and Vasilis Syrgkanis.
\newblock Automatic debiased machine learning via riesz regression.
\newblock \emph{arXiv preprint arXiv:2104.14737}, 2021.

\bibitem[Chernozhukov et~al.(2022)Chernozhukov, Newey, and Singh]{chernozhukov2022automatic}
Victor Chernozhukov, Whitney~K Newey, and Rahul Singh.
\newblock Automatic debiased machine learning of causal and structural effects.
\newblock \emph{Econometrica}, 90\penalty0 (3):\penalty0 967--1027, 2022.

\bibitem[Chernozhukov et~al.(2023)Chernozhukov, Newey, Newey, Singh, and Syrgkanis]{chernozhukov2023automatic}
Victor Chernozhukov, Michael Newey, Whitney~K Newey, Rahul Singh, and Vasilis Syrgkanis.
\newblock Automatic debiased machine learning for covariate shifts.
\newblock \emph{arXiv preprint arXiv:2307.04527}, 2023.

\bibitem[Chetverikov et~al.(2021)Chetverikov, Liao, and Chernozhukov]{chetverikov2021cross}
Denis Chetverikov, Zhipeng Liao, and Victor Chernozhukov.
\newblock On cross-validated lasso in high dimensions.
\newblock \emph{The Annals of Statistics}, 49\penalty0 (3):\penalty0 1300--1317, 2021.

\bibitem[Dudley(2014)]{dudley2014uniform}
Richard~M Dudley.
\newblock \emph{Uniform central limit theorems}, volume 142.
\newblock Cambridge university press, 2014.

\bibitem[D{\v{z}}eroski and {\v{Z}}enko(2004)]{dvzeroski2004combining}
Saso D{\v{z}}eroski and Bernard {\v{Z}}enko.
\newblock Is combining classifiers with stacking better than selecting the best one?
\newblock \emph{Machine learning}, 54:\penalty0 255--273, 2004.

\bibitem[Farrell et~al.(2021)Farrell, Liang, and Misra]{farrell2021deep}
Max~H Farrell, Tengyuan Liang, and Sanjog Misra.
\newblock Deep neural networks for estimation and inference.
\newblock \emph{Econometrica}, 89\penalty0 (1):\penalty0 181--213, 2021.

\bibitem[Foster and Syrgkanis(2023)]{foster2023orthogonal}
Dylan~J Foster and Vasilis Syrgkanis.
\newblock Orthogonal statistical learning.
\newblock \emph{The Annals of Statistics}, 51\penalty0 (3):\penalty0 879--908, 2023.

\bibitem[Freund and Schapire(1997)]{freund1997decision}
Yoav Freund and Robert~E Schapire.
\newblock A decision-theoretic generalization of on-line learning and an application to boosting.
\newblock \emph{Journal of computer and system sciences}, 55\penalty0 (1):\penalty0 119--139, 1997.

\bibitem[Friedman(2001)]{friedman2001greedy}
Jerome~H Friedman.
\newblock Greedy function approximation: a gradient boosting machine.
\newblock \emph{Annals of statistics}, pages 1189--1232, 2001.

\bibitem[Gu(2025)]{pmlr-v291-gu25b}
Yihong Gu.
\newblock Open problem: Structure-agnostic minimax risk for partial linear model.
\newblock In Nika Haghtalab and Ankur Moitra, editors, \emph{Proceedings of Thirty Eighth Conference on Learning Theory}, volume 291 of \emph{Proceedings of Machine Learning Research}, pages 6220--6224. PMLR, 30 Jun--04 Jul 2025.

\bibitem[H{\"a}rdle et~al.(1991)H{\"a}rdle, Hildenbrand, and Jerison]{hardle1991empirical}
Wolfgang H{\"a}rdle, Werner Hildenbrand, and Michael Jerison.
\newblock Empirical evidence on the law of demand.
\newblock \emph{Econometrica: Journal of the Econometric Society}, pages 1525--1549, 1991.

\bibitem[Hastie et~al.(2009)Hastie, Tibshirani, Friedman, Hastie, Tibshirani, and Friedman]{hastie2009random}
Trevor Hastie, Robert Tibshirani, Jerome Friedman, Trevor Hastie, Robert Tibshirani, and Jerome Friedman.
\newblock Random forests.
\newblock \emph{The elements of statistical learning: Data mining, inference, and prediction}, pages 587--604, 2009.

\bibitem[Heckman et~al.(1998)Heckman, Ichimura, and Todd]{heckman1998matching}
James~J Heckman, Hidehiko Ichimura, and Petra Todd.
\newblock Matching as an econometric evaluation estimator.
\newblock \emph{The review of economic studies}, 65\penalty0 (2):\penalty0 261--294, 1998.

\bibitem[Hirano et~al.(2003)Hirano, Imbens, and Ridder]{hirano2003efficient}
Keisuke Hirano, Guido~W Imbens, and Geert Ridder.
\newblock Efficient estimation of average treatment effects using the estimated propensity score.
\newblock \emph{Econometrica}, 71\penalty0 (4):\penalty0 1161--1189, 2003.

\bibitem[Imbens(2004)]{imbens2004nonparametric}
Guido~W Imbens.
\newblock Nonparametric estimation of average treatment effects under exogeneity: A review.
\newblock \emph{Review of Economics and statistics}, 86\penalty0 (1):\penalty0 4--29, 2004.

\bibitem[Imbens and Newey(2009)]{imbens2009identification}
Guido~W Imbens and Whitney~K Newey.
\newblock Identification and estimation of triangular simultaneous equations models without additivity.
\newblock \emph{Econometrica}, 77\penalty0 (5):\penalty0 1481--1512, 2009.

\bibitem[Imbens et~al.(2003)Imbens, Newey, and Ridder]{imbens2003mean}
GW~Imbens, W~Newey, and G~Ridder.
\newblock Mean-squared-error calculations for average treatment effects. department of economics, uc berkeley, 2003.

\bibitem[Ingster(1994)]{ingster1994minimax}
Yu~I Ingster.
\newblock Minimax detection of a signal in $\ell_p$ metrics.
\newblock \emph{Journal of Mathematical Sciences}, 68:\penalty0 503--515, 1994.

\bibitem[Jin and Syrgkanis(2024)]{jin2024structure}
Jikai Jin and Vasilis Syrgkanis.
\newblock Structure-agnostic optimality of doubly robust learning for treatment effect estimation.
\newblock \emph{arXiv preprint arXiv:2402.14264}, 2024.

\bibitem[Jin et~al.(2025)Jin, Mackey, and Syrgkanis]{jin2025its}
Jikai Jin, Lester Mackey, and Vasilis Syrgkanis.
\newblock It{\textquoteright}s hard to be normal: The impact of noise on structure-agnostic estimation.
\newblock In \emph{The Thirty-ninth Annual Conference on Neural Information Processing Systems}, 2025.

\bibitem[Karmaker et~al.(2021)Karmaker, Hassan, Smith, Xu, Zhai, and Veeramachaneni]{karmaker2021automl}
Shubhra~Kanti Karmaker, Md~Mahadi Hassan, Micah~J Smith, Lei Xu, Chengxiang Zhai, and Kalyan Veeramachaneni.
\newblock Automl to date and beyond: Challenges and opportunities.
\newblock \emph{ACM Computing Surveys (CSUR)}, 54\penalty0 (8):\penalty0 1--36, 2021.

\bibitem[Kennedy et~al.(2022)Kennedy, Balakrishnan, Robins, and Wasserman]{kennedy2022minimax}
Edward~H Kennedy, Sivaraman Balakrishnan, James~M Robins, and Larry Wasserman.
\newblock Minimax rates for heterogeneous causal effect estimation.
\newblock \emph{arXiv preprint arXiv:2203.00837}, 2022.

\bibitem[Krantz and Parks(2002)]{krantz2002implicit}
Steven~G. Krantz and Harold~R. Parks.
\newblock \emph{The Implicit Function Theorem: History, Theory, and Applications}.
\newblock Birkh{\"a}user, Boston, MA, 2002.

\bibitem[LeDell and Poirier(2020)]{ledell2020h2o}
Erin LeDell and Sebastien Poirier.
\newblock H2o automl: Scalable automatic machine learning.
\newblock In \emph{Proceedings of the AutoML Workshop at ICML}, volume 2020. ICML, 2020.

\bibitem[Little and Rubin(2000)]{little2000causal}
Roderick~J Little and Donald~B Rubin.
\newblock Causal effects in clinical and epidemiological studies via potential outcomes: concepts and analytical approaches.
\newblock \emph{Annual review of public health}, 21\penalty0 (1):\penalty0 121--145, 2000.

\bibitem[Liu et~al.(2017)Liu, Mukherjee, Newey, and Robins]{liu2017semiparametric}
Lin Liu, Rajarshi Mukherjee, Whitney~K Newey, and James~M Robins.
\newblock Semiparametric efficient empirical higher order influence function estimators.
\newblock \emph{arXiv preprint arXiv:1705.07577}, 2017.

\bibitem[Mayer(2011)]{mayer2011does}
Alexander~K Mayer.
\newblock Does education increase political participation?
\newblock \emph{The Journal of Politics}, 73\penalty0 (3):\penalty0 633--645, 2011.

\bibitem[Newey(1994)]{newey1994asymptotic}
Whitney~K Newey.
\newblock The asymptotic variance of semiparametric estimators.
\newblock \emph{Econometrica: Journal of the Econometric Society}, pages 1349--1382, 1994.

\bibitem[Newey and Stoker(1993)]{newey1993efficiency}
Whitney~K Newey and Thomas~M Stoker.
\newblock Efficiency of weighted average derivative estimators and index models.
\newblock \emph{Econometrica: Journal of the Econometric Society}, pages 1199--1223, 1993.

\bibitem[Oreopoulos(2006)]{oreopoulos2006estimating}
Philip Oreopoulos.
\newblock Estimating average and local average treatment effects of education when compulsory schooling laws really matter.
\newblock \emph{American Economic Review}, 96\penalty0 (1):\penalty0 152--175, 2006.

\bibitem[Polley et~al.(2019)Polley, LeDell, Kennedy, Lendle, and van~der Laan]{polley2019package}
Eric Polley, Erin LeDell, Chris Kennedy, Sam Lendle, and Mark van~der Laan.
\newblock Package ‘superlearner’.
\newblock \emph{CRAN}, 2019.

\bibitem[Powell et~al.(1989)Powell, Stock, and Stoker]{powell1989semiparametric}
James~L Powell, James~H Stock, and Thomas~M Stoker.
\newblock Semiparametric estimation of index coefficients.
\newblock \emph{Econometrica: Journal of the Econometric Society}, pages 1403--1430, 1989.

\bibitem[Reddi et~al.(2015)Reddi, Poczos, and Smola]{reddi2015doubly}
Sashank Reddi, Barnabas Poczos, and Alex Smola.
\newblock Doubly robust covariate shift correction.
\newblock In \emph{Proceedings of the AAAI conference on artificial intelligence}, volume~29, 2015.

\bibitem[Robins et~al.(2008)Robins, Li, Tchetgen, van~der Vaart, et~al.]{robins2008higher}
James Robins, Lingling Li, Eric Tchetgen, Aad van~der Vaart, et~al.
\newblock Higher order influence functions and minimax estimation of nonlinear functionals.
\newblock In \emph{Probability and statistics: essays in honor of David A. Freedman}, volume~2, pages 335--422. Institute of Mathematical Statistics, 2008.

\bibitem[Robins et~al.(2009)Robins, Tchetgen, Li, and van~der Vaart]{robins2009semiparametric}
James Robins, Eric~Tchetgen Tchetgen, Lingling Li, and Aad van~der Vaart.
\newblock Semiparametric minimax rates.
\newblock \emph{Electronic journal of statistics}, 3:\penalty0 1305, 2009.

\bibitem[Robins and Rotnitzky(1995)]{robins1995semiparametric}
James~M Robins and Andrea Rotnitzky.
\newblock Semiparametric efficiency in multivariate regression models with missing data.
\newblock \emph{Journal of the American Statistical Association}, 90\penalty0 (429):\penalty0 122--129, 1995.

\bibitem[Robins et~al.(1994)Robins, Rotnitzky, and Zhao]{robins1994estimation}
James~M Robins, Andrea Rotnitzky, and Lue~Ping Zhao.
\newblock Estimation of regression coefficients when some regressors are not always observed.
\newblock \emph{Journal of the American statistical Association}, 89\penalty0 (427):\penalty0 846--866, 1994.

\bibitem[Robins et~al.(1995)Robins, Rotnitzky, and Zhao]{robins1995analysis}
James~M Robins, Andrea Rotnitzky, and Lue~Ping Zhao.
\newblock Analysis of semiparametric regression models for repeated outcomes in the presence of missing data.
\newblock \emph{Journal of the american statistical association}, 90\penalty0 (429):\penalty0 106--121, 1995.

\bibitem[Robins et~al.(2017)Robins, Li, and Mukherjee]{robins2017minimax}
James~M Robins, Lingling Li, and Rajarshi Mukherjee.
\newblock Minimax estimation of a functional on a structured high-dimensional model.
\newblock \emph{The Annals of Statistics}, 45\penalty0 (5):\penalty0 1951--1987, 2017.

\bibitem[Rosenbaum(1989)]{rosenbaum1989optimal}
Paul~R Rosenbaum.
\newblock Optimal matching for observational studies.
\newblock \emph{Journal of the American Statistical Association}, 84\penalty0 (408):\penalty0 1024--1032, 1989.

\bibitem[Rosenbaum and Rubin(1983)]{rosenbaum1983central}
Paul~R Rosenbaum and Donald~B Rubin.
\newblock The central role of the propensity score in observational studies for causal effects.
\newblock \emph{Biometrika}, 70\penalty0 (1):\penalty0 41--55, 1983.

\bibitem[Rotnitzky et~al.(2021)Rotnitzky, Smucler, and Robins]{rotnitzky2021characterization}
Andrea Rotnitzky, Ezequiel Smucler, and James~M Robins.
\newblock Characterization of parameters with a mixed bias property.
\newblock \emph{Biometrika}, 108\penalty0 (1):\penalty0 231--238, 2021.

\bibitem[Sasaki et~al.(2022)Sasaki, Ura, and Zhang]{sasaki2022unconditional}
Yuya Sasaki, Takuya Ura, and Yichong Zhang.
\newblock Unconditional quantile regression with high-dimensional data.
\newblock \emph{Quantitative Economics}, 13\penalty0 (3):\penalty0 955--978, 2022.

\bibitem[Schmidt-Hieber(2020)]{schmidt2020nonparametric}
Anselm~Johannes Schmidt-Hieber.
\newblock Nonparametric regression using deep neural networks with relu activation function.
\newblock \emph{Annals of statistics}, 48\penalty0 (4):\penalty0 1875--1897, 2020.

\bibitem[Sill et~al.(2009)Sill, Tak{\'a}cs, Mackey, and Lin]{sill2009feature}
Joseph Sill, G{\'a}bor Tak{\'a}cs, Lester Mackey, and David Lin.
\newblock Feature-weighted linear stacking.
\newblock \emph{arXiv preprint arXiv:0911.0460}, 2009.

\bibitem[Stock(1989)]{stock1989nonparametric}
James~H Stock.
\newblock Nonparametric policy analysis.
\newblock \emph{Journal of the American Statistical Association}, 84\penalty0 (406):\penalty0 567--575, 1989.

\bibitem[Sugiyama et~al.(2007)Sugiyama, Krauledat, and M{\"u}ller]{sugiyama2007covariate}
Masashi Sugiyama, Matthias Krauledat, and Klaus-Robert M{\"u}ller.
\newblock Covariate shift adaptation by importance weighted cross validation.
\newblock \emph{Journal of Machine Learning Research}, 8\penalty0 (5), 2007.

\bibitem[Syrgkanis and Zampetakis(2020)]{syrgkanis2020estimation}
Vasilis Syrgkanis and Manolis Zampetakis.
\newblock Estimation and inference with trees and forests in high dimensions.
\newblock In \emph{Conference on learning theory}, pages 3453--3454. PMLR, 2020.

\bibitem[Tsybakov(2008)]{tsybakov2008introduction}
Alexandre~B Tsybakov.
\newblock \emph{Introduction to nonparametric estimation}.
\newblock Springer Science \& Business Media, 2008.

\bibitem[van~de Geer et~al.(2014)van~de Geer, B{\"u}hlmann, Ritov, and Dezeure]{van2014asymptotically}
Sara van~de Geer, Peter B{\"u}hlmann, Ya’acov Ritov, and Ruben Dezeure.
\newblock On asymptotically optimal confidence regions and tests for high-dimensional models.
\newblock \emph{The Annals of Statistics}, 42\penalty0 (3), 2014.

\bibitem[Van~der Laan et~al.(2007)Van~der Laan, Polley, and Hubbard]{van2007super}
Mark~J Van~der Laan, Eric~C Polley, and Alan~E Hubbard.
\newblock Super learner.
\newblock \emph{Statistical applications in genetics and molecular biology}, 6\penalty0 (1), 2007.

\bibitem[van~der Vaart(2014)]{van2014higher}
Aad van~der Vaart.
\newblock Higher order tangent spaces and influence functions.
\newblock \emph{Statistical science}, 29\penalty0 (4):\penalty0 679--686, 2014.

\bibitem[Wager and Athey(2018)]{wager2018estimation}
Stefan Wager and Susan Athey.
\newblock Estimation and inference of heterogeneous treatment effects using random forests.
\newblock \emph{Journal of the American Statistical Association}, 113\penalty0 (523):\penalty0 1228--1242, 2018.

\bibitem[Wager and Walther(2015)]{wager2015adaptive}
Stefan Wager and Guenther Walther.
\newblock Adaptive concentration of regression trees, with application to random forests.
\newblock \emph{arXiv preprint arXiv:1503.06388}, 2015.

\bibitem[Wang et~al.(2021)Wang, Wu, Weimer, and Zhu]{wang2021flaml}
Chi Wang, Qingyun Wu, Markus Weimer, and Erkang Zhu.
\newblock Flaml: A fast and lightweight automl library.
\newblock \emph{Proceedings of Machine Learning and Systems}, 3:\penalty0 434--447, 2021.

\bibitem[Wegkamp(2003)]{wegkamp2003model}
Marten Wegkamp.
\newblock Model selection in nonparametric regression.
\newblock \emph{The Annals of Statistics}, 31\penalty0 (1):\penalty0 252--273, 2003.

\bibitem[Wolpert(1992)]{wolpert1992stacked}
David~H Wolpert.
\newblock Stacked generalization.
\newblock \emph{Neural networks}, 5\penalty0 (2):\penalty0 241--259, 1992.

\bibitem[Wood et~al.(2008)Wood, Egger, Gluud, Schulz, J{\"u}ni, Altman, Gluud, Martin, Wood, and Sterne]{wood2008empirical}
Lesley Wood, Matthias Egger, Lise~Lotte Gluud, Kenneth~F Schulz, Peter J{\"u}ni, Douglas~G Altman, Christian Gluud, Richard~M Martin, Anthony~JG Wood, and Jonathan~AC Sterne.
\newblock Empirical evidence of bias in treatment effect estimates in controlled trials with different interventions and outcomes: meta-epidemiological study.
\newblock \emph{bmj}, 336\penalty0 (7644):\penalty0 601--605, 2008.

\bibitem[Zhang(1993)]{zhang1993model}
Ping Zhang.
\newblock Model selection via multifold cross validation.
\newblock \emph{The annals of statistics}, pages 299--313, 1993.

\bibitem[Zhang and Yu(2005)]{zhang2005boosting}
Tong Zhang and Bin Yu.
\newblock Boosting with early stopping: Convergence and consistency.
\newblock \emph{Annals of statistics}, 33\penalty0 (4):\penalty0 1538--1579, 2005.

\bibitem[Zou and Hastie(2005)]{zou2005regularization}
Hui Zou and Trevor Hastie.
\newblock Regularization and variable selection via the elastic net.
\newblock \emph{Journal of the Royal Statistical Society Series B: Statistical Methodology}, 67\penalty0 (2):\penalty0 301--320, 2005.

\end{thebibliography}

\newpage
\appendix

\addcontentsline{toc}{section}{Appendix} %
\part{Appendix} %
\parttoc %

\section{The method of fuzzy hypotheses}
\label{subsec:prelim}
Our proof uses the method of fuzzy hypotheses. For two probability measures $P,Q$ that are absolutely continuous with respect to a common measure $\mu$, with densities $p=\dd P/\dd\mu$ and $q=\dd Q/\dd\mu$, we define the squared Hellinger distance
\begin{equation}
    \label{eq:hellinger-distance}
    H^2(P,Q):=\int \left(\sqrt{p}-\sqrt{q}\right)^2\dd\mu
    =2-2\int \sqrt{p q}\dd\mu
    \in [0,2].
\end{equation}
We also define the associated \emph{Hellinger affinity}
\begin{equation}
    \label{eq:hellinger-affinity}
    \rho(P,Q):=\int \sqrt{p q}\,\dd\mu = 1 - \frac{1}{2}H^2(P,Q)\in[0,1].
\end{equation}
\paragraph{Notation.}
In this appendix, $\rho(P,Q)$ denotes the Hellinger affinity \eqref{eq:hellinger-affinity} and should not be confused
with the score function $\rho(o,\gamma)$ used elsewhere in the paper. Similarly, the symbol $\pi$ denotes a mixing
measure on $\Lambda$ (and $\pi_j$ on $\Lambda_j$), unrelated to propensity-score notation.

\paragraph{A multi-sample Hellinger bound for hypercube mixtures.}
The first result we use is a \emph{multi-sample} extension of the Hellinger bound for hypercube mixtures
appearing, for instance, as a special case of \citet[Theorem 2.1]{robins2009semiparametric}.
Unlike the classical i.i.d. setting, we allow for \emph{multiple independent samples} drawn from
potentially different distributions (e.g., a training and a target sample under covariate shift).
We give a self-contained proof.

\begin{theorem}[Multi-sample Hellinger bound for hypercube mixtures]
\label{semi-param-thm}
\label{thm:semi-param-thm}
Fix an integer $S\ge 1$ (the number of samples) and sample sizes $n_1,\ldots,n_S\in\mathbb{N}$.
For each $s\in\{1,\ldots,S\}$, let $(\mathcal{X}^{(s)},\mathcal{A}^{(s)})$ be a measurable space with
a $\sigma$-finite dominating measure $\mu_s$, and let $P^{(s)}$ be a probability measure on $\mathcal{X}^{(s)}$
with density $p^{(s)}:=\dd P^{(s)}/\dd\mu_s$.
Let $m\in\mathbb{N}$ and let $\Lambda=\Lambda_1\times\cdots\times \Lambda_m$ be a product parameter space,
equipped with a product probability measure $\pi=\pi_1\otimes\cdots\otimes \pi_m$.

For each $\lambda=(\lambda_1,\ldots,\lambda_m)\in\Lambda$ and each $s\in\{1,\ldots,S\}$, let $Q^{(s)}_\lambda$
be a probability measure on $\mathcal{X}^{(s)}$ with density $q^{(s)}_\lambda:=\dd Q^{(s)}_\lambda/\dd\mu_s$.
Assume the following.

\begin{enumerate}[label=(A.\arabic*)]
    \item \textbf{Hypercube/partition structure.}
    For each $s\in\{1,\ldots,S\}$, there exists a measurable partition
    $\{\mathcal{X}^{(s)}_1,\ldots,\mathcal{X}^{(s)}_m\}$ of $\mathcal{X}^{(s)}$ such that:
    \begin{enumerate}[label=(\roman*)]
        \item \emph{Cell probabilities are fixed:} for every $j\in\{1,\ldots,m\}$,
        \[
            p_{s,j}:=P^{(s)}(\mathcal{X}^{(s)}_j)=Q^{(s)}_\lambda(\mathcal{X}^{(s)}_j)\in(0,1)
            \qquad\text{for all }\lambda\in\Lambda;
        \]
        \item \emph{Only the $j$-th coordinate matters on the $j$-th cell:} if $x\in \mathcal{X}^{(s)}_j$,
        then $q^{(s)}_\lambda(x)$ depends on $\lambda$ only through $\lambda_j$.
    \end{enumerate}
    \item \textbf{Mixture equals baseline (centering).} For each $s\in\{1,\ldots,S\}$,
    \begin{equation}
        \label{eq:centering-single-sample}
        p^{(s)}(x) = \int q^{(s)}_\lambda(x)\,\pi(\dd\lambda)\qquad\text{for $\mu_s$-a.e. }x\in\mathcal{X}^{(s)}.
    \end{equation}
\end{enumerate}

Define
\begin{equation}
\label{eq:def-b}
b \;:=\; \max_{1\le s\le S}\max_{1\le j\le m}\; p_{s,j}^{-1}\,\sup_{\lambda\in\Lambda}\int_{\mathcal{X}^{(s)}_j}\frac{\bigl(q^{(s)}_\lambda-p^{(s)}\bigr)^2}{p^{(s)}}\,\dd\mu_s,
\qquad
p_{\max}:=\max_{1\le s\le S}\max_{1\le j\le m}p_{s,j},
\qquad
n_{\mathrm{tot}}:=\sum_{s=1}^S n_s.
\end{equation}
Assume that for some constant $A>0$,
\begin{equation}
\label{eq:hellinger-assumption-A}
n_{\mathrm{tot}}\cdot p_{\max}\cdot \max\{1,b\}\le A.
\end{equation}

Let $\mathbb{P}:=\bigotimes_{s=1}^S (P^{(s)})^{\otimes n_s}$ be the joint law of $S$ independent samples
$(X^{(s)}_1,\ldots,X^{(s)}_{n_s})$ with $X^{(s)}_i\overset{\mathrm{i.i.d.}}{\sim}P^{(s)}$,
and let $\mathbb{Q}_\lambda:=\bigotimes_{s=1}^S (Q^{(s)}_\lambda)^{\otimes n_s}$.
Define the mixture $\overline{\mathbb{Q}}:=\int \mathbb{Q}_\lambda\,\pi(\dd\lambda)$.

Then the squared Hellinger distance between $\mathbb{P}$ and $\overline{\mathbb{Q}}$ satisfies
\begin{equation}
\label{eq:hellinger-bound-multisample}
H^2\bigl(\mathbb{P},\overline{\mathbb{Q}}\bigr)\;\le\; C(A)\, n_{\mathrm{tot}}^2\,p_{\max}\, b^2,
\end{equation}
where one may take $C(A)=\exp(A)/2$.
\end{theorem}

\begin{proof}
Throughout, all integrals are with respect to the appropriate dominating measures, and we freely use Tonelli/Fubini
whenever integrands are nonnegative.

\paragraph{Reduce to a bound on the Hellinger affinity.}
Let $\mu^{(\boldsymbol{n})}:=\bigotimes_{s=1}^S \mu_s^{\otimes n_s}$ denote a dominating measure for both
$\mathbb{P}$ and $\overline{\mathbb{Q}}$ on $\prod_{s=1}^S (\mathcal{X}^{(s)})^{n_s}$.
Let $p^{(\boldsymbol{n})}:=\dd\mathbb{P}/\dd\mu^{(\boldsymbol{n})}$ and
$\bar q^{(\boldsymbol{n})}:=\dd\overline{\mathbb{Q}}/\dd\mu^{(\boldsymbol{n})}$ denote the corresponding densities.
By \eqref{eq:hellinger-distance}--\eqref{eq:hellinger-affinity},
\begin{equation}
\label{eq:H2-vs-affinity}
H^2(\mathbb{P},\overline{\mathbb{Q}})=2-2\rho(\mathbb{P},\overline{\mathbb{Q}}),
\qquad
\rho(\mathbb{P},\overline{\mathbb{Q}})=\int \sqrt{p^{(\boldsymbol{n})}\,\bar q^{(\boldsymbol{n})}}\,\dd\mu^{(\boldsymbol{n})}.
\end{equation}
Since $H^2(\mathbb{P},\overline{\mathbb{Q}})\le 2\{1-\rho(\mathbb{P},\overline{\mathbb{Q}})\}$, it suffices to lower bound
$\rho(\mathbb{P},\overline{\mathbb{Q}})$.

\paragraph{Factorize the mixture likelihood ratio over cells and condition on cell indices.}
For each sample $s$ and observation $i\in\{1,\ldots,n_s\}$, define the (random) cell index
\[
I^{(s)}_i \;:=\; j \quad\Longleftrightarrow\quad X^{(s)}_i\in \mathcal{X}^{(s)}_j,
\qquad j\in\{1,\ldots,m\},
\]
and the corresponding cell counts
\[
N_{s,j}:=\sum_{i=1}^{n_s}\mathbf{1}\{I^{(s)}_i=j\},
\qquad
M_j:=\sum_{s=1}^S N_{s,j}.
\]

For each $s$ and $j$, define the conditional densities (supported on $\mathcal{X}^{(s)}_j$)
\[
p^{(s)}_j(x):=\frac{p^{(s)}(x)\mathbf{1}\{x\in \mathcal{X}^{(s)}_j\}}{p_{s,j}},
\qquad
q^{(s)}_{j,\lambda_j}(x):=\frac{q^{(s)}_\lambda(x)\mathbf{1}\{x\in \mathcal{X}^{(s)}_j\}}{p_{s,j}},
\]
where $q^{(s)}_{j,\lambda_j}$ is well-defined because, by Assumption (A.1)(ii), $q^{(s)}_\lambda(x)$ depends on $\lambda$
only through $\lambda_j$ when $x\in\mathcal{X}^{(s)}_j$.
Then, for $x\in\mathcal{X}^{(s)}_j$,
\begin{equation}
\label{eq:cell-factorization-single}
p^{(s)}(x)=p_{s,j}\,p^{(s)}_j(x),
\qquad
q^{(s)}_\lambda(x)=p_{s,j}\,q^{(s)}_{j,\lambda_j}(x).
\end{equation}

Define the mixture likelihood ratio
\[
L(\mathbf{x}) \;:=\; \frac{\bar q^{(\boldsymbol{n})}(\mathbf{x})}{p^{(\boldsymbol{n})}(\mathbf{x})},
\qquad \mathbf{x}\in\prod_{s=1}^S (\mathcal{X}^{(s)})^{n_s}.
\]
By definition of $\overline{\mathbb{Q}}$ and Tonelli,
\[
\bar q^{(\boldsymbol{n})}(\mathbf{x})
=\int \prod_{s=1}^S\prod_{i=1}^{n_s} q^{(s)}_\lambda(x^{(s)}_i)\,\pi(\dd\lambda).
\]
Using \eqref{eq:cell-factorization-single} and the product structure of $\pi=\bigotimes_{j=1}^m\pi_j$, we obtain
\begin{align}
\label{eq:L-factorization}
L(\mathbf{x})
&=\frac{\int \prod_{s=1}^S\prod_{i=1}^{n_s} q^{(s)}_\lambda(x^{(s)}_i)\,\pi(\dd\lambda)}
{\prod_{s=1}^S\prod_{i=1}^{n_s} p^{(s)}(x^{(s)}_i)}\\
\nonumber
&=\int \prod_{s=1}^S\prod_{i=1}^{n_s} \frac{q^{(s)}_\lambda(x^{(s)}_i)}{p^{(s)}(x^{(s)}_i)}\,\pi(\dd\lambda)\\
\nonumber
&=\int \prod_{j=1}^m\;\prod_{s=1}^S\;\prod_{i:\,I^{(s)}_i=j}\frac{q^{(s)}_{j,\lambda_j}(x^{(s)}_i)}{p^{(s)}_j(x^{(s)}_i)}\,
\bigotimes_{j=1}^m\pi_j(\dd\lambda_j)\\
\nonumber
&=\prod_{j=1}^m \left\{ \int \prod_{s=1}^S\;\prod_{i:\,I^{(s)}_i=j}\frac{q^{(s)}_{j,\lambda_j}(x^{(s)}_i)}{p^{(s)}_j(x^{(s)}_i)}\,\pi_j(\dd\lambda_j)\right\}.
\end{align}
In the last step we used Tonelli and the fact that the integrand is a product over $j$ of nonnegative functions of
$\lambda_j$.

Taking square roots yields
\begin{equation}
\label{eq:sqrtL-factorization}
\sqrt{L(\mathbf{x})}
=\prod_{j=1}^m \left\{ \int \prod_{s=1}^S\;\prod_{i:\,I^{(s)}_i=j}\frac{q^{(s)}_{j,\lambda_j}(x^{(s)}_i)}{p^{(s)}_j(x^{(s)}_i)}\,\pi_j(\dd\lambda_j)\right\}^{1/2}.
\end{equation}

Now use \eqref{eq:H2-vs-affinity} together with $\rho(\mathbb{P},\overline{\mathbb{Q}})=\mathbb{E}_{\mathbb{P}}[\sqrt{L(\mathbf{X})}]$,
where $\mathbf{X}$ denotes the full collection of observations.
Conditioning on the cell indices $\mathbf{I}:=(I^{(s)}_i)_{s,i}$ and using \eqref{eq:sqrtL-factorization}, we get
\begin{equation}
\label{eq:rho-conditional}
\rho(\mathbb{P},\overline{\mathbb{Q}})
=\mathbb{E}_{\mathbb{P}}\left[\,\mathbb{E}_{\mathbb{P}}\left[\prod_{j=1}^m U_j(\mathbf{X})\ \Big|\ \mathbf{I}\right]\right],
\end{equation}
where
\[
U_j(\mathbf{X})
:=\left\{ \int \prod_{s=1}^S\;\prod_{i:\,I^{(s)}_i=j}\frac{q^{(s)}_{j,\lambda_j}(X^{(s)}_i)}{p^{(s)}_j(X^{(s)}_i)}\,\pi_j(\dd\lambda_j)\right\}^{1/2}.
\]
Under $\mathbb{P}$, conditional on $\mathbf{I}$ (equivalently on the counts $(N_{s,j})_{s,j}$), the collections of observations
$\{X^{(s)}_i: I^{(s)}_i=j,\ 1\le s\le S\}$ are independent across $j$, and $U_j(\mathbf{X})$ is a measurable function
of the observations in cell $j$ only. Therefore, conditional on $\mathbf{I}$,
\[
\mathbb{E}_{\mathbb{P}}\left[\prod_{j=1}^m U_j(\mathbf{X})\ \Big|\ \mathbf{I}\right]
=\prod_{j=1}^m \mathbb{E}_{\mathbb{P}}\left[U_j(\mathbf{X})\ \Big|\ \mathbf{I}\right].
\]
Plugging this into \eqref{eq:rho-conditional} and using the tower property yields
\begin{equation}
\label{eq:rho-factorized}
\rho(\mathbb{P},\overline{\mathbb{Q}})
=\mathbb{E}\left[\prod_{j=1}^m \rho_j(N_{1,j},\ldots,N_{S,j})\right],
\end{equation}
where $\rho_j(n_1,\ldots,n_S)$ denotes the Hellinger affinity between the \emph{within-cell} baseline and mixture laws:
\begin{equation}
\label{eq:rhoj-def}
\rho_j(n_1,\ldots,n_S)
:=\rho\left(\bigotimes_{s=1}^S (P^{(s)}_j)^{\otimes n_s},\ \int \bigotimes_{s=1}^S (Q^{(s)}_{j,\lambda_j})^{\otimes n_s}\,\pi_j(\dd\lambda_j)\right).
\end{equation}
In \eqref{eq:rhoj-def}, $P^{(s)}_j$ is the law with density $p^{(s)}_j$ and $Q^{(s)}_{j,\lambda_j}$ is the law with density
$q^{(s)}_{j,\lambda_j}$.

\paragraph{Within-cell affinity bound for a fixed count vector.}
Fix a cell $j$ and integers $n_1,\ldots,n_S\ge 0$, and let $n:=n_1+\cdots+n_S$.
We claim that
\begin{equation}
\label{eq:within-cell-affinity-bound}
1-\rho_j(n_1,\ldots,n_S)\;\le\;\frac{1}{2}\sum_{r=2}^{n}\binom{n}{r} b^r
=\frac{1}{2}\Bigl\{(1+b)^{n}-1-nb\Bigr\}.
\end{equation}
\emph{Proof of \eqref{eq:within-cell-affinity-bound}.}
Let $\mathbb{P}_j$ and $\overline{\mathbb{Q}}_j$ denote the within-cell baseline and mixture laws, respectively, where
\[
\mathbb{P}_j:=\bigotimes_{s=1}^S (P^{(s)}_j)^{\otimes n_s},
\qquad
\overline{\mathbb{Q}}_j:=\int \bigotimes_{s=1}^S (Q^{(s)}_{j,\lambda_j})^{\otimes n_s}\,\pi_j(\dd\lambda_j).
\]
\emph{Remark.} Fix $j$. For each $s$ and $x\in\mathcal{X}^{(s)}_j$, Assumption (A.1)(ii) and \eqref{eq:centering-single-sample} give
$p^{(s)}_j(x)=\int q^{(s)}_{j,\lambda_j}(x)\,\pi_j(\dd\lambda_j)$. Since the integrand is nonnegative, this implies
$p^{(s)}_j(x)=0\Rightarrow q^{(s)}_{j,\lambda_j}(x)=0$ for $\pi_j$-a.e.\ $\lambda_j$, hence $\overline{\mathbb{Q}}_j\ll \mathbb{P}_j$.
Therefore the likelihood ratio $L_j$ defined below exists $\mathbb{P}_j$-a.s., is nonnegative, and satisfies $\mathbb{E}_{\mathbb{P}_j}[L_j]=1$.

Let $\nu_j$ be any $\sigma$-finite measure dominating both $\mathbb{P}_j$ and $\overline{\mathbb{Q}}_j$ and let
$L_j:=\dd\overline{\mathbb{Q}}_j/\dd\mathbb{P}_j$ denote the likelihood ratio (which exists $\mathbb{P}_j$-a.s. by absolute continuity).

First note the elementary inequality
\begin{equation}
\label{eq:sqrt-ineq}
\sqrt{1+y}\ \ge\ 1+\frac{y}{2}-\frac{y^2}{2}\qquad\text{for all }y\ge -1.
\end{equation}
To verify \eqref{eq:sqrt-ineq}, first note that if $y\ge 2$ then
$1+\frac{y}{2}-\frac{y^2}{2}\le 0$, while $\sqrt{1+y}\ge 0$, so the inequality holds.
Now assume $y\in[-1,2]$. In this range, both sides of \eqref{eq:sqrt-ineq} are nonnegative, so we may square them.
A direct expansion yields
\[
(1+y)-\left(1+\frac{y}{2}-\frac{y^2}{2}\right)^2
=\frac{y^2}{4}(y+1)(3-y)\ \ge\ 0
\qquad\text{for all }y\in[-1,2],
\]
which proves \eqref{eq:sqrt-ineq}.
Apply \eqref{eq:sqrt-ineq} with $y=L_j-1$ (note that $L_j\ge 0$ implies $y\ge -1$). Since $\mathbb{E}_{\mathbb{P}_j}[L_j]=1$,
we obtain
\begin{equation}
\label{eq:affinity-chi2-ineq}
\rho_j(n_1,\ldots,n_S)=\mathbb{E}_{\mathbb{P}_j}[\sqrt{L_j}]
\ge 1-\frac{1}{2}\mathbb{E}_{\mathbb{P}_j}\bigl[(L_j-1)^2\bigr]
=1-\frac{1}{2}\Bigl(\mathbb{E}_{\mathbb{P}_j}[L_j^2]-1\Bigr).
\end{equation}

Next we bound $\mathbb{E}_{\mathbb{P}_j}[L_j^2]$.
Write $\lambda$ for $\lambda_j$ to simplify notation.
For each $s$ and $\lambda$, let $r^{(s)}_\lambda(x):=q^{(s)}_{j,\lambda}(x)/p^{(s)}_j(x)$ be the within-cell density ratio.
Then, by definition of $\overline{\mathbb{Q}}_j$,
\[
L_j(\mathbf{x})
=\int \prod_{s=1}^S\prod_{i=1}^{n_s} r^{(s)}_\lambda(x^{(s)}_i)\,\pi_j(\dd\lambda).
\]
Using Tonelli (nonnegative integrands) and independence under $\mathbb{P}_j$ gives
\begin{align}
\label{eq:Lj2-expansion-1}
\mathbb{E}_{\mathbb{P}_j}[L_j^2]
&=\mathbb{E}_{\mathbb{P}_j}\left[\left(\int \prod_{s=1}^S\prod_{i=1}^{n_s} r^{(s)}_\lambda(X^{(s)}_i)\,\pi_j(\dd\lambda)\right)
\left(\int \prod_{s=1}^S\prod_{i=1}^{n_s} r^{(s)}_{\lambda'}(X^{(s)}_i)\,\pi_j(\dd\lambda')\right)\right]\\
&=\iint \mathbb{E}_{\mathbb{P}_j}\left[\prod_{s=1}^S\prod_{i=1}^{n_s} r^{(s)}_\lambda(X^{(s)}_i)\,r^{(s)}_{\lambda'}(X^{(s)}_i)\right]\pi_j(\dd\lambda)\pi_j(\dd\lambda')\nonumber\\
&=\iint \prod_{s=1}^S \left(\int r^{(s)}_\lambda(x)\,r^{(s)}_{\lambda'}(x)\,p^{(s)}_j(x)\,\mu_s(\dd x)\right)^{n_s}\pi_j(\dd\lambda)\pi_j(\dd\lambda').\nolinebreak\nonumber
\end{align}
For each $s$, define
\[
D_s(\lambda,\lambda'):=\int \bigl(r^{(s)}_\lambda(x)-1\bigr)\bigl(r^{(s)}_{\lambda'}(x)-1\bigr)\,p^{(s)}_j(x)\,\mu_s(\dd x).
\]
Since $\int (r^{(s)}_\lambda-1)p^{(s)}_j\,\dd\mu_s= \int (q^{(s)}_{j,\lambda}-p^{(s)}_j)\,\dd\mu_s=0$, we have
\begin{equation}
\label{eq:inner-prod-1plusD}
\int r^{(s)}_\lambda r^{(s)}_{\lambda'}\,p^{(s)}_j\,\dd\mu_s
=1 + D_s(\lambda,\lambda').
\end{equation}
Moreover, by Cauchy--Schwarz and the definition of $b$ in \eqref{eq:def-b}, for every $s$ and all $\lambda,\lambda'$,
\begin{equation}
\label{eq:Ds-bound-by-b}
\begin{aligned}
|D_s(\lambda,\lambda')|
&\le
\left(\int (r^{(s)}_\lambda-1)^2\,p^{(s)}_j\,\dd\mu_s\right)^{1/2}\\
&\qquad\times
\left(\int (r^{(s)}_{\lambda'}-1)^2\,p^{(s)}_j\,\dd\mu_s\right)^{1/2}\\
&\le b.
\end{aligned}
\end{equation}
Finally, Assumption (A.2) implies that
\[
\int r^{(s)}_\lambda(x)\,\pi_j(\dd\lambda)=1
\qquad\text{for $\mu_s$-a.e.\ $x\in\mathcal{X}^{(s)}_j$,}
\]
and hence, by Fubini,
\[
\iint D_s(\lambda,\lambda')\,\pi_j(\dd\lambda)\pi_j(\dd\lambda')=0.
\]

Plugging \eqref{eq:inner-prod-1plusD} into \eqref{eq:Lj2-expansion-1}, we obtain
\begin{equation}
\label{eq:Lj2-expansion-2}
\mathbb{E}_{\mathbb{P}_j}[L_j^2]
=\iint \prod_{s=1}^S \bigl(1+D_s(\lambda,\lambda')\bigr)^{n_s}\,\pi_j(\dd\lambda)\pi_j(\dd\lambda').
\end{equation}
Expand each factor via the binomial theorem:
\[
\bigl(1+D_s(\lambda,\lambda')\bigr)^{n_s}=\sum_{k_s=0}^{n_s}\binom{n_s}{k_s}D_s(\lambda,\lambda')^{k_s}.
\]
Multiplying these expansions over $s$ yields
\[
\prod_{s=1}^S \bigl(1+D_s\bigr)^{n_s}
=\sum_{k_1=0}^{n_1}\cdots\sum_{k_S=0}^{n_S}\left(\prod_{s=1}^S\binom{n_s}{k_s}\right)\prod_{s=1}^S D_s^{k_s}.
\]
Taking expectation with respect to $\pi_j(\dd\lambda)\pi_j(\dd\lambda')$ and using that $\mathbb{E}[D_s]=0$ shows that all terms
with total degree $k_1+\cdots+k_S=1$ vanish. Therefore, using \eqref{eq:Ds-bound-by-b} and absolute values,
\begin{align}
\label{eq:Lj2-bound}
\mathbb{E}_{\mathbb{P}_j}[L_j^2]-1
&\le \sum_{r=2}^{n}\ \sum_{\substack{k_1,\ldots,k_S\ge 0:\\ k_1+\cdots+k_S=r}}\left(\prod_{s=1}^S\binom{n_s}{k_s}\right)b^{r}.\nolinebreak
\end{align}
The inner sum is the coefficient of $t^r$ in $\prod_{s=1}^S(1+t)^{n_s}=(1+t)^{n}$, and hence equals $\binom{n}{r}$.
Thus \eqref{eq:Lj2-bound} implies
\[
\mathbb{E}_{\mathbb{P}_j}[L_j^2]-1\le \sum_{r=2}^{n}\binom{n}{r}b^{r}.
\]
Plugging this into \eqref{eq:affinity-chi2-ineq} yields \eqref{eq:within-cell-affinity-bound}. This completes the proof of
\eqref{eq:within-cell-affinity-bound}. \hfill $\blacksquare$

\paragraph{Bound the total affinity loss by summing over cells.}
Returning to \eqref{eq:rho-factorized} and using $0\le \rho_j(\cdot)\le 1$, we have the elementary inequality
\begin{equation}
\label{eq:prod-lower-bound}
1-\prod_{j=1}^m \rho_j \le \sum_{j=1}^m (1-\rho_j),
\end{equation}
which holds for any numbers $\rho_j\in[0,1]$. Taking expectations in \eqref{eq:prod-lower-bound} and using \eqref{eq:rho-factorized} gives
\[
1-\rho(\mathbb{P},\overline{\mathbb{Q}})
\le \sum_{j=1}^m \mathbb{E}\bigl[\,1-\rho_j(N_{1,j},\ldots,N_{S,j})\,\bigr].
\]
Applying \eqref{eq:within-cell-affinity-bound} with $n=M_j$ yields
\begin{equation}
\label{eq:affinity-loss-sum}
1-\rho(\mathbb{P},\overline{\mathbb{Q}})
\le \frac{1}{2}\sum_{j=1}^m \mathbb{E}\Bigl[\,(1+b)^{M_j}-1-M_j b\,\Bigr].
\end{equation}

We now bound each expectation in \eqref{eq:affinity-loss-sum}. For each fixed $j$, note that for each sample $s$,
$N_{s,j}$ has a $\mathrm{Binomial}(n_s,p_{s,j})$ distribution under $\mathbb{P}$. Moreover, because the $S$ samples are independent,
the random variables $(N_{s,j})_{s=1}^S$ are independent for each fixed $j$, and therefore so are the $(1+b)^{N_{s,j}}$.
Consequently,
\begin{align}
\label{eq:mgf-Mj}
\mathbb{E}\bigl[(1+b)^{M_j}\bigr]
&=\mathbb{E}\Bigl[\prod_{s=1}^S (1+b)^{N_{s,j}}\Bigr]
=\prod_{s=1}^S \mathbb{E}\bigl[(1+b)^{N_{s,j}}\bigr]
=\prod_{s=1}^S (1+p_{s,j}b)^{n_s}.
\end{align}
Also, $\mathbb{E}[M_j]=\sum_{s=1}^S n_s p_{s,j}$.

Define $m_j:=\sum_{s=1}^S n_s p_{s,j}$ and $x_j:=b m_j$. Then by \eqref{eq:mgf-Mj} and the inequality
$\log(1+u)\le u$ for $u>-1$,
\begin{equation}
\label{eq:exp-bound-Mj}
\mathbb{E}\bigl[(1+b)^{M_j}\bigr]
=\prod_{s=1}^S (1+p_{s,j}b)^{n_s}
\le \exp\left(\sum_{s=1}^S n_s p_{s,j}b\right)
=\exp(x_j).
\end{equation}
Moreover, by \eqref{eq:hellinger-assumption-A}, we have
\begin{equation}
\label{eq:xj-bound-by-A}
0\le x_j=b m_j \le b\cdot n_{\mathrm{tot}}\cdot p_{\max} \le A.
\end{equation}
For $x\ge 0$, the Taylor remainder formula implies
\begin{equation}
\label{eq:exp-remainder-bound}
e^x-1-x \le \frac{x^2}{2}e^x.
\end{equation}
Indeed, $e^x=1+x+\frac{x^2}{2}e^\xi$ for some $\xi\in[0,x]$, so $e^x-1-x=\frac{x^2}{2}e^\xi\le \frac{x^2}{2}e^x$.
Combining \eqref{eq:exp-bound-Mj}, \eqref{eq:exp-remainder-bound}, and \eqref{eq:xj-bound-by-A} yields
\[
\mathbb{E}\Bigl[(1+b)^{M_j}-1-M_j b\Bigr]
=\mathbb{E}\bigl[(1+b)^{M_j}\bigr]-1-x_j
\le e^{x_j}-1-x_j
\le \frac{x_j^2}{2}e^{x_j}
\le \frac{e^A}{2}x_j^2
=\frac{e^A}{2}b^2 m_j^2.
\]
Plugging this bound into \eqref{eq:affinity-loss-sum} gives
\begin{equation}
\label{eq:affinity-loss-prelim}
1-\rho(\mathbb{P},\overline{\mathbb{Q}})
\le \frac{e^A}{4}b^2\sum_{j=1}^m m_j^2.
\end{equation}

Finally, we bound $\sum_{j=1}^m m_j^2$. Since $m_j\ge 0$ and $\sum_{j=1}^m m_j=\sum_{s=1}^S n_s\sum_{j=1}^m p_{s,j}=n_{\mathrm{tot}}$, we have
\[
\sum_{j=1}^m m_j^2 \le \Bigl(\max_{1\le j\le m}m_j\Bigr)\sum_{j=1}^m m_j
=\Bigl(\max_{1\le j\le m}m_j\Bigr)\,n_{\mathrm{tot}}.
\]
Moreover, $m_j=\sum_{s=1}^S n_s p_{s,j}\le (\sum_{s=1}^S n_s)\,p_{\max}=n_{\mathrm{tot}}p_{\max}$, so $\max_j m_j\le n_{\mathrm{tot}}p_{\max}$ and therefore
\begin{equation}
\label{eq:mu-sum-squares-bound}
\sum_{j=1}^m m_j^2 \le n_{\mathrm{tot}}^2 p_{\max}.
\end{equation}
Combining \eqref{eq:affinity-loss-prelim} and \eqref{eq:mu-sum-squares-bound} yields
\[
1-\rho(\mathbb{P},\overline{\mathbb{Q}})
\le \frac{e^A}{4}b^2\,n_{\mathrm{tot}}^2\,p_{\max}.
\]
Finally, using \eqref{eq:H2-vs-affinity}, we obtain
\[
H^2(\mathbb{P},\overline{\mathbb{Q}})
=2-2\rho(\mathbb{P},\overline{\mathbb{Q}})
\le 2\{1-\rho(\mathbb{P},\overline{\mathbb{Q}})\}
\le \frac{e^A}{2}b^2\,n_{\mathrm{tot}}^2\,p_{\max},
\]
which is exactly \eqref{eq:hellinger-bound-multisample} with $C(A)=e^A/2$.
\end{proof}

\begin{remark}
\label{rem:hellinger-multisample}
When $S=1$, Theorem~\ref{thm:semi-param-thm} reduces to the usual one-sample hypercube bound
(cf. the simplified form of \citet[Theorem 2.1]{robins2009semiparametric} used in earlier drafts).
The multi-sample formulation is needed in covariate-shift settings, where one observes multiple independent samples
(e.g., a training sample and a target sample) whose component distributions may both vary under the local alternatives.
\end{remark}

Finally, we use the following theorem from \citet[Theorem 2.15]{tsybakov2008introduction}, which gives a lower bound based on the Hellinger distance. It is reproduced here for the reader's convenience.

\begin{theorem}[Lower bound via fuzzy hypotheses]
\label{fuzzy-thm}
\label{fano-method}
\label{thm:fano-method}
(\cite{tsybakov2008introduction}, Theorem 2.15) Let $\pi$ be a probability distribution on a set (measure space) of
distributions $\mathcal{P}$ with common support $\gX$, which induces the mixture distribution
$$
Q_1(A)=\int Q^{\otimes n}(A)\,\pi(\dd Q), \quad \forall A \subset \gX^n.
$$
Suppose that there exist $P\in\mathcal{P}$ and a functional $T: \mathcal{P}\mapsto\mathbb{R}$ satisfying
\begin{equation}
\label{fano:separation-condition}
T(P)\leq c, \quad \pi(\{Q: T(Q) \geq c+2s\})=1
\end{equation}
for some $s>0$. If $H^2\left(P^{\otimes n}, Q_1\right) \leq \delta<2$, then
\[
\inf_{\hat{T}:\,\gX^n\mapsto\mathbb{R}} \sup_{P' \in \mathcal{P}}
P'\left[\left|\hat{T}(X_1,\ldots,X_n)-T(P')\right|\geq s\right]
\geq \frac{1-\sqrt{\delta(1-\delta / 4)}}{2}.
\]
\end{theorem}

\section{Technical lemmas}

In this section, we present and prove several technical lemmas that will be used in the main proof.

First, we state a version of the Ham--sandwich theorem. For completeness, we also provide a proof via the Borsuk--Ulam theorem.

\begin{theorem}[Ham-sandwich via a nondegenerate function family]
\label{thm:ham-sandwich}
Let $(\gZ,\mu)$ be a measure space. Assume there are bounded $\mu$-measurable functions
$f_0,f_1,\dots,f_q:\gZ\to\R$ that are linearly independent modulo $\mu$-null sets, i.e.,
for any $(\lambda_0,\dots,\lambda_q)\neq 0$,
\[
\mu\left(\Big\{z\in\gZ:\ \sum_{i=0}^q \lambda_i f_i(z)=0\Big\}\right)=0.
\]
Let $w_1,\dots,w_q\in L^1(\mu)$ be $\mu$-integrable (not necessarily nonnegative).
Then there exists a vector $\alpha=(\alpha_0,\alpha_1,\dots,\alpha_q)\in\R^{q+1}\setminus\{0\}$
such that with
\[
\gZ^{(1)}(\alpha):=\Big\{z\in\gZ: \ \alpha_0 f_0(z)+\sum_{i=1}^q \alpha_i f_i(z)\ge 0\Big\}
\]
we have, for each $j=1,\dots,q$,
\[
\int_{\gZ^{(1)}(\alpha)} w_j\,d\mu \;=\; \frac12\int_{\gZ} w_j\,d\mu.
\]
Equivalently, the two parts $\gZ^{(1)}(\alpha)$ and $\gZ\setminus \gZ^{(1)}(\alpha)$ give equal
$w_j$-mass for all $j=1,\dots,q$.
\end{theorem}

\begin{proof}
Define the $(q{+}1)$-tuple of functions $F=(f_0,f_1,\dots,f_q)$ and, for $\alpha\in\R^{q+1}$,
\[
h_\alpha(z):=\alpha\cdot F(z)=\alpha_0 f_0(z)+\sum_{i=1}^q \alpha_i f_i(z).
\]
By linear independence modulo null sets, for any $\alpha\neq 0$ the zero set
$\{z:h_\alpha(z)=0\}$ is $\mu$-null.

We work on the unit sphere $S^{q}=\{\alpha\in\R^{q+1}:\|\alpha\|_2=1\}$. For
$\alpha\in S^{q}$, define
\[
\Phi(\alpha)\;:=\;\Big(\Phi_1(\alpha),\dots,\Phi_q(\alpha)\Big)\in\R^q,\qquad
\Phi_j(\alpha)\;:=\;\int_{\gZ}\operatorname{sgn}\big(h_\alpha(z)\big)\,w_j(z)\,d\mu(z),
\]
where $\operatorname{sgn}(t)=\mathbf{1}\{t\ge 0\}-\mathbf{1}\{t<0\}$.
Because $\{h_\alpha=0\}$ is $\mu$-null for each $\alpha\neq 0$, we may (and do) treat
$\mathbf{1}\{h_\alpha\ge 0\}$ and $\mathbf{1}\{h_\alpha> 0\}$ as indistinguishable in $L^1(\mu)$.

\smallskip\noindent\emph{Claim 1 (continuity).}
$\Phi:S^{q}\to\R^q$ is continuous.

\emph{Proof of Claim 1.} Fix $\alpha\in S^{q}$ and let $\alpha^{(n)}\to\alpha$.
Then $h_{\alpha^{(n)}}(z)\to h_\alpha(z)$ for each $z$, hence
$\mathbf{1}\{h_{\alpha^{(n)}}(z)\ge 0\}\to\mathbf{1}\{h_\alpha(z)\ge 0\}$
for all $z$ with $h_\alpha(z)\neq 0$. Since the exceptional set $\{h_\alpha=0\}$
is $\mu$-null, the convergence holds $\mu$-a.e.  By dominated convergence
(because $|\operatorname{sgn}(h_{\alpha^{(n)}})|\le 1$ and $w_j\in L^1(\mu)$),
$\Phi_j(\alpha^{(n)})\to\Phi_j(\alpha)$ for each $j$. \qed

\smallskip\noindent\emph{Claim 2 (oddness).}
$\Phi(-\alpha)=-\Phi(\alpha)$ for all $\alpha\in S^{q}$.

\emph{Proof of Claim 2.} Since $h_{-\alpha}=-h_\alpha$, we have
$\operatorname{sgn}(h_{-\alpha})=-\operatorname{sgn}(h_\alpha)$ pointwise
away from the $\mu$-null set $\{h_\alpha=0\}$. Hence the integrals change sign. \qed

\smallskip
By the Borsuk--Ulam theorem, there exists $\alpha^\star\in S^{q}$ such that
$\Phi(\alpha^\star)=\Phi(-\alpha^\star)$. Since $\Phi$ is odd, this forces $\Phi(\alpha^\star)=0$.

Finally, for each $j$,
\[
0=\Phi_j(\alpha^\star)
=\int_{\gZ}\operatorname{sgn}(h_{\alpha^\star})\,w_j\,d\mu
=\int_{\gZ}\big(\mathbf{1}\{h_{\alpha^\star}\ge 0\}-\mathbf{1}\{h_{\alpha^\star}<0\}\big)\,w_j\,d\mu.
\]
Since $\{h_{\alpha^\star}=0\}$ is $\mu$-null, the last display equals
\[
\Big(\int_{\{h_{\alpha^\star}\ge 0\}}w_j\,d\mu\Big)
-\Big(\int_{\{h_{\alpha^\star}< 0\}}w_j\,d\mu\Big)
=2\int_{\{h_{\alpha^\star}\ge 0\}}w_j\,d\mu-\int_{\gZ}w_j\,d\mu,
\]
which yields
\(
\displaystyle \int_{\{h_{\alpha^\star}\ge 0\}} w_j\,d\mu=\tfrac12\int_{\gZ} w_j\,d\mu.
\)
Setting $\gZ^{(1)}=\{z:h_{\alpha^\star}(z)\ge 0\}$ completes the proof.
\end{proof}

The following corollary can be obtained via applying the general Ham-sandwich theorem multiple times.

\begin{corollary}[Iterated ham-sandwich partition]
\label{cor:ham-sandwich-implication}
    Under the assumptions in Theorem \ref{thm:ham-sandwich}, for any positive integer $m$ there exists a partition
    $\{B_j\}_{j=1}^M$ of $\gZ$ with $M=2^m$ such that
    \begin{equation}
        \int_{B_j}w_i(z)\,d\mu = \frac{1}{M}\int_{\gZ}w_i(z)\,d\mu,\quad \forall 1\leq i\leq q, 1\leq j\leq M.
    \end{equation}
\end{corollary}

\begin{proof}
We prove the result by induction on $m$.

\emph{Base case ($m=1$).}
By Theorem \ref{thm:ham-sandwich} there exists a measurable set
$\gZ^{(1)}\subseteq\gZ$ such that, for each $1\le i\le q$,
\[
\int_{\gZ^{(1)}} w_i\,d\mu=\frac12\int_{\gZ} w_i\,d\mu.
\]
Setting $B_1=\gZ^{(1)}$ and $B_2=\gZ\setminus \gZ^{(1)}$ gives $M=2$ with the stated property.

\emph{Inductive step.}
Assume the statement holds for some $m\ge 1$, i.e., there exists a partition
$\{\hat B_r\}_{r=1}^{2^m}$ of $\gZ$ such that for all $1\le i\le q$ and
$1\le r\le 2^m$,
\[
\int_{\hat B_r} w_i\,d\mu=\frac{1}{2^m}\int_{\gZ} w_i\,d\mu.
\]
For each $r\in\{1,\dots,2^m\}$, define the restricted measure
$\mu_r(A):=\mu(A\cap \hat B_r)$.
Let $f_0,f_1,\dots,f_q$ be the functions appearing in Theorem \ref{thm:ham-sandwich}, and write
$\hat f_{i,r}:=f_i|_{\hat B_r}$ for $0\le i\le q$ and $\hat w_{j,r}:=w_j|_{\hat B_r}$.

\emph{Claim (nondegeneracy is inherited).}
If $f_0,f_1,\dots,f_q$ are linearly independent modulo $\mu$-null sets on $\gZ$, then
$\hat f_{0,r},\hat f_{1,r},\dots,\hat f_{q,r}$ are linearly independent modulo $\mu_r$-null sets on $\hat B_r$.
Indeed, for any nonzero $(\lambda_0,\lambda_1,\dots,\lambda_q)$,
\begin{equation}
    \Big\{ z\in \hat{B}_r: \sum_{i=0}^q\lambda_i \hat f_{i,r}(z)=0\Big\}
    \subseteq \Big\{ z\in \gZ: \sum_{i=0}^q\lambda_i f_i(z)=0\Big\}
\end{equation}
is a $\mu$-null subset of $\hat B_r$, hence also a $\mu_r$-null subset. Therefore
$(\hat B_r,\mu_r)$ is $(q{+}1)$-nondegenerate.

Applying Theorem \ref{thm:ham-sandwich} on each $(\hat B_r,\mu_r)$ with the functions
$\hat f_{0,r},\hat f_{1,r},\dots,\hat f_{q,r}$ and weights $\hat w_{j,r}$, we obtain a bipartition
$\hat B_r=B_{2r-1}\cup B_{2r}$ such that, for every $1\le i\le q$,
\begin{equation}
    \int_{B_{2r-1}} w_i(z)\,d\mu_r=\frac{1}{2}\int_{\hat B_r} w_i(z)\,d\mu_r.
\end{equation}
Since $\mu_r$ is the restriction of $\mu$ to $\hat B_r$, the above equality is equivalent to
\begin{equation}
    \int_{B_{2r-1}} w_i(z)\,d\mu=\frac{1}{2}\int_{\hat B_r} w_i(z)\,d\mu.
\end{equation}
(And of course the same holds with $B_{2r}$ in place of $B_{2r-1}$.)

Summing up, from the inductive hypothesis
$\int_{\hat B_r} w_i\,d\mu=\frac{1}{2^m}\int_{\gZ} w_i\,d\mu$ we conclude
\[
\int_{B_{2r-1}} w_i\,d\mu
=\frac{1}{2}\cdot\frac{1}{2^m}\int_{\gZ} w_i\,d\mu
=\frac{1}{2^{m+1}}\int_{\gZ} w_i\,d\mu,
\quad
\int_{B_{2r}} w_i\,d\mu=\frac{1}{2^{m+1}}\int_{\gZ} w_i\,d\mu,
\]
for every $1\le i\le q$. Since $r$ was arbitrary, the $2^{m+1}$ sets
$B_1,\dots,B_{2^{m+1}}$ form a partition of $\gZ$ with the desired property for $m{+}1$.

This completes the induction.
\end{proof}

We move on to derive some formulae for calculating the values and derivatives of a functional.

\begin{lemma}[Bumped perturbations preserve feasibility and first derivatives]
    \label{lemma:bump-derivative}
    Let $(G,K)$ be a feasible joint perturbation at $(\hat{P},\hat{Q})$ in the sense of
    Definition~\ref{def:joint-perturbation}, and suppose that $(G,K)$ is $Z_1$-modulation closed at $(\hat{P},\hat{Q})$
    with modulation radius $r_{G,K}^{\mathrm{mod}}>0$ (Definition~\ref{def:z1-modulation-closure}).
    For any function $\psi:\gZ_1\to\R$ uniformly bounded by $C_{\psi}>0$, define the $Z_1$-modulated signed measures
    $(G_{\psi},K_{\psi})$ by
    \[
        \frac{\dd G_{\psi}}{\dd\mu}(o):=\psi(z_1)\,\frac{\dd G}{\dd\mu}(o),
        \qquad
        \frac{\dd K_{\psi}}{\dd\mu_Z}(z):=\psi(z_1)\,\frac{\dd K}{\dd\mu_Z}(z).
    \]
    Then under the assumptions in Section~\ref{subsec:main-asmp}, $(G_{\psi},K_{\psi})$ is a feasible joint perturbation
    at $(\hat{P},\hat{Q})$ with feasible radius $C_{\psi}^{-1}r_{G,K}^{\mathrm{mod}}$ as long as the centering
    conditions $G_{\psi}(\gO)=0$ and $K_{\psi}(\gZ)=0$ hold.
    In this case, for all $z=(z_1,z_2)\in\gZ$ and all $|s|\leq C_{\psi}^{-1}r_{G,K}^{\mathrm{mod}}$, we have
    \begin{equation}
        \gamma\big(z;\hat{P}+sG_{\psi}\big) = \gamma\big(z;\hat{P}+(s\psi(z_1))G\big),\quad
        \alpha\big(z;\hat{P}+sG_{\psi},\hat{Q}+sK_{\psi}\big) = \alpha\big(z;\hat{P}+(s\psi(z_1))G,\hat{Q}+(s\psi(z_1))K\big),
    \end{equation}
    where, on the right-hand side, $\eta(z;\hat{P}+(s\psi(z_1))G,\hat{Q}+(s\psi(z_1))K)$ is a shorthand for evaluating
    $\eta$ along the original path $(\hat{P}+tG,\hat{Q}+tK)$ at $t=s\psi(z_1)$ (and for $\gamma$ the $Q$-argument is
	    ignored).
	    Moreover, whenever the directional (G\^ateaux) derivatives exist (e.g., under Assumption~\ref{asmp:regularity}(1)), we have
	    \begin{equation}
	        \begin{aligned}
	            \gamma_P'(z;\hat{P})[G_{\psi}]
	            &=
	            \psi(z_1)\gamma_P'(z;\hat{P})[G],\\
	            \alpha_{(P,Q)}'(z;\hat{P},\hat{Q})[G_{\psi},K_{\psi}]
	            &=
	            \psi(z_1)\alpha_{(P,Q)}'(z;\hat{P},\hat{Q})[G,K],
	            \qquad\forall z=(z_1,z_2)\in\gZ.
	        \end{aligned}
	    \end{equation}
\end{lemma}

\begin{proof}
    Write $g=\dd G/\dd\mu$ and $k=\dd K/\dd\mu_Z$, and define $g_{\psi}:=\dd G_{\psi}/\dd\mu$ and
    $k_{\psi}:=\dd K_{\psi}/\dd\mu_Z$ so that $g_{\psi}(o)=\psi(z_1)g(o)$ and $k_{\psi}(z)=\psi(z_1)k(z)$.
    Essential boundedness of $g$ and $k$ and boundedness of $\psi$ imply that $g_{\psi}$ and $k_{\psi}$ are also
    essentially bounded.

    \paragraph{Feasibility of the modulated pair.}
    Let $\tilde\psi:=\psi/C_{\psi}$ so that $\|\tilde\psi\|_{\infty}\le 1$ and note that
    $G_{\psi}=C_{\psi}G_{\tilde\psi}$ and $K_{\psi}=C_{\psi}K_{\tilde\psi}$ in the notation of
    Definition~\ref{def:z1-modulation-closure}.
    The centering conditions $G_{\psi}(\gO)=0$ and $K_{\psi}(\gZ)=0$ are equivalent to
    $G_{\tilde\psi}(\gO)=0$ and $K_{\tilde\psi}(\gZ)=0$.
    Since $(G,K)$ is $Z_1$-modulation closed at $(\hat{P},\hat{Q})$ with modulation radius $r_{G,K}^{\mathrm{mod}}$,
    it follows that
    $(\hat{P}+tG_{\tilde\psi},\hat{Q}+tK_{\tilde\psi})\in\gP_0$ for all $|t|\le r_{G,K}^{\mathrm{mod}}$.
    Setting $t:=sC_{\psi}$ gives
    \[
        (\hat{P}+sG_{\psi},\hat{Q}+sK_{\psi})
        =
        (\hat{P}+tG_{\tilde\psi},\hat{Q}+tK_{\tilde\psi})
        \in\gP_0
        \qquad\text{for all }|s|\le C_{\psi}^{-1}r_{G,K}^{\mathrm{mod}}.
    \]
    Thus $(G_{\psi},K_{\psi})$ is a feasible joint perturbation at $(\hat{P},\hat{Q})$ with feasible radius
    $C_{\psi}^{-1}r_{G,K}^{\mathrm{mod}}$.

    \paragraph{Effect on slice-based maps and derivative scaling.}
    For each fixed $z_1\in\gZ_1$ and any $s$ with $|s|\le C_{\psi}^{-1}r_{G,K}^{\mathrm{mod}}$,
    \[
        (\hat p+s g_{\psi})_{z_1} = \hat p_{z_1}+s\,\psi(z_1)\,g_{z_1},
        \qquad
        (\hat q+s k_{\psi})_{z_1} = \hat q_{z_1}+s\,\psi(z_1)\,k_{z_1}.
    \]
    By Assumption~\ref{asmp:cond-prob-functional}(1), $\gamma(z_1,z_2;P)$ depends on $P$ only through the slice
    $p_{z_1}$, while $\alpha(z_1,z_2;P,Q)$ depends on $(P,Q)$ only through the pair of slices $(p_{z_1},q_{z_1})$.
    Therefore, for each $z=(z_1,z_2)$,
    \[
        \gamma\big(z;\hat{P}+sG_{\psi}\big) = \gamma\big(z;\hat{P}+(s\psi(z_1))G\big),
        \qquad
        \alpha\big(z;\hat{P}+sG_{\psi},\hat{Q}+sK_{\psi}\big)
        =\alpha\big(z;\hat{P}+(s\psi(z_1))G,\hat{Q}+(s\psi(z_1))K\big),
    \]
    where the right-hand side is interpreted as evaluation along the original path
    $(\hat{P}+tG,\hat{Q}+tK)$ at $t=s\psi(z_1)$.

	    Finally, whenever the directional (G\^ateaux) derivatives exist at $(\hat{P},\hat{Q})$ (in particular, under
    Assumption~\ref{asmp:regularity}(1)), the derivative identities follow by differentiating the previous display
    at $s=0$ and using linearity of the first derivative in its direction arguments:
    \[
        \gamma_P'(z;\hat{P})[G_{\psi}]
        = \psi(z_1)\gamma_P'(z;\hat{P})[G],
        \qquad
        \alpha_{(P,Q)}'(z;\hat{P},\hat{Q})[G_{\psi},K_{\psi}]
        = \psi(z_1)\alpha_{(P,Q)}'(z;\hat{P},\hat{Q})[G,K].
    \]
\end{proof}

The following corollary is an immediate consequence of Lemma \ref{lemma:bump-derivative}.

\begin{corollary}[Bumped invariant directions remain invariant]
\label{cor:bump-invariance}
    Let $(G_0,K_0)$ be the $\gamma$-invariant feasible joint perturbation given by Assumption~\ref{asmp:main}, and suppose
    $(G_0,K_0)$ is $Z_1$-modulation closed at $(\hat{P},\hat{Q})$ with modulation radius
    $r_{G_0,K_0}^{\mathrm{mod}}>0$ (Definition~\ref{def:z1-modulation-closure}).
    For any measurable $\psi:\gZ_1\to\R$ with $\|\psi\|_{\infty}<\infty$, define the modulated pair
    $(G_{0\psi},K_{0\psi})$ as in Lemma~\ref{lemma:bump-derivative}.
    If $\psi\equiv 0$, then $\gamma(z;\hat{P}+sG_{0\psi})=\gamma(z;\hat{P})$ holds trivially for all $s$.
    Otherwise, if the centering conditions $G_{0\psi}(\gO)=0$ and $K_{0\psi}(\gZ)=0$ hold, then
    \[
        \gamma(z;\hat{P}+sG_{0\psi})=\gamma(z;\hat{P}),\quad
        \forall |s|\leq\|\psi\|_{\infty}^{-1}\min\big\{c_t,r_{G_0,K_0}^{\mathrm{mod}}\big\},
    \]
    where $c_t$ is the constant defined in Assumption~\ref{asmp:main}.
\end{corollary}

\begin{proof}
    If $\psi\equiv 0$, then $G_{0\psi}=0$ and the claim holds for all $s$. Hence assume $\|\psi\|_{\infty}>0$.
    By Lemma~\ref{lemma:bump-derivative}, for each $z=(z_1,z_2)\in\gZ$ and each
    $|s|\le \|\psi\|_{\infty}^{-1}r_{G_0,K_0}^{\mathrm{mod}}$,
    \[
        \gamma(z;\hat{P}+sG_{0\psi})
        =
        \gamma\big(z;\hat{P}+(s\psi(z_1))G_0\big).
    \]
    If additionally $|s|\le \|\psi\|_{\infty}^{-1}c_t$, then $|s\psi(z_1)|\le c_t$ for all $z_1$.
    Since $G_0$ is $\gamma$-invariant in Assumption~\ref{asmp:main}(1), the right-hand side equals $\gamma(z;\hat{P})$.
    Combining the two bounds on $|s|$ yields the claim.
\end{proof}

\section{Proof of Theorem \ref{thm:ate-upper-bound}}
\label{sec:proof:ate-upper-bound}

We prove the high-probability bound by decomposing the error into a sampling term and a bias term.

\paragraph{Reduce to bounding a sampling term and a bias term.}
Write
\[
\psi_{\ate}\big(o;\hat g,\hat m\big)
:=
\hat g(1,x)-\hat g(0,x)
\;+\;
\frac{d-\hat m(x)}{\hat m(x)\,(1-\hat m(x))}\Big(y-\hat g(d,x)\Big),
\qquad o=(x,d,y),
\]
so that $\hat\theta^{\ate}=\frac1n\sum_{i=1}^n \psi_{\ate}(O_i;\hat g,\hat m)$.
Let
\[
\bar\theta^{\ate}:=\E\big[\psi_{\ate}(O;\hat g,\hat m)\big],
\]
where the expectation is under the true data-generating distribution $P_0$.
Then
\[
|\hat\theta^{\ate}-\theta^{\ate}|
\le
\big|\hat\theta^{\ate}-\bar\theta^{\ate}\big|
\;+\;
\big|\bar\theta^{\ate}-\theta^{\ate}\big|.
\]

\paragraph{Bound the sampling term.}
Since $|Y|\le G$ a.s., we have $|g_0(d,x)|=|\E[Y\mid D=d,X=x]|\le G$.
Define the clipped estimator $\tilde g(d,x):=\min\{G,\max\{-G,\hat g(d,x)\}\}$. Since $g_0(d,x)\in[-G,G]$, clipping cannot increase
$\|\hat g(d,X)-g_0(d,X)\|_{P_X,2}$, so we may replace $\hat g$ by $\tilde g$ and (by abuse of notation) continue to write $\hat g$, with $|\hat g(d,x)|\le G$ for all $(d,x)$.
Under $c\le \hat m(x)\le 1-c$, we have $\hat m(x)(1-\hat m(x))\ge c(1-c)$ and $|d-\hat m(x)|\le 1$, hence
\[
\big|\psi_{\ate}(O;\hat g,\hat m)\big|
\le
2G+\frac{2G}{c(1-c)}
:\;B.
\]
Conditioning on the nuisance estimates under sample splitting (so that the evaluation sample is independent of $\hat g,\hat m$),
the summands $\psi_{\ate}(O_i;\hat g,\hat m)$ are i.i.d.\ with mean $\bar\theta^{\ate}$ and are bounded in $[-B,B]$.
(Under $K$-fold cross-fitting, the same bound follows by applying Hoeffding's inequality within each fold conditional on the fold-specific nuisance fits, and taking a union bound over folds.)
Hoeffding's inequality therefore implies that for any $\delta\in(0,1)$,
\[
\big|\hat\theta^{\ate}-\bar\theta^{\ate}\big|
\le
B\,\sqrt{2\log(2/\delta)}\,n^{-1/2}
\qquad\text{with probability at least }1-\delta.
\]

\paragraph{Bound the bias term.}
Let $m_0(x)=\E[D\mid X=x]$ and $g_0(d,x)=\E[Y\mid D=d,X=x]$.
A direct conditional expectation calculation yields
\[
\bar\theta^{\ate}-\theta^{\ate}
=
\E\left[
\frac{m_0(X)-\hat m(X)}{\hat m(X)}\Big(g_0(1,X)-\hat g(1,X)\Big)
\;+\;
\frac{m_0(X)-\hat m(X)}{1-\hat m(X)}\Big(g_0(0,X)-\hat g(0,X)\Big)
\right].
\]
Using $c\le \hat m(X)\le 1-c$ and Cauchy--Schwarz,
\[
\big|\bar\theta^{\ate}-\theta^{\ate}\big|
\le
\frac{1}{c}\,\|m_0(X)-\hat m(X)\|_{P_X,2}\Big(\|g_0(1,X)-\hat g(1,X)\|_{P_X,2}+\|g_0(0,X)-\hat g(0,X)\|_{P_X,2}\Big)
\le
\frac{2}{c}\,\eps_{n,m}\eps_{n,g}.
\]

\textit{Conclusion.}
Combining the sampling and bias bounds and absorbing constants into $C_\delta$ yields
$
|\hat\theta^{\ate}-\theta^{\ate}|
\le
C_\delta\big(\eps_{n,m}\eps_{n,g}+n^{-1/2}\big)
$
with probability at least $1-\delta$, as claimed.

\section{Proof of Theorem \ref{thm:ate-only}}
\label{sec:proof:ate-only}

In this section, we prove Theorem \ref{thm:ate-only}, which asserts the optimality of first-order debiasing in the special case of ATE. Although this statement is a corollary of Theorem \ref{thm:main-mixed-bias}, we develop a stand-alone proof both because of ATE's prominence and because the construction here motivates the general strategy.

\paragraph{Choosing the bump function (via ham--sandwich pairing).}
Choose an integer \(r\ge 1\) and set \(M:=2^r\). Apply Corollary \ref{cor:ham-sandwich-implication} on \((\gX,\mu)\) with the two weight functions \(w_1(x)\equiv 1\) and \(w_2(x):=2\hat m(x)-1\) (with the corollary parameter set to \(r\), so the number of blocks is \(2^r=M\)) to obtain a partition \(B_1,B_2,\dots,B_M\) of \(\gX=[0,1]^K\) such that, for every \(j=1,\dots,M\),
\[
\mu(B_j)=\frac{1}{M}
\quad\text{and}\quad
\int_{B_j}w_2(x)\,d\mu(x)=\frac{1}{M}\int_{\gX}w_2(x)\,d\mu(x).
\]
Pair the blocks as \((B_1,B_2),(B_3,B_4),\dots,(B_{M-1},B_M)\). The choice \(w_1\equiv 1\) enforces uniform block measure (used in Lemma \ref{lemma:hellinger-bound-ate} to get \(p_j=2/M\)), while balancing \(w_2\) ensures \(\E[\Delta(\lambda,X)\,(2\hat m(X)-1)]=0\), which cancels the linear term in Lemma \ref{lemma:ate-separation}.

For \(\lambda=(\lambda_1,\dots,\lambda_{M/2})\in\{-1,+1\}^{M/2}\), define the “bump”
\[
\Delta(\lambda,x)=\sum_{i=1}^{M/2}\lambda_i\big(\mathbbm{1}\{x\in B_{2i-1}\}-\mathbbm{1}\{x\in B_{2i}\}\big).
\]
By construction, for every fixed \(\lambda\) we have
\[
\begin{aligned}
\int \Delta(\lambda,x)\,d\mu(x)
&=\sum_{i=1}^{M/2}\lambda_i\big(\mu(B_{2i-1})-\mu(B_{2i})\big)=0,\\
\int \Delta(\lambda,x)\,\big(2\hat m(x)-1\big)\,d\mu(x)
&=\sum_{i=1}^{M/2}\lambda_i\left(\int_{B_{2i-1}}w_2\,d\mu-\int_{B_{2i}}w_2\,d\mu\right)\\
&=0,
\end{aligned}
\]
and pointwise \(\Delta(\lambda,x)\in\{-1,+1\}\), hence \(\Delta(\lambda,x)^2=1\) \(\mu\)-a.e.

\paragraph{Defining the local alternatives.} We let
\begin{equation}
\label{eq:hypothesis-construction}
    \begin{aligned}
        g_{\lambda}(0,x) &= \hat{g}(0,x) + \eps_{n,g}\Delta(\lambda,x) (1-\hat{m}(x)+\eps_{n,m}\Delta(\lambda,x)) \\
        m_{\lambda}(x) &= \hat{m}(x) + \eps_{n,m}\Delta(\lambda,x) \\
        g_{\lambda}(1,x) &= \hat{g}(1,x) + \eps_{n,g}\Delta(\lambda,x) (\hat{m}(x)-\eps_{n,m}\Delta(\lambda,x)).
    \end{aligned}
\end{equation}

\begin{lemma}
    \label{lemma:mixture-dist-equal}
    Let $\pi$ be the uniform distribution over $\{-1,+1\}^{M/2}$. Then, for every $o\in\gX\times\{0,1\}^2$, we have
    \[
    \hat{p}(o)=\int p_{\lambda}(o)\dd\pi(\lambda).
    \]
\end{lemma}

\begin{proof}
Fix $x\in\gX$ and let $j(x)$ be the unique index such that $x\in B_{2j(x)-1}\cup B_{2j(x)}$. Then $\Delta(\lambda,x)=\pm\lambda_{j(x)}$, so under $\pi$ we have $\E_{\pi}[\Delta(\lambda,x)]=0$ and $\Delta(\lambda,x)^2=1$.

By definition, for $(d,y)\in\{0,1\}^2$ we have
\[
p_{\lambda}(x,d,y)=m_{\lambda}(x)^d(1-m_{\lambda}(x))^{1-d}\,g_{\lambda}(d,x)^y(1-g_{\lambda}(d,x))^{1-y},
\]
and similarly for $\hat{p}(x,d,y)$ using $\hat m,\hat g$. Using \eqref{eq:hypothesis-construction} and $\Delta(\lambda,x)^2=1$, each $p_{\lambda}(x,d,y)$ is affine in $\Delta(\lambda,x)$ with constant term $\hat{p}(x,d,y)$. For example,
\[
\begin{aligned}
p_{\lambda}(x,1,1)&=m_{\lambda}(x)g_{\lambda}(1,x)\\
&=\big(\hat m(x)+\eps_{n,m}\Delta(\lambda,x)\big)\Big(\hat g(1,x)+\eps_{n,g}\Delta(\lambda,x)\big(\hat m(x)-\eps_{n,m}\Delta(\lambda,x)\big)\Big)\\
&=\hat m(x)\hat g(1,x)+\Delta(\lambda,x)\Big(\eps_{n,m}\hat g(1,x)+\eps_{n,g}\hat m(x)^2-\eps_{n,g}\eps_{n,m}^2\Big),
\end{aligned}
\]
and the other three cases $(d,y)\in\{0,1\}^2$ are analogous. Therefore, taking $\E_{\pi}$ kills the $\Delta(\lambda,x)$ term pointwise in $x$ and yields $\int p_{\lambda}(x,d,y)\dd\pi(\lambda)=\hat{p}(x,d,y)$ for all $(d,y)$.
\end{proof}

The next lemma provides sufficient conditions for $\eps_{n,g}$ and $\eps_{n,m}$ such that all $m_{\lambda}(\cdot)$ and $g_{\lambda}(\cdot,\cdot)$ lie in the designated uncertainty set.

\begin{lemma}[Alternatives lie in the uncertainty set]
    \label{lemma:inside-uncertainty-set}
    If $\eps_{n,m},\eps_{n,g}\leq c$, then for all $x\in\gX$ and $d\in\{0,1\}$ we have $0\leq m_{\lambda}(x)\leq 1$ and $0\leq g_{\lambda}(d,x)\leq 1$, and
    \[
    \|g_{\lambda}(d,X)-\hat{g}(d,X)\|_{P_X,2}\leq\eps_{n,g},
    \qquad
    \|m_{\lambda}(X)-\hat{m}(X)\|_{P_X,2}\leq\eps_{n,m}.
    \]
\end{lemma}

\begin{proof}
Since $c\leq\hat{m}(x) \leq 1-c$ and $|\eps_{n,m}\Delta(\lambda,x)|\leq c$ by assumption, we have $0\leq m_{\lambda}(x)\leq 1$ for all $x$. Moreover, since $\Delta(\lambda,X)^2=1$ $P_X$-a.s.,
\[
\|\hat{m}(X)-m_{\lambda}(X)\|_{P_X,2}=\|\eps_{n,m}\Delta(\lambda,X)\|_{P_X,2}=\eps_{n,m}.
\]
Similarly, $0\leq \hat{m}(x)-\eps_{n,m}\Delta(\lambda,x)\leq 1$ and $0\leq 1-\hat{m}(x)+\eps_{n,m}\Delta(\lambda,x)\leq 1$. Plugging into \eqref{eq:hypothesis-construction}, we have for $d\in\{0,1\}$ that
\[
|g_{\lambda}(d,x)-\hat{g}(d,x)|\leq \eps_{n,g},\qquad \forall x\in\gX,
\]
and therefore $\|g_{\lambda}(d,X)-\hat{g}(d,X)\|_{P_X,2}\leq \eps_{n,g}$. Since $c\leq\hat{g}(d,x)\leq 1-c$ and $\eps_{n,g}\leq c$, we also have $0\leq g_{\lambda}(d,x)\leq 1$ for all $x$ and $d\in\{0,1\}$.
\end{proof}

Lemma \ref{lemma:mixture-dist-equal} allows us to apply Theorem \ref{semi-param-thm} to derive a Hellinger distance bound.

\begin{lemma}[Hellinger bound for ATE mixtures]
    \label{lemma:hellinger-bound-ate}
    For any $\delta>0$, if $M\geq\max\{n, (2Cn^2)/(c^4\delta)\}$ where $C$ is the constant in Theorem \ref{semi-param-thm} with $A=2c^{-2}$, then we have
    \begin{equation}
        H^2(\hat{P}^{\otimes n}, \int P_{\lambda}^{\otimes n}\dd\pi(\lambda))\leq\delta.
    \end{equation}
\end{lemma}

\begin{proof}
    We apply Theorem \ref{semi-param-thm} to the partition $\gX_j=(B_{2j-1}\cup B_{2j})\times\{0,1\}^2, j=1,2,\cdots,M/2$ of $\gX\times\{0,1\}^2$, $P=\hat{P}$ and $Q_{\lambda}=P_{\lambda}$ as constructed above. Since $\mu_{\gX}(B_j)=1/M$, we have $p_j=2/M$ and
    \begin{equation}
        \begin{aligned}
            b &= \frac{M}{2}\max_j\sup_{\lambda}\int_{\gX_j}\frac{(p_{\lambda}-\hat{p})^2}{\hat{p}}\dd\mu \\
            &\leq \frac{M}{2}\frac{2}{M}\sup_{o=(x,d,y)\in\gX_j}\frac{(p_{\lambda}(o)-\hat{p}(o))^2}{\hat{p}(o)}
            &\leq c^{-2},
        \end{aligned}
    \end{equation}
    where the last step follows from $\hat{p}(o)\geq c^2$. Hence we have $Cn^2\max_j p_j\cdot b^2 \leq (2Cn^2)/(c^4M)\leq\delta$, concluding the proof.
\[
n\max\{1,b\}\max_j p_j \leq n\cdot c^{-2}\cdot \frac{2}{M} \leq 2c^{-2}=A,
\]
so all conditions of Theorem \ref{semi-param-thm} are satisfied.
\end{proof}

Next we quantify the separation in ATE. Here the balanced pairing with respect to \(2\hat m-1\) is exactly what removes the linear term in \(\Delta\).

\begin{lemma}[ATE separation]
    \label{lemma:ate-separation}
    Let \(\theta_{\lambda}\) be the ATE of \(P_{\lambda}\). Then \(\theta_{\lambda}= \hat{\theta}-2\eps_{n,m}\eps_{n,g}\).
\end{lemma}

\begin{proof}
By definition,
\[
\theta_\lambda-\hat\theta=\E\big[g_\lambda(1,X)-g_\lambda(0,X)-(\hat g(1,X)-\hat g(0,X))\big].
\]
From \eqref{eq:hypothesis-construction},
\[
g_\lambda(1,X)-g_\lambda(0,X)
=\hat g(1,X)-\hat g(0,X)\;+\;\eps_{n,g}\Delta(\lambda,X)\big(2\hat m(X)-1-2\eps_{n,m}\Delta(\lambda,X)\big).
\]
Taking \(\E_{\hat P}\) and using the pairing properties,
\(\E[\Delta(\lambda,X)]=0\) and \(\E[\Delta(\lambda,X)\,(2\hat m(X)-1)]=0\), while \(\Delta(\lambda,X)^2=1\) \(P_X\)-a.s. Hence
\[
\theta_\lambda-\hat\theta
=\E\left[\eps_{n,g}\Delta(\lambda,X)\big(2\hat m(X)-1\big)\right]
-2\eps_{n,m}\eps_{n,g}\E\left[\Delta(\lambda,X)^2\right]
=-2\eps_{n,m}\eps_{n,g}.
\]
\end{proof}

\textit{Concluding the proof (the $\eps_{n,m}\eps_{n,g}$ term).} With Lemmas \ref{lemma:mixture-dist-equal}--\ref{lemma:hellinger-bound-ate}--\ref{lemma:ate-separation} in place, the proof of Theorem \ref{thm:ate-only} is standard. For any \(\gamma>1/2\), pick \(\delta\in(0,2)\) with \(\gamma=\big(1+\sqrt{\delta(1-\delta/4)}\big)/2\) and choose \(M=2^r\) large enough to satisfy Lemma \ref{lemma:hellinger-bound-ate}. Define the functional $T(P):=-\theta^{\ate}(P)$, set $c:=-\hat{\theta}$ and $s:=\eps_{n,m}\eps_{n,g}$. Then $T(\hat P)=c$ and, by Lemma \ref{lemma:ate-separation}, for all $\lambda$ we have $T(P_\lambda)=c+2s$. Therefore, the separation condition \eqref{fano:separation-condition} holds. Applying Theorem \ref{fano-method} yields the desired lower bound $\Omega(\eps_{n,m}\eps_{n,g})$.

\paragraph{The $n^{-1/2}$ term.} We briefly explain the $\min\{\eps_{n,g},n^{-1/2}\}$ component in Theorem \ref{thm:ate-only}. Fix $m=\hat m$ and consider the two-point subfamily with
\[
g_{\pm}(1,x):=\hat g(1,x)\pm \delta_n,\qquad g_{\pm}(0,x):=\hat g(0,x)\mp \delta_n,
\]
where $\delta_n:=\min\{t/\sqrt{n},\eps_{n,g}\}$ for a small constant $t\in(0,c/2]$. For all large $n$, both alternatives lie in $\gM_1(\hat{P};\eps_{n,m},\eps_{n,g})$ and satisfy $0\leq g_{\pm}(d,x)\leq 1$ by Assumption \ref{asmp:estimator-bounded}. Moreover, they are separated by
\(
|\theta_+-\theta_-|=4\delta_n.
\)
A standard two-point (Le Cam) argument for estimating a Bernoulli mean (applied to $Y\mid(D,X)$ under this subfamily) yields a minimax quantile risk of order $\delta_n=\min\{\eps_{n,g},n^{-1/2}\}$, giving the second term.

\section{Directional derivatives of the functional \texorpdfstring{$\chi(P)$}{chi(P)}}

\begin{proposition}[Derivative formulas for $\chi$ under covariate shift]
\label{prop:compute-2nd-order-differential}

Suppose Assumptions~5.1--5.5 hold.
Fix an anchor pair $(\hat P,\hat Q)$ and recall the covariate shift functional
\[
\chi(P,Q)=\E_{Z\sim Q}\left[m_1\left(Z,\gamma(Z;P)\right)\right],
\]
where for each fixed $z$ the map $a\mapsto m_1(z,a)$ is linear in $a$.
Write $\hat\gamma(\cdot):=\gamma(\cdot;\hat P)$.
All appearances of $\gamma_P'(\cdot;\hat P)$ and $\gamma_P''(\cdot;\hat P)$ below are understood on the set
$\{z:\hat p_Z(z)>0\}$; in particular, the expectations in
\eqref{eq:chi-first-derivative-pair}--\eqref{eq:chi-second-derivative-square-pair} are well-defined whenever
$\hat Q(\hat p_Z(Z)>0)=1$ and all target perturbations $K$ considered are supported on $\{z:\hat p_Z(z)>0\}$ (e.g.\ under an overlap/density-boundedness assumption on the anchor pair and perturbations).
We use the shorthand $\E_{K}[f(Z)]:=\int f(z)\,\dd K(z)$ for finite signed measures $K$ on $\gZ$.
For a \emph{joint perturbation pair} $(H,K)$ consisting of a training perturbation $H$ on $\gO$ and a target
perturbation $K$ on $\gZ$, define the first and mixed second directional (G\^ateaux) derivatives at $(\hat P,\hat Q)$ by
\[
\chi_{(P,Q)}'(\hat P,\hat Q)[H,K]
:=
\left.\frac{\partial}{\partial t}\right|_{t=0}\chi(\hat P+tH,\hat Q+tK)
\]
and
\[
\begin{aligned}
\chi''(\hat P,\hat Q)[(H_0,K_0),(H_1,K_1)]
&:=
\left.\frac{\partial^2}{\partial t\,\partial s}\right|_{t=s=0}
\chi(\hat P+tH_0+sH_1,\\
&\qquad \hat Q+tK_0+sK_1),\\
\chi''(\hat P,\hat Q)[(H,K)]
&:= \chi''(\hat P,\hat Q)[(H,K),(H,K)].
\end{aligned}
\]
Then:
\begin{enumerate}
\item For any joint perturbation pair $(H,K)$,
\begin{equation}
\label{eq:chi-first-derivative-pair}
\chi_{(P,Q)}'(\hat P,\hat Q)[H,K]
=
\E_{\hat Q}\left[m_1\left(Z,\gamma_P'(Z;\hat P)[H]\right)\right]
\;+
\E_{K}\left[m_1\left(Z,\hat\gamma(Z)\right)\right].
\end{equation}

\item For any joint perturbation pairs $(H_0,K_0)$ and $(H_1,K_1)$,
\begin{equation}
\label{eq:chi-second-derivative-pair}
\begin{aligned}
\chi''(\hat P,\hat Q)[(H_0,K_0),(H_1,K_1)]
&=
\E_{\hat Q}\left[m_1\left(Z,\gamma_P''(Z;\hat P)[H_0,H_1]\right)\right]\\
&\quad+
\E_{K_0}\left[m_1\left(Z,\gamma_P'(Z;\hat P)[H_1]\right)\right]\\
&\quad+
\E_{K_1}\left[m_1\left(Z,\gamma_P'(Z;\hat P)[H_0]\right)\right].
\end{aligned}
\end{equation}
In particular,
\begin{equation}
\label{eq:chi-second-derivative-square-pair}
\chi''(\hat P,\hat Q)[(H,K)]
=
\E_{\hat Q}\left[m_1\left(Z,\gamma_P''(Z;\hat P)[H,H]\right)\right]
\;+
2\,\E_{K}\left[m_1\left(Z,\gamma_P'(Z;\hat P)[H]\right)\right].
\end{equation}

\item If $\gamma(\cdot;\hat P+tG_0)=\gamma(\cdot;\hat P)$ for all $|t|\le c_t$, then
\begin{equation}
\label{eq:chi-second-derivative-invariant-direction}
\chi''(\hat P,\hat Q)[(G_0,K_0)]=0\qquad\text{for every target perturbation $K_0$.}
\end{equation}

\end{enumerate}
\end{proposition}

\begin{proof}
We prove the derivative identities in the order they are stated.

\paragraph{Derivatives of $\chi$ along joint perturbations (proof of \eqref{eq:chi-first-derivative-pair} and \eqref{eq:chi-second-derivative-pair}).}
Let $(H,K)$ be a joint perturbation of $(\hat P,\hat Q)$ and consider the one-dimensional path
$(P_t,Q_t):=(\hat P+tH,\hat Q+tK)$ for $t$ in a neighborhood of $0$ on which $P_t$ and $Q_t$ are probability measures.
Write $q_t:=\dd Q_t/\dd\mu_Z=\hat q+t k$ where $k:=\dd K/\dd\mu_Z$.
Write $\gamma_t:=\gamma(\cdot;P_t)$ and define the remainder
\[
r_t := \gamma_t-\hat\gamma - t\,\gamma_P'(\cdot;\hat P)[H].
\]
By Assumption~\ref{asmp:regularity}(2), $\|r_t\|_{\hat P_Z,2}=o(t)$ as $t\to 0$.
Since $h\mapsto \E_{\hat Q}[m_1(Z,h)]$ is a bounded linear functional on $L^2(\hat P_Z)$ with Riesz representer
$\nu_m(\cdot;\hat P,\hat Q)$ (cf.\ \eqref{eq:riesz-representer}), we have
\[
\big|\E_{\hat Q}[m_1(Z,r_t(Z))]\big|
=
\big|\E_{\hat P}[r_t(Z)\,\nu_m(Z;\hat P,\hat Q)]\big|
\le \|r_t\|_{\hat P_Z,2}\,\|\nu_m(\cdot;\hat P,\hat Q)\|_{\hat P_Z,2}
=o(t).
\]
Using $Q_t=\hat Q+tK$ and linearity of $m_1$ in its second argument,
\begin{align*}
\chi(P_t,Q_t)
&=\E_{\hat Q+tK}\Big[m_1\big(Z,\hat\gamma+t\,\gamma_P'(\cdot;\hat P)[H]+r_t\big)\Big]\\
&=\chi(\hat P,\hat Q)
\;+\;
t\,\E_{\hat Q}\left[m_1\left(Z,\gamma_P'(Z;\hat P)[H]\right)\right]
\;+\;
t\,\E_K\left[m_1\left(Z,\hat\gamma(Z)\right)\right]
\;+\;
\E_{\hat Q}\left[m_1\left(Z,r_t\right)\right]
\;+\;
o(t),
\end{align*}
which yields \eqref{eq:chi-first-derivative-pair}.

For the mixed second derivative, fix two joint perturbation pairs $(H_0,K_0)$ and $(H_1,K_1)$ and consider the
two-parameter path
$(P_{t,s},Q_{t,s}):=(\hat P+tH_0+sH_1,\hat Q+tK_0+sK_1)$.
Write $q_{t,s}=\hat q+t k_0+s k_1$.
Write $\gamma_{t,s}(z):=\gamma(z;P_{t,s})$.
Differentiate the identity
\[
\chi(P_{t,s},Q_{t,s})
=\int m_1\bigl(z,\gamma_{t,s}(z)\bigr)\,q_{t,s}(z)\,\dd\mu_Z(z),
\qquad
q_{t,s}(z)=\hat q(z)+t k_0(z)+s k_1(z),
\]
with respect to $(t,s)$ at $(0,0)$.
Since $q_{t,s}$ is affine in $(t,s)$, it has no mixed $ts$ term, and by linearity of $m_1$ we have
\begin{align*}
\partial_t\,m_1\bigl(z,\gamma_{t,0}(z)\bigr)\big|_{t=0} &= m_1\bigl(z,\gamma_P'(z;\hat P)[H_0]\bigr),\\
\partial_s\,m_1\bigl(z,\gamma_{0,s}(z)\bigr)\big|_{s=0} &= m_1\bigl(z,\gamma_P'(z;\hat P)[H_1]\bigr),\\
\partial_{ts}\,m_1\bigl(z,\gamma_{t,s}(z)\bigr)\big|_{(0,0)} &= m_1\bigl(z,\gamma_P''(z;\hat P)[H_0,H_1]\bigr).
\end{align*}
Therefore,
\[
\begin{aligned}
\chi''(\hat P,\hat Q)[(H_0,K_0),(H_1,K_1)]
&=
\E_{\hat Q}\left[m_1\left(Z,\gamma_P''(Z;\hat P)[H_0,H_1]\right)\right]\\
&\quad+
\E_{K_0}\left[m_1\left(Z,\gamma_P'(Z;\hat P)[H_1]\right)\right]\\
&\quad+
\E_{K_1}\left[m_1\left(Z,\gamma_P'(Z;\hat P)[H_0]\right)\right],
\end{aligned}
\]
which is \eqref{eq:chi-second-derivative-pair}. Equation \eqref{eq:chi-second-derivative-square-pair} follows by
specializing $(H_0,K_0)=(H_1,K_1)=(H,K)$.

\paragraph{Invariance implication (proof of \eqref{eq:chi-second-derivative-invariant-direction}).}
If $\gamma(\cdot;\hat P+tG_0)$ is constant for $|t|\le c_t$, then the map $t\mapsto \gamma(\cdot;\hat P+tG_0)$ is
identically constant in a neighborhood of $0$, hence its first and second derivatives at $t=0$ vanish:
$\gamma_P'(\cdot;\hat P)[G_0]\equiv 0$ and $\gamma_P''(\cdot;\hat P)[G_0,G_0]\equiv 0$.
Substituting into \eqref{eq:chi-second-derivative-square-pair} yields
$\chi''(\hat P,\hat Q)[(G_0,K_0)]=0$ for every $K_0$, proving \eqref{eq:chi-second-derivative-invariant-direction}.

\end{proof}

\section{Proofs of Theorems~\ref{thm:main-mixed-bias} and \ref{thm:main}}
\label{sec:proof-main-thms}
\label{subsec:main-proof}

Throughout this section we work on the anchored class
$\gM((\hat P,\hat Q);\eps_{N,\gamma},\eps_{N,\alpha})$ defined in \eqref{eq:candidate-distribution}.
We write
\[
\hat\gamma(\cdot):=\gamma(\cdot;\hat P),
\qquad
\hat\alpha(\cdot):=\alpha(\cdot;\hat P,\hat Q),
\qquad
\hat p:=d\hat P/d\mu,
\qquad
\hat q:=d\hat Q/d\mu_Z.
\]
All derivatives of $\gamma$ that appear below are directional (G\^ateaux) derivatives with respect to the \emph{training} law $P$,
as in Assumption~\ref{asmp:regularity}.
Derivatives of $\alpha$ and of the target functional $\chi(P,Q)$ are taken along \emph{feasible joint perturbations}
of $(P,Q)$ (cf.~Assumptions~\ref{asmp:regularity}--\ref{asmp:main}).
Throughout, a ``directional derivative'' is meant in the G\^ateaux sense (hence linear in the direction argument), and a
``mixed second directional derivative'' is bilinear in its direction arguments.

We observe an i.i.d.\ training sample $(O_t)_{t=1}^N\sim P$ and an independent i.i.d.\ target sample
$(Z_i)_{i=1}^N\sim Q$.
In the lower bound constructions below we perturb $P$ and $Q$ \emph{jointly} according to the feasible perturbation
pairs provided by Assumption~\ref{asmp:main}. Consequently, we work with the full experiment
$P^{\otimes N}\otimes Q^{\otimes N}$, and we must control divergences between \emph{joint} product measures.

We split the argument into two cases depending on the relative sizes of $\eps_{N,\gamma}$ and $\eps_{N,\alpha}$.
Case~1 yields the mixed-bias (product) separation $\Omega(\eps_{N,\gamma}\eps_{N,\alpha})$.
Case~2 handles the regime $\eps_{N,\alpha}<\eps_{N,\gamma}$:
for Theorem~\ref{thm:main-mixed-bias} (which assumes the mixed-bias property) we obtain the product rate by swapping
the roles of $\gamma$ and $\alpha$, whereas for Theorem~\ref{thm:main} (non-affine $\rho$) we obtain the larger
quadratic separation $\Omega(\eps_{N,\gamma}^2)$.

\subsubsection{Auxiliary constructions}

\paragraph{Bumps and bumped perturbations.}
Let $m\ge 1$ and set $M:=2m$.
Given a partition $\{\mathcal{X}_j\}_{j=1}^M$ of $\gZ_1$ and a vector $\lambda\in\{-1,1\}^m$, define the paired bump
\[
\Delta(\lambda,z_1)
:=\sum_{i=1}^m \lambda_i\Big(\mathbf{1}\{z_1\in \mathcal{X}_{2i-1}\}-\mathbf{1}\{z_1\in \mathcal{X}_{2i}\}\Big),
\qquad z_1\in\gZ_1.
\]
Note that $\Delta(\lambda,z_1)\in\{-1,1\}$ and $\Delta(\lambda,z_1)^2\equiv 1$.
Given a training perturbation $G$ with density $g=\dd G/\dd\mu$, define its bumped version $G_\lambda$ by
$\dd G_\lambda/\dd\mu := g_\lambda$, where $g_\lambda(o):=\Delta(\lambda,z_1)\,g(o)$.
Given a target perturbation $K$ with density $k=\dd K/\dd\mu_Z$, define its bumped version $K_\lambda$ by
$\dd K_\lambda/\dd\mu_Z := k_\lambda$, where $k_\lambda(z):=\Delta(\lambda,z_1)\,k(z)$.
In coupled model classes $\gP_0$, multiplying a feasible direction by a bounded $Z_1$-measurable factor need not
preserve feasibility.
Accordingly, our lower bound constructions only apply bumping to perturbation pairs that are $Z_1$-modulation closed
(Definition~\ref{def:z1-modulation-closure}); combined with the centering equalities enforced by the ham-sandwich
partition, this guarantees that the bumped pairs remain feasible joint perturbations.

\begin{lemma}[Effect of bumping on slice-based maps]
\label{lemma:bumped-slice}
Suppose $\eta$ satisfies the slice dependence in Assumption~\ref{asmp:cond-prob-functional}(1), i.e.\ there exists a
map $\Gamma_\eta$ such that
\[
\eta(z_1,z_2;P,Q)=\Gamma_\eta(p_{z_1},q_{z_1})(z_2),
\]
with the understanding that $\Gamma_\eta$ may ignore its second argument for objects (like $\gamma$) that only
depend on $P$.
Let $(G,K)$ be a feasible joint perturbation at $(\hat P,\hat Q)$ with densities $g=\dd G/\dd\mu$ and
$k=\dd K/\dd\mu_Z$, and define $(G_\lambda,K_\lambda)$ as above.
Then for all $t$ such that $(\hat P+tG_\lambda,\hat Q+tK_\lambda)$ is well-defined,
\[
(\hat p+t g_\lambda)_{z_1}
~=~
\hat p_{z_1}+t\,\Delta(\lambda,z_1)\,g_{z_1},
\qquad
(\hat q+t k_\lambda)_{z_1}
~=~
\hat q_{z_1}+t\,\Delta(\lambda,z_1)\,k_{z_1}.
\]
In particular:
\begin{enumerate}
\item If $\eta(\cdot;\hat P+tG,\hat Q+tK)=\eta(\cdot;\hat P,\hat Q)$ for all $|t|\le c_t$, then also
$\eta(\cdot;\hat P+tG_\lambda,\hat Q+tK_\lambda)=\eta(\cdot;\hat P,\hat Q)$ for all $|t|\le c_t$.
\item If $\eta$ is directionally (G\^ateaux) differentiable at $(\hat P,\hat Q)$ along feasible joint perturbations, with derivative denoted by
$\eta_{(P,Q)}'$, then
\[
\eta_{(P,Q)}'(\cdot;\hat P,\hat Q)[G_\lambda,K_\lambda]
~=~
\Delta(\lambda,\cdot)\,\eta_{(P,Q)}'(\cdot;\hat P,\hat Q)[G,K],
\]
where $\Delta(\lambda,\cdot)$ is understood as a function of $z=(z_1,z_2)$ via its dependence on $z_1$.
For objects depending only on $P$ (such as $\gamma$), the identity specializes to
$\gamma_P'(\cdot;\hat P)[G_\lambda]=\Delta(\lambda,\cdot)\,\gamma_P'(\cdot;\hat P)[G]$.
\item If $\eta$ is twice directionally (G\^ateaux) differentiable at $(\hat P,\hat Q)$ along feasible joint perturbations, with second
derivative $\eta_{(P,Q)}''(\cdot;\hat P,\hat Q)$, then for any two feasible perturbations
$(G_0,K_0)$ and $(G_1,K_1)$,
\[
\eta_{(P,Q)}''(\cdot;\hat P,\hat Q)[G_{0,\lambda},K_{0,\lambda};\,G_{1,\lambda},K_{1,\lambda}]
=
\Delta(\lambda,\cdot)^2\,\eta_{(P,Q)}''(\cdot;\hat P,\hat Q)[G_0,K_0;\,G_1,K_1].
\]
For objects depending only on $P$ (such as $\gamma$), this specializes to
\[
\gamma_P''(\cdot;\hat P)[G_{0,\lambda},G_{1,\lambda}]
=\Delta(\lambda,\cdot)^2\,\gamma_P''(\cdot;\hat P)[G_0,G_1].
\]
\end{enumerate}
\end{lemma}

\begin{proof}
Fix $\lambda\in\{-1,1\}^m$ and write $\psi_\lambda(z_1):=\Delta(\lambda,z_1)\in\{-1,1\}$.
By definition of bumping,
\[
g_\lambda(o)=\psi_\lambda(z_1)\,g(o),
\qquad
k_\lambda(z)=\psi_\lambda(z_1)\,k(z).
\]
Therefore, for every $z_1\in\gZ_1$ and every $t$ for which the perturbed densities are well-defined,
taking $z_1$-slices yields
\[
(\hat p+t g_\lambda)_{z_1}
=\hat p_{z_1}+t\,\psi_\lambda(z_1)\,g_{z_1}
=\hat p_{z_1}+t\,\Delta(\lambda,z_1)\,g_{z_1},
\qquad
(\hat q+t k_\lambda)_{z_1}
=\hat q_{z_1}+t\,\psi_\lambda(z_1)\,k_{z_1}
=\hat q_{z_1}+t\,\Delta(\lambda,z_1)\,k_{z_1},
\]
which proves the first display.

\medskip
Next we record the basic $Z_1$-modulation identity that underlies Lemma~B.1 and Corollary~B.2.
See Lemma~\ref{lemma:bump-derivative} and Corollary~\ref{cor:bump-invariance}.
Fix $z=(z_1,z_2)\in\gZ$ and let $t$ be such that $(\hat P+tG_\lambda,\hat Q+tK_\lambda)$ is well-defined.
Define
\[
s:=t\,\psi_\lambda(z_1)=t\,\Delta(\lambda,z_1),
\qquad\text{so that}\qquad |s|=|t|.
\]
Using the slice representation $\eta(z;P,Q)=\Gamma_\eta(p_{z_1},q_{z_1})(z_2)$ together with the slice identities above,
we obtain
\begin{align*}
\eta(z;\hat P+tG_\lambda,\hat Q+tK_\lambda)
&=\Gamma_\eta\big((\hat p+t g_\lambda)_{z_1},\,(\hat q+t k_\lambda)_{z_1}\big)(z_2)\\
&=\Gamma_\eta\big(\hat p_{z_1}+t\psi_\lambda(z_1)g_{z_1},\,\hat q_{z_1}+t\psi_\lambda(z_1)k_{z_1}\big)(z_2)\\
&=\Gamma_\eta\big(\hat p_{z_1}+s g_{z_1},\,\hat q_{z_1}+s k_{z_1}\big)(z_2)\\
&=\Gamma_\eta\big((\hat p+s g)_{z_1},\,(\hat q+s k)_{z_1}\big)(z_2)\\
&=\eta(z;\hat P+sG,\hat Q+sK).
\end{align*}
We will use this identity repeatedly.

\medskip
\noindent\textbf{(1) Invariance.}
Assume $\eta(\cdot;\hat P+tG,\hat Q+tK)=\eta(\cdot;\hat P,\hat Q)$ for all $|t|\le c_t$.
Fix $z=(z_1,z_2)\in\gZ$ and fix any $t$ with $|t|\le c_t$ such that $(\hat P+tG_\lambda,\hat Q+tK_\lambda)$
is well-defined. Set $s=t\psi_\lambda(z_1)$, so $|s|=|t|\le c_t$.
By the assumed invariance along $(G,K)$,
\[
\eta(z;\hat P+sG,\hat Q+sK)=\eta(z;\hat P,\hat Q).
\]
Applying the modulation identity above yields
\[
\eta(z;\hat P+tG_\lambda,\hat Q+tK_\lambda)=\eta(z;\hat P,\hat Q).
\]
Since $z$ was arbitrary, this proves $\eta(\cdot;\hat P+tG_\lambda,\hat Q+tK_\lambda)=\eta(\cdot;\hat P,\hat Q)$
for all $|t|\le c_t$ (whenever the bumped segment is well-defined), exactly as in Corollary~B.2.

\medskip
\noindent\textbf{(2) First-derivative scaling.}
Assume $\eta$ is directionally (G\^ateaux) differentiable at $(\hat P,\hat Q)$ along feasible joint perturbations, with derivative
$\eta'_{(P,Q)}(\cdot;\hat P,\hat Q)$.
Fix $z=(z_1,z_2)\in\gZ$ and abbreviate $\psi:=\psi_\lambda(z_1)=\Delta(\lambda,z_1)\in\{-1,1\}$.
By definition of the directional derivative and the modulation identity established above,
\begin{align*}
\eta'_{(P,Q)}(z;\hat P,\hat Q)[G_\lambda,K_\lambda]
&=\lim_{t\to 0}\frac{\eta(z;\hat P+tG_\lambda,\hat Q+tK_\lambda)-\eta(z;\hat P,\hat Q)}{t}\\
&=\lim_{t\to 0}\frac{\eta(z;\hat P+t\psi\,G,\hat Q+t\psi\,K)-\eta(z;\hat P,\hat Q)}{t}.
\end{align*}
Let $s=t\psi$. Then $t=s/\psi$ and $t\to 0$ iff $s\to 0$, hence
\begin{align*}
\eta'_{(P,Q)}(z;\hat P,\hat Q)[G_\lambda,K_\lambda]
&=\lim_{s\to 0}\frac{\eta(z;\hat P+sG,\hat Q+sK)-\eta(z;\hat P,\hat Q)}{s/\psi}\\
&=\psi\lim_{s\to 0}\frac{\eta(z;\hat P+sG,\hat Q+sK)-\eta(z;\hat P,\hat Q)}{s}\\
&=\psi\,\eta'_{(P,Q)}(z;\hat P,\hat Q)[G,K]\\
&=\Delta(\lambda,z_1)\,\eta'_{(P,Q)}(z;\hat P,\hat Q)[G,K].
\end{align*}
Viewing $\Delta(\lambda,z_1)$ as the function $\Delta(\lambda,\cdot)$ of $z=(z_1,z_2)$ through its dependence on $z_1$
gives the asserted identity
\(
\eta'_{(P,Q)}(\cdot;\hat P,\hat Q)[G_\lambda,K_\lambda]
=
\Delta(\lambda,\cdot)\,\eta'_{(P,Q)}(\cdot;\hat P,\hat Q)[G,K].
\)
If $\eta$ depends only on $P$ (e.g.\ $\eta=\gamma$), the same argument applies with $K\equiv 0$, yielding
$\gamma_P'(\cdot;\hat P)[G_\lambda]=\Delta(\lambda,\cdot)\,\gamma_P'(\cdot;\hat P)[G]$.

\medskip
\noindent\textbf{(3) Second-derivative scaling.}
Assume $\eta$ is twice directionally (G\^ateaux) differentiable at $(\hat P,\hat Q)$ along feasible joint perturbations, with
second derivative $\eta''_{(P,Q)}(\cdot;\hat P,\hat Q)$.
Fix $z=(z_1,z_2)\in\gZ$ and set $\psi:=\psi_\lambda(z_1)=\Delta(\lambda,z_1)$.
Fix two feasible perturbations $(G_0,K_0)$ and $(G_1,K_1)$.
For $t$ sufficiently close to $0$, all the pairs below are well-defined by feasibility.

We will use the standard second-difference characterization of the bilinear mixed second directional derivative:
for any two direction pairs $(H_0,L_0)$ and $(H_1,L_1)$,
\begin{equation}\label{eq:second-difference-fr}
\begin{aligned}
\eta''_{(P,Q)}(z;\hat P,\hat Q)[H_0,L_0;\,H_1,L_1]
&=
\lim_{t\to 0}
\frac{1}{t^2}\Big[
\eta\big(z;\hat P+t(H_0+H_1),\,\hat Q+t(L_0+L_1)\big)
-\eta\big(z;\hat P+tH_0,\,\hat Q+tL_0\big)\\
&\qquad
-\eta\big(z;\hat P+tH_1,\,\hat Q+tL_1\big)
+\eta(z;\hat P,\hat Q)
\Big].
\end{aligned}
\end{equation}
For completeness, \eqref{eq:second-difference-fr} follows by applying the second-order directional expansion
to the three perturbations $(H_0+H_1,L_0+L_1)$, $(H_0,L_0)$, and $(H_1,L_1)$, and then subtracting:
the $O(t)$ terms cancel by linearity of $\eta'_{(P,Q)}$, while the $O(t^2)$ terms reduce to
$t^2\,\eta''_{(P,Q)}(z;\hat P,\hat Q)[H_0,L_0;\,H_1,L_1]$ by bilinearity of $\eta''_{(P,Q)}$, leaving an $o(t^2)$ remainder.

Apply \eqref{eq:second-difference-fr} with $(H_j,L_j)=(G_{j,\lambda},K_{j,\lambda})$, $j\in\{0,1\}$.
Using linearity of bumping, $G_{0,\lambda}+G_{1,\lambda}=(G_0+G_1)_\lambda$ and $K_{0,\lambda}+K_{1,\lambda}=(K_0+K_1)_\lambda$.
Then, applying the modulation identity to each of the three perturbed terms gives, with $s=t\psi$,
\begin{align*}
&\eta\big(z;\hat P+t(G_{0,\lambda}+G_{1,\lambda}),\,\hat Q+t(K_{0,\lambda}+K_{1,\lambda})\big)
=\eta\big(z;\hat P+s(G_0+G_1),\,\hat Q+s(K_0+K_1)\big),\\
&\eta\big(z;\hat P+tG_{0,\lambda},\,\hat Q+tK_{0,\lambda}\big)
=\eta\big(z;\hat P+sG_0,\,\hat Q+sK_0\big),\\
&\eta\big(z;\hat P+tG_{1,\lambda},\,\hat Q+tK_{1,\lambda}\big)
=\eta\big(z;\hat P+sG_1,\,\hat Q+sK_1\big).
\end{align*}
Substituting these identities into \eqref{eq:second-difference-fr} yields
\begin{align*}
&\eta''_{(P,Q)}(z;\hat P,\hat Q)[G_{0,\lambda},K_{0,\lambda};\,G_{1,\lambda},K_{1,\lambda}]\\
&\qquad=
\lim_{t\to 0}\frac{1}{t^2}\left[
\begin{aligned}[t]
&\eta\big(z;\hat P+s(G_0+G_1),\,\hat Q+s(K_0+K_1)\big)
-\eta\big(z;\hat P+sG_0,\,\hat Q+sK_0\big)\\
&\quad
-\eta\big(z;\hat P+sG_1,\,\hat Q+sK_1\big)
+\eta(z;\hat P,\hat Q)
\end{aligned}
\right].
\end{align*}
Since $s=t\psi$, we have $t^2=s^2/\psi^2$, and $t\to 0$ iff $s\to 0$. Therefore
\begin{align*}
&\eta''_{(P,Q)}(z;\hat P,\hat Q)[G_{0,\lambda},K_{0,\lambda};\,G_{1,\lambda},K_{1,\lambda}]\\
&\qquad=
\psi^2
\lim_{s\to 0}\frac{1}{s^2}\left[
\begin{aligned}[t]
&\eta\big(z;\hat P+s(G_0+G_1),\,\hat Q+s(K_0+K_1)\big)
-\eta\big(z;\hat P+sG_0,\,\hat Q+sK_0\big)\\
&\quad
-\eta\big(z;\hat P+sG_1,\,\hat Q+sK_1\big)
+\eta(z;\hat P,\hat Q)
\end{aligned}
\right]\\
&\qquad=
\psi^2\,\eta''_{(P,Q)}(z;\hat P,\hat Q)[G_0,K_0;\,G_1,K_1]\\
&\qquad=
\Delta(\lambda,z_1)^2\,\eta''_{(P,Q)}(z;\hat P,\hat Q)[G_0,K_0;\,G_1,K_1].
\end{align*}
Interpreting $\Delta(\lambda,z_1)^2$ as $\Delta(\lambda,\cdot)^2$ via $z=(z_1,z_2)$ gives the asserted function identity.
The specialization to objects depending only on $P$ (such as $\gamma$) follows by taking $K_0\equiv K_1\equiv 0$.
\end{proof}

\paragraph{Ham-sandwich partition.}
We need partitions of $\gZ_1$ that equalize integrals of finitely many integrable functions under both the training
base measure $\mu$ and the target base measure $\mu_Z$.

\begin{corollary}[Ham-sandwich partition for training and target weights]
\label{cor:ham-sandwich}
Let $M=2^r$ for some $r\ge 1$.
Let $\{\psi_\ell\}_{\ell=1}^L$ be $\mu$-integrable functions on $\gO$ and let $\{\varphi_k\}_{k=1}^K$ be
$\mu_Z$-integrable functions on $\gZ$.
Under Assumption~\ref{asmp:cond-space-nontrivial} and $L+K\le 9$, there exists a partition
$\{\mathcal{X}_j\}_{j=1}^M$ of $\gZ_1$ such that for every $j\in[M]$,
\begin{align}
\int \psi_\ell(o)\,\mathbf{1}\{z_1\in \mathcal{X}_j\}\,d\mu(o)
&=\frac{1}{M}\int \psi_\ell(o)\,d\mu(o),
\qquad \forall~\ell\in[L],
\label{eq:hs-training}\\
\int \varphi_k(z)\,\mathbf{1}\{z_1\in \mathcal{X}_j\}\,d\mu_Z(z)
&=\frac{1}{M}\int \varphi_k(z)\,d\mu_Z(z),
\qquad \forall~k\in[K].
\label{eq:hs-target}
\end{align}
\end{corollary}

\begin{proof}
For each $\psi_\ell$ define the induced weight on $\gZ_1$:
$w_{\psi_\ell}(z_1):=\int \psi_\ell(z_1,z_2,w)\,d(\mu_{Z_2}\otimes\mu_W)(z_2,w)$.
For each $\varphi_k$ define $w_{\varphi_k}(z_1):=\int \varphi_k(z_1,z_2)\,d\mu_{Z_2}(z_2)$.
By Fubini's theorem, \eqref{eq:hs-training} and \eqref{eq:hs-target} are equivalent to equalizing the integrals of
the finite family $\{w_{\psi_\ell}\}_{\ell=1}^L\cup \{w_{\varphi_k}\}_{k=1}^K$ over a partition of $\gZ_1$.
Let $q:=L+K\le 9$. By Assumption~\ref{asmp:cond-space-nontrivial}, the measure space $(\gZ_1,\mu_{Z_1})$ admits a
$(q+1)$-nondegenerate function family in the sense of Definition~\ref{def:nondegenerate}. Therefore,
Corollary~\ref{cor:ham-sandwich-implication} applied on $(\gZ_1,\mu_{Z_1})$ with weights
$\{w_{\psi_\ell}\}_{\ell=1}^L\cup \{w_{\varphi_k}\}_{k=1}^K$ yields the desired partition.
\end{proof}

\subsubsection{Case 1: $\eps_{N,\gamma}\le \eps_{N,\alpha}$}

Fix feasible joint perturbations $(G_0,K_0)$ and $(G_1,K_1)$ from Assumption~\ref{asmp:main}$,$ and assume they are
$Z_1$-modulation closed at $(\hat P,\hat Q)$ in the sense of Definition~\ref{def:z1-modulation-closure}.
Define
\[
I_1:=\chi''(\hat P,\hat Q)[(G_0,K_0),(G_1,K_1)]\neq 0.
\]
Without loss of generality assume $I_1>0$ (otherwise replace $(G_1,K_1)$ by $-(G_1,K_1)$).
Write $g_0:=\dd G_0/\dd\mu$, $g_1:=\dd G_1/\dd\mu$, $k_0:=\dd K_0/\dd\mu_Z$, and $k_1:=\dd K_1/\dd\mu_Z$.

\paragraph{Local radii.}
Recall that $\eps_{N,\gamma}$ and $\eps_{N,\alpha}$ may depend on $N$; the same convention applies to the derived
radii $\tilde\eps_{N,\gamma}$ and $\tilde\eps_{N,\alpha}$.
Fix constants $C_{\gamma},C_{\alpha}>0$ as in Lemma~\ref{lemma:case1-functional-value-gap}.
Define
\begin{align}
\label{eq:def-tilde-eps-case1}
\tilde \eps_{N,\gamma}
&\;:=
\min\left\{
\frac{\eps_{N,\gamma}}{8\,(L_{1}+1)(L_{2}+1)}, \frac{r}{8\,\max\{\|g_0\|_{\mu,\infty},\|g_1\|_{\mu,\infty}\}}, \frac{r}{8\,\max\{\|k_0\|_{\mu_Z,\infty},\|k_1\|_{\mu_Z,\infty}\}}, \frac{c_t}{2}, \frac{I_1}{8C_{\gamma}}\tilde\eps_{N,\alpha}
\right\}, \\
\tilde \eps_{N,\alpha}
&\;:=
\min\left\{
\frac{\eps_{N,\alpha}}{4\,(L_{1}+1)}, \frac{r}{8\,\max\{\|g_0\|_{\mu,\infty},\|g_1\|_{\mu,\infty}\}}, \frac{r}{8\,\max\{\|k_0\|_{\mu_Z,\infty},\|k_1\|_{\mu_Z,\infty}\}}, \frac{c_t}{2}, \frac{I_1}{8C_{\alpha}}
\right\}.
\end{align}
Set $\bar b:=r^2/(4b_0^2)$ and $C_m:=\max\{2\max\{1,\bar b\},\ e\bar b^2/(1-e^{-3/2})\}$.
Fix $m:=2^{\lceil \log_2(C_mN^2)\rceil}$ so that $m$ is a power of two and $m\ge C_mN^2$, and set $M:=2m$ (so $M$ is still a power of two).

\paragraph{Partition.}
In all applications below, the total number of training and target weight functions is at most $9$, so the condition
$L+K\le 9$ in Corollary~\ref{cor:ham-sandwich} is satisfied.
Apply Corollary~\ref{cor:ham-sandwich} with training weights
\[
\psi\in\{\hat p,g_0,g_1\}
\]
and target weights
\[
\varphi\in\Big\{\hat q,~k_0,~k_1,~m_1\big(z,\hat\gamma(z)\big)\,k_0(z),~m_1\big(z,\gamma_P'(z;\hat P)[G_1]\big)\,\hat q(z),~m_1\big(z,\hat\gamma(z)\big)\,k_1(z)\Big\},
\]
to obtain a partition $\{\mathcal{X}_j\}_{j=1}^M$ of $\gZ_1$.

\paragraph{Alternatives.}
For each $\lambda\in\{-1,1\}^m$ define bumped perturbations $G_{0,\lambda},G_{1,\lambda}$ and
$K_{0,\lambda},K_{1,\lambda}$ and set
\begin{equation}
\label{eq:case1-alternatives}
\hat P_\lambda
\;:=\;\hat P+\tilde\eps_{N,\alpha}\,G_{0,\lambda}+\tilde\eps_{N,\gamma}\,G_{1,\lambda},
\qquad
\hat Q_\lambda
\;:=\;\hat Q+\tilde\eps_{N,\alpha}\,K_{0,\lambda}+\tilde\eps_{N,\gamma}\,K_{1,\lambda}.
\end{equation}
Write $\hat\gamma_\lambda:=\gamma(\cdot;\hat P_\lambda)$ and
$\hat\alpha_\lambda:=\alpha(\cdot;\hat P_\lambda,\hat Q_\lambda)$.

\begin{lemma}[Case 1: feasibility]
\label{lemma:case1-bump-perturbations-feasible}
Under Assumptions~\ref{asmp:density-bounded}--\ref{asmp:main}, for all sufficiently large $N$ and every
$\lambda\in\{-1,1\}^m$, we have
\[
    \max\left\{d_{\mu,\infty}(\hat P_\lambda,\hat P),d_{\mu_Z,\infty}(\hat Q_\lambda,\hat Q)\right\} \le r/2
\]
and
\[
(\hat P_\lambda,\hat Q_\lambda)\in \gM((\hat P,\hat Q);\eps_{N,\gamma},\eps_{N,\alpha}).
\]
\end{lemma}

\begin{proof}
Fix $\lambda$.

\paragraph{Feasibility of bumped directions.}
By \eqref{eq:hs-training}--\eqref{eq:hs-target} applied to $g_0,g_1,k_0,k_1$ and the paired form of
$\Delta(\lambda,\cdot)$, we have
$\int g_{k,\lambda}\,\dd\mu=0$ and $\int k_{k,\lambda}\,\dd\mu_Z=0$ for $k\in\{0,1\}$.
Thus the bump $\psi(\cdot)=\Delta(\lambda,\cdot)$ satisfies the centering conditions in
Definition~\ref{def:z1-modulation-closure} for each pair $(G_k,K_k)$.
Since $(G_0,K_0)$ and $(G_1,K_1)$ are $Z_1$-modulation closed, it follows that
for each $k\in\{0,1\}$ the bumped pair $(G_{k,\lambda},K_{k,\lambda})$ is itself a feasible joint perturbation at
$(\hat P,\hat Q)$.

\paragraph{(i) $\hat P_\lambda$ is a probability distribution and $d_{\mu,\infty}(\hat P_\lambda,\hat P)\le r/2$.}
Since $\int g_k\,d\mu=G_k(\gO)=0$ and \eqref{eq:hs-training} holds for $\psi=g_k$, each cell integral
$\int g_k(o)\mathbf{1}\{z_1\in\mathcal{X}_j\}\,d\mu(o)$ equals $(1/M)\int g_k\,d\mu=0$.
Hence $\int g_{k,\lambda}\,d\mu=0$ and $\int d\hat P_\lambda=1$.
Moreover, since $g_{k,\lambda}=\Delta(\lambda,z_1)g_k$ and $\Delta\in\{-1,1\}$,
$d_{\mu,\infty}(\hat P_\lambda,\hat P)=\|\tilde\eps_{N,\alpha}g_{0,\lambda}+\tilde\eps_{N,\gamma}g_{1,\lambda}\|_{\mu,\infty}$
$\le (\tilde\eps_{N,\alpha}+\tilde\eps_{N,\gamma})\max\{\|g_0\|_{\mu,\infty},\|g_1\|_{\mu,\infty}\}\le r/4$,
and hence $d_{\mu,\infty}(\hat P_\lambda,\hat P)\le r/2$ for all large $N$.

\paragraph{(ii) $\hat Q_\lambda$ is a probability distribution and $d_{\mu_Z,\infty}(\hat Q_\lambda,\hat Q)\le r/2$.}
Since $\int k_i\,d\mu_Z = K_i(\gZ)=0$ and \eqref{eq:hs-target} holds for $\varphi=k_i$, each cell integral
$\int k_i(z)\mathbf{1}\{z_1\in\mathcal{X}_j\}\,d\mu_Z(z)$ equals $(1/M)\int k_i\,d\mu_Z = 0$.
Hence $\int k_{i,\lambda}\,d\mu_Z=0$ and therefore $\int d\hat Q_\lambda=1$.
Moreover, since $k_{i,\lambda}=\Delta(\lambda,z_1)k_i$ and $\Delta\in\{-1,1\}$,
$d_{\mu_Z,\infty}(\hat Q_\lambda,\hat Q)=\|\tilde\eps_{N,\alpha}k_{0,\lambda}+\tilde\eps_{N,\gamma}k_{1,\lambda}\|_{\mu_Z,\infty}$
$\le (\tilde\eps_{N,\alpha}+\tilde\eps_{N,\gamma})\max\{\|k_0\|_{\mu_Z,\infty},\|k_1\|_{\mu_Z,\infty}\}\le r/4$,
and hence $d_{\mu_Z,\infty}(\hat Q_\lambda,\hat Q)\le r/2$ for all large $N$.

\paragraph{(iii) Nuisance constraints.}
Define the intermediate pair
$(\hat P_{\lambda,0},\hat Q_{\lambda,0}) := (\hat P+\tilde\eps_{N,\alpha}G_{0,\lambda},\ \hat Q+\tilde\eps_{N,\alpha}K_{0,\lambda})$.
By the two-step feasibility condition in Assumption~\ref{asmp:main} (applied with $\psi(\cdot)=\Delta(\lambda,\cdot)$), both
$(\hat P_{\lambda,0},\hat Q_{\lambda,0})$ and $(\hat P_\lambda,\hat Q_\lambda)$ belong to $\gP_0$.
By Lemma~\ref{lemma:bumped-slice} and the $\gamma$-invariance of $G_0$ in Assumption~\ref{asmp:main},
$\gamma(\cdot;\hat P_{\lambda,0})=\hat\gamma(\cdot)$.

\smallskip

\emph{Constraint for $\gamma$.}
Applying Assumption~\ref{asmp:regularity}(2) with the training perturbation $G_{1,\lambda}$ at base point
$\hat P_{\lambda,0}$ gives
\begin{align*}
\|\hat\gamma_\lambda-\hat\gamma\|_{\hat P_Z,2}
&=\|\gamma(\cdot;\hat P_{\lambda,0}+\tilde\eps_{N,\gamma}G_{1,\lambda})-\gamma(\cdot;\hat P_{\lambda,0})\|_{\hat P_Z,2}\\
&\le \tilde\eps_{N,\gamma}\,\|\gamma_P'(\cdot;\hat P_{\lambda,0})[G_{1,\lambda}]\|_{\hat P_Z,2}
\;+
L_{2}\,\tilde\eps_{N,\gamma}^2\,\|G_1\|_{\mathrm{TV}}^2.
\end{align*}
Using Assumption~\ref{asmp:regularity}(3) (with $t=\tilde\eps_{N,\alpha}$ and direction $G_{0,\lambda}$) to control
$\|\gamma_P'(\cdot;\hat P_{\lambda,0})[G_{1,\lambda}]\|$ by its value at $\hat P$, and the definition of
$\tilde\eps_{N,\gamma}$ in \eqref{eq:def-tilde-eps-case1}, yields
$\|\hat\gamma_\lambda-\hat\gamma\|_{\hat P_Z,2}\le \eps_{N,\gamma}$ for all large $N$.

\smallskip

\emph{Constraint for $\alpha$.}
We bound $\|\hat\alpha_\lambda-\hat\alpha\|_{\hat P_Z,2}$ along the two-step path
$(\hat P,\hat Q)\to(\hat P_{\lambda,0},\hat Q_{\lambda,0})\to(\hat P_\lambda,\hat Q_\lambda)$.
First, applying Assumption~\ref{asmp:regularity}(2) with joint perturbation $(G_{0,\lambda},K_{0,\lambda})$ at base point
$(\hat P,\hat Q)$ gives
\begin{align*}
\|\alpha(\cdot;\hat P_{\lambda,0},\hat Q_{\lambda,0})-\hat\alpha\|_{\hat P_Z,2}
&\le \tilde\eps_{N,\alpha}\,\|\alpha_{(P,Q)}'(\cdot;\hat P,\hat Q)[G_{0,\lambda},K_{0,\lambda}]\|_{\hat P_Z,2}
\;+
L_{2}\,\tilde\eps_{N,\alpha}^2\,\big(\|G_0\|_{\mathrm{TV}}+\|K_0\|_{\mathrm{TV}}\big)^2.
\end{align*}
Second, by the two-step feasibility condition in Assumption~\ref{asmp:main}, $(G_{1,\lambda},K_{1,\lambda})$ is a feasible joint perturbation at
$(\hat P_{\lambda,0},\hat Q_{\lambda,0})$, so Assumption~\ref{asmp:regularity}(2) yields
\begin{align*}
\|\hat\alpha_\lambda-\alpha(\cdot;\hat P_{\lambda,0},\hat Q_{\lambda,0})\|_{\hat P_Z,2}
&\le \tilde\eps_{N,\gamma}\,\|\alpha_{(P,Q)}'(\cdot;\hat P_{\lambda,0},\hat Q_{\lambda,0})[G_{1,\lambda},K_{1,\lambda}]\|_{\hat P_Z,2}
\;+
L_{2}\,\tilde\eps_{N,\gamma}^2\,\big(\|G_1\|_{\mathrm{TV}}+\|K_1\|_{\mathrm{TV}}\big)^2.
\end{align*}
Using the Lipschitz control in Assumption~\ref{asmp:regularity}(3) (with $t=\tilde\eps_{N,\alpha}$ and direction
$(G_{0,\lambda},K_{0,\lambda})$) to bound $\|\alpha_{(P,Q)}'(\cdot;\hat P_{\lambda,0},\hat Q_{\lambda,0})[G_{1,\lambda},K_{1,\lambda}]\|$ by its value at
$(\hat P,\hat Q)$, and combining the above displays with the triangle inequality, yields
$\|\hat\alpha_\lambda-\hat\alpha\|_{\hat P_Z,2}\le \eps_{N,\alpha}$ for all large $N$ by \eqref{eq:def-tilde-eps-case1} (using that $\eps_{N,\gamma}\le \eps_{N,\alpha}$ in Case~1).

\smallskip

Finally, because $d_{\mu,\infty}(\hat P_\lambda,\hat P)\le r/2$ and the anchor density is bounded below
(Assumption~\ref{asmp:density-bounded}), the $L^2$ norms under $\hat P_Z$ and under $\hat P_{\lambda,Z}$ are
equivalent up to a constant factor.
Thus the above bounds also yield
$\|\hat\gamma_\lambda-\hat\gamma\|_{\hat P_{\lambda,Z},2}\le \eps_{N,\gamma}$ and
$\|\hat\alpha_\lambda-\hat\alpha\|_{\hat P_{\lambda,Z},2}\le \eps_{N,\alpha}$.
\end{proof}

\begin{lemma}[Case 1: Hellinger bound]
\label{lemma:case1-hellinger-bound}
Let $\hat\Pi:=\mathrm{Unif}(\{-1,1\}^m)$ and define the (joint) mixture distribution
\[
\overline{\mathbb{P}}
:=\E_{\lambda\sim\hat\Pi}\big[\hat P_\lambda^{\otimes N}\otimes \hat Q_\lambda^{\otimes N}\big]
\qquad\text{on }\gO^N\times\gZ^N.
\]
Then, for all sufficiently large $N$,
\[
\HH^2\Big(\hat P^{\otimes N}\otimes \hat Q^{\otimes N},\,\overline{\mathbb{P}}\Big)
\le 2\left(1-e^{-3/2}\right).
\]
\end{lemma}

\begin{proof}
We apply the multi-sample Hellinger bound in Theorem~\ref{thm:semi-param-thm} with $S=2$, $n_1=N$ and $n_2=N$.
For the training sample, take $(\mathcal{X}^{(1)},\mu_1)=(\gO,\mu)$, $P^{(1)}=\hat P$ and $Q^{(1)}_\lambda=\hat P_\lambda$.
For the target sample, take $(\mathcal{X}^{(2)},\mu_2)=(\gZ,\mu_Z)$, $P^{(2)}=\hat Q$ and $Q^{(2)}_\lambda=\hat Q_\lambda$.
For both samples use the coarsened partition into $m$ \emph{pair-cells}
$\widetilde{\mathcal{X}}_i:=\mathcal{X}_{2i-1}\cup\mathcal{X}_{2i}$, $i\in[m]$:
\[
\mathcal{X}^{(1)}_i:=\{o\in\gO:\ z_1(o)\in\widetilde{\mathcal{X}}_i\},
\qquad
\mathcal{X}^{(2)}_i:=\{z\in\gZ:\ z_1\in\widetilde{\mathcal{X}}_i\},
\qquad i=1,\ldots,m.
\]

\paragraph{Verification of Theorem~\ref{thm:semi-param-thm}(A.1).}
By the ham-sandwich equalization with weights $\psi=\hat p$ and $\varphi=\hat q$, each cell has fixed probability
under the anchors, and therefore each pair-cell satisfies
$\hat P(\mathcal{X}^{(1)}_i)=\hat Q(\mathcal{X}^{(2)}_i)=2/M=1/m$.
Moreover, since $\eqref{eq:hs-training}$ holds for $\psi\in\{g_0,g_1\}$ and $\eqref{eq:hs-target}$ holds for
$\varphi\in\{k_0,k_1\}$, each of the perturbation densities has zero integral on every block $\mathcal{X}_j$ and
therefore also on each pair-cell $\widetilde{\mathcal{X}}_i$. Consequently,
$\hat P_\lambda(\mathcal{X}^{(1)}_i)=\hat P(\mathcal{X}^{(1)}_i)$ and
$\hat Q_\lambda(\mathcal{X}^{(2)}_i)=\hat Q(\mathcal{X}^{(2)}_i)$ for all $\lambda$.
Finally, on $\mathcal{X}_{2i-1}$ we have $\Delta(\lambda,z_1)=\lambda_i$ and on $\mathcal{X}_{2i}$ we have
$\Delta(\lambda,z_1)=-\lambda_i$, so the perturbed densities $\hat p_\lambda$ and $\hat q_\lambda$ depend on $\lambda$
only through $\lambda_i$ on the $i$th pair-cell.

\paragraph{Verification of Theorem~\ref{thm:semi-param-thm}(A.2).}
Because $\hat\Pi=\mathrm{Unif}(\{-1,1\}^m)$ is a product measure with mean-zero coordinates and $\Delta(\lambda,z_1)$ is
linear in $\lambda$, we have the centering identities
\(
\hat p=\E_{\lambda\sim\hat\Pi}[\hat p_\lambda]
\)
and
\(
\hat q=\E_{\lambda\sim\hat\Pi}[\hat q_\lambda]
\)
$\mu$-a.e.\ and $\mu_Z$-a.e., respectively.

\textit{Conclusion.}
For the constants in \eqref{eq:def-b}--\eqref{eq:hellinger-assumption-A}, we have $p_{\max}=1/m$ and
$n_{\mathrm{tot}}=2N$.
Moreover, Lemma~\ref{lemma:case1-bump-perturbations-feasible} implies
\(\|\hat p_\lambda-\hat p\|_{\mu,\infty}\le r/2\) and \(\|\hat q_\lambda-\hat q\|_{\mu_Z,\infty}\le r/2\) uniformly over
$\lambda\in\{-1,1\}^m$. Since the anchor densities are bounded away from zero
(Assumption~\ref{asmp:density-bounded}), the constant $b$ in \eqref{eq:def-b} satisfies
\(b\le \bar b:=r^2/(4b_0^2)\).
By our choice of $m$ (so that $m\ge 2N\max\{1,\bar b\}$), we have
$(2N)\cdot p_{\max}\cdot \max\{1,b\}\le 1$, and Theorem~\ref{thm:semi-param-thm} applies with $A=1$.
Therefore,
\[
\HH^2(\hat P^{\otimes N}\otimes \hat Q^{\otimes N},\overline{\mathbb{P}})
\le \frac{e}{2}\,(2N)^2\,\frac{1}{m}\,b^2
\le \frac{e}{2}\,(2N)^2\,\frac{1}{m}\,\bar b^2
\le 2\,(1-e^{-3/2}),
\]
where the last inequality uses $m\ge C_mN^2$ and the definition of $C_m$.
\end{proof}

\begin{lemma}[Case 1: functional separation]
\label{lemma:case1-functional-value-gap}
Under Assumptions~\ref{asmp:density-bounded}--\ref{asmp:main}, for all sufficiently large $N$ and every
$\lambda\in\{-1,1\}^m$,
\[
\chi(\hat P_\lambda,\hat Q_\lambda)-\chi(\hat P,\hat Q)
\ge \frac{I_1}{4}\,\tilde\eps_{N,\alpha}\tilde\eps_{N,\gamma}.
\]
\end{lemma}

\begin{proof}
Fix $\lambda$ and define the intermediate pair
\[
(\hat P_{\lambda,0},\hat Q_{\lambda,0})
:=\bigl(\hat P+\tilde\eps_{N,\alpha}G_{0,\lambda},\;\hat Q+\tilde\eps_{N,\alpha}K_{0,\lambda}\bigr).
\]
By Lemma~\ref{lemma:bumped-slice} and the $\gamma$-invariance of $G_0$, we have
\begin{equation}
\label{eq:case1-gamma-invariance-cs}
\gamma(\cdot;\hat P_{\lambda,0})=\hat\gamma(\cdot).
\end{equation}

\paragraph{The $O(\tilde\eps_{N,\alpha})$ term vanishes by construction.}
Using \eqref{eq:cs-target-functional} and \eqref{eq:case1-gamma-invariance-cs},
\[
\chi(\hat P_{\lambda,0},\hat Q_{\lambda,0})-\chi(\hat P,\hat Q)
= \E_{\hat Q_{\lambda,0}}\big[m_1(Z,\hat\gamma(Z))\big]-\E_{\hat Q}\big[m_1(Z,\hat\gamma(Z))\big]
=\tilde\eps_{N,\alpha}\,\E_{K_{0,\lambda}}\big[m_1(Z,\hat\gamma(Z))\big].
\]
By Lemma~\ref{lemma:bumped-slice},
$\E_{K_{0,\lambda}}[m_1(Z,\hat\gamma(Z))]$ equals
\(\int \Delta(\lambda,z_1)m_1(z,\hat\gamma(z))k_0(z)\,d\mu_Z(z)\).
The ham-sandwich condition \eqref{eq:hs-target} applied to
$\varphi(z)=m_1(z,\hat\gamma(z))k_0(z)$ implies that for every $j\in[M]$,
\[
\int \varphi(z)\,\mathbf{1}\{z_1\in\mathcal{X}_j\}\,d\mu_Z(z)=\frac{1}{M}\int \varphi(z)\,d\mu_Z(z).
\]
Using the paired form of $\Delta(\lambda,\cdot)$ and $M=2m$, we obtain
\[
\int \Delta(\lambda,z_1)\,\varphi(z)\,d\mu_Z(z)
=
\sum_{i=1}^m \lambda_i
\left(
\int \varphi(z)\,\mathbf{1}\{z_1\in\mathcal{X}_{2i-1}\}\,d\mu_Z(z)
-
\int \varphi(z)\,\mathbf{1}\{z_1\in\mathcal{X}_{2i}\}\,d\mu_Z(z)
\right)=0,
\]
since each difference vanishes. Therefore
\begin{equation}
\label{eq:case1-linear-alpha-zero}
\chi(\hat P_{\lambda,0},\hat Q_{\lambda,0})=\chi(\hat P,\hat Q).
\end{equation}

\paragraph{First-order expansion in the $(G_{1,\lambda},K_{1,\lambda})$ direction.}
Consider the path
\[
(P_t,Q_t):=(\hat P_{\lambda,0}+tG_{1,\lambda},\hat Q_{\lambda,0}+tK_{1,\lambda}).
\]
Since $d_{\mu,\infty}(\hat P_{\lambda,0},\hat P)\le r/2$ and $d_{\mu_Z,\infty}(\hat Q_{\lambda,0},\hat Q)\le r/2$
(Lemma~\ref{lemma:case1-bump-perturbations-feasible}) and $\tilde\eps_{N,\gamma}\le c_t$ for all large $N$, we can
apply Assumption~\ref{asmp:regularity}(4) at base point $(\hat P_{\lambda,0},\hat Q_{\lambda,0})$ in direction
$(G_{1,\lambda},K_{1,\lambda})$ to obtain
\begin{equation}
\label{eq:case1-taylor-in-gamma-dir}
\chi(\hat P_\lambda,\hat Q_\lambda)-\chi(\hat P_{\lambda,0},\hat Q_{\lambda,0})
\;=\;\tilde\eps_{N,\gamma}\,
\chi_{(P,Q)}'\big(\hat P_{\lambda,0},\hat Q_{\lambda,0}\big)\big[G_{1,\lambda},K_{1,\lambda}\big]
+\mathrm{Rem}^{(1)}_\lambda,
\end{equation}
where the remainder satisfies
\begin{equation}
\label{eq:case1-rem1-bound}
|\mathrm{Rem}^{(1)}_\lambda|
\le
L_{\chi,2}\,\tilde\eps_{N,\gamma}^2\big(\|G_{1,\lambda}\|_{\mathrm{TV}}+\|K_{1,\lambda}\|_{\mathrm{TV}}\big)^2
\le C\,\tilde\eps_{N,\gamma}^2,
\end{equation}
where $C:=L_{\chi,2}\big(\|G_{1}\|_{\mathrm{TV}}+\|K_{1}\|_{\mathrm{TV}}\big)^2$ and we used
$|\Delta(\lambda,\cdot)|=1$ so that the total variation norms do not depend on $\lambda$.

\paragraph{The $O(\tilde\eps_{N,\gamma})$ score term vanishes by construction.}
Using Proposition~\ref{prop:compute-2nd-order-differential} and \eqref{eq:case1-gamma-invariance-cs},
\[
\chi_{(P,Q)}'\big(\hat P_{\lambda,0},\hat Q_{\lambda,0}\big)\big[G_{1,\lambda},K_{1,\lambda}\big]
=\E_{\hat Q_{\lambda,0}}\left[m_1\left(Z,\gamma_P'(Z;\hat P_{\lambda,0})[G_{1,\lambda}]\right)\right]
+\E_{K_{1,\lambda}}\left[m_1\left(Z,\hat\gamma(Z)\right)\right].
\]
Expand the first expectation under $\hat Q_{\lambda,0}=\hat Q+\tilde\eps_{N,\alpha}K_{0,\lambda}$ and retain only
the leading $O(1)$ term:
\[
\E_{\hat Q_{\lambda,0}}\left[m_1\left(Z,\gamma_P'(Z;\hat P_{\lambda,0})[G_{1,\lambda}]\right)\right]
=\E_{\hat Q}\left[m_1\left(Z,\gamma_P'(Z;\hat P_{\lambda,0})[G_{1,\lambda}]\right)\right]
+O(\tilde\eps_{N,\alpha}).
\]
By Lemma~\ref{lemma:bumped-slice},
$\gamma_P'(\cdot;\hat P)[G_{1,\lambda}]=\Delta(\lambda,\cdot)\gamma_P'(\cdot;\hat P)[G_1]$ and
\[
\E_{K_{1,\lambda}}\left[m_1\left(Z,\hat\gamma(Z)\right)\right]
=\int \Delta(\lambda,z_1)\,m_1\bigl(z,\hat\gamma(z)\bigr)\,k_1(z)\,\dd\mu_Z(z).
\]
The ham-sandwich conditions \eqref{eq:hs-target} applied to the target weights
\[
\varphi(z)=m_1\bigl(z,\gamma_P'(z;\hat P)[G_1]\bigr)\,\hat q(z),
\qquad
\varphi(z)=m_1\bigl(z,\hat\gamma(z)\bigr)\,k_1(z),
\]
imply that the integrals of each weight over $\mathcal X_j$ are constant in $j$, and hence (by the same paired
calculation as above) both integrals against $\Delta(\lambda,z_1)$ vanish. Consequently,
\begin{equation}
\label{eq:case1-linear-gamma-zero}
\E_{\hat Q}\left[m_1\left(Z,\gamma_P'(Z;\hat P)[G_{1,\lambda}]\right)\right]
+\E_{K_{1,\lambda}}\left[m_1\left(Z,\hat\gamma(Z)\right)\right]=0.
\end{equation}

\paragraph{The mixed $O(\tilde\eps_{N,\alpha}\tilde\eps_{N,\gamma})$ term equals $I_1$ up to a negligible remainder.}
By Assumption~\ref{asmp:regularity}(3), the map $P\mapsto \gamma_P'(\cdot;P)[G_{1,\lambda}]$ is locally Lipschitz,
so
\[
\gamma_P'(\cdot;\hat P_{\lambda,0})[G_{1,\lambda}]
=\gamma_P'(\cdot;\hat P)[G_{1,\lambda}]
+\tilde\eps_{N,\alpha}\,\gamma_P''(\cdot;\hat P)[G_{0,\lambda},G_{1,\lambda}]
+\mathrm{Rem}'_\lambda,
\qquad
\|\mathrm{Rem}'_\lambda\|_{\hat P_Z,2}\le L_1\,\tilde\eps_{N,\alpha}^2.
\]
Plugging this into the first term of \eqref{eq:case1-linear-gamma-zero} and using the same cancellation argument as
above yields
\[
\chi'\big(\hat P_{\lambda,0},\hat Q_{\lambda,0}\big)\big[(G_{1,\lambda},K_{1,\lambda})\big]
=
\begin{aligned}[t]
\tilde\eps_{N,\alpha}\Big\{&
\E_{\hat Q}\big[m_1(Z,\gamma_P''(Z;\hat P)[G_{0,\lambda},G_{1,\lambda}])\big]\\
&+\E_{K_{0,\lambda}}\big[m_1(Z,\gamma_P'(Z;\hat P)[G_{1,\lambda}])\big]\Big\}
+O(\tilde\eps_{N,\alpha}^2).
\end{aligned}
\]
By Lemma~\ref{lemma:bumped-slice} and $\Delta(\lambda,\cdot)^2\equiv 1$,
\[
\gamma_P''(\cdot;\hat P)[G_{0,\lambda},G_{1,\lambda}]
=
\gamma_P''(\cdot;\hat P)[G_0,G_1],
\]
and, using also linearity of $m_1$ in its second argument,
\[
\begin{aligned}
\E_{K_{0,\lambda}}\big[m_1(Z,\gamma_P'(Z;\hat P)[G_{1,\lambda}])\big]
&=
\int \Delta(\lambda,z_1)^2\,m_1\bigl(z,\gamma_P'(z;\hat P)[G_1]\bigr)\,k_0(z)\,\dd\mu_Z(z)\\
&=
\E_{K_0}\big[m_1(Z,\gamma_P'(Z;\hat P)[G_1])\big].
\end{aligned}
\]
Therefore the bracketed term equals
\[
\E_{\hat Q}\big[m_1(Z,\gamma_P''(Z;\hat P)[G_0,G_1])\big]
\;+\;
\E_{K_0}\big[m_1(Z,\gamma_P'(Z;\hat P)[G_1])\big]
=\chi''(\hat P,\hat Q)[(G_0,K_0),(G_1,K_1)]
=I_1,
\]
where we used Proposition~\ref{prop:compute-2nd-order-differential} and that $\gamma_P'(\cdot;\hat P)[G_0]\equiv 0$
since $\gamma(\cdot;\hat P+tG_0)$ is constant in a neighborhood of $t=0$ (Assumption~\ref{asmp:main}).

\paragraph{Collect terms and control remainders.}
Combining \eqref{eq:case1-linear-alpha-zero}, \eqref{eq:case1-taylor-in-gamma-dir} and the expansion above gives
\[
\chi(\hat P_\lambda,\hat Q_\lambda)-\chi(\hat P,\hat Q)
=I_1\,\tilde\eps_{N,\alpha}\tilde\eps_{N,\gamma}+\mathrm{Rem}^{\chi}_\lambda,
\]
where there exist constants $C_{\gamma},C_{\alpha}>0$ (depending only on the constants in
Assumptions~\ref{asmp:density-bounded}--\ref{asmp:main} and the fixed perturbations $(G_0,K_0)$ and $(G_1,K_1)$)
such that
\[
|\mathrm{Rem}^{\chi}_\lambda|\le C_{\gamma}\tilde\eps_{N,\gamma}^2+C_{\alpha}\tilde\eps_{N,\alpha}^2\tilde\eps_{N,\gamma}.
\]
By the choice \eqref{eq:def-tilde-eps-case1}, we have $\tilde\eps_{N,\alpha}\le I_1/(8C_{\alpha})$ and
$\tilde\eps_{N,\gamma}\le (I_1/(8C_{\gamma}))\tilde\eps_{N,\alpha}$, whence
\[
|\mathrm{Rem}^{\chi}_\lambda|
\le \frac{I_1}{8}\tilde\eps_{N,\alpha}\tilde\eps_{N,\gamma}+\frac{I_1}{8}\tilde\eps_{N,\alpha}\tilde\eps_{N,\gamma}
=\frac{I_1}{4}\tilde\eps_{N,\alpha}\tilde\eps_{N,\gamma},
\]
which implies the claim.
\end{proof}

\begin{lemma}[A parametric $N^{-1/2}$ lower bound]
\label{lemma:parametric-rootn}
Fix any $\xi\in(1/2,1)$.
Assume Assumptions~\ref{asmp:density-bounded} and \ref{asmp:regularity} hold, and suppose that
Assumption~\ref{asmp:main}(3) holds with a feasible joint perturbation $(G_{\mathrm{LC}},K_{\mathrm{LC}})$.
Assume further that there exists a constant $c_{\min}>0$ such that, for all sufficiently large $N$,
\[
\eps_{N,\gamma}\ge c_{\min}N^{-1/2}
\qquad\text{and}\qquad
\eps_{N,\alpha}\ge c_{\min}N^{-1/2}.
\]
Then there exists $\delta_{\mathrm{LC}}>0$ such that for all sufficiently large $N$,
\[
\mathfrak{M}^{\chi}_{N,\xi}\left(\gM((\hat P,\hat Q);\eps_{N,\gamma},\eps_{N,\alpha})\right)
\ge
\delta_{\mathrm{LC}}N^{-1/2}.
\]
\end{lemma}

\begin{proof}
Let $(G_{\mathrm{LC}},K_{\mathrm{LC}})$ be as in Assumption~\ref{asmp:main}(3).
By replacing $(G_{\mathrm{LC}},K_{\mathrm{LC}})$ with $-(G_{\mathrm{LC}},K_{\mathrm{LC}})$ if needed, we may assume
without loss of generality that the directional derivative
\[
D_{\mathrm{LC}}:=\chi_{(P,Q)}'(\hat P,\hat Q)[G_{\mathrm{LC}},K_{\mathrm{LC}}]
\]
is strictly positive.
Write $g_{\mathrm{LC}}:=\dd G_{\mathrm{LC}}/\dd\mu$ and $k_{\mathrm{LC}}:=\dd K_{\mathrm{LC}}/\dd\mu_Z$.
By feasibility, there exists $r_{\mathrm{LC}}>0$ such that
$(\hat P+tG_{\mathrm{LC}},\hat Q+tK_{\mathrm{LC}})\in\gP_0$ for all $|t|\le r_{\mathrm{LC}}$.

For a constant $c>0$ to be chosen below, set $t_N:=c/\sqrt{N}$ and define the two local alternatives
\[
(\hat P^{(0)},\hat Q^{(0)}):=(\hat P-t_NG_{\mathrm{LC}},\hat Q-t_NK_{\mathrm{LC}}),
\qquad
(\hat P^{(1)},\hat Q^{(1)}):=(\hat P+t_NG_{\mathrm{LC}},\hat Q+t_NK_{\mathrm{LC}}).
\]
For all sufficiently large $N$ and sufficiently small $c$, we have $t_N\le r_{\mathrm{LC}}$, so both pairs lie in
$\gP_0$.

\paragraph{Step 1: both alternatives lie in the anchored class.}
We verify that, for each $j\in\{0,1\}$,
\[
(\hat P^{(j)},\hat Q^{(j)})\in\gM((\hat P,\hat Q);\eps_{N,\gamma},\eps_{N,\alpha}).
\]

\emph{(i) $d_{\mu,\infty}$-constraints.}
By construction,
\[
d_{\mu,\infty}(\hat P^{(j)},\hat P)=t_N\|g_{\mathrm{LC}}\|_{\infty},
\qquad
d_{\mu_Z,\infty}(\hat Q^{(j)},\hat Q)=t_N\|k_{\mathrm{LC}}\|_{\infty}.
\]
Thus, choosing $c$ so that
$c\max\{\|g_{\mathrm{LC}}\|_{\infty},\|k_{\mathrm{LC}}\|_{\infty}\}\le r/2$
ensures these constraints hold for all sufficiently large $N$.

\emph{(ii) $\gamma$-constraint.}
Assumption~\ref{asmp:regularity}(2) (with $P=\hat P$ and direction $\pm G_{\mathrm{LC}}$) gives
\[
\gamma(\cdot;\hat P^{(j)})-\gamma(\cdot;\hat P)
=(-1)^j t_N\,\gamma_P'(\cdot;\hat P)[G_{\mathrm{LC}}]+\mathrm{Rem}^{\gamma}_{\hat P^{(j)},\hat P},
\qquad
\big\|\mathrm{Rem}^{\gamma}_{\hat P^{(j)},\hat P}\big\|_{L^2(\hat P_Z)}\le L_2 t_N^2\|G_{\mathrm{LC}}\|_{TV}^2.
\]
Using the pointwise derivative bound \eqref{eq:asmp-regularity-derivative-bounded},
$|\gamma_P'(z;\hat P)[G_{\mathrm{LC}}]|\le L_1$ for all $z$, hence
$\|\gamma_P'(\cdot;\hat P)[G_{\mathrm{LC}}]\|_{L^2(\hat P_Z)}\le L_1$.
Therefore, for all sufficiently large $N$,
\[
\|\gamma(\cdot;\hat P^{(j)})-\gamma(\cdot;\hat P)\|_{L^2(\hat P_Z)}
\le L_1 t_N + L_2 t_N^2\|G_{\mathrm{LC}}\|_{TV}^2
\le 2L_1 t_N.
\]
If we additionally choose $c\le c_{\min}/(2L_1)$, then $2L_1 t_N\le \eps_{N,\gamma}$ for all sufficiently large $N$.

\emph{(iii) $\alpha$-constraint.}
The same argument, using Assumption~\ref{asmp:regularity}(2) for $\alpha$ with direction
$(\pm G_{\mathrm{LC}},\pm K_{\mathrm{LC}})$ and the derivative bound \eqref{eq:asmp-regularity-derivative-bounded}, yields
for all sufficiently large $N$,
\[
\|\alpha(\cdot;\hat P^{(j)},\hat Q^{(j)})-\alpha(\cdot;\hat P,\hat Q)\|_{L^2(\hat P_Z)}\le 2L_1 t_N.
\]
With the same choice $c\le c_{\min}/(2L_1)$, this is at most $\eps_{N,\alpha}$ for all sufficiently large $N$.
This completes the membership verification.

\paragraph{Step 2: $N^{-1/2}$ separation in $\chi$.}
Assumption~\ref{asmp:regularity}(4) gives
\[
\big|\chi(\hat P^{(j)},\hat Q^{(j)})-\chi(\hat P,\hat Q)-(-1)^j t_N D_{\mathrm{LC}}\big|
\le
L_{\chi,2}t_N^2\big(\|G_{\mathrm{LC}}\|_{TV}+\|K_{\mathrm{LC}}\|_{TV}\big)^2.
\]
Hence,
\[
\chi(\hat P^{(1)},\hat Q^{(1)})-\chi(\hat P^{(0)},\hat Q^{(0)})
\ge 2t_ND_{\mathrm{LC}}-2L_{\chi,2}t_N^2\big(\|G_{\mathrm{LC}}\|_{TV}+\|K_{\mathrm{LC}}\|_{TV}\big)^2.
\]
Since $t_N\to 0$, for all sufficiently large $N$ the second term is at most $t_ND_{\mathrm{LC}}$, and thus
\[
\chi(\hat P^{(1)},\hat Q^{(1)})-\chi(\hat P^{(0)},\hat Q^{(0)})\ge t_ND_{\mathrm{LC}}.
\]
Define
\[
s_N:=\frac{1}{2}\Big\{\chi(\hat P^{(1)},\hat Q^{(1)})-\chi(\hat P^{(0)},\hat Q^{(0)})\Big\}
\ge \frac{D_{\mathrm{LC}}}{2}t_N
=\frac{D_{\mathrm{LC}}c}{2}N^{-1/2}.
\]

\paragraph{Step 3: small Hellinger distance between the two joint laws.}
Let $\mathbb{P}_j:=(\hat P^{(j)})^{\otimes N}\otimes(\hat Q^{(j)})^{\otimes N}$ denote the joint laws of the training
and target samples.
Using the product property of Hellinger affinity and the inequality $1-ab\le (1-a)+(1-b)$ for $a,b\in[0,1]$, we have
\[
H^2(\mathbb{P}_1,\mathbb{P}_0)
\le N H^2(\hat P^{(1)},\hat P^{(0)})+N H^2(\hat Q^{(1)},\hat Q^{(0)}).
\]
We bound $H^2(\hat P^{(1)},\hat P^{(0)})$.
Write $\hat p:=\dd\hat P/\dd\mu$ and $p^{(j)}:=\dd\hat P^{(j)}/\dd\mu=\hat p+(-1)^j t_N g_{\mathrm{LC}}$.
By Assumption~\ref{asmp:density-bounded}, $\hat p\ge b_0$ $\mu$-a.e.
Since $t_N\to 0$ and $g_{\mathrm{LC}}$ is bounded, for all sufficiently large $N$ we have $|t_N g_{\mathrm{LC}}|\le b_0/2$ and hence
$p^{(j)}\ge b_0/2$.
Therefore,
\[
\big(\sqrt{p^{(1)}}-\sqrt{p^{(0)}}\big)^2
=\frac{(p^{(1)}-p^{(0)})^2}{(\sqrt{p^{(1)}}+\sqrt{p^{(0)}})^2}
=\frac{(2t_N g_{\mathrm{LC}})^2}{(\sqrt{p^{(1)}}+\sqrt{p^{(0)}})^2}
\le \frac{4t_N^2 g_{\mathrm{LC}}^2}{b_0}.
\]
Integrating yields
\[
H^2(\hat P^{(1)},\hat P^{(0)})
\le \frac{4t_N^2}{b_0}\int g_{\mathrm{LC}}(o)^2\,\dd\mu(o)
\le \frac{4t_N^2}{b_0}\|g_{\mathrm{LC}}\|_{\infty}\int |g_{\mathrm{LC}}(o)|\,\dd\mu(o)
=\frac{4t_N^2}{b_0}\|g_{\mathrm{LC}}\|_{\infty}\|G_{\mathrm{LC}}\|_{TV}.
\]
An identical argument gives
\[
H^2(\hat Q^{(1)},\hat Q^{(0)})
\le \frac{4t_N^2}{b_0}\|k_{\mathrm{LC}}\|_{\infty}\|K_{\mathrm{LC}}\|_{TV}.
\]
Consequently,
\[
H^2(\mathbb{P}_1,\mathbb{P}_0)
\le C_{\mathrm{LC}}\,N t_N^2
=C_{\mathrm{LC}}c^2,
\]
where
$C_{\mathrm{LC}}:=(4/b_0)\big(\|g_{\mathrm{LC}}\|_{\infty}\|G_{\mathrm{LC}}\|_{TV}+\|k_{\mathrm{LC}}\|_{\infty}\|K_{\mathrm{LC}}\|_{TV}\big)$.

\paragraph{Step 4: apply the two-point fuzzy-hypotheses bound.}
Choose $c>0$ small enough that $\delta_{\xi}:=C_{\mathrm{LC}}c^2<2$ and
\[
1-\sqrt{\frac{\delta_{\xi}(1-\delta_{\xi}/4)}{2}}\ge \xi.
\]
Apply Theorem~\ref{thm:fano-method} with base distribution $P=\mathbb{P}_0$, mixing measure $\pi=\delta_{\mathbb{P}_1}$,
and functional $T(P,Q)=\chi(P,Q)$.
By Step 2, we have $T(\mathbb{P}_0)\le c_0$ and $T(\mathbb{P}_1)=c_0+2s_N$ for $c_0:=\chi(\hat P^{(0)},\hat Q^{(0)})$.
The theorem then yields that for every estimator $\hat\chi$,
\[
\sup_{(P,Q)\in\{(\hat P^{(0)},\hat Q^{(0)}),(\hat P^{(1)},\hat Q^{(1)})\}}
\Pr_{(P,Q)}\big(|\hat\chi-\chi(P,Q)|\ge s_N\big)
\ge \xi.
\]
Because both alternatives belong to $\gM((\hat P,\hat Q);\eps_{N,\gamma},\eps_{N,\alpha})$, this implies
\[
\mathfrak{M}^{\chi}_{N,\xi}\left(\gM((\hat P,\hat Q);\eps_{N,\gamma},\eps_{N,\alpha})\right)
\ge s_N
\ge \frac{D_{\mathrm{LC}}c}{2}N^{-1/2}.
\]
Setting $\delta_{\mathrm{LC}}:=D_{\mathrm{LC}}c/2$ completes the proof.
\end{proof}

\begin{lemma}[Case 1: minimax lower bound]
\label{lemma:case1-asymptotic-separation}
Fix any $\xi\in(1/2,1)$.
Under Assumptions~\ref{asmp:density-bounded}--\ref{asmp:main}, and assuming that there exists a constant $c_{\min}>0$
such that $\eps_{N,\gamma}\ge c_{\min}N^{-1/2}$ and $\eps_{N,\alpha}\ge c_{\min}N^{-1/2}$ for all sufficiently large $N$,
there exists $\delta>0$ such that for all sufficiently large $N$,
\[
\mathfrak{M}^{\chi}_{N,\xi}\left(\gM((\hat P,\hat Q);\eps_{N,\gamma},\eps_{N,\alpha})\right)
\ge
\delta\left(\tilde\eps_{N,\alpha}\tilde\eps_{N,\gamma}+N^{-1/2}\right).
\]
\end{lemma}

\begin{proof}
Combine Lemma~\ref{lemma:case1-bump-perturbations-feasible},
Lemma~\ref{lemma:case1-hellinger-bound}, and Lemma~\ref{lemma:case1-functional-value-gap}
with Theorem~\ref{thm:fano-method} applied to the joint laws $P^{\otimes N}\otimes \hat Q^{\otimes N}$.
Thus, there exists a constant $\delta_{\mathrm{prod}}>0$ such that for all sufficiently large $N$,
\[
\mathfrak{M}^{\chi}_{N,\xi}\left(\gM((\hat P,\hat Q);\eps_{N,\gamma},\eps_{N,\alpha})\right)
\ge \delta_{\mathrm{prod}}\,\tilde\eps_{N,\alpha}\tilde\eps_{N,\gamma}.
\]
Separately, Lemma~\ref{lemma:parametric-rootn} yields a constant $\delta_{\mathrm{LC}}>0$ such that for all sufficiently
large $N$,
\[
\mathfrak{M}^{\chi}_{N,\xi}\left(\gM((\hat P,\hat Q);\eps_{N,\gamma},\eps_{N,\alpha})\right)
\ge \delta_{\mathrm{LC}}\,N^{-1/2}.
\]
Therefore,
\[
\mathfrak{M}^{\chi}_{N,\xi}\left(\gM((\hat P,\hat Q);\eps_{N,\gamma},\eps_{N,\alpha})\right)
\ge \max\{\delta_{\mathrm{prod}}\,\tilde\eps_{N,\alpha}\tilde\eps_{N,\gamma},\;\delta_{\mathrm{LC}}\,N^{-1/2}\}
\ge \frac{\min\{\delta_{\mathrm{prod}},\delta_{\mathrm{LC}}\}}{2}\left(\tilde\eps_{N,\alpha}\tilde\eps_{N,\gamma}+N^{-1/2}\right),
\]
which is the claimed bound.
\end{proof}

\subsubsection{Case 2: $\eps_{N,\alpha}<\eps_{N,\gamma}$}

\paragraph{Part (a): product lower bound under mixed-bias.}
Assume the mixed-bias property \eqref{eq:mixed-bias} holds, so that there exists a mapping
$m_2:\gO\times L^2(\mu_Z)\to\mathbb{R}$ which is linear in its second argument and such that
\[
\chi(P,Q)=\E_{O\sim P}\left[m_2\left(O,\alpha(Z;P,Q)\right)\right].
\]
In this representation the outer expectation is with respect to the training law $P$.
We repeat the Case~1 construction with the replacements
\[
m_1~\rightsquigarrow~ m_2,\qquad
\gamma~\rightsquigarrow~ \alpha,\qquad
(G_0,K_0,G_1,K_1)~\rightsquigarrow~ (H_0,L_0,H_1,L_1),
\]
where $(H_0,L_0),(H_1,L_1)$ are the feasible perturbation pairs from Assumption~\ref{asmp:main}$,$ and are assumed to
be $Z_1$-modulation closed at $(\hat P,\hat Q)$ in the sense of Definition~\ref{def:z1-modulation-closure}.

Define
\[
I_2:=\chi''(\hat P,\hat Q)[(H_0,L_0),(H_1,L_1)]\neq 0
\]
and assume $I_2>0$ without loss of generality.
Write $h_i:=\dd H_i/\dd\mu$ and $l_i:=\dd L_i/\dd\mu_Z$ for $i\in\{0,1\}$.
Choose local radii
\begin{align*}
\tilde \eps_{N,\alpha}^{(2)}
&:=
\min\left\{\frac{\eps_{N,\alpha}}{8\,(L_{1}+1)(L_{2}+1)}, \frac{r}{8\,\max\{\|h_0\|_{\mu,\infty},\|h_1\|_{\mu,\infty}\}}, \frac{r}{8\,\max\{\|l_0\|_{\mu_Z,\infty},\|l_1\|_{\mu_Z,\infty}\}}, \frac{c_t}{2}
\right\},\\
\tilde \eps_{N,\gamma}^{(2)}
&:=
\min\left\{
\frac{\eps_{N,\gamma}}{4\,(L_{1}+1)}, \frac{r}{8\,\max\{\|h_0\|_{\mu,\infty},\|h_1\|_{\mu,\infty}\}}, \frac{r}{8\,\max\{\|l_0\|_{\mu_Z,\infty},\|l_1\|_{\mu_Z,\infty}\}}, \frac{c_t}{2}
\right\},
\end{align*}
so that $\tilde \eps_{N,\alpha}^{(2)}\le \tilde \eps_{N,\gamma}^{(2)}$ in Case~2.

Apply Corollary~\ref{cor:ham-sandwich} with training weights
\[
\psi\in\Big\{\hat p,h_0,h_1,~m_2\big(\cdot,\hat\alpha(\cdot)\big)\,h_0(\cdot),~m_2\big(\cdot,\hat\alpha(\cdot)\big)\,h_1(\cdot),
~m_2\big(\cdot,\alpha_{(P,Q)}'(\cdot;\hat P,\hat Q)[(H_1,L_1)]\big)\,\hat p(\cdot)\Big\},
\]
and target weights (to ensure $\hat Q_\lambda'$ is a probability distribution)
\[
\varphi\in\{\hat q,l_0,l_1\},
\]
to obtain a partition $\{\mathcal{X}_j\}_{j=1}^M$.
Construct bumped perturbations $H_{0,\lambda},H_{1,\lambda},L_{0,\lambda},L_{1,\lambda}$ and define
\[
\hat P_\lambda'
:=\hat P+\tilde\eps_{N,\gamma}^{(2)}\,H_{0,\lambda}+\tilde\eps_{N,\alpha}^{(2)}\,H_{1,\lambda},
\qquad
\hat Q_\lambda'
:=\hat Q+\tilde\eps_{N,\gamma}^{(2)}\,L_{0,\lambda}+\tilde\eps_{N,\alpha}^{(2)}\,L_{1,\lambda}.
\]
Define $\hat\gamma_\lambda':=\gamma(\cdot;\hat P_\lambda')$ and $\hat\alpha_\lambda':=\alpha(\cdot;\hat P_\lambda',\hat Q_\lambda')$.

We verify the conditions of Theorem~\ref{thm:fano-method}.

(i) \emph{Feasibility and nuisance constraints.}\par\noindent
As in Case~1, the ham-sandwich equalization implies $\int h_{i,\lambda}\,d\mu=0$ for all $i\in\{0,1\}$.
It also implies $\int l_{i,\lambda}\,d\mu_Z=0$ for all $i\in\{0,1\}$.
Therefore, $\hat P_\lambda'$ and $\hat Q_\lambda'$ are probability distributions.
Moreover, $h_{i,\lambda}=\Delta(\lambda,z_1)h_i$ and $l_{i,\lambda}=\Delta(\lambda,z_1)l_i$, where $\Delta(\lambda,z_1)\in\{-1,1\}$.
Hence,
\begin{align*}
d_{\mu,\infty}(\hat P_\lambda',\hat P)
&=\left\|\tilde\eps_{N,\gamma}^{(2)}h_{0,\lambda}+\tilde\eps_{N,\alpha}^{(2)}h_{1,\lambda}\right\|_{\mu,\infty}\\
&\le (\tilde\eps_{N,\gamma}^{(2)}+\tilde\eps_{N,\alpha}^{(2)})\max\{\|h_0\|_{\mu,\infty},\|h_1\|_{\mu,\infty}\}
\le r/4,\\
d_{\mu_Z,\infty}(\hat Q_\lambda',\hat Q)
&=\left\|\tilde\eps_{N,\gamma}^{(2)}l_{0,\lambda}+\tilde\eps_{N,\alpha}^{(2)}l_{1,\lambda}\right\|_{\mu_Z,\infty}\\
&\le (\tilde\eps_{N,\gamma}^{(2)}+\tilde\eps_{N,\alpha}^{(2)})\max\{\|l_0\|_{\mu_Z,\infty},\|l_1\|_{\mu_Z,\infty}\}
\le r/4.
\end{align*}
By the two-step feasibility condition in Assumption~\ref{asmp:main}, $(\hat P_\lambda',\hat Q_\lambda')\in\gP_0$ for all large $N$.

Define the intermediate pair
$(\hat P'_{\lambda,0},\hat Q'_{\lambda,0}) := (\hat P+\tilde\eps_{N,\gamma}^{(2)}H_{0,\lambda},\ \hat Q+\tilde\eps_{N,\gamma}^{(2)}L_{0,\lambda})$.
By Lemma~\ref{lemma:bumped-slice} and the $\alpha$-invariance of $(H_0,L_0)$ in Assumption~\ref{asmp:main},
$\alpha(\cdot;\hat P'_{\lambda,0},\hat Q'_{\lambda,0})=\hat\alpha(\cdot)$.
Applying Assumption~\ref{asmp:regularity}(2) at base point $(\hat P'_{\lambda,0},\hat Q'_{\lambda,0})$ along the feasible perturbation $(H_{1,\lambda},L_{1,\lambda})$, together with Assumption~\ref{asmp:regularity}(3), yields $\|\hat\alpha_\lambda'-\hat\alpha\|_{\hat P_Z,2}\le \eps_{N,\alpha}$ for all large $N$.
A two-step argument identical to the one above, but applied to $\gamma(\cdot;P)$ along $H_{0,\lambda}$ and then $H_{1,\lambda}$, yields $\|\hat\gamma_\lambda'-\hat\gamma\|_{\hat P_Z,2}\le \eps_{N,\gamma}$ for all large $N$.
Hence $(\hat P_\lambda',\hat Q_\lambda')\in\gM((\hat P,\hat Q);\eps_{N,\gamma},\eps_{N,\alpha})$.

(ii) \emph{Hellinger bound.}
Let $\hat\Pi=\mathrm{Unif}(\{-1,1\}^m)$ and $\overline{\mathbb P}':=\E_{\lambda\sim\hat\Pi}[(\hat P_\lambda')^{\otimes N}\otimes(\hat Q_\lambda')^{\otimes N}]$.
The proof of Lemma~\ref{lemma:case1-hellinger-bound} applies verbatim (with $h_0,h_1,l_0,l_1$ in place of $g_0,g_1,k_0,k_1$), giving
$\HH^2(\hat P^{\otimes N}\otimes \hat Q^{\otimes N},\overline{\mathbb P}')\le 2(1-e^{-3/2})$ for all large $N$.

(iii) \emph{Separation.}
The proof of Lemma~\ref{lemma:case1-functional-value-gap} applies verbatim after the replacements $m_1\rightsquigarrow m_2$ and $\gamma\rightsquigarrow \alpha$, so that
$\chi(\hat P_\lambda',\hat Q_\lambda')-\chi(\hat P,\hat Q)\ge \frac{I_2}{4}\,\tilde\eps_{N,\alpha}^{(2)}\tilde\eps_{N,\gamma}^{(2)}$ for all large $N$.

Combining (i)–(iii) with Theorem~\ref{thm:fano-method} (and the parametric lower bound in Lemma~\ref{lemma:parametric-rootn}) yields
\[
\mathfrak{M}^{\chi}_{N,\xi}\left(\gM((\hat P,\hat Q);\eps_{N,\gamma},\eps_{N,\alpha})\right)
=\Omega\left(\eps_{N,\gamma}\eps_{N,\alpha}+N^{-1/2}\right),
\]
in the regime $\eps_{N,\alpha}<\eps_{N,\gamma}$, completing the proof of Theorem~\ref{thm:main-mixed-bias}.

\paragraph{Part (b): quadratic lower bound under non-affine $\rho$.}
For Theorem~\ref{thm:main} we use the $\alpha$-invariant direction $H_0$ to obtain a separation of order
$\eps_{N,\gamma}^2$.
Define
\[
I_3:=\chi''(\hat P,\hat Q)\big[(H_0,L_0),(H_0,L_0)\big]\neq 0,
\]
and assume $I_3>0$ without loss of generality.
Fix a constant $t_0>0$ as in Lemma~\ref{lemma:case2-quadratic-gap}.
Let $h_0:=dH_0/d\mu$ and $l_0:=dL_0/d\mu_Z$ and set
\[
\bar \eps_{N,\gamma}
:=\min\left\{
\frac{\eps_{N,\gamma}}{8\,(L_{1}+1)(L_{2}+1)},~
\frac{r}{8\,\max\{\|h_0\|_{\mu,\infty},\|l_0\|_{\mu_Z,\infty}\}},~
\frac{c_t}{2},~
t_0
\right\}.
\]
Apply Corollary~\ref{cor:ham-sandwich} with training weights $\psi\in\{\hat p,h_0\}$ and target weights
\[
\varphi\in\Big\{\hat q,~l_0,~m_1\big(z,\gamma_P'(z;\hat P)[H_0]\big)\,\hat q(z),~m_1\big(z,\hat\gamma(z)\big)\,l_0(z)\Big\},
\]
and construct bumped perturbations $H_{0,\lambda}$ and $L_{0,\lambda}$.
Define the alternatives
\[
\hat P_\lambda^{(2)}:=\hat P+\bar\eps_{N,\gamma}\,H_{0,\lambda},
\qquad
\hat Q_\lambda^{(2)}:=\hat Q+\bar\eps_{N,\gamma}\,L_{0,\lambda}.
\]

\begin{lemma}[Case 2: quadratic functional separation]
\label{lemma:case2-quadratic-gap}
Under Assumptions~\ref{asmp:density-bounded}--\ref{asmp:main}, for all sufficiently large $N$ and every
$\lambda\in\{-1,1\}^m$,
\[
\chi(\hat P_\lambda^{(2)},\hat Q_\lambda^{(2)})-\chi(\hat P,\hat Q)
\ge \frac{I_3}{4}\,\bar\eps_{N,\gamma}^2.
\]
\end{lemma}

\begin{proof}
Fix $\lambda$ and write $(P_t,Q_t):=(\hat P+tH_{0,\lambda},\hat Q+tL_{0,\lambda})$.
By Proposition~\ref{prop:compute-2nd-order-differential}, $\chi(P_t,Q_t)$ admits the second-order expansion
\[
\chi(P_t,Q_t)
=
\chi(\hat P,\hat Q)
+t\,\chi_{(P,Q)}'(\hat P,\hat Q)\big[H_{0,\lambda},L_{0,\lambda}\big]
+\frac{t^2}{2}\,\chi''(\hat P,\hat Q)\big[(H_{0,\lambda},L_{0,\lambda}),(H_{0,\lambda},L_{0,\lambda})\big]
+o(t^2),
\]
as $t\to 0$, uniformly over $\lambda$.
Using \eqref{eq:chi-first-derivative-pair} together with Lemma~\ref{lemma:bumped-slice}, we have
\[
\chi_{(P,Q)}'(\hat P,\hat Q)\big[H_{0,\lambda},L_{0,\lambda}\big]
=
\E_{\hat Q}\left[m_1\left(Z,\gamma_P'(Z;\hat P)[H_{0,\lambda}]\right)\right]
+\E_{L_{0,\lambda}}\left[m_1\left(Z,\hat\gamma(Z)\right)\right].
\]
By Lemma~\ref{lemma:bumped-slice} and linearity of $m_1$ in its second argument,
\[
\E_{\hat Q}\left[m_1\left(Z,\gamma_P'(Z;\hat P)[H_{0,\lambda}]\right)\right]
=
\int \Delta(\lambda,z_1)\,m_1\left(z,\gamma_P'(z;\hat P)[H_0]\right)\,\hat q(z)\,d\mu_Z(z),
\]
and
\(
\E_{L_{0,\lambda}}\left[m_1\left(Z,\hat\gamma(Z)\right)\right]
=\int \Delta(\lambda,z_1)\,m_1(z,\hat\gamma(z))\,l_0(z)\,d\mu_Z(z).
\)
The ham-sandwich condition \eqref{eq:hs-target} applied to the target weights
$m_1(z,\gamma_P'(z;\hat P)[H_0])\,\hat q(z)$ and $m_1(z,\hat\gamma(z))\,l_0(z)$ implies that each weight has equal
integral over every block $\mathcal X_j$, and therefore (by the same paired computation as in
\eqref{eq:case1-linear-alpha-zero}) both integrals against $\Delta(\lambda,z_1)$ are zero. Hence the first-order term
vanishes.

Next, use \eqref{eq:chi-second-derivative-square-pair} together with Lemma~\ref{lemma:bumped-slice} and
$\Delta(\lambda,\cdot)^2\equiv 1$ to obtain
\[
\chi''(\hat P,\hat Q)\big[(H_{0,\lambda},L_{0,\lambda}),(H_{0,\lambda},L_{0,\lambda})\big]
=\chi''(\hat P,\hat Q)\big[(H_0,L_0),(H_0,L_0)\big]
=I_3,
\]
so the leading term equals $(t^2/2)I_3$ uniformly over $\lambda$.
By the definition of the $o(t^2)$ remainder, there exists $t_0>0$ such that for all $|t|\le t_0$,
\[
\big|o(t^2)\big|\le \frac{I_3}{4}t^2.
\]
Since $\bar\eps_{N,\gamma}\le t_0$ by definition, taking $t=\bar\eps_{N,\gamma}$ yields the
claim.
\end{proof}

\begin{lemma}[Case 2: minimax lower bound]
\label{lemma:case2-asymptotic-separation}
Fix any $\xi\in(1/2,1)$.
Under Assumptions~\ref{asmp:density-bounded}--\ref{asmp:main}, and assuming that there exists a constant $c_{\min}>0$
such that $\eps_{N,\gamma}\ge c_{\min}N^{-1/2}$ and $\eps_{N,\alpha}\ge c_{\min}N^{-1/2}$ for all sufficiently large $N$,
there exists $\delta>0$ such that for all sufficiently large $N$,
\[
\mathfrak{M}^{\chi}_{N,\xi}\left(\gM((\hat P,\hat Q);\eps_{N,\gamma},\eps_{N,\alpha})\right)
\ge
\delta\left(\bar\eps_{N,\gamma}^2+N^{-1/2}\right).
\]
\end{lemma}

\begin{proof}
The feasibility of $(\hat P_\lambda^{(2)},\hat Q_\lambda^{(2)})$ follows as in Lemma~\ref{lemma:case1-bump-perturbations-feasible}
using that $(H_0,L_0)$ is $\alpha$-invariant (Assumption~\ref{asmp:main}).
The Hellinger bound follows from Theorem~\ref{thm:semi-param-thm} applied to the family
$\{(\hat P_\lambda^{(2)},\hat Q_\lambda^{(2)})\}$ with $m\ge 2N$.
Finally, Lemma~\ref{lemma:case2-quadratic-gap} gives uniform separation of order $\bar\eps_{N,\gamma}^2$.
Applying Theorem~\ref{thm:fano-method} therefore yields a constant $\delta_{\mathrm{quad}}>0$ such that for all
sufficiently large $N$,
\[
\mathfrak{M}^{\chi}_{N,\xi}\left(\gM((\hat P,\hat Q);\eps_{N,\gamma},\eps_{N,\alpha})\right)
\ge \delta_{\mathrm{quad}}\,\bar\eps_{N,\gamma}^2.
\]
Separately, Lemma~\ref{lemma:parametric-rootn} yields a constant $\delta_{\mathrm{LC}}>0$ such that for all sufficiently
large $N$,
\[
\mathfrak{M}^{\chi}_{N,\xi}\left(\gM((\hat P,\hat Q);\eps_{N,\gamma},\eps_{N,\alpha})\right)
\ge \delta_{\mathrm{LC}}\,N^{-1/2}.
\]
Therefore,
\[
\mathfrak{M}^{\chi}_{N,\xi}\left(\gM((\hat P,\hat Q);\eps_{N,\gamma},\eps_{N,\alpha})\right)
\ge \max\{\delta_{\mathrm{quad}}\,\bar\eps_{N,\gamma}^2,\;\delta_{\mathrm{LC}}\,N^{-1/2}\}
\ge \frac{\min\{\delta_{\mathrm{quad}},\delta_{\mathrm{LC}}\}}{2}\left(\bar\eps_{N,\gamma}^2+N^{-1/2}\right),
\]
which is the claimed bound.
\end{proof}

This completes the proofs of Theorems~\ref{thm:main-mixed-bias} and \ref{thm:main}.

\section{Proof of lower bounds of the examples in Section \ref{sec:examples}}
\label{sec:proof:examples}

\subsection{Proof of Theorem \ref{thm:ate}}
\label{sec:proof:ate}

We verify that the assumptions of Theorem \ref{thm:main-mixed-bias} hold under the conditions imposed in Theorem \ref{thm:ate}. 
Throughout, we let $O=(X,D,Y)\in\gO=\gX\times\gD\times\gY$ with $\gD=\gY=\{0,1\}$, and we take the dominating measure to be
\[
    \mu := \mu_{\gX}\otimes \mathrm{count}_{\gD}\otimes \mathrm{count}_{\gY}.
\]
For any $P\ll \mu$ we write $p:=\dd P/\dd\mu$ and similarly $\hat p := \dd \hat P/\dd\mu$. We also use the shorthand
\[
    p(x,d,\cdot):=\sum_{y\in\{0,1\}}p(x,d,y),\qquad p_X(x):=\sum_{d\in\{0,1\}}p(x,d,\cdot),
\]
and analogously for $\hat p$.

\textit{Identifying the objects $(m_1,\rho,\gamma,\alpha)$ and checking affineness of $\rho$.}
With $Z_1=X\in\gX$, $Z_2=D\in\gD$ and $W=Y\in\gY$, we set $Z=(Z_1,Z_2)=(X,D)$ and define, for any $\gM$-feasible $P$,
\[
    \gamma(z;P) := g(d,x;P) := \E_P[Y\mid X=x,D=d] = \frac{p(x,d,1)}{p(x,d,\cdot)},
\]
\[
    \rho(o,\gamma):= y-\gamma(x,d), 
    \qquad 
    m_1(o,h):=h(x,1)-h(x,0).
\]
The mapping $\gamma\mapsto \rho(o,\gamma)$ is affine (indeed linear with slope $-1$), so the ``mixed-bias'' theorem (Theorem \ref{thm:main-mixed-bias}) is the appropriate main result to invoke.

The Riesz representer for the ATE functional is the usual inverse-propensity weight
\begin{equation}
    \label{eq:ate-alpha-def}
    \alpha(z;P)=\alpha(x,d;P)
    :=
    \frac{d}{\pi(x;P)}-\frac{1-d}{1-\pi(x;P)},
    \qquad 
    \pi(x;P):=\mathbb{P}(D=1\mid X=x)=\frac{p(x,1,\cdot)}{p_X(x)}.
\end{equation}
We will verify below that this $\alpha$ satisfies the weighted Riesz representer requirement in Assumption \ref{asmp:cond-prob-functional}.

\paragraph{Verifying Assumption \ref{asmp:density-bounded}.}
Under the factorization in Theorem \ref{thm:ate} we may write, for $\mu$-a.e.\ $(x,d,y)$,
\[
    \hat p(x,d,y) 
    = \hat p_X(x)\,
    \pi(x;\hat P)^d\big(1-\pi(x;\hat P)\big)^{1-d}\,
    g(d,x;\hat P)^y\big(1-g(d,x;\hat P)\big)^{1-y}.
\]
By assumption, each of $\pi(x;\hat P)$, $1-\pi(x;\hat P)$, $g(d,x;\hat P)$ and $1-g(d,x;\hat P)$ is at least $c$.
If $\hat p_X$ satisfies $l_X\le \hat p_X(x)\le u_X$ on $\gX$ (the density-boundedness hypothesis in Theorem \ref{thm:ate}), then for $\mu$-a.e.\ $(x,d,y)$,
\[
    l_X c^2 \;\le\; \hat p(x,d,y)\;\le\; u_X.
\]
Therefore Assumption \ref{asmp:density-bounded} holds for $\hat P$ with $l_{\hat p}=l_Xc^2$ and $u_{\hat p}=u_X$.

\paragraph{Verifying Assumption \ref{asmp:cond-prob-functional}.}
Both nuisances $\gamma(z;P)=g(d,x;P)$ and $\alpha(z;P)$ in \eqref{eq:ate-alpha-def} depend on $P$ only through the conditional law of $(D,Y)$ given $X=x$, equivalently through the collection $\{p(x,d,y):d,y\in\{0,1\}\}$ up to the multiplicative factor $p_X(x)$ which cancels in the ratios defining $\gamma$ and $\pi$.

Next, we verify the weighted Riesz representer identity for $m_1$ and $\rho$. 
Fix any square-integrable $h:\gZ\to\R$.
Using iterated expectation and the definition of $\alpha$,
\begin{align*}
    \E_P\big[h(Z)\alpha(Z;P)\big]
    &=\E_P\Big[\E_P\big[h(X,D)\alpha(X,D;P)\mid X\big]\Big] \\
    &= \E_P\Big[ h(X,1)\frac{\mathbb{P}(D=1\mid X)}{\pi(X;P)} - h(X,0)\frac{\mathbb{P}(D=0\mid X)}{1-\pi(X;P)}\Big] \\
    &= \E_P\big[ h(X,1)-h(X,0)\big]
    = \E_P\big[m_1(O,h)\big].
\end{align*}
Moreover, since $\rho(o,\gamma)=y-\gamma(x,d)$ and $\gamma(z;P)=\E_P[Y\mid Z=z]$, we have
\[
    \E_P[\rho(O,\gamma(Z;P)+a)\mid Z=z]=\E_P[Y\mid Z=z]-(\gamma(z;P)+a)=-a,
\]
so that $\nu_\rho(z;P)=\frac{\dd}{\dd a}\E_P[\rho(O,\gamma(Z;P)+a)\mid Z=z]\big|_{a=0}=-1$ is well-defined and uniformly bounded.
This verifies Assumption \ref{asmp:cond-prob-functional} for the ATE specification.

\paragraph{Verifying Assumption \ref{asmp:regularity}.}
Let $l_{\hat p}=l_Xc^2$ be the lower bound from the verification of Assumption \ref{asmp:density-bounded} and set $r:=\tfrac12 l_{\hat p}$.
If $P$ satisfies $d_{\mu,\infty}(P,\hat P)\le r$, then $p(o)\ge \hat p(o)-r\ge \tfrac12 l_{\hat p}$ for $\mu$-a.e.\ $o$.
In particular, for all $x\in\gX$ and $d\in\{0,1\}$,
\[
    p(x,d,\cdot)\ge \tfrac12 l_{\hat p},\qquad p_X(x)\ge l_{\hat p}.
\]
Now let $H$ be any perturbation with density $h=\dd H/\dd\mu$ satisfying $\|h\|_\infty\le C_P$ (as in Assumption \ref{asmp:regularity}).

\textit{Derivative bounds for $\gamma$.}
Write $a=p(x,d,1)$ and $b=p(x,d,0)$ so that $\gamma=a/(a+b)$ and $a+b=p(x,d,\cdot)\ge \tfrac12 l_{\hat p}$.
A direct differentiation yields, for $\mu$-a.e.\ $(x,d)$,
\begin{align}
    \label{eq:ate-gamma-derivative}
    \gamma_P'(x,d;P)[H]
    &= \frac{b\,h(x,d,1)-a\,h(x,d,0)}{(a+b)^2}.
\end{align}
Hence,
\[
    \big|\gamma_P'(x,d;P)[H]\big|
    \le 
    \frac{|b|\,|h(x,d,1)|+|a|\,|h(x,d,0)|}{(a+b)^2}
    \le 
    \frac{2u_{\hat p}C_P}{(\tfrac12 l_{\hat p})^2},
\]
using $a,b\le u_{\hat p}$ and $\|h\|_\infty\le C_P$.
Similarly, since $\gamma$ is a smooth rational function of $(a,b)$ on the set $\{a+b\ge \tfrac12 l_{\hat p}\}$, its second directional derivative $\gamma_P''(z;P)[H,H']$ exists and is uniformly bounded by a constant depending only on $(l_{\hat p},u_{\hat p},C_P)$; one may bound it explicitly by differentiating \eqref{eq:ate-gamma-derivative} once more and using $|h|,|h'|\le C_P$.

\textit{Derivative bounds for $\alpha$.}
Using \eqref{eq:ate-alpha-def}, $\alpha$ is a smooth function of $\pi(x;P)$ on $\pi\in[c,1-c]$.
Moreover, $\pi(x;P)=p(x,1,\cdot)/p_X(x)$ is a smooth rational function of $(p(x,1,\cdot),p(x,0,\cdot))$ on the set where $p_X(x)\ge l_{\hat p}$.
Therefore $\pi_P'(x;P)[H]$ and $\pi_P''(x;P)[H,H']$ exist and are uniformly bounded for all $P$ with $d_{\mu,\infty}(P,\hat P)\le r$ and all perturbations with bounded densities.
Combining with the boundedness of the derivatives of $\pi\mapsto d/\pi-(1-d)/(1-\pi)$ on $[c,1-c]$ gives uniform bounds for $\alpha_P'(z;P)[H]$ and $\alpha_P''(z;P)[H,H']$ as required in \eqref{eq:asmp-regularity-derivative-bounded}.

Finally, $\rho(o,\gamma)=y-\gamma(x,d)$ is uniformly bounded by $1$ and $\upsilon_\rho(\cdot;P)\equiv 0$ for all $P$ because $\rho$ is affine in $\gamma$.
Thus Assumption \ref{asmp:regularity} holds.

\paragraph{Verifying Assumption \ref{asmp:main} by constructing perturbations.}
We now construct $\gM$-feasible perturbations $G_0,G_1,H_0,H_1$ of $\hat P$ satisfying the invariance conditions and the nondegeneracy of the mixed second derivatives required in Assumption \ref{asmp:main}.

Fix any measurable set $B\subseteq \gX$ with $0<\mu_{\gX}(B)<1$ and define $\varphi(x):=\mathbf{1}\{x\in B\}-\mathbf{1}\{x\notin B\}$ so that $\varphi(x)\in\{-1,1\}$ and $\varphi^2(x)\equiv 1$.
(Existence of such a $B$ is trivial since $\mu_{\gX}$ is non-atomic; e.g.\ for $\gX=[0,1]^d$ one can take $B=\{x:x_1\le 1/2\}$.)

\textit{(a) A $\gamma$-invariant direction $G_0$ and a companion direction $G_1$ with $\chi''(\hat P)[G_0,G_1]\neq 0$.}
Define $\dd G_0=g_0\,\dd\mu$ and $\dd G_1=g_1\,\dd\mu$ by
\begin{equation}
    \label{eq:ate-G0G1-def}
    \begin{aligned}
        g_0(x,1,y)&:=\varphi(x)\hat p(x,1,y), \\
        g_0(x,0,y)&:=-\varphi(x)\hat p(x,1,\cdot)\frac{\hat p(x,0,y)}{\hat p(x,0,\cdot)},\\
        g_1(x,1,1)&:=\varphi(x)\hat p(x,1,\cdot),\qquad g_1(x,1,0):=-\varphi(x)\hat p(x,1,\cdot),\\
        g_1(x,0,1)&:=g_1(x,0,0):=0.
    \end{aligned}
\end{equation}
First note that for each $x$, $g_0(x,1,\cdot)+g_0(x,0,\cdot)=0$, hence $\int g_0\,\dd\mu=0$.
Also $\int g_1\,\dd\mu=0$ since $g_1(x,1,1)+g_1(x,1,0)=0$ and $g_1(x,0,\cdot)\equiv 0$.
Both $g_0$ and $g_1$ are uniformly bounded because $\hat p$ is uniformly bounded away from $0$ and $\infty$ (Assumption \ref{asmp:density-bounded}), so $G_0$ and $G_1$ are valid perturbations; moreover, for sufficiently small $|t|$ the perturbed densities $\hat p+t g_0$ and $\hat p+t g_1$ remain nonnegative and uniformly bounded, hence define $\gM$-feasible distributions.

\textit{$\gamma$-invariance along $G_0$.}
Fix $z=(x,d)$.
For $d=1$, we have for all sufficiently small $t$ and both $y\in\{0,1\}$,
\[
    \hat p_t(x,1,y):=\hat p(x,1,y)+t g_0(x,1,y)=\hat p(x,1,y)\big(1+t\varphi(x)\big),
\]
so the conditional law of $Y$ given $(X,D)=(x,1)$ is unchanged, and thus $\gamma(x,1;\hat P+tG_0)=\gamma(x,1;\hat P)$.
For $d=0$, similarly,
\[
    \hat p_t(x,0,y)=\hat p(x,0,y)\Big(1-t\varphi(x)\frac{\hat p(x,1,\cdot)}{\hat p(x,0,\cdot)}\Big),
\]
so again the conditional law of $Y$ given $(X,D)=(x,0)$ is unchanged, and $\gamma(x,0;\hat P+tG_0)=\gamma(x,0;\hat P)$.
Hence $\gamma(z;\hat P+tG_0)=\gamma(z;\hat P)$ for all $z$ and all sufficiently small $|t|$.

\textit{Computing $\chi''(\hat P)[G_0,G_1]$.}
For $s,t$ small, let $P_{s,t}:=\hat P+sG_0+tG_1$ and denote its density by $p_{s,t}$.
By construction, $p_{s,t,X}(x)=\sum_{d,y}p_{s,t}(x,d,y)=\hat p_X(x)$, i.e.\ the marginal of $X$ is unchanged.
Moreover, $g_{s,t}(0,x):=\gamma(x,0;P_{s,t})=\gamma(x,0;\hat P)$ since $G_1$ does not perturb the $d=0$ slice and $G_0$ preserves $\gamma$.
For $d=1$,
\[
    g_{s,t}(1,x)
    =\frac{\hat p(x,1,1)(1+s\varphi(x))+t\varphi(x)\hat p(x,1,\cdot)}
    {\hat p(x,1,\cdot)(1+s\varphi(x))}
    =\hat g(1,x)+t\frac{\varphi(x)}{1+s\varphi(x)}.
\]
Therefore
\begin{align*}
    \chi_{\ate}(P_{s,t})
    &= \int_{\gX} \Big(g_{s,t}(1,x)-g_{s,t}(0,x)\Big)\,\dd \hat P_X(x) \\
    &= \chi_{\ate}(\hat P) + t\int_{\gX}\frac{\varphi(x)}{1+s\varphi(x)}\,\dd \hat P_X(x).
\end{align*}
Expanding $(1+s\varphi)^{-1}=1-s\varphi+O(s^2)$ gives
\[
    \chi_{\ate}(P_{s,t})
    = \chi_{\ate}(\hat P) + t\int \varphi\,\dd\hat P_X - st \int \varphi^2\,\dd\hat P_X + O(s^2t).
\]
Since $\varphi^2\equiv 1$ and $\hat P_X$ is a probability measure, $\int \varphi^2\,\dd\hat P_X = 1$, hence the coefficient of $st$ is $-1$.
It follows that
\[
    \chi''_{\ate}(\hat P)[G_0,G_1] = \frac{\partial^2}{\partial s\,\partial t}\chi_{\ate}(\hat P+sG_0+tG_1)\Big|_{s=t=0} = -1\neq 0.
\]

\textit{(b) An $\alpha$-invariant direction $H_0$ and a companion direction $H_1$ with $\chi''(\hat P)[H_0,H_1]\neq 0$.}
Define $\dd H_0=h_0\,\dd\mu$ and $\dd H_1=h_1\,\dd\mu$ by
\begin{equation}
    \label{eq:ate-H0H1-def}
    \begin{aligned}
        h_0(x,1,1)&:=\varphi(x)\hat p(x,1,\cdot),\qquad h_0(x,1,0):=-\varphi(x)\hat p(x,1,\cdot),\\
        h_0(x,0,1)&:=h_0(x,0,0):=0,\\
        h_1(x,d,y)&:=g_0(x,d,y)\qquad (\text{i.e. }H_1:=G_0).
    \end{aligned}
\end{equation}
As before, $\int h_0\,\dd\mu=0$ and $h_0$ is bounded, so $H_0$ is a valid perturbation; $H_1$ is already known to be $\gM$-feasible.

\textit{$\alpha$-invariance along $H_0$.}
Along the path $P_t:=\hat P+tH_0$, we have $p_t(x,d,\cdot)=\hat p(x,d,\cdot)$ for both $d=0,1$ because $h_0(x,d,1)+h_0(x,d,0)=0$ for each $(x,d)$.
Hence $\pi(x;P_t)=\pi(x;\hat P)$ for all $x$ and all sufficiently small $|t|$, and therefore $\alpha(z;P_t)=\alpha(z;\hat P)$ for all $z$.

\textit{Computing $\chi''(\hat P)[H_0,H_1]$.}
Let $P_{s,t}:=\hat P+sH_0+tH_1$.
Since $H_1=G_0$ preserves the marginal of $X$, we again have $p_{s,t,X}=\hat p_X$.
Moreover, the same calculation as in part (a) (with the roles of $(s,t)$ swapped) yields
\[
    \chi_{\ate}(P_{s,t})
    =\chi_{\ate}(\hat P)+ s\int_{\gX}\frac{\varphi(x)}{1+t\varphi(x)}\,\dd\hat P_X(x),
\]
so the coefficient of $st$ equals $- \int \varphi^2\,\dd\hat P_X=-1$.
Thus $\chi''_{\ate}(\hat P)[H_0,H_1]=-1\neq 0$.

\textit{Conclusion.}
The arguments above verify Assumptions \ref{asmp:density-bounded}, \ref{asmp:cond-prob-functional}, \ref{asmp:regularity}, \ref{asmp:cond-space-nontrivial} (assumed in Theorem \ref{thm:ate}) and \ref{asmp:main}. Since $\rho$ is affine in $\gamma$, Theorem \ref{thm:main-mixed-bias} applies and yields the desired lower bound in Theorem \ref{thm:ate}.

\subsection{Proof of Theorem \ref{thm:att}}\label{sec:proof:att}

\begin{proof}[Proof of Theorem~\ref{thm:att}]
We verify Assumptions~\ref{asmp:density-bounded}, \ref{asmp:cond-prob-functional}, \ref{asmp:regularity}, and
\ref{asmp:main} for the functional $\chi_{\att}$ in \eqref{eq:att-cs-functional}, and then invoke
Theorem~\ref{thm:main-mixed-bias}.

\textit{Setup and notation.}
Let $O=(X,D,Y)$, where $X\in\gX$, $D\in\{0,1\}$, and $Y\in\{0,1\}$.
Fix a dominating measure $\mu:=\mu_{\gX}\otimes\mathrm{count}_{\{0,1\}}\otimes\mathrm{count}_{\{0,1\}}$ on $\gO$,
where $\mu_{\gX}$ is the uniform distribution on $\gX$.
Write the anchor training law $\hat P$ by its density $\hat p:=\dd \hat P/\dd\mu$.
Write the anchor target law $\hat Q$ by its $X$-density $\hat q_{\gX}:=\dd \hat Q/\dd\mu_{\gX}$.
For any $P\ll \mu$ with density $p:=\dd P/\dd\mu$, write
\[
p_{\gX}(x):=\sum_{d\in\{0,1\}}\sum_{y\in\{0,1\}}p(x,d,y),\qquad
p_{d,\cdot}(x):=\sum_{y\in\{0,1\}}p(x,d,y),
\]
and define the propensity and outcome regressions by
\[
\pi(x;P):=P(D=1\mid X=x)=\frac{p_{1,\cdot}(x)}{p_{\gX}(x)},\qquad
g(d,x;P):=P(Y=1\mid X=x,D=d)=\frac{p(x,d,1)}{p_{d,\cdot}(x)}.
\]
The regression nuisance of interest is the control regression $\gamma(x;P):=g(0,x;P)$.

\paragraph{Verifying Assumption~\ref{asmp:density-bounded}.}
Condition (2) in Theorem~\ref{thm:att} gives constants $0<l\le u<\infty$ such that
$l\le \hat p_{\gX}(x)\le u$ and $l\le \hat q_{\gX}(x)\le u$ for all $x\in\gX$.
Condition (1) gives $c\le \pi(x;\hat P)\le 1-c$ and $c\le g(d,x;\hat P)\le 1-c$ for all $x$ and $d$.
Therefore, for all $(x,d,y)$,
\[
\hat p(x,d,y)
=\hat p_{\gX}(x)\,P_{\hat P}(D=d\mid X=x)\,P_{\hat P}(Y=y\mid X=x,D=d)
\]
is bounded above and below by positive constants depending only on $(c,l,u)$.
Thus Assumption~\ref{asmp:density-bounded} holds at $(\hat P,\hat Q)$ with some $0<b_0<b_1<\infty$.

\paragraph{Verifying Assumption~\ref{asmp:cond-prob-functional} (conditional moment and Riesz representer).}
We cast the ATT second term into the general form \eqref{eq:cs-target-functional}.
Set $Z=X$ and $W=(D,Y)$ so that $O=(Z,W)$, and define $m_1(z,h)=h(z)$.
Define the residual
\[
\rho(O,\gamma):=(1-D)\{Y-\gamma(X)\}.
\]
Then for any measurable $\gamma:\gX\to\R$,
\[
\E_P[\rho(O,\gamma)\mid X=x]=(1-\pi(x;P))\{g(0,x;P)-\gamma(x)\}.
\]
Hence $\gamma(\cdot;P)=g(0,\cdot;P)$ is the unique solution of the conditional moment restriction
$\E_P[\rho(O,\gamma)\mid X]=0$, which is \eqref{eq:att-gamma-moment}.

Next, we compute $\nu_\rho(\cdot;P)$ and the Riesz representer $\nu_m(\cdot;P,\hat Q)$.
For any $a\in\R$,
\[
\E_P[\rho(O,\gamma+a)\mid X=x]=-(1-\pi(x;P))\,a,
\]
so $\nu_\rho(x;P)=-(1-\pi(x;P))$.
At the anchor, $|\nu_\rho(x;\hat P)|\ge c$ by Condition (1).

For $\nu_m$, note that for any $\Delta\in L^2(P_X)$,
\[
\E_{\hat Q}[\Delta(X)]
=\int_{\gX}\Delta(x)\hat q_{\gX}(x)\,\mu_{\gX}(\dd x)
=\E_P\!\left[\Delta(X)\,\frac{\hat q_{\gX}(X)}{p_{\gX}(X)}\right].
\]
By Assumption~\ref{asmp:density-bounded}, $\hat q_{\gX}/p_{\gX}$ is essentially bounded in a small
$d_{\mu,\infty}$-neighborhood of $\hat P$, so this functional is continuous on $L^2(P_X)$ and the Riesz representer is
\[
\nu_m(x;P,\hat Q)=\frac{\hat q_{\gX}(x)}{p_{\gX}(x)}.
\]
Therefore, the debiasing weight is
\[
\alpha(x;P,\hat Q)=-\frac{\nu_m(x;P,\hat Q)}{\nu_\rho(x;P)}
=\frac{\hat q_{\gX}(x)}{p_{\gX}(x)}\cdot \frac{1}{1-\pi(x;P)},
\]
as stated in Theorem~\ref{thm:att}. This verifies Assumption~\ref{asmp:cond-prob-functional}.

\textit{Mixed-bias property and affinity of $\rho$.}
The map $\gamma\mapsto \rho(O,\gamma)$ is affine in $\gamma$, since $\rho(O,\gamma)=(1-D)Y-(1-D)\gamma(X)$.
Moreover, using $\nu_m(x;P,\hat Q)=\hat q_{\gX}(x)/p_{\gX}(x)$ and iterated expectations,
\begin{align*}
\chi_{\att}(P,\hat Q)
&=\E_{\hat Q}[\gamma(X;P)]
=\E_P[\gamma(X;P)\nu_m(X;P,\hat Q)]\\
&=\E_P\!\left[\gamma(X;P)\frac{\hat q_{\gX}(X)}{p_{\gX}(X)}\right]
=\E_P\!\left[\gamma(X;P)(1-\pi(X;P))\alpha(X;P,\hat Q)\right]\\
&=\E_P\!\left[\E_P[(1-D)Y\mid X]\;\alpha(X;P,\hat Q)\right]
=\E_P[(1-D)Y\,\alpha(X;P,\hat Q)].
\end{align*}
Thus $\chi_{\att}$ satisfies the mixed-bias property with $m_2(o,a)=a(x)(1-d)y$, which is linear in $a$.

\paragraph{Verifying Assumption~\ref{asmp:regularity}.}
We check differentiability and local boundedness of the first and second derivatives of $\gamma(\cdot;P)$ and
$\alpha(\cdot;P,\hat Q)$ at $\hat P$.

Let $G$ be a signed measure on $\gO$ with density $g=\dd G/\dd\mu$ such that
$\|g\|_{\mu,\infty}\le 1$ and $\int g\,\dd\mu=0$. For $|t|$ small, define $P_t=\hat P+tG$ with density
$p_t=\hat p+t g$.

\textit{Derivative of $\gamma$.}
For each $x\in\gX$, set
\[
A_t(x):=p_t(x,0,1),\qquad B_t(x):=p_t(x,0,0),\qquad S_t(x):=A_t(x)+B_t(x)=p_{0,\cdot,t}(x).
\]
Then $\gamma(x;P_t)=A_t(x)/S_t(x)$. Since $S_0(x)\ge b_0$ for all $x$ and
$|S_t(x)-S_0(x)|\le 2|t|$ whenever $\|g\|_{\mu,\infty}\le 1$, there exists $t_0>0$ such that
$S_t(x)\ge b_0/2$ for all $x$ and $|t|\le t_0$.
For such $t$, quotient differentiation gives the Gateaux derivative
\[
\gamma_P'(x;\hat P)[G]
=\frac{g(x,0,1)S_0(x)-\hat p(x,0,1)\{g(x,0,1)+g(x,0,0)\}}{S_0(x)^2},
\]
and hence
\[
|\gamma_P'(x;\hat P)[G]|
\le \frac{|g(x,0,1)|S_0(x)+\hat p(x,0,1)(|g(x,0,1)|+|g(x,0,0)|)}{S_0(x)^2}
\le \frac{3b_1}{b_0^2}\|g\|_{\mu,\infty}.
\]
Differentiating again yields, for signed measures $G_1,G_2$ with densities $g_1,g_2$,
\[
\gamma_{PP}''(x;\hat P)[G_1,G_2]
=\frac{2\hat p(x,0,1)\,S_0'(x;G_1)\,S_0'(x;G_2)}{S_0(x)^3}
-\frac{g_1(x,0,1)\,S_0'(x;G_2)+g_2(x,0,1)\,S_0'(x;G_1)}{S_0(x)^2},
\]
where $S_0'(x;G)=g(x,0,1)+g(x,0,0)$, so
$|\gamma_{PP}''(x;\hat P)[G_1,G_2]|\lesssim \|g_1\|_{\mu,\infty}\|g_2\|_{\mu,\infty}$ uniformly in $x$.

\textit{Derivative of $\alpha$.}
Write
\[
\alpha(x;P,\hat Q)=\frac{\hat q_{\gX}(x)}{p_{\gX}(x)(1-\pi(x;P))}.
\]
Let $U_t(x):=p_{\gX,t}(x)=\sum_{d,y}p_t(x,d,y)$ and $V_t(x):=1-\pi_t(x)$, where
$\pi_t(x)=\frac{\sum_y p_t(x,1,y)}{U_t(x)}$.
Then $\alpha(x;P_t,\hat Q)=\hat q_{\gX}(x)/(U_t(x)V_t(x))$.
Since $U_0(x)\ge l$ and $V_0(x)\ge c$ uniformly in $x$, and $U_t,V_t$ vary Lipschitzly in $t$ under the
$d_{\mu,\infty}$-constraint, there exists $t_1>0$ such that $U_t(x)V_t(x)\ge lc/2$ for all $x$ and $|t|\le t_1$.
For such $t$, the product/quotient rule yields a Gateaux derivative
\[
\alpha_P'(x;\hat P,\hat Q)[G]
=-\alpha(x;\hat P,\hat Q)\left\{\frac{U_0'(x;G)}{U_0(x)}+\frac{V_0'(x;G)}{V_0(x)}\right\},
\]
where $U_0'(x;G)=\sum_{d,y}g(x,d,y)$ and
\[
V_0'(x;G)=\frac{\sum_y g(x,0,y)}{U_0(x)}-\frac{\sum_y \hat p(x,0,y)}{U_0(x)^2}\,U_0'(x;G).
\]
All denominators are bounded away from $0$, and $|U_0'|,|V_0'|\lesssim \|g\|_{\mu,\infty}$ pointwise, so
$|\alpha_P'(x;\hat P,\hat Q)[G]|\lesssim \|g\|_{\mu,\infty}$ uniformly in $x$.
A second differentiation gives $\alpha_{PP}''(x;\hat P,\hat Q)[G_1,G_2]$ and the same reasoning shows
$|\alpha_{PP}''(x;\hat P,\hat Q)[G_1,G_2]|\lesssim \|g_1\|_{\mu,\infty}\|g_2\|_{\mu,\infty}$ uniformly in $x$.
These bounds verify Assumption~\ref{asmp:regularity}.

\paragraph{Verifying Assumption~\ref{asmp:main} (invariant directions and non-degenerate mixed curvature).}
We construct explicit perturbations around $\hat P$ with $\hat Q$ held fixed.

By Assumption~\ref{asmp:cond-space-nontrivial}, there exists a measurable set $B\subseteq\gX$ with
$0<\mu_{\gX}(B)<1$. Define the bounded, mean-zero function
\[
\varphi(x):=\1\{x\in B\}-\hat P(X\in B\mid D=0),
\qquad\text{so that}\qquad \E_{\hat P}[\varphi(X)\mid D=0]=0.
\]

\textit{(i) A $\gamma$-invariant direction.}
Define a signed measure $G_0$ with density
\[
g_0(x,d,y):=\varphi(x)\,\hat p(x,0,y)\,\1\{d=0\}.
\]
Then $\int g_0\,\dd\mu=0$ since $\E_{\hat P}[\varphi(X)\mid D=0]=0$ and $\sum_y \hat p(x,0,y)\le \hat p_{\gX}(x)$ is bounded.
For $|t|$ small, $p_t=\hat p+t g_0$ remains nonnegative, so $\hat P+tG_0$ is a probability measure.
Moreover, for every $x$ and $|t|$ small,
\[
p_t(x,0,y)=\hat p(x,0,y)\{1+t\varphi(x)\},\qquad p_t(x,1,y)=\hat p(x,1,y),
\]
so the ratio $p_t(x,0,1)/\sum_y p_t(x,0,y)$ is unchanged. Hence
\[
\gamma(x;\hat P+tG_0)=\gamma(x;\hat P)\qquad\text{for all }x\in\gX\text{ and all }|t|\le c_t
\]
for a sufficiently small constant $c_t>0$, verifying the first part of Assumption~\ref{asmp:main}.

\textit{(ii) A companion direction producing mixed curvature.}
Define a signed measure $G_1$ with density
\[
g_1(x,d,y):=\varphi(x)\,\hat p(x,0,0)\,\1\{d=0\}\{\,\1\{y=1\}-\1\{y=0\}\,\}.
\]
Again $\int g_1\,\dd\mu=0$ since the $y$-terms cancel, and for small $|t|$ the density $\hat p+t g_1$ is nonnegative.
Consider the two-parameter perturbation $P_{s,t}:=\hat P+sG_0+tG_1$ and write
\[
A:=\hat p(x,0,1),\qquad B:=\hat p(x,0,0),\qquad S:=A+B.
\]
Then for each $x$,
\[
p_{s,t}(x,0,1)=A(1+s\varphi(x))+t\varphi(x)B,\qquad
p_{s,t}(x,0,0)=B(1+s\varphi(x))-t\varphi(x)B,
\]
so $p_{s,t}(x,0,1)+p_{s,t}(x,0,0)=S(1+s\varphi(x))$ and therefore
\[
\gamma(x;P_{s,t})
=\frac{A(1+s\varphi(x))+t\varphi(x)B}{S(1+s\varphi(x))}
=\frac{A}{S}+\frac{t\,\varphi(x)\,B}{S(1+s\varphi(x))}.
\]
Consequently,
\[
\chi_{\att}(P_{s,t},\hat Q)
=\E_{\hat Q}\!\left[\frac{A}{S}\right]
+t\,\E_{\hat Q}\!\left[\frac{\varphi(X)\,B(X)}{S(X)\{1+s\varphi(X)\}}\right].
\]
Differentiating at $(s,t)=(0,0)$ yields
\[
\frac{\partial^2}{\partial s\,\partial t}\chi_{\att}(P_{s,t},\hat Q)\Big|_{s=t=0}
=-\E_{\hat Q}\!\left[\frac{\varphi(X)^2\,B(X)}{S(X)}\right].
\]
Since $B/S=P_{\hat P}(Y=0\mid X,D=0)\ge c$ by Condition (1), and $\varphi^2$ is nonzero $\hat Q$-a.s.\ (because $B$
has positive measure under $\mu_{\gX}$ and $\hat q_{\gX}$ is bounded away from $0$), the right-hand side is strictly
negative. Hence $\chi_{PP}''(\hat P,\hat Q)[G_0,G_1]\neq 0$.

\textit{(iii) An $\alpha$-invariant direction and its mixed curvature.}
Set $H_0:=G_1$. Along $P_t=\hat P+tH_0$, we have for each $x$,
$p_t(x,0,1)+p_t(x,0,0)=\hat p(x,0,1)+\hat p(x,0,0)$ and $p_t(x,1,y)=\hat p(x,1,y)$.
Therefore $p_{\gX,t}(x)=\hat p_{\gX}(x)$ and $\pi(x;P_t)=\pi(x;\hat P)$ for all $x$, so
$\alpha(x;P_t,\hat Q)=\alpha(x;\hat P,\hat Q)$ for all $x$ and all $|t|$ small.
Thus $H_0$ is an $\alpha$-invariant direction in the sense of Assumption~\ref{asmp:main}.
	With $H_1:=G_0$, the calculation in (ii) shows
	$\chi_{PP}''(\hat P,\hat Q)[H_0,H_1]=\chi_{PP}''(\hat P,\hat Q)[G_1,G_0]\neq 0$.
	This completes the verification of Assumption~\ref{asmp:main}.

	\textit{$Z_1$-modulation closure.}
	We verify that the perturbation pairs used above are $Z_1$-modulation closed in the sense of
	Definition~\ref{def:z1-modulation-closure}.
	Here $Z_1=X$ (and $Z_2$ is empty), and we take $K_0=K_1=0$ and $L_0=L_1=0$ since the coupled ATT target law
	$Q(\cdot)\allowbreak=P(X\in\cdot\mid D=1)$ is unchanged along our perturbations.
	Indeed, each of $g_0,g_1,h_0,h_1$ is supported on $\{D=0\}$, so for any bounded $\psi:\gX\to\R$ the modulated density
	$\psi(X)g(x,d,y)$ leaves the joint law of $(X,D)$ on $\{D=1\}$ unchanged; consequently $P_t(X\in\cdot\mid D=1)\allowbreak=\hat Q$
	for all sufficiently small $|t|$.
	Moreover, since $|\psi|\le 1$ and the perturbation densities are bounded, there exists $r^{\mathrm{mod}}>0$ (depending
	only on $\|\hat p\|_{\mu,\infty}$ and the $L^\infty$ bounds on $g_0,g_1,h_0,h_1$) such that
	$\hat p+t\,\psi\,g\ge 0$ $\mu$-a.e.\ for all $|t|\le r^{\mathrm{mod}}$.
	Hence, whenever $\psi$ also satisfies the centering condition $\int \psi\,\dd G=0$ (so that $G^\psi(\gO)=0$), the
	pairs $(G,0)$ and $(G^\psi,0)$ are feasible joint perturbations in the coupled ATT class for all $|t|\le r^{\mathrm{mod}}$,
	as required by Definition~\ref{def:z1-modulation-closure}.

	\textit{Conclusion.}
	We have shown that $(\chi_{\att},\rho,m_1)$ satisfies Assumptions~\ref{asmp:density-bounded},
	\ref{asmp:cond-prob-functional}, \ref{asmp:regularity}, and \ref{asmp:main} at the anchor pair $(\hat P,\hat Q)$.
	Since $\rho$ is affine and $\chi_{\att}$ satisfies the mixed-bias property,
Theorem~\ref{thm:main-mixed-bias} yields
\[
\mathfrak{M}^{\chi_{\att}}_{N,\xi}\!\left(\gM\!\left((\hat P,\hat Q);\eps_{N,\gamma},\eps_{N,\alpha}\right)\right)
=\Omega\!\left(\eps_{N,\gamma}\eps_{N,\alpha}+N^{-1/2}\right),
\]
as claimed.
Finally, $\theta_{\att}(P_0)=\E_{P_0}[Y\mid D=1]-\chi_{\att}(P_0,Q_0)$ differs from $\chi_{\att}$ only by the regular term
$\E_{P_0}[Y\mid D=1]$, which is estimable at the parametric rate $N^{-1/2}$ under the same boundedness/overlap
conditions, so the same lower bound applies to estimating $\theta_{\att}$ up to addition of a parametric term.
\end{proof}

\subsection{Proof of Theorem \ref{thm:wad}}
\label{sec:proof:wad}

We verify that the assumptions of Theorem \ref{thm:main-mixed-bias} hold for the weighted average derivative (WAD) estimand in Theorem \ref{thm:wad}.
Throughout, we take $O=(X,D,Y)\in\gO=\gX\times \gD\times\gY$ with $\gD=[0,1]$ and $\gY=\{0,1\}$, and we use the dominating measure
\[
    \mu := \mu_{\gX}\otimes \mathrm{Leb}_{[0,1]}\otimes \mathrm{count}_{\{0,1\}}.
\]
For any $P\ll \mu$, we write $p:=\dd P/\dd\mu$ and similarly $\hat p := \dd \hat P/\dd\mu$, and we use the shorthand
\[
    p(x,d,\cdot):=\sum_{y\in\{0,1\}}p(x,d,y),
    \qquad 
    p_X(x):=\int_0^1 p(x,u,\cdot)\,\dd u,
\]
and analogously for $\hat p$.

\textit{Identifying $(m_1,\rho,\gamma,\alpha)$ and checking affineness of $\rho$.}
We set $Z_1=X\in\gX$, $Z_2=D\in[0,1]$, $W=Y\in\{0,1\}$, and $Z=(Z_1,Z_2)=(X,D)$.
For any $\gM$-feasible $P$, define the outcome regression
\[
    \gamma(z;P)=\gamma(x,d;P):=g(x,d;P):=\E_P[Y\mid X=x,D=d]=\frac{p(x,d,1)}{p(x,d,\cdot)},
\]
the residual
\[
    \rho(o,\gamma):=y-\gamma(x,d),
\]
and the linear functional
\[
    m_1(o,h):=\int_0^1 s(u)\,\omega(u)\,h(x,u)\,\dd u,
    \qquad s(u):=-\frac{\omega'(u)}{\omega(u)}.
\]
Again, $\gamma\mapsto \rho(o,\gamma)$ is affine, so Theorem \ref{thm:main-mixed-bias} is the correct main theorem to invoke.

The Riesz representer for $m_1$ under the $L^2(P_Z)$ inner product is
\begin{equation}
    \label{eq:wad-alpha-def}
    \alpha(z;P)=\alpha(x,d;P)
    :=
    \frac{s(d)\omega(d)}{p(d\mid x;P)}
    =
    -\frac{\omega'(d)}{p(d\mid x;P)},
    \qquad 
    p(d\mid x;P):=\frac{p(x,d,\cdot)}{p_X(x)}.
\end{equation}
We verify the Riesz identity below.

\paragraph{Verifying Assumption \ref{asmp:density-bounded}.}
Theorem \ref{thm:wad} assumes uniform bounds $l_{\hat p}\le \hat p\le u_{\hat p}$ on $\gO$ (and an additional $d$-smoothness condition).
This directly implies Assumption \ref{asmp:density-bounded}.
Moreover, if $P$ satisfies $d_{\mu,\infty}(P,\hat P)\le r$ for $r:=\tfrac12 l_{\hat p}$, then $p\ge \hat p-r\ge \tfrac12 l_{\hat p}$ $\mu$-a.e.\ on $\gO$, so all denominators below remain uniformly bounded away from $0$.

\paragraph{Verifying Assumption \ref{asmp:cond-prob-functional}.}
Both nuisances $\gamma(x,d;P)$ and $\alpha(x,d;P)$ in \eqref{eq:wad-alpha-def} depend on $P$ only through the conditional law of $(D,Y)$ given $X=x$, equivalently through the conditional density $p(d,y\mid x)$.

We now verify the Riesz identity for $m_1$. 
Fix any square-integrable $h:\gZ\to\R$.
By iterated expectation and Fubini's theorem,
\begin{align*}
    \E_P[m_1(O,h)]
    &= \E_P\Big[\int_0^1 s(u)\omega(u)h(X,u)\,\dd u\Big] \\
    &= \int_{\gX}\Big(\int_0^1 s(u)\omega(u)h(x,u)\,\dd u\Big)\,p_X(x)\,\dd \mu_{\gX}(x).
\end{align*}
On the other hand,
\begin{align*}
    \E_P[h(Z)\alpha(Z;P)]
    &= \int_{\gX}\int_0^1 h(x,d)\,\alpha(x,d;P)\,p(x,d,\cdot)\,\dd d\,\dd \mu_{\gX}(x)\\
    &= \int_{\gX}\int_0^1 h(x,d)\,\frac{s(d)\omega(d)}{p(d\mid x;P)}\,\frac{p(x,d,\cdot)}{1}\,\dd d\,\dd \mu_{\gX}(x)\\
    &= \int_{\gX}\int_0^1 h(x,d)\,s(d)\omega(d)\,p_X(x)\,\dd d\,\dd \mu_{\gX}(x)
    = \E_P[m_1(O,h)].
\end{align*}
Finally, as in the ATE case,
\[
    \E_P[\rho(O,\gamma(Z;P)+a)\mid Z=z]= -a,
    \qquad 
    \nu_\rho(z;P)=-1,
\]
so the weighted-Riesz requirements in Assumption \ref{asmp:cond-prob-functional} are satisfied.

\paragraph{Verifying Assumption \ref{asmp:regularity}.}
Fix $r=\tfrac12 l_{\hat p}$ (as in the verification of Assumption \ref{asmp:density-bounded}) and let $P$ satisfy $d_{\mu,\infty}(P,\hat P)\le r$.
Then $p(x,d,\cdot)\ge l_{\hat p}$ and $p_X(x)\ge l_{\hat p}$ for all $(x,d)$, since $p(x,d,\cdot)\ge p(x,d,0)+p(x,d,1)\ge l_{\hat p}$ and $p_X(x)=\int_0^1 p(x,u,\cdot)\dd u\ge l_{\hat p}$.

Let $H$ be any perturbation with density $h=\dd H/\dd\mu$ satisfying $\|h\|_\infty\le C_P$, and define the shorthand
\[
    h(x,d,\cdot):=\sum_{y\in\{0,1\}}h(x,d,y),
    \qquad 
    h_X(x):=\int_0^1 h(x,u,\cdot)\,\dd u.
\]

\textit{Derivative bounds for $\gamma$.}
Write $a=p(x,d,1)$ and $b=p(x,d,0)$ so that $\gamma=a/(a+b)$.
A direct calculation yields, for $\mu$-a.e.\ $(x,d)$,
\begin{equation}
    \label{eq:wad-gamma-derivative}
    \gamma_P'(x,d;P)[H]
    =
    \frac{b\,h(x,d,1)-a\,h(x,d,0)}{(a+b)^2},
\end{equation}
and therefore $|\gamma_P'(x,d;P)[H]|\lesssim (u_{\hat p}C_P)/l_{\hat p}^2$ uniformly over $P$ in the $r$-ball.
The existence and boundedness of $\gamma_P''(z;P)[H,H']$ follow similarly because $\gamma$ is smooth in $(a,b)$ on $\{a+b\ge l_{\hat p}\}$.

\textit{Derivative bounds for $\alpha$.}
Let $q(x,d):=p(x,d,\cdot)$ and $q_X(x):=p_X(x)=\int_0^1 q(x,u)\dd u$.
Then $p(d\mid x;P)=q(x,d)/q_X(x)$ and
\[
    \alpha(x,d;P)=s(d)\omega(d)\frac{q_X(x)}{q(x,d)}.
\]
Since $q,q_X$ are bounded away from $0$ on the $r$-ball, $\alpha$ is smooth in $(q,q_X)$.
Differentiating in the direction $H$ gives, for $\mu$-a.e.\ $(x,d)$,
\begin{align}
    \label{eq:wad-alpha-derivative}
    \alpha_P'(x,d;P)[H]
    &= s(d)\omega(d)\Big\{\frac{h_X(x)}{q(x,d)}-\frac{q_X(x)\,h(x,d,\cdot)}{q(x,d)^2}\Big\}.
\end{align}
Using $|h_X(x)|\le \int_0^1 |h(x,u,\cdot)|\dd u\le C_P$ and $|h(x,d,\cdot)|\le 2C_P$, we obtain the uniform bound
\[
    |\alpha_P'(x,d;P)[H]|
    \le 
    |s(d)\omega(d)|\Big(\frac{C_P}{l_{\hat p}}+\frac{2u_{\hat p}C_P}{l_{\hat p}^2}\Big),
\]
and the existence/boundedness of $\alpha_P''(z;P)[H,H']$ follow by differentiating \eqref{eq:wad-alpha-derivative} once more and using again that $q,q_X$ are bounded away from $0$.

Finally, $\rho(o,\gamma)$ is bounded by $1$, $\nu_\rho\equiv -1$, and $\upsilon_\rho(\cdot;P)\equiv 0$ since $\rho$ is affine in $\gamma$.
This verifies Assumption \ref{asmp:regularity}.

\paragraph{Verifying Assumption \ref{asmp:main} by constructing perturbations.}
We construct $\gM$-feasible perturbations $G_0,G_1,H_0,H_1$ of $\hat P$ satisfying the invariance conditions and the nondegeneracy of the mixed second derivatives.

\textit{Choosing a nontrivial mean-zero function of $d$.}
Since $\omega$ is continuously differentiable on $[0,1]$ and the WAD estimand is nontrivial only when $\omega'$ is not identically zero, we assume $\omega'\not\equiv 0$.
Then there exists an open interval $I\subset(0,1)$ on which $\omega'$ has a constant sign.
Choose any nonconstant $b\in C_c^\infty(I)$ and define $\varphi(d):=b'(d)$.
Then $\varphi\in C_c^\infty(I)$, $\int_0^1 \varphi(d)\dd d = b(1)-b(0)=0$, and $\int_0^1 \omega'(d)\varphi(d)^2\dd d\neq 0$ because $\varphi^2$ is strictly positive on a subset of $I$ and $\omega'$ has constant sign on $I$.

\textit{(a) A $\gamma$-invariant direction $G_0$ and a companion direction $G_1$ with $\chi''(\hat P)[G_0,G_1]\neq 0$.}
Define $\dd G_0=g_0\,\dd\mu$ and $\dd G_1=g_1\,\dd\mu$ by
\begin{equation}
    \label{eq:wad-G0G1-def}
    \begin{aligned}
        g_0(x,d,y) &:= \varphi(d)\frac{\hat p(x,d,y)}{\hat p(x,d,\cdot)},\\
        g_1(x,d,1) &:= \varphi(d)\hat p(x,d,\cdot),\qquad 
        g_1(x,d,0):=-\varphi(d)\hat p(x,d,\cdot).
    \end{aligned}
\end{equation}
First, $\int g_0\,\dd\mu=0$ since $\sum_y g_0(x,d,y)=\varphi(d)$ and $\int_0^1\varphi(d)\dd d=0$.
Also $\int g_1\,\dd\mu=0$ because $g_1(x,d,1)+g_1(x,d,0)=0$ for every $(x,d)$.
Both $g_0$ and $g_1$ are bounded and continuously differentiable in $d$ because $\hat p$ is bounded and $C^1$ in $d$ and $\varphi\in C^\infty$.
Thus, for sufficiently small $|t|$, the perturbed density $\hat p+t g_i$ remains nonnegative, uniformly bounded, and $C^1$ in $d$ with bounded derivative; hence $G_0$ and $G_1$ are $\gM$-feasible perturbations under the $\gM$ defined in Theorem \ref{thm:wad}.

\textit{$\gamma$-invariance along $G_0$.}
For $P_t:=\hat P+tG_0$ we have, for all $(x,d)$,
\[
    p_t(x,d,\cdot)=\hat p(x,d,\cdot)+t\varphi(d),
    \qquad 
    p_t(x,d,1)=\hat p(x,d,1)+t\varphi(d)\frac{\hat p(x,d,1)}{\hat p(x,d,\cdot)}.
\]
Therefore
\[
    \gamma(x,d;P_t)
    =
    \frac{p_t(x,d,1)}{p_t(x,d,\cdot)}
    =
    \frac{\hat p(x,d,1)\big(1+t\varphi(d)/\hat p(x,d,\cdot)\big)}
         {\hat p(x,d,\cdot)\big(1+t\varphi(d)/\hat p(x,d,\cdot)\big)}
    =
    \gamma(x,d;\hat P),
\]
for all sufficiently small $|t|$.

\textit{Computing $\chi''(\hat P)[G_0,G_1]$.}
Let $P_{s,t}:=\hat P+sG_0+tG_1$ and denote the corresponding regression function by $\gamma_{s,t}$.
Since $\sum_y g_0(x,d,y)=\varphi(d)$ and $\int_0^1\varphi(d)\dd d=0$, the marginal of $X$ is unchanged along $G_0$; similarly, $G_1$ does not change the marginal of $X$ because it has zero $y$-sum.
Hence $P_{s,t,X}=\hat P_X$ for all sufficiently small $(s,t)$.

Moreover, for fixed $(x,d)$, the $d$-marginal $p_{s,t}(x,d,\cdot)=\hat p(x,d,\cdot)+s\varphi(d)$ is unaffected by $G_1$ since $g_1(x,d,1)+g_1(x,d,0)=0$, while the numerator $p_{s,t}(x,d,1)$ changes by $t\varphi(d)\hat p(x,d,\cdot)$.
Thus
\[
    \gamma_{s,t}(x,d)
    = \gamma(x,d;\hat P) + t\,\frac{\varphi(d)\hat p(x,d,\cdot)}{\hat p(x,d,\cdot)+s\varphi(d)}.
\]
Plugging this into the WAD functional $\chi_{\wad}(P)=\E_P\big[\int_0^1 s(d)\omega(d)\gamma(X,d;P)\dd d\big]$ yields
\begin{align*}
    \chi_{\wad}(P_{s,t})
    &= \chi_{\wad}(\hat P)
    + t \int_{\gX}\int_0^1 s(d)\omega(d)\,\frac{\varphi(d)\hat p(x,d,\cdot)}{\hat p(x,d,\cdot)+s\varphi(d)}\,\dd d\,\dd \hat P_X(x).
\end{align*}
Expanding $(\hat p+s\varphi)^{-1} = \hat p^{-1} - s\varphi\,\hat p^{-2} + O(s^2)$ gives
\[
    \begin{aligned}
    \chi_{\wad}(P_{s,t})
    = \chi_{\wad}(\hat P)
    &+ t \int_{\gX}\int_0^1 s(d)\omega(d)\varphi(d)\,\dd d\,\dd \hat P_X(x)\\
    &- st \int_{\gX}\int_0^1 s(d)\omega(d)\frac{\varphi(d)^2}{\hat p(x,d,\cdot)}\,\dd d\,\dd \hat P_X(x)\\
    &+ O(s^2t).
    \end{aligned}
\]
The coefficient of $st$ is
\[
    - \int_{\gX}\int_0^1 s(d)\omega(d)\frac{\varphi(d)^2}{\hat p(x,d,\cdot)}\,\dd d\,\dd \hat P_X(x).
\]
By construction, $\varphi$ is supported on an interval where $s(d)\omega(d)=-\omega'(d)$ has a constant nonzero sign, and $\hat p(x,d,\cdot)>0$ everywhere, so this coefficient is nonzero.
Therefore $\chi_{\wad}''(\hat P)[G_0,G_1]\neq 0$.

\textit{(b) An $\alpha$-invariant direction $H_0$ and a companion direction $H_1$ with $\chi''(\hat P)[H_0,H_1]\neq 0$.}
Define $\dd H_0=h_0\,\dd\mu$ and $\dd H_1=h_1\,\dd\mu$ by
\begin{equation}
    \label{eq:wad-H0H1-def}
    \begin{aligned}
        h_0(x,d,1)&:= \varphi(d)\hat p(x,d,\cdot),\qquad h_0(x,d,0):=-\varphi(d)\hat p(x,d,\cdot),\\
        h_1(x,d,y)&:= g_0(x,d,y)\qquad (\text{i.e. }H_1:=G_0).
    \end{aligned}
\end{equation}
We already know $H_1$ is $\gM$-feasible, and $H_0$ is $\gM$-feasible for the same reasons as $G_1$.

\textit{$\alpha$-invariance along $H_0$.}
Along $P_t:=\hat P+tH_0$, we have $p_t(x,d,\cdot)=\hat p(x,d,\cdot)$ for every $(x,d)$, since $h_0(x,d,1)+h_0(x,d,0)=0$.
Consequently $p_t(d\mid x)=\hat p(d\mid x)$ and therefore $\alpha(z;P_t)=\alpha(z;\hat P)$ for all $z$ and all sufficiently small $|t|$.

\textit{Nondegeneracy of $\chi''(\hat P)[H_0,H_1]$.}
Let $P_{s,t}:=\hat P+sH_0+tH_1$.
As in part (a), $P_{s,t,X}=\hat P_X$.
Moreover, since $H_1=G_0$ changes $\hat p(x,d,\cdot)$ by $t\varphi(d)$ and $H_0$ changes only $p(x,d,1)$ by $s\varphi(d)\hat p(x,d,\cdot)$ while keeping $p(x,d,\cdot)$ fixed, we obtain
\[
    \gamma_{s,t}(x,d)
    = \gamma(x,d;\hat P) + s\,\frac{\varphi(d)\hat p(x,d,\cdot)}{\hat p(x,d,\cdot)+t\varphi(d)}.
\]
Repeating the same expansion as in part (a) (with $(s,t)$ swapped) shows that the coefficient of $st$ in $\chi_{\wad}(P_{s,t})$ is again proportional to
\[
    - \int_{\gX}\int_0^1 s(d)\omega(d)\frac{\varphi(d)^2}{\hat p(x,d,\cdot)}\,\dd d\,\dd \hat P_X(x)\neq 0,
\]
hence $\chi_{\wad}''(\hat P)[H_0,H_1]\neq 0$.

\textit{Conclusion.}
The arguments above verify Assumptions \ref{asmp:density-bounded}, \ref{asmp:cond-prob-functional}, \ref{asmp:regularity}, \ref{asmp:cond-space-nontrivial} (assumed in Theorem \ref{thm:wad}) and \ref{asmp:main}. Since $\rho$ is affine in $\gamma$, Theorem \ref{thm:main-mixed-bias} applies and yields the desired lower bound in Theorem \ref{thm:wad}.

\subsection{Proof of Theorem \ref{thm:ape}}
\label{sec:proof:ape}

For this example we write $o=(x,d,y)$ and $z=(x,d)$. For any density $p$ on
$\gO=\gX\times[0,1]\times\{0,1\}$ (w.r.t.~$\mu=\mu_{\gX}\otimes\mathrm{Leb}\otimes\mu_{\{0,1\}}$), we use the shorthand
\[
  p(x,d,\cdot):=p(x,d,0)+p(x,d,1),\qquad 
  p(x,\cdot,\cdot):=\int_0^1 p(x,u,\cdot)\,\dd u.
\]
We denote by $\tau^{-1}$ the inverse of $\tau$ (well-defined since $\tau$ is strictly monotone on $[0,1]$).

\paragraph{Verifying Assumption~\ref{asmp:cond-prob-functional}.}
By definition, $\gamma(z;P)=g(x,d;P)=\E_P[Y\mid X=x,D=d]$ depends on $P$ only through the conditional law $P(\cdot\mid X=x)$ (equivalently, through the conditional density $p(x,\cdot,\cdot)$).

Next, for any square-integrable test function $h:\gX\times[0,1]\to\R$,
\[
  \E_P\big[h\{X,\tau(D)\}\big]
  =\int_{\gX}\int_0^1 h\{x,\tau(d)\}\,p(x,d,\cdot)\,\dd d\,\dd\mu_{\gX}(x).
\]
Assuming $\tau$ is a $C^1$-bijection of $[0,1]$ onto itself, we may change variables $u=\tau(d)$ to obtain
\[
  \E_P\big[h\{X,\tau(D)\}\big]
  =\int_{\gX}\int_0^1 h(x,u)\,\frac{p\{x,\tau^{-1}(u),\cdot\}}{|\tau'|\{\tau^{-1}(u)\}}\,\dd u\,\dd\mu_{\gX}(x).
\]
Therefore,
\begin{equation}
\label{eq:ape-riesz}
  \E_P\big[h\{X,\tau(D)\}-h(X,D)\big]
  =\E_P\big[ h(X,D)\,\nu_m\{X,D;P\}\big],
\end{equation}
where the Riesz representer is
\begin{equation}
\label{eq:ape-nu-m}
  \nu_m(x,d;P)
  :=\frac{p\{x,\tau^{-1}(d),\cdot\}}{|\tau'|\{\tau^{-1}(d)\}\,p(x,d,\cdot)}-1
  \;=\;\frac{p_{\tau}(d\mid x)}{p(d\mid x)}-1,
\end{equation}
and $p_\tau(\cdot\mid x)$ denotes the conditional density of $\tau(D)$ given $X=x$.

Since $\rho(o,\gamma)=y-\gamma(x,d)$ is affine in $\gamma$, we have $\nu_\rho\equiv-1$ and hence
\begin{equation}
\label{eq:ape-alpha}
  \alpha(z;P)=-\frac{\nu_m(z;P)}{\nu_\rho(z;P)}=\nu_m(z;P).
\end{equation}
Finally, $m_1(o,h)=h\{x,\tau(d)\}-h(x,d)$ is linear in $h$ and $\chi_{\ape}(P)=\E_P[m_1\{O,\gamma(Z;P)\}]$ by definition.
This verifies Assumption~\ref{asmp:cond-prob-functional}.

\paragraph{Verifying Assumption~\ref{asmp:regularity}.}
Fix $\hat P$ as in Theorem~\ref{thm:ape} and write $\hat p$ for its density.
Assume in addition that
\begin{equation}
\label{eq:ape-tau-bilipschitz}
  0<\underline\tau\leq |\tau'(d)|\leq \overline\tau<\infty,\qquad \forall d\in[0,1].
\end{equation}
(Equivalently, $\tau$ is bi-Lipschitz and $\tau^{-1}$ is Lipschitz.)

Under the density boundedness assumption $l_{\hat P}\leq \hat p\leq u_{\hat P}$, we have
$\hat p(x,d,\cdot)\ge 2l_{\hat P}$ for all $(x,d)$.
Choose $r:=l_{\hat P}/2$.
Then for any $\gM$-feasible $P$ with $d_{\mu,\infty}(P,\hat P)\le r$, we have
$p(x,d,y)\ge l_{\hat P}/2$ and hence $p(x,d,\cdot)\ge l_{\hat P}$ for all $(x,d)$.

Let $H$ be a feasible perturbation with density $h=\dd H/\dd\mu$ satisfying $\|h\|_{\mu,\infty}\le C_P$.
A direct quotient-rule calculation gives, for $z=(x,d)$,
\begin{align}
\label{eq:ape-gamma-derivs}
  \gamma_P'(z;P)[H]
  &=\frac{h(x,d,1)}{p(x,d,\cdot)}-\frac{p(x,d,1)\,h(x,d,\cdot)}{p(x,d,\cdot)^2},\\
  \gamma_P''(z;P)[H,H']
  &= -\frac{h(x,d,1)h'(x,d,\cdot)+h'(x,d,1)h(x,d,\cdot)}{p(x,d,\cdot)^2}\\
  &\quad +\frac{2p(x,d,1)\,h(x,d,\cdot)h'(x,d,\cdot)}{p(x,d,\cdot)^3}.
\end{align}
Using $p(x,d,\cdot)\ge l_{\hat P}$ and $|h|,|h'|\le C_P$, we obtain uniform bounds
\[
|\gamma_P'(z;P)[H]|\le 3C_P l_{\hat P}^{-1},
\qquad
|\gamma_P''(z;P)[H,H']|\le 4C_P^2 l_{\hat P}^{-2}.
\]

For $\alpha$, using \eqref{eq:ape-alpha}--\eqref{eq:ape-nu-m} we can write
\[
  \alpha(x,d;P)=\frac{p\{x,\tau^{-1}(d),\cdot\}}{|\tau'|\{\tau^{-1}(d)\}\,p(x,d,\cdot)}-1.
\]
Let $H,H'$ be feasible perturbations with densities $h,h'$.
Since $\tau$ is fixed and does not depend on $P$, we have
\begin{align}
\label{eq:ape-alpha-derivs}
  \alpha_P'(x,d;P)[H]
  &= \frac{h\{x,\tau^{-1}(d),\cdot\}}{|\tau'|\{\tau^{-1}(d)\}\,p(x,d,\cdot)}
     -\frac{p\{x,\tau^{-1}(d),\cdot\}\,h(x,d,\cdot)}{|\tau'|\{\tau^{-1}(d)\}\,p(x,d,\cdot)^2},\\
  \alpha_P''(x,d;P)[H,H']
  &= -\frac{h\{x,\tau^{-1}(d),\cdot\}h'(x,d,\cdot)+h'\{x,\tau^{-1}(d),\cdot\}h(x,d,\cdot)}{|\tau'|\{\tau^{-1}(d)\}\,p(x,d,\cdot)^2} \\
  &\hspace{2em}
     +\frac{2p\{x,\tau^{-1}(d),\cdot\}\,h(x,d,\cdot)h'(x,d,\cdot)}{|\tau'|\{\tau^{-1}(d)\}\,p(x,d,\cdot)^3}.
\end{align}
Combining $p(x,d,\cdot)\ge l_{\hat P}$ with \eqref{eq:ape-tau-bilipschitz} yields the uniform bounds
$|\alpha_P'(x,d;P)[H]|\le 2\underline\tau^{-1} C_P l_{\hat P}^{-1}$ and
$|\alpha_P''(x,d;P)[H,H']|\le 4\underline\tau^{-1} C_P^2 l_{\hat P}^{-2}$.

Moreover, since $Y\in\{0,1\}$ and $\gamma\in[0,1]$, we have
$|\rho(o,\gamma)|=|y-\gamma(x,d)|\le 1$.
Because $\rho$ is affine in $\gamma$, $\nu_{\rho}\equiv-1$ and $\upsilon_{\rho}\equiv 0$.
Therefore Assumption~\ref{asmp:regularity} holds.

\paragraph{Verifying Assumption~\ref{asmp:main}.}
We now construct perturbations $G_0,G_1,H_0,H_1$ at $\hat P$.

\smallskip
\textit{A $\gamma$-invariant direction $G_0$.}
Let $\phi$ be a bounded (e.g.~smooth) function on $[0,1]$ satisfying $\int_0^1\phi(u)\,\dd u=0$.
Define the perturbation density
\begin{equation}
\label{eq:ape-g0}
  g_0(x,d,y):=\phi(d)\,\frac{\hat p(x,d,y)}{\hat p(x,d,\cdot)}.
\end{equation}
Then $g_0(x,d,\cdot)=\phi(d)$, and hence $\int g_0\,\dd\mu = 0$ because $\int_0^1\phi(u)\,\dd u=0$.
Thus $G_0$ is a valid perturbation. Moreover, for any $s$ such that $\hat p+s g_0\ge 0$,
\[
  \gamma\{x,d;\hat P+sG_0\}
  =\frac{\hat p(x,d,1)+s g_0(x,d,1)}{\hat p(x,d,\cdot)+s g_0(x,d,\cdot)}
  =\frac{\hat p(x,d,1)}{\hat p(x,d,\cdot)}
  =\gamma(x,d;\hat P),
\]
so $\gamma(\cdot;\hat P+sG_0)$ is exactly invariant along $G_0$.

\smallskip
\textit{Choosing $G_1$ so that $\chi_{\ape}''(\hat P)[G_0,G_1]\neq 0$.}
Let $\psi$ be a bounded measurable function on $\gX\times[0,1]$ and define
\begin{equation}
\label{eq:ape-g1}
  g_1(x,d,1):=\psi(x,d),\qquad g_1(x,d,0):=-\psi(x,d).
\end{equation}
Then $g_1(x,d,\cdot)=0$ for all $(x,d)$, so perturbing along $G_1$ does not change
$p(x,d,\cdot)$ and hence does not change $\alpha$ (cf.~\eqref{eq:ape-alpha}).

For $s,t$ small, write $P_{s,t}:=\hat P+sG_0+tG_1$.
Since $g_1(x,d,\cdot)=0$ we have
$p_{s,t}(x,d,\cdot)=\hat p(x,d,\cdot)+s\phi(d)$.
A direct calculation yields, for every $(x,d)$,
\begin{equation}
\label{eq:ape-gamma-st}
  \gamma\{x,d;P_{s,t}\}=\gamma(x,d;\hat P)+t\,\frac{g_1(x,d,1)}{\hat p(x,d,\cdot)+s\phi(d)}.
\end{equation}
Using the definition $\chi_{\ape}(P)=\E_P\big[\gamma\{X,\tau(D);P\}-\gamma(X,D;P)\big]$ and the fact that $p_{s,t}(x,d,\cdot)$ is the $(X,D)$-marginal density of $P_{s,t}$, we can write
\begin{equation}
\label{eq:ape-chi-st}
  \chi_{\ape}(P_{s,t})
  =\int_{\gX}\int_0^1 \Big(\gamma\{x,\tau(d);P_{s,t}\}-\gamma\{x,d;P_{s,t}\}\Big)\,\big\{\hat p(x,d,\cdot)+s\phi(d)\big\}\,\dd d\,\dd\mu_{\gX}(x).
\end{equation}
Differentiate \eqref{eq:ape-chi-st} with respect to $t$ at $t=0$ and use \eqref{eq:ape-gamma-st}:
\[
  \partial_t\chi_{\ape}(P_{s,t})\big|_{t=0}
  =\int_{\gX}\int_0^1
  \Bigg[\frac{g_1\{x,\tau(d),1\}}{\hat p\{x,\tau(d),\cdot\}+s\phi\{\tau(d)\}}
        -\frac{g_1(x,d,1)}{\hat p(x,d,\cdot)+s\phi(d)}\Bigg]\,\big\{\hat p(x,d,\cdot)+s\phi(d)\big\}\,\dd d\,\dd\mu_{\gX}(x).
\]
The second term inside the brackets cancels with the factor $\hat p(x,d,\cdot)+s\phi(d)$.
Hence the mixed derivative at $(s,t)=(0,0)$ is
\begin{align}
\label{eq:ape-mixed-deriv}
  \chi_{\ape}''(\hat P)[G_0,G_1]
  &=\partial_s\partial_t\chi_{\ape}(P_{s,t})\big|_{(s,t)=(0,0)}\\
  &=\int_{\gX}\int_0^1 g_1\{x,\tau(d),1\}\,\frac{\phi(d)\,\hat p\{x,\tau(d),\cdot\}-\phi\{\tau(d)\}\,\hat p(x,d,\cdot)}{\hat p\{x,\tau(d),\cdot\}^2}\,\dd d\,\dd\mu_{\gX}(x).
\end{align}
(Justification: the integrand is uniformly bounded in a neighborhood of $s=0$ since $\hat p$ is bounded away from $0$, hence we may differentiate under the integral sign by dominated convergence.)

Now choose
\begin{equation}
\label{eq:ape-psi-choice}
  \psi(x,d)
  :=\lambda\Big(\phi\{\tau^{-1}(d)\}\,\hat p(x,d,\cdot)-\phi(d)\,\hat p\{x,\tau^{-1}(d),\cdot\}\Big),
\end{equation}
with $\lambda>0$ small enough so that $\|g_1\|_{\mu,\infty}$ is uniformly bounded and $\hat p+t g_1\ge 0$ for all $|t|\le c_t$.
Then $g_1\{x,\tau(d),1\}=\psi\{x,\tau(d)\}=\lambda\big(\phi(d)\hat p\{x,\tau(d),\cdot\}-\phi\{\tau(d)\}\hat p(x,d,\cdot)\big)$.
Plugging this into \eqref{eq:ape-mixed-deriv} gives
\begin{equation}
\label{eq:ape-mixed-square}
  \begin{aligned}
  \chi_{\ape}''(\hat P)[G_0,G_1]
  &=\lambda\int_{\gX}\int_0^1
  \frac{\big(\phi(d)\hat p\{x,\tau(d),\cdot\}-\phi\{\tau(d)\}\hat p(x,d,\cdot)\big)^2}{\hat p\{x,\tau(d),\cdot\}^2}\\
  &\qquad\qquad\qquad\qquad\qquad\qquad\qquad\qquad\,\dd d\,\dd\mu_{\gX}(x).
  \end{aligned}
\end{equation}
This quantity is strictly positive provided the continuous function
\[
(x,d)\mapsto \phi(d)\hat p\{x,\tau(d),\cdot\}-\phi\{\tau(d)\}\hat p(x,d,\cdot)
\]
is not identically zero.
If $\tau$ is not the identity map, such a $\phi$ exists:
pick $d_0\in(0,1)$ with $\tau(d_0)\neq d_0$ and choose disjoint neighborhoods $U$ of $d_0$ and $V$ of $\tau(d_0)$.
Let $\phi_1$ be a smooth bump supported on $U$ with $\phi_1(d_0)=1$ and let $\phi_2$ be a smooth function supported on $[0,1]\setminus(U\cup V)$ with $\int_0^1\phi_2\neq 0$.
Setting $\phi:=\phi_1-c\phi_2$ with $c$ chosen so that $\int_0^1\phi=0$, we have $\phi(d_0)=1$ and $\phi\{\tau(d_0)\}=0$, so the integrand in \eqref{eq:ape-mixed-square} is strictly positive on a set of positive $\mu$-measure.
Hence $\chi_{\ape}''(\hat P)[G_0,G_1]\neq 0$.

\smallskip
\textit{Constructing $H_0,H_1$ for the $\alpha$-invariance condition.}
Since $g_1(x,d,\cdot)=0$, the $(X,D)$-marginal density $p(x,d,\cdot)$ is unchanged along $G_1$, and therefore $\alpha(z;\hat P+tG_1)=\alpha(z;\hat P)$ for all $z$ and all small $t$.
Consequently, we may take
\[
  H_0:=G_1,\qquad H_1:=G_0.
\]
Then Assumption~\ref{asmp:main}(1) holds for $H_0$ and
$\chi_{\ape}''(\hat P)[H_0,H_1]=\chi_{\ape}''(\hat P)[G_1,G_0]\neq 0$ by \eqref{eq:ape-mixed-square}.
This completes the Verifying Assumption~\ref{asmp:main}.

\textit{Conclusion.}
All assumptions required to invoke Theorem~\ref{thm:main-mixed-bias} are satisfied for $\chi_{\ape}$, and since $\rho$ is affine in $\gamma$ we obtain the claimed
$\Omega(\eps_{n,\gamma}\eps_{n,\alpha}+1/\sqrt{n})$ lower bound.

\subsection{Proof of Theorem \ref{thm:plm-numerator}}
\label{sec:proof:plm-numerator}

Write $O=(X,T,Y)$ with $X\in\gX=[0,1]^K$ and $(T,Y)\in\{0,1\}^2$, and let $\mu:=\mu_X\otimes\mu_{\{0,1\}}\otimes\mu_{\{0,1\}}$ where $\mu_X$ is Lebesgue measure on $\gX$ and $\mu_{\{0,1\}}$ is counting measure.
Let $\hat P\in\mathcal P_{\mathrm{PLM}}$ be the anchor distribution in Section~\ref{subsec:ecc} with density $\hat p$.

\textit{Reduce $\chi_{\ecc}$ to an affine-score functional.}
Recall from \eqref{eq:ecc-plm-num} that
\[
\chi_{\ecc}(P)=\mathbb E_P\Big[(T-g(X;P))(Y-q(X;P))\Big].
\]
Using $\mathbb E_P[T-g(X;P)\mid X]=0$ and the law of total expectation,
\begin{align}
\chi_{\ecc}(P)
&=\mathbb E_P\Big[(T-g(X;P))\,Y\Big]-\mathbb E_P\Big[(T-g(X;P))\,q(X;P)\Big]\notag\\
&=\mathbb E_P[TY]-\mathbb E_P\Big[g(X;P)\,Y\Big]\label{eq:plm-num-decomp}\\
&=\mathbb E_P[TY]-\widetilde\chi(P),\notag
\end{align}
where we define the auxiliary functional
\[
\widetilde\chi(P):=\mathbb E_P\big[Y\,g(X;P)\big]=\mathbb E_P\big[g(X;P)\,q(X;P)\big].
\]
Since $TY\in[0,1]$, the empirical mean $\mathbb P_n[TY]$ achieves the parametric rate $n^{-1/2}$ uniformly over $P$.
Therefore it suffices to prove the mixed-bias lower bound for $\widetilde\chi$; combining with~\eqref{eq:plm-num-decomp} then yields~\eqref{eq:plm-num-lb}.

Let $Z_1:=X$, let $Z_2$ be trivial, and let $W:=(T,Y)$.
For any $P\ll\mu$, define
\[
\gamma(x;P):=g(x;P)=\mathbb P_P(T=1\mid X=x),
\qquad 
\alpha(x;P):=q(x;P)=\mathbb E_P[Y\mid X=x].
\]
Define the regression score and linear functional
\[
\rho\{o,\gamma\}:=t-\gamma(x),
\qquad 
m_1\{o,h\}:=y\,h(x),
\qquad o=(x,t,y).
\]
Then $\mathbb E_P[\rho\{O,\gamma(Z;P)\}\mid Z]=0$ and $\partial\rho/\partial\gamma\equiv-1$, so $\nu_\rho\equiv-1$.
Moreover, for any square-integrable test function $h$,
\[
\mathbb E_P[m_1\{O,h(Z)\}]
=\mathbb E_P\big[Y\,h(X)\big]
=\mathbb E_P\big[q(X;P)\,h(X)\big],
\]
so the unweighted Riesz representer is $\nu_m(x;P)=q(x;P)$ and the weighted representer is
$\alpha(x;P)=-\nu_m(x;P)/\nu_\rho(x;P)=q(x;P)$ as claimed.
Finally, the map $\gamma\mapsto\rho(o,\gamma)=t-\gamma(x)$ is affine, so $\widetilde\chi$ is in the affine-score regime of Theorem~\ref{thm:main-mixed-bias}.
It remains to verify the perturbation condition in Assumption~\ref{asmp:main} within the PLM model.

\textit{PLM-preserving perturbation of anchor distribution.}
We now construct a two-parameter family of \emph{PLM-feasible} perturbations of $\hat p$ that is compatible with the
sign-flip hypercube construction used in the other examples.

\smallskip
\noindent\textit{(a) A $\pm1$ bump function.}
Fix an integer $M\ge1$ and partition $\gX$ into $2M$ measurable sets $B_1,\dots,B_{2M}$ of equal $\mu_X$-measure.
For $\lambda=(\lambda_1,\dots,\lambda_M)\in\{-1,+1\}^M$, define
\[
\Delta(\lambda,x):=\sum_{j=1}^M \lambda_j\Big(\mathbf 1\{x\in B_{2j-1}\}-\mathbf 1\{x\in B_{2j}\}\Big),
\]
so that $\Delta(\lambda,x)\in\{-1,+1\}$ for all $x$ and hence $\Delta(\lambda,x)^2\equiv 1$.

\smallskip
\noindent\textit{(b) Define the perturbed density.}
Let $\hat g(x):=g(x;\hat P)$ and $\hat q(x):=q(x;\hat P)$, and write
\[
s(x):=\sqrt{\hat g(x)\big(1-\hat g(x)\big)}.
\]
For scalars $u,v$ and each $\lambda$, define the constant
\begin{equation}\label{eq:plm-pert-theta}
\theta^{u,v}:=\frac{\hat\theta+u v}{1-u^2}.
\end{equation}
Define a density $p_{\lambda}^{u,v}$ on $\gO=\gX\times\{0,1\}^2$ by setting, for each $x\in\gX$,
\begin{align}
p_{\lambda}^{u,v}(x,1,1)
&:=\hat p(x,1,1)+\Big(u\,\hat q(x)-v\,\hat g(x)+\theta^{u,v}u\big(1-2\hat g(x)\big)\Big)\,s(x)\,\Delta(\lambda,x),\label{eq:plm-pert-p11}\\
p_{\lambda}^{u,v}(x,1,0)
&:=\hat p(x,1,0)+\Big(u\,(1-\hat q(x))+v\,\hat g(x)-\theta^{u,v}u\big(1-2\hat g(x)\big)\Big)\,s(x)\,\Delta(\lambda,x),\label{eq:plm-pert-p10}\\
p_{\lambda}^{u,v}(x,0,1)
&:=\hat p(x,0,1)-\Big(u\,\hat q(x)+v\,(1-\hat g(x))+\theta^{u,v}u\big(1-2\hat g(x)\big)\Big)\,s(x)\,\Delta(\lambda,x),\label{eq:plm-pert-p01}\\
p_{\lambda}^{u,v}(x,0,0)
&:=\hat p(x,0,0)+\Big(u\,(\hat q(x)-1)+v\,(1-\hat g(x))+\theta^{u,v}u\big(1-2\hat g(x)\big)\Big)\,s(x)\,\Delta(\lambda,x).\label{eq:plm-pert-p00}
\end{align}

\smallskip
\noindent\textit{(c) Validity as a density.}
Summing~\eqref{eq:plm-pert-p11}--\eqref{eq:plm-pert-p00} over $(t,y)\in\{0,1\}^2$ cancels all perturbation terms, so for every $x$,
\[
\sum_{t,y}p_{\lambda}^{u,v}(x,t,y)=\sum_{t,y}\hat p(x,t,y)=\hat p_X(x).
\]
Since $\hat P$ has $X\sim\mathrm{Unif}(\gX)$, we have $\hat p_X(x)=1$ and thus $\int p_{\lambda}^{u,v}\,\dd\mu=1$.
Moreover, by the bounded-density assumption in Theorem~\ref{thm:plm-numerator}, $\hat p(x,t,y)\ge c_0>0$ uniformly.
Because $\hat g\in[c,1-c]$, we have $\sup_x s(x)\le 1/2$.
Also, for $|u|\le 1/2$ we have $|\theta^{u,v}|\le 2(|\hat\theta|+|u v|)$, hence $|\theta^{u,v}|$ is bounded uniformly for small $|u|,|v|$.
Consequently, there exists $\delta>0$ such that whenever $|u|,|v|\le \delta$ we have
$|p_{\lambda}^{u,v}(x,t,y)-\hat p(x,t,y)|\le c_0/2$ for all $(x,t,y)$ and hence $p_{\lambda}^{u,v}(x,t,y)\ge 0$.
Thus $p_{\lambda}^{u,v}$ is a valid joint density.

\smallskip
\noindent\textit{(d) Induced nuisance perturbations and invariance.}
Let $P_{\lambda}^{u,v}$ denote the distribution with density $p_{\lambda}^{u,v}$.
Summing~\eqref{eq:plm-pert-p11}--\eqref{eq:plm-pert-p10} over $y$ yields
\[
p_{\lambda}^{u,v}(x,1,\cdot)=\hat p(x,1,\cdot)+u\,s(x)\,\Delta(\lambda,x),
\]
and since $p_{\lambda}^{u,v}(x,\cdot,\cdot)=\hat p_X(x)=1$, we obtain
\begin{equation}\label{eq:plm-nuis-g-pert}
g(x;P_{\lambda}^{u,v})=\hat g(x)+u\,s(x)\,\Delta(\lambda,x).
\end{equation}
Similarly, summing~\eqref{eq:plm-pert-p11}--\eqref{eq:plm-pert-p01} over $t$ yields
\[
p_{\lambda}^{u,v}(x,\cdot,1)=\hat p(x,\cdot,1)-v\,s(x)\,\Delta(\lambda,x),
\]
so
\begin{equation}\label{eq:plm-nuis-q-pert}
q(x;P_{\lambda}^{u,v})=\hat q(x)-v\,s(x)\,\Delta(\lambda,x).
\end{equation}
In particular:
\begin{itemize}
\item (\emph{$\gamma$-invariance}) if $u=0$ then $g(\cdot;P_{\lambda}^{0,v})\equiv \hat g(\cdot)$ for all $v$;
\item (\emph{$\alpha$-invariance}) if $v=0$ then $q(\cdot;P_{\lambda}^{u,0})\equiv \hat q(\cdot)$ for all $u$.
\end{itemize}

\smallskip
\noindent\textit{(e) PLM feasibility (factorization check).}
Define the induced conditional mean functions
\[
g_{\lambda}^{u}(x):=g(x;P_{\lambda}^{u,v}),\qquad q_{\lambda}^{v}(x):=q(x;P_{\lambda}^{u,v}),
\qquad f_{\lambda}^{u,v}(x):=q_{\lambda}^{v}(x)-\theta^{u,v}g_{\lambda}^{u}(x).
\]
We now verify that, under $P_{\lambda}^{u,v}$, $Y\mid(T=t,X=x)$ is Bernoulli with mean $f_{\lambda}^{u,v}(x)+\theta^{u,v}t$.
It suffices to show that the cell probability $p_{\lambda}^{u,v}(x,1,1)$ factorizes as
\begin{equation}\label{eq:plm-factor-p11}
p_{\lambda}^{u,v}(x,1,1)
=
g_{\lambda}^{u}(x)\Big(q_{\lambda}^{v}(x)+\theta^{u,v}\big(1-g_{\lambda}^{u}(x)\big)\Big),
\end{equation}
since then
\[
\mathbb P_{P_{\lambda}^{u,v}}(Y=1\mid T=1,X=x)
=
\frac{p_{\lambda}^{u,v}(x,1,1)}{p_{\lambda}^{u,v}(x,1,\cdot)}
=
q_{\lambda}^{v}(x)+\theta^{u,v}\big(1-g_{\lambda}^{u}(x)\big)
=
f_{\lambda}^{u,v}(x)+\theta^{u,v},
\]
and similarly $\mathbb P(Y=1\mid T=0,X=x)=f_{\lambda}^{u,v}(x)$.
To prove~\eqref{eq:plm-factor-p11}, use~\eqref{eq:plm-nuis-g-pert}--\eqref{eq:plm-nuis-q-pert} to write
$g_{\lambda}^{u}(x)=\hat g(x)+u s(x)\Delta(\lambda,x)$ and $q_{\lambda}^{v}(x)=\hat q(x)-v s(x)\Delta(\lambda,x)$.
Expanding the right-hand side of~\eqref{eq:plm-factor-p11} and using $\Delta(\lambda,x)^2=1$ gives
\begin{align*}
&g_{\lambda}^{u}(x)\Big(q_{\lambda}^{v}(x)+\theta^{u,v}\big(1-g_{\lambda}^{u}(x)\big)\Big)\\
&\qquad
=
\hat g(x)\Big(\hat q(x)+\hat\theta\big(1-\hat g(x)\big)\Big)
+\Big(u\,\hat q(x)-v\,\hat g(x)+\theta^{u,v}u\big(1-2\hat g(x)\big)\Big)s(x)\Delta(\lambda,x),
\end{align*}
where we used the identity $\theta^{u,v}(1-u^2)=\hat\theta+u v$ to simplify the $\Delta$-free term.
Finally, since $\hat P$ is PLM with slope $\hat\theta$, we have
$\hat p(x,1,1)=\hat g(x)\big(\hat q(x)+\hat\theta(1-\hat g(x))\big)$,
and comparing with~\eqref{eq:plm-pert-p11} proves~\eqref{eq:plm-factor-p11}.
Therefore $P_{\lambda}^{u,v}\in\mathcal P_{\mathrm{PLM}}$ with constant slope $\theta^{u,v}$.

\smallskip
\noindent\textit{(f) Matching the neighborhood radii.}
Since $\Delta(\lambda,X)^2\equiv1$ and $X\sim\mathrm{Unif}(\gX)$, we have
\[
\begin{aligned}
\|g(X;P_{\lambda}^{u,v})-\hat g(X)\|_{2}
&=|u|\,\Big\{\mathbb E_{\mu_X}\big[\hat g(X)\big(1-\hat g(X)\big)\big]\Big\}^{1/2},\\
\|q(X;P_{\lambda}^{u,v})-\hat q(X)\|_{2}
&=|v|\,\Big\{\mathbb E_{\mu_X}\big[\hat g(X)\big(1-\hat g(X)\big)\big]\Big\}^{1/2}.
\end{aligned}
\]
Thus, for any prescribed radii $\eps_{n,g},\eps_{n,q}$ we may choose $u,v$ of order $\eps_{n,g},\eps_{n,q}$ so that
\[
P_{\lambda}^{u,v}\in\gM_{\mathrm{PLM}}(\hat P;\eps_{n,g},\eps_{n,q}).
\]

\textit{Nondegenerate mixed second derivative.}
For any $(u,v,\lambda)$, since $X$ is uniform,
\[
\widetilde\chi(P_{\lambda}^{u,v})
=\mathbb E_{P_{\lambda}^{u,v}}\big[Y\,g(X;P_{\lambda}^{u,v})\big]
=\mathbb E_{\mu_X}\big[g_{\lambda}^{u}(X)\,q_{\lambda}^{v}(X)\big].
\]
Using~\eqref{eq:plm-nuis-g-pert}--\eqref{eq:plm-nuis-q-pert} and $\Delta(\lambda,X)^2\equiv1$,
\[
\widetilde\chi(P_{\lambda}^{u,v})
=\widetilde\chi(\hat P)
+u\,A_\lambda
-v\,B_\lambda
-u v\,\mathbb E_{\mu_X}\big[\hat g(X)\big(1-\hat g(X)\big)\big],
\]
where
$A_\lambda:=\mathbb E_{\mu_X}\big[\hat q(X)s(X)\Delta(\lambda,X)\big]$
and
$B_\lambda:=\mathbb E_{\mu_X}\big[\hat g(X)s(X)\Delta(\lambda,X)\big]$.
Therefore the mixed second derivative at the origin is
\begin{equation}\label{eq:plm-cross-deriv}
\left.\frac{\partial^2}{\partial u\,\partial v}\widetilde\chi(P_{\lambda}^{u,v})\right|_{(0,0)}
=
-\mathbb E_{\mu_X}\big[\hat g(X)\big(1-\hat g(X)\big)\big],
\end{equation}
which is nonzero (and bounded away from $0$) under the overlap condition $c\le \hat g\le 1-c$.

\textit{Apply Theorem~\ref{thm:main-mixed-bias} and conclude.}
The arguments above verify Assumption~\ref{asmp:main} for the affine-score functional $\widetilde\chi$.
Moreover, the lower bound argument underlying Theorem~\ref{thm:main-mixed-bias} is based on evaluating the risk over a finite hypercube of alternatives,
and we have verified that the corresponding alternatives $\{P_{\lambda}^{u,v}\}$ lie inside the PLM-restricted neighborhood
$\gM_{\mathrm{PLM}}(\hat P;\eps_{n,g},\eps_{n,q})$.
Therefore Theorem~\ref{thm:main-mixed-bias} yields
\[
\mathfrak{M}_{n,\xi}^{\widetilde\chi}\Big(\gM_{\mathrm{PLM}}(\hat P;\eps_{n,g},\eps_{n,q})\Big)
=
\Omega\Big(\eps_{n,g}\eps_{n,q}+\frac{1}{\sqrt n}\Big).
\]
Combining this with the decomposition~\eqref{eq:plm-num-decomp} and the parametric estimability of $\mathbb E_P[TY]$
establishes~\eqref{eq:plm-num-lb} for $\chi_{\ecc}$.

\subsection{Proof of Theorem \ref{thm:ds}}
\label{sec:proof:ds}

For this example we write $o=(x,y)$ and $z=x$. For any density $p$ on
$\gO=\gX\times\{0,1\}$ (w.r.t.~$\mu=\mu_{\gX}\otimes\mu_{\{0,1\}}$), we use the shorthand
\[
  p(x,\cdot):=p(x,0)+p(x,1),
\]
so that $p(x,\cdot)$ is the $X$-marginal density of $P$ at $x$.

\paragraph{Verifying Assumption~\ref{asmp:cond-prob-functional}.}
By definition,
\[
  \gamma(x;P)=\E_P[Y\mid X=x]=\frac{p(x,1)}{p(x,\cdot)}.
\]
Hence $\gamma(\cdot;P)$ depends on $P$ only through the conditional law $P(\cdot\mid X=x)$ (equivalently, through the function $y\mapsto p(x,y)$).
Moreover, for any square-integrable test function $h:\gX\to\R$,
\[
  \E_P\big[m_1\{O,h(Z)\}\big]
  =\int_{\gX} h(x)\,\{f_2(x)-f_1(x)\}\,\dd\mu_{\gX}(x),
\]
since $m_1(o,h)=\int h(x)\{f_2(x)-f_1(x)\}\dd\mu_{\gX}(x)$ does not depend on $o$.
Writing $f(x):=p(x,\cdot)$ for the $X$-marginal density under $P$, we can also express the same functional as
\[
  \int_{\gX} h(x)\,\{f_2(x)-f_1(x)\}\,\dd\mu_{\gX}(x)
  =\E_P\Big[h(X)\,\frac{f_2(X)-f_1(X)}{f(X)}\Big],
\]
so the (unweighted) Riesz representer is
\[
  \nu_m(x;P)=\frac{f_2(x)-f_1(x)}{p(x,\cdot)}.
\]
With $\rho(o,\gamma)=y-\gamma(x)$ we have $\nu_\rho\equiv-1$ and therefore
\[
  \alpha(x;P)=-\frac{\nu_m(x;P)}{\nu_\rho(x;P)}=\nu_m(x;P)=\frac{f_2(x)-f_1(x)}{p(x,\cdot)}.
\]
This verifies Assumption~\ref{asmp:cond-prob-functional}.

\paragraph{Verifying Assumption~\ref{asmp:regularity}.}
Fix $\hat P$ as in Theorem~\ref{thm:ds} and write $\hat p$ for its density.
Under the density boundedness assumption $l_{\hat P}\le \hat p\le u_{\hat P}$, we have
$\hat p(x,\cdot)\ge 2l_{\hat P}$.
Choose $r:=l_{\hat P}/2$.
Then for any $\gM$-feasible $P$ with $d_{\mu,\infty}(P,\hat P)\le r$, we have
$p(x,y)\ge l_{\hat P}/2$ and hence $p(x,\cdot)\ge l_{\hat P}$ for all $x$.

Let $H$ be a feasible perturbation with density $h=\dd H/\dd\mu$ satisfying $\|h\|_{\mu,\infty}\le C_P$ and write $h(x,\cdot):=h(x,0)+h(x,1)$.
A quotient-rule calculation yields, for each $x$,
\begin{align}
\label{eq:dis-gamma-derivs}
  \gamma_P'(x;P)[H]
  &=\frac{h(x,1)}{p(x,\cdot)}-\frac{p(x,1)\,h(x,\cdot)}{p(x,\cdot)^2},\\
  \gamma_P''(x;P)[H,H']
  &= -\frac{h(x,1)h'(x,\cdot)+h'(x,1)h(x,\cdot)}{p(x,\cdot)^2}
     +\frac{2p(x,1)\,h(x,\cdot)h'(x,\cdot)}{p(x,\cdot)^3}.
\end{align}
Using $p(x,\cdot)\ge l_{\hat P}$ and $|h|,|h'|\le C_P$, we obtain the uniform bounds
$|\gamma_P'(x;P)[H]|\le 3C_P l_{\hat P}^{-1}$ and $|\gamma_P''(x;P)[H,H']|\le 4C_P^2 l_{\hat P}^{-2}$.

Next, since $f_1,f_2$ are fixed and bounded and $\alpha(x;P)=\{f_2(x)-f_1(x)\}/p(x,\cdot)$, we have
\begin{align}
\label{eq:dis-alpha-derivs}
  \alpha_P'(x;P)[H]
  &= -\{f_2(x)-f_1(x)\}\,\frac{h(x,\cdot)}{p(x,\cdot)^2},\\
  \alpha_P''(x;P)[H,H']
  &= 2\{f_2(x)-f_1(x)\}\,\frac{h(x,\cdot)h'(x,\cdot)}{p(x,\cdot)^3}.
\end{align}
Hence, using $|f_2-f_1|\le 2C_F$ and $p(x,\cdot)\ge l_{\hat P}$,
$|\alpha_P'(x;P)[H]|\le 2C_F C_P l_{\hat P}^{-2}$ and
$|\alpha_P''(x;P)[H,H']|\le 4C_F C_P^2 l_{\hat P}^{-3}$.

Finally, since $Y\in\{0,1\}$ and $\gamma\in[0,1]$, we have $|\rho(o,\gamma)|\le 1$, and because $\rho$ is affine in $\gamma$ we have $\nu_\rho\equiv -1$ and $\upsilon_\rho\equiv 0$.
Therefore Assumption~\ref{asmp:regularity} holds.

\paragraph{Verifying Assumption~\ref{asmp:main}.}
We construct perturbations $G_0,G_1,H_0,H_1$ at $\hat P$.

\smallskip
\textit{Choosing a bounded function $\zeta$ with two properties.}
Since $F_1\neq F_2$ and both are absolutely continuous w.r.t.~$\mu_{\gX}$, the signed density $f_2-f_1$ is not a.e.~zero.
Hence there exists a measurable set $S\subseteq \gX$ of positive $\mu_{\gX}$-measure on which $f_2-f_1$ has a constant sign.
Without loss of generality, assume $f_2-f_1<0$ on $S$ (otherwise replace $S$ by a subset where $f_2-f_1>0$).

Pick two disjoint measurable subsets $A,B\subseteq S$ such that
$\int_A \hat p(x,\cdot)\dd\mu_{\gX}(x)>0$ and $\int_B \hat p(x,\cdot)\dd\mu_{\gX}(x)>0$.
Define
\[
  c:=\frac{\int_A \hat p(x,\cdot)\dd\mu_{\gX}(x)}{\int_B \hat p(x,\cdot)\dd\mu_{\gX}(x)},\qquad
  \zeta(x):=\1\{x\in A\}-c\,\1\{x\in B\}.
\]
Then $\zeta$ is bounded and satisfies the mean-zero constraint
\begin{equation}
\label{eq:dis-zeta-mean0}
  \int_{\gX} \zeta(x)\,\hat p(x,\cdot)\,\dd\mu_{\gX}(x)=0.
\end{equation}
Moreover, since $A,B\subseteq S$ and $f_2-f_1<0$ on $S$, we have
\begin{equation}
\label{eq:dis-zeta-nonzero}
  \int_{\gX}\frac{\zeta(x)^2}{\hat p(x,\cdot)^2}\,\dd(F_2-F_1)(x)
  =\int_A\frac{f_2(x)-f_1(x)}{\hat p(x,\cdot)^2}\dd\mu_{\gX}(x)
   +c^2\int_B\frac{f_2(x)-f_1(x)}{\hat p(x,\cdot)^2}\dd\mu_{\gX}(x)\;<\;0,
\end{equation}
and in particular the left-hand side is nonzero.

\smallskip
\textit{A $\gamma$-invariant direction $G_0$.}
Define
\begin{equation}
\label{eq:dis-g0}
  g_0(x,y):=\zeta(x)\,\hat p(x,y).
\end{equation}
Then, using \eqref{eq:dis-zeta-mean0},
\[
  \int g_0\,\dd\mu=\int_{\gX}\zeta(x)\hat p(x,\cdot)\dd\mu_{\gX}(x)=0,
\]
so $G_0$ is a valid perturbation.
For any $s$ such that $\hat p+s g_0\ge 0$, we have
\[
  \gamma\{x;\hat P+sG_0\}
  =\frac{\hat p(x,1)+s g_0(x,1)}{\hat p(x,\cdot)+s g_0(x,\cdot)}
  =\frac{\hat p(x,1)\{1+s\zeta(x)\}}{\hat p(x,\cdot)\{1+s\zeta(x)\}}
  =\gamma(x;\hat P),
\]
so $\gamma(\cdot;\hat P+sG_0)$ is exactly invariant along $G_0$.

\smallskip
\textit{Choosing $G_1$ so that $\chi_{\dis}''(\hat P)[G_0,G_1]\neq 0$.}
Define
\begin{equation}
\label{eq:dis-g1}
  g_1(x,y):=(-1)^y\,\zeta(x).
\end{equation}
Then $g_1(x,\cdot)=g_1(x,0)+g_1(x,1)=0$ for all $x$, so perturbing along $G_1$ does not change the $X$-marginal density $p(x,\cdot)$ and therefore does not change $\alpha(x;P)$.

For $s,t$ small, write $P_{s,t}:=\hat P+sG_0+tG_1$.
Then
\[
  p_{s,t}(x,y)=\hat p(x,y)\{1+s\zeta(x)\}+t(-1)^y\zeta(x),\qquad
  p_{s,t}(x,\cdot)=\hat p(x,\cdot)\{1+s\zeta(x)\}.
\]
Therefore,
\[
  \gamma\{x;P_{s,t}\}
  =\frac{p_{s,t}(x,1)}{p_{s,t}(x,\cdot)}
  =\frac{\hat p(x,1)\{1+s\zeta(x)\}-t\zeta(x)}{\hat p(x,\cdot)\{1+s\zeta(x)\}}
  =\gamma(x;\hat P)-t\,\frac{\zeta(x)}{\hat p(x,\cdot)\{1+s\zeta(x)\}}.
\]
Since $\chi_{\dis}(P)=\int_{\gX}\gamma(x;P)\dd(F_2-F_1)(x)$, it follows that
\[
  \chi_{\dis}(P_{s,t})
  =\chi_{\dis}(\hat P)-t\int_{\gX}\frac{\zeta(x)}{\hat p(x,\cdot)\{1+s\zeta(x)\}}\,\dd(F_2-F_1)(x).
\]
Differentiating in $t$ and then in $s$ gives the mixed derivative
\begin{align}
\label{eq:dis-mixed-deriv}
  \chi_{\dis}''(\hat P)[G_0,G_1]
  &=\partial_s\partial_t\chi_{\dis}(P_{s,t})\big|_{(s,t)=(0,0)}
    =\int_{\gX}\frac{\zeta(x)^2}{\hat p(x,\cdot)^2}\,\dd(F_2-F_1)(x),
\end{align}
which is nonzero by \eqref{eq:dis-zeta-nonzero}.

\smallskip
\textit{Constructing $H_0,H_1$ for the $\alpha$-invariance condition.}
As noted above, $g_1(x,\cdot)\equiv 0$, so $p(x,\cdot)$ and hence $\alpha(x;P)$ are invariant along $G_1$.
Thus we may take
\[
  H_0:=G_1,\qquad H_1:=G_0.
\]
Then Assumption~\ref{asmp:main}(1) holds for $H_0$, and
$\chi_{\dis}''(\hat P)[H_0,H_1]=\chi_{\dis}''(\hat P)[G_1,G_0]\neq 0$ by \eqref{eq:dis-mixed-deriv}.
This completes the Verifying Assumption~\ref{asmp:main}.

\textit{Conclusion.}
All assumptions required to invoke Theorem~\ref{thm:main-mixed-bias} are satisfied for $\chi_{\dis}$, and since $\rho$ is affine in $\gamma$ we obtain the claimed
$\Omega(\eps_{n,\gamma}\eps_{n,\alpha}+1/\sqrt{n})$ lower bound.

\subsection{Proof of Theorem \ref{thm:lod}}
\label{sec:proof:lod}

We prove Theorem \ref{thm:lod}. Throughout, let $O=(X,D,Y)\in \gX\times\{0,1\}\times\{0,1\}$ and write $Z=(X,D)$. Let $\mu=\mu_X\otimes\mu_D\otimes\mu_Y$, where $\mu_X$ is Lebesgue measure on $\gX=[0,1]^K$ and $\mu_D,\mu_Y$ are counting measures on $\{0,1\}$.

\paragraph{Verifying Assumption \ref{asmp:cond-prob-functional}.}
Let $P\ll \mu$ with density $p=\dd P/\dd\mu$ and define, for $(x,d)\in\gX\times\{0,1\}$,
\[
p_{dy}(x):=p(x,d,y),\qquad p_{d\cdot}(x):=\sum_{y\in\{0,1\}}p_{dy}(x),\qquad p_{\cdot\cdot}(x):=\sum_{d\in\{0,1\}}p_{d\cdot}(x).
\]
We also define the conditional mean and propensity score
\[
g(d,x;P):=\E_P[Y\mid D=d,X=x]=\frac{p_{d1}(x)}{p_{d\cdot}(x)},\qquad 
\pi(x;P):=\E_P[D\mid X=x]=\frac{p_{1\cdot}(x)}{p_{\cdot\cdot}(x)}.
\]
Since $Y\in\{0,1\}$, the log-odds function can be written equivalently as
\begin{equation}\label{eq:lod-gamma-density}
\gamma(d,x;P)=\log\left(\frac{g(d,x;P)}{1-g(d,x;P)}\right)=\log\left(\frac{p_{d1}(x)}{p_{d0}(x)}\right).
\end{equation}
Let $\Lambda(t):=(1+\exp(-t))^{-1}$ denote the logistic link. Define the generalized regression score, for $o=(x,d,y)$ and scalar $\gamma\in\R$,
\begin{equation}\label{eq:lod-rho-def}
\rho(o,\gamma):=\frac{y-\Lambda(\gamma)}{\Lambda(\gamma)\{1-\Lambda(\gamma)\}}.
\end{equation}
Then, for any measurable function $\tilde\gamma:\gX\times\{0,1\}\to\R$ and $z=(x,d)$,
\begin{align*}
\E_P\left[\rho\{O,\tilde\gamma(Z)\}\mid Z=z\right]
&=\frac{\E_P[Y\mid Z=z]-\Lambda\{\tilde\gamma(z)\}}{\Lambda\{\tilde\gamma(z)\}\{1-\Lambda\{\tilde\gamma(z)\}\}}\\
&=\frac{g(z;P)-\Lambda\{\tilde\gamma(z)\}}{\Lambda\{\tilde\gamma(z)\}\{1-\Lambda\{\tilde\gamma(z)\}\}},
\end{align*}
where $g(z;P):=\E_P[Y\mid Z=z]=g(d,x;P)$. Since $\Lambda$ is strictly increasing, the unique solution to $\E_P[\rho\{O,\tilde\gamma(Z)\}\mid Z]=0$ is $\tilde\gamma(z)=\log\left(\frac{g(z;P)}{1-g(z;P)}\right)$, i.e.\ \eqref{eq:lod-gamma-density}. This verifies the conditional moment condition in Assumption \ref{asmp:cond-prob-functional}.

Next, we verify the Riesz representer required by Assumption \ref{asmp:cond-prob-functional}. Let $m_1(o,h):=h(1,x)-h(0,x)$, which is linear in $h$. For any square-integrable $h$,
\[
\E_P[m_1\{O,h(Z)\}]=\E_P[h(1,X)]-\E_P[h(0,X)].
\]
A direct calculation shows that the Riesz representer $\nu_m(\cdot;P)$ is the unique function satisfying
\[
\E_P[m_1\{O,h(Z)\}]=\E_P[h(Z)\nu_m(Z;P)]
\qquad\text{for all square-integrable $h$,}
\]
and is given by
\begin{equation}\label{eq:lod-nu-m}
\nu_m(z;P)=\frac{d}{\pi(x;P)}-\frac{1-d}{1-\pi(x;P)},\qquad z=(x,d).
\end{equation}

Finally, we compute $\nu_\rho$ and $\upsilon_\rho$. Fix $z=(x,d)$ and let $\gamma(z;P)$ denote the solution above. For $a\in\R$ define
\[
\psi_z(a):=\E_P\left[\rho\{O,\gamma(z;P)+a\}\mid Z=z\right]
=\frac{g(z;P)-\Lambda\{\gamma(z;P)+a\}}{\Lambda\{\gamma(z;P)+a\}\{1-\Lambda\{\gamma(z;P)+a\}\}}.
\]
Since $g(z;P)=\Lambda\{\gamma(z;P)\}$, differentiating $\psi_z(a)$ at $a=0$ yields
\begin{equation}\label{eq:lod-nu-rho}
\nu_\rho(z;P):=\left.\frac{\dd}{\dd a}\psi_z(a)\right|_{a=0}=-1.
\end{equation}
A second derivative calculation gives
\begin{equation}\label{eq:lod-upsilon-rho}
\upsilon_\rho(z;P):=\left.\frac{\dd^2}{\dd a^2}\psi_z(a)\right|_{a=0}
=1-2\Lambda\{\gamma(z;P)\}=1-2g(z;P).
\end{equation}
In particular, $\nu_\rho(\cdot;P)\equiv -1$ and $|\upsilon_\rho(z;P)|\leq 1$.

Combining \eqref{eq:lod-nu-m}--\eqref{eq:lod-nu-rho}, we obtain
\begin{equation}\label{eq:lod-alpha}
\alpha(z;P):=-\frac{\nu_m(z;P)}{\nu_\rho(z;P)}=\nu_m(z;P),
\end{equation}
so that $\alpha(\cdot;P)$ depends on $P$ only through $\pi(\cdot;P)$.

\paragraph{Verifying Assumption \ref{asmp:regularity}.}
Fix $\hat P\ll \mu$ satisfying the conditions of Theorem \ref{thm:lod}, and write $\hat p=\dd\hat P/\dd\mu$. Let $H$ be any signed measure with $H\ll \mu$ and density $h=\dd H/\dd\mu$ satisfying $\int h\,\dd\mu=0$ and $\|h\|_\infty<\infty$. For $t$ such that $\hat p+t h\geq 0$ $\mu$-a.e., define $P_t:=\hat P+tH$ with density $p_t:=\hat p+t h$.

\smallskip
\noindent\textit{Directional derivatives of $\gamma$.}
For $z=(x,d)$, using \eqref{eq:lod-gamma-density} with $P=P_t$, we have
\[
\gamma(z;P_t)=\log\left(\frac{p_t(x,d,1)}{p_t(x,d,0)}\right).
\]
Since $t\mapsto \log(\hat p(x,d,y)+t h(x,d,y))$ is twice continuously differentiable for each $(x,d,y)$ in the region where $\hat p(x,d,y)+t h(x,d,y)>0$, it follows that $\gamma(z;P_t)$ is twice differentiable in $t$ with
\begin{align}
\gamma_P'(z;\hat P)[H]
&:=\left.\frac{\dd}{\dd t}\gamma(z;P_t)\right|_{t=0}
=\frac{h(x,d,1)}{\hat p(x,d,1)}-\frac{h(x,d,0)}{\hat p(x,d,0)},\label{eq:lod-gamma-deriv}\\[0.25em]
\gamma_P''(z;\hat P)[H,H]
&:=\left.\frac{\dd^2}{\dd t^2}\gamma(z;P_t)\right|_{t=0}
=-\frac{h(x,d,1)^2}{\hat p(x,d,1)^2}+\frac{h(x,d,0)^2}{\hat p(x,d,0)^2}.\label{eq:lod-gamma-second}
\end{align}
Moreover, by the density lower bound $\hat p\geq p_{\mathrm{lb}}>0$, we have the uniform bounds
\begin{equation}\label{eq:lod-gamma-deriv-bounds}
\sup_{z\in\gX\times\{0,1\}}|\gamma_P'(z;\hat P)[H]|
\leq \frac{2}{p_{\mathrm{lb}}}\|h\|_\infty,\qquad
\sup_{z\in\gX\times\{0,1\}}|\gamma_P''(z;\hat P)[H,H]|
\leq \frac{2}{p_{\mathrm{lb}}^2}\|h\|_\infty^2.
\end{equation}

\smallskip
\noindent\textit{Directional derivatives of $\alpha$.}
For $x\in\gX$, define $\hat p_{1\cdot}(x):=\sum_{y}\hat p(x,1,y)$, $\hat p_{0\cdot}(x):=\sum_{y}\hat p(x,0,y)$, and $\hat p_{\cdot\cdot}(x):=\hat p_{1\cdot}(x)+\hat p_{0\cdot}(x)$. Likewise define $h_{1\cdot}(x):=\sum_{y}h(x,1,y)$, $h_{0\cdot}(x):=\sum_{y}h(x,0,y)$, and $h_{\cdot\cdot}(x):=h_{1\cdot}(x)+h_{0\cdot}(x)$.

Using \eqref{eq:lod-nu-m}--\eqref{eq:lod-alpha} and $\pi(x;P_t)=p_{t,1\cdot}(x)/p_{t,\cdot\cdot}(x)$, we may write
\[
\alpha\{(x,1);\hat P\}=\frac{\hat p_{\cdot\cdot}(x)}{\hat p_{1\cdot}(x)},\qquad 
\alpha\{(x,0);\hat P\}=-\frac{\hat p_{\cdot\cdot}(x)}{\hat p_{0\cdot}(x)}.
\]
Elementary differentiation yields, for $d\in\{0,1\}$,
\begin{align}
\alpha_P'\{(x,1);\hat P\}[H]
&=\left.\frac{\dd}{\dd t}\frac{p_{t,\cdot\cdot}(x)}{p_{t,1\cdot}(x)}\right|_{t=0}
=\frac{h_{\cdot\cdot}(x)\hat p_{1\cdot}(x)-\hat p_{\cdot\cdot}(x)h_{1\cdot}(x)}{\hat p_{1\cdot}(x)^2},\label{eq:lod-alpha-deriv-1}\\[0.25em]
\alpha_P'\{(x,0);\hat P\}[H]
&=-\left.\frac{\dd}{\dd t}\frac{p_{t,\cdot\cdot}(x)}{p_{t,0\cdot}(x)}\right|_{t=0}
=-\frac{h_{\cdot\cdot}(x)\hat p_{0\cdot}(x)-\hat p_{\cdot\cdot}(x)h_{0\cdot}(x)}{\hat p_{0\cdot}(x)^2}.\label{eq:lod-alpha-deriv-0}
\end{align}
Similarly, $\alpha_P''(z;\hat P)[H,H]$ exists and can be computed explicitly. For $d=1$,
\begin{align}
\alpha_P''\{(x,1);\hat P\}[H,H]
&:=\left.\frac{\dd^2}{\dd t^2}\frac{p_{t,\cdot\cdot}(x)}{p_{t,1\cdot}(x)}\right|_{t=0}
=-2\,\frac{h_{\cdot\cdot}(x)\hat p_{1\cdot}(x)-\hat p_{\cdot\cdot}(x)h_{1\cdot}(x)}{\hat p_{1\cdot}(x)^3}\,h_{1\cdot}(x),\label{eq:lod-alpha-second-1}\\[0.25em]
\alpha_P''\{(x,0);\hat P\}[H,H]
&:=\left.\frac{\dd^2}{\dd t^2}\Big(-\frac{p_{t,\cdot\cdot}(x)}{p_{t,0\cdot}(x)}\Big)\right|_{t=0}
=2\,\frac{h_{\cdot\cdot}(x)\hat p_{0\cdot}(x)-\hat p_{\cdot\cdot}(x)h_{0\cdot}(x)}{\hat p_{0\cdot}(x)^3}\,h_{0\cdot}(x).\label{eq:lod-alpha-second-0}
\end{align}
In view of the density bounds $\hat p\in[p_{\mathrm{lb}},p_{\mathrm{ub}}]$, we have $\hat p_{1\cdot}(x),\hat p_{0\cdot}(x)\in[2p_{\mathrm{lb}},2p_{\mathrm{ub}}]$ and $\hat p_{\cdot\cdot}(x)\in[4p_{\mathrm{lb}},4p_{\mathrm{ub}}]$ for all $x\in\gX$. Moreover, $|h_{\cdot\cdot}(x)|\leq 4\|h\|_\infty$ and $|h_{d\cdot}(x)|\leq 2\|h\|_\infty$ for $d\in\{0,1\}$. Therefore, from \eqref{eq:lod-alpha-deriv-1}--\eqref{eq:lod-alpha-second-0} we obtain the uniform bounds
\begin{equation}\label{eq:lod-alpha-deriv-bounds}
\sup_{z\in\gX\times\{0,1\}}\big|\alpha_P'(z;\hat P)[H]\big|
\leq \frac{4\,p_{\mathrm{ub}}}{p_{\mathrm{lb}}^2}\|h\|_\infty,\qquad
\sup_{z\in\gX\times\{0,1\}}\big|\alpha_P''(z;\hat P)[H,H]\big|
\leq \frac{8\,p_{\mathrm{ub}}}{p_{\mathrm{lb}}^3}\|h\|_\infty^2.
\end{equation}
This verifies the differentiability and boundedness requirements in Assumption \ref{asmp:regularity}.

\paragraph{Construction of perturbations required by Assumption \ref{asmp:main}.}
We construct perturbations $G_0$ and $G_1$ (and set $H_0:=G_1$) satisfying the exact-invariance conditions in Assumption \ref{asmp:main} and such that $\chi_{\mathrm{LOD}}''(\hat P)[G_0,G_1]\neq 0$ and $\chi_{\mathrm{LOD}}''(\hat P)[H_0,H_0]\neq 0$.

\smallskip
\noindent\textit{Choice of a set where $\upsilon_\rho$ has fixed sign.}
By assumption, $\hat P_X\{x:g(1,x;\hat P)\neq 1/2\}>0$. Define
\[
A_+:=\{x\in\gX: g(1,x;\hat P)>1/2\},\qquad A_-:=\{x\in\gX: g(1,x;\hat P)<1/2\}.
\]
Then $\hat P_X(A_+\cup A_-)>0$, so at least one of $\hat P_X(A_+)$ or $\hat P_X(A_-)$ is strictly positive. Let $A$ denote either $A_+$ or $A_-$ such that $\hat P_X(A)>0$, and define $b(x):=\mathbf{1}\{x\in A\}$.

\smallskip
\noindent\textit{Definition of the perturbation $G_0$.}
Let $a(x)\equiv 1$ and define a bounded function $\phi_0:\gX\times\{0,1\}\times\{0,1\}\to\R$ by
\begin{equation}\label{eq:lod-phi0}
\phi_0(x,1,y):=\delta_0\,a(x)\,\hat p(x,1,y),\qquad 
\phi_0(x,0,y):=-\delta_0\,a(x)\,\frac{\hat p_{1\cdot}(x)}{\hat p_{0\cdot}(x)}\,\hat p(x,0,y),
\end{equation}
where $\delta_0>0$ is a constant chosen below. Let $G_0$ be the signed measure with density $\phi_0$ with respect to $\mu$, i.e.\ $\dd G_0=\phi_0\,\dd\mu$.

By construction, for each $x$,
\[
\sum_{d\in\{0,1\}}\sum_{y\in\{0,1\}}\phi_0(x,d,y)
=\delta_0\,\hat p_{1\cdot}(x)-\delta_0\,\frac{\hat p_{1\cdot}(x)}{\hat p_{0\cdot}(x)}\hat p_{0\cdot}(x)=0,
\]
and hence $\int \phi_0\,\dd\mu=0$, so $\hat P+tG_0$ has total mass one for all $t$ for which the density is nonnegative.

Moreover, for each $(x,d)$, the perturbation $\phi_0(x,d,\cdot)$ scales both $y=0$ and $y=1$ by the same multiplicative factor. Consequently, for any $t$ such that $\hat p+t\phi_0\geq 0$ we have
\[
g(d,x;\hat P+tG_0)=g(d,x;\hat P),\qquad \gamma(d,x;\hat P+tG_0)=\gamma(d,x;\hat P),
\]
i.e.\ $\gamma(\cdot;\hat P+tG_0)$ is exactly invariant in $t$, as required in Assumption \ref{asmp:main}(1).

\smallskip
\noindent\textit{Definition of the perturbation $G_1$ and $H_0$.}
Define a bounded function $\phi_1:\gX\times\{0,1\}\times\{0,1\}\to\R$ by
\begin{equation}\label{eq:lod-phi1}
\phi_1(x,1,1):=\delta_1\,b(x)\,\hat p(x,1,0),\qquad
\phi_1(x,1,0):=-\delta_1\,b(x)\,\hat p(x,1,0),\qquad 
\phi_1(x,0,y):=0,
\end{equation}
where $\delta_1>0$ is a constant chosen below. Let $G_1$ be the signed measure with density $\phi_1$ and set $H_0:=G_1$.

Since $\phi_1(x,1,1)+\phi_1(x,1,0)=0$ for all $x$, we have $\int\phi_1\,\dd\mu=0$. Moreover, for any $t$ such that $\hat p+t\phi_1\geq 0$, the $(X,D)$-marginal remains unchanged:
\[
\sum_{y\in\{0,1\}}(\hat p+t\phi_1)(x,d,y)=\sum_{y\in\{0,1\}}\hat p(x,d,y)\qquad\text{for all }(x,d).
\]
Therefore $\pi(x;\hat P+tH_0)=\pi(x;\hat P)$ for all such $t$, and hence $\alpha(\cdot;\hat P+tH_0)=\alpha(\cdot;\hat P)$ by \eqref{eq:lod-alpha}. This gives the required exact invariance of $\alpha$ along $H_0$ in Assumption \ref{asmp:main}(1).

\smallskip
\noindent\textit{Feasibility of the perturbations.}
We now choose $\delta_0,\delta_1>0$ so that $G_0$ and $H_0$ are $\gM$-feasible perturbations around $\hat P$ (in the sense of Assumption \ref{asmp:main}). Since $\hat p\in[p_{\mathrm{lb}},p_{\mathrm{ub}}]$ and $\pi(\cdot;\hat P)\in[\eta,1-\eta]$, we have
\[
\frac{\hat p_{1\cdot}(x)}{\hat p_{0\cdot}(x)}=\frac{\pi(x;\hat P)}{1-\pi(x;\hat P)}\in\left[\frac{\eta}{1-\eta},\frac{1-\eta}{\eta}\right]\qquad\text{for $\hat P_X$-a.e.\ }x.
\]
Therefore, for all $(x,d,y)$,
\[
|\phi_0(x,d,y)|\leq \delta_0\,\frac{1-\eta}{\eta}\,\hat p(x,d,y),\qquad 
|\phi_1(x,d,y)|\leq \delta_1\,\hat p(x,d,y).
\]
Choose
\begin{equation}\label{eq:lod-delta-choice}
\delta_0:=\frac{\eta}{8(1-\eta)}\qquad\text{and}\qquad \delta_1:=\frac{1}{8},
\end{equation}
and set $c_0:=1$.
Then, for all $t\in[-c_0,c_0]$, we have $\hat p+t\phi_0\geq \hat p/2\geq p_{\mathrm{lb}}/2$ and $\hat p+t\phi_1\geq \hat p/2\geq p_{\mathrm{lb}}/2$, and also $\hat p+t\phi_0\leq 2\hat p\leq 2p_{\mathrm{ub}}$ and $\hat p+t\phi_1\leq 2\hat p\leq 2p_{\mathrm{ub}}$. In particular, $\hat P+tG_0$ and $\hat P+tH_0$ remain in a density-bounded neighborhood of $\hat P$.

Moreover, by construction $g(\cdot;\hat P+tG_0)=g(\cdot;\hat P)$ for all $t$, and for $H_0$ we have, on the set $A$,
\[
g(1,x;\hat P+tH_0)
=\frac{\hat p(x,1,1)+t\delta_1\hat p(x,1,0)}{\hat p_{1\cdot}(x)}
=g(1,x;\hat P)+t\delta_1\{1-g(1,x;\hat P)\},
\]
while $g(1,x;\hat P+tH_0)=g(1,x;\hat P)$ on $A^c$ and $g(0,x;\hat P+tH_0)=g(0,x;\hat P)$ everywhere. Since $g(d,x;\hat P)\in[\eta,1-\eta]$ and $\delta_1=1/8$, it follows that for all $t\in[-1,1]$,
\[
\frac{\eta}{2}\leq g(d,x;\hat P+tH_0)\leq 1-\frac{\eta}{2},\qquad d\in\{0,1\},
\]
so overlap is preserved along these perturbations (possibly with a smaller constant).

This proves that $G_0,G_1,H_0$ are $\gM$-feasible perturbations around $\hat P$.

\textit{Non-vanishing of the second-order derivatives.}
We compute
\[
\chi_{\mathrm{LOD}}''(\hat P)[G_0,G_1]
\qquad\text{and}\qquad
\chi_{\mathrm{LOD}}''(\hat P)[H_0,H_0].
\]

\smallskip
\noindent\textit{Computation of $\chi_{\mathrm{LOD}}''(\hat P)[G_0,G_1]$.}
For $(s,t)$ in a neighborhood of $(0,0)$, let $P_{s,t}:=\hat P+sG_0+tG_1$ and denote its density by $p_{s,t}:=\hat p+s\phi_0+t\phi_1$. By the construction of $\phi_0$ and $\phi_1$, the $X$-marginal of $P_{s,t}$ equals that of $\hat P$, i.e.\ $P_{s,t,X}=\hat P_X$ for all such $(s,t)$. Moreover, $\gamma(0,x;P_{s,t})=\gamma(0,x;\hat P)$ for all $(s,t)$ because neither $\phi_0$ nor $\phi_1$ changes the odds ratio within $D=0$.

For $D=1$, writing $\hat p_{11}(x):=\hat p(x,1,1)$ and $\hat p_{10}(x):=\hat p(x,1,0)$, \eqref{eq:lod-phi0}--\eqref{eq:lod-phi1} imply
\[
p_{s,t}(x,1,1)=\hat p_{11}(x)\{1+s\delta_0\}+t\delta_1 b(x)\hat p_{10}(x),\qquad 
p_{s,t}(x,1,0)=\hat p_{10}(x)\{1+s\delta_0\}-t\delta_1 b(x)\hat p_{10}(x).
\]
Therefore, for each $x$,
\begin{equation}\label{eq:lod-gamma-st}
\gamma(1,x;P_{s,t})=\log\{p_{s,t}(x,1,1)\}-\log\{p_{s,t}(x,1,0)\}.
\end{equation}
Differentiating \eqref{eq:lod-gamma-st} with respect to $t$ and then $s$, and evaluating at $(s,t)=(0,0)$, we obtain
\begin{align*}
\left.\frac{\partial}{\partial t}\gamma(1,x;P_{s,t})\right|_{t=0}
&=\delta_1 b(x)\hat p_{10}(x)\Big\{\frac{1}{\hat p_{11}(x)\{1+s\delta_0\}}+\frac{1}{\hat p_{10}(x)\{1+s\delta_0\}}\Big\},\\
\left.\frac{\partial^2}{\partial s\,\partial t}\gamma(1,x;P_{s,t})\right|_{(s,t)=(0,0)}
&=-\delta_0\delta_1\,b(x)\hat p_{10}(x)\Big\{\frac{1}{\hat p_{11}(x)}+\frac{1}{\hat p_{10}(x)}\Big\}.
\end{align*}
Since $\hat p_{10}(x)\{1/\hat p_{11}(x)+1/\hat p_{10}(x)\}=(\hat p_{10}(x)+\hat p_{11}(x))/\hat p_{11}(x)=1/g(1,x;\hat P)$, it follows that
\begin{equation}\label{eq:lod-chi-second-G0G1}
\chi_{\mathrm{LOD}}''(\hat P)[G_0,G_1]
=\left.\frac{\partial^2}{\partial s\,\partial t}\chi_{\mathrm{LOD}}(P_{s,t})\right|_{(s,t)=(0,0)}
=-\delta_0\delta_1\,\E_{\hat P_X}\left[\frac{b(X)}{g(1,X;\hat P)}\right].
\end{equation}
Because $b=\mathbf{1}\{X\in A\}$ and $\hat P_X(A)>0$, and since $g(1,x;\hat P)\in[\eta,1-\eta]$, the expectation in \eqref{eq:lod-chi-second-G0G1} is strictly positive. Hence $\chi_{\mathrm{LOD}}''(\hat P)[G_0,G_1]\neq 0$.

\smallskip
\noindent\textit{Computation of $\chi_{\mathrm{LOD}}''(\hat P)[H_0,H_0]$.}
Let $P_t:=\hat P+tH_0$ and write $p_t=\hat p+t\phi_1$. As above, $P_{t,X}=\hat P_X$ for all $t$ and $\gamma(0,x;P_t)=\gamma(0,x;\hat P)$. For $D=1$ we have
\[
p_t(x,1,1)=\hat p_{11}(x)+t\delta_1 b(x)\hat p_{10}(x),\qquad 
p_t(x,1,0)=\hat p_{10}(x)-t\delta_1 b(x)\hat p_{10}(x).
\]
Therefore, $\gamma(1,x;P_t)=\log\{p_t(x,1,1)\}-\log\{p_t(x,1,0)\}$ and, by direct differentiation,
\[
\left.\frac{\dd^2}{\dd t^2}\gamma(1,x;P_t)\right|_{t=0}
=(\delta_1 b(x)\hat p_{10}(x))^2\Big\{\frac{1}{\hat p_{10}(x)^2}-\frac{1}{\hat p_{11}(x)^2}\Big\}.
\]
Using $\hat p_{10}/\hat p_{11}=(1-g)/g$ with $g=g(1,x;\hat P)$, we obtain
\[
\left.\frac{\dd^2}{\dd t^2}\gamma(1,x;P_t)\right|_{t=0}
=\delta_1^2\,b(x)^2\,\frac{2g(1,x;\hat P)-1}{g(1,x;\hat P)^2}.
\]
Consequently,
\begin{equation}\label{eq:lod-chi-second-H0}
\chi_{\mathrm{LOD}}''(\hat P)[H_0,H_0]
=\left.\frac{\dd^2}{\dd t^2}\chi_{\mathrm{LOD}}(P_t)\right|_{t=0}
=\delta_1^2\,\E_{\hat P_X}\left[b(X)^2\,\frac{2g(1,X;\hat P)-1}{g(1,X;\hat P)^2}\right].
\end{equation}
By construction, $b(X)=\mathbf{1}\{X\in A\}$ and $2g(1,x;\hat P)-1$ has a constant nonzero sign on $A$. Therefore, \eqref{eq:lod-chi-second-H0} is nonzero, i.e.\ $\chi_{\mathrm{LOD}}''(\hat P)[H_0,H_0]\neq 0$.

\textit{Conclusion.}
The score function $\rho$ in \eqref{eq:lod-rho-def} is not affine in $\gamma$ (because $\Lambda(\gamma)$ is nonlinear), so this example falls under the general case of Theorem \ref{thm:main}. The arguments above verify Assumptions \ref{asmp:density-bounded}, \ref{asmp:cond-prob-functional}, \ref{asmp:cond-space-nontrivial}, \ref{asmp:regularity}, and \ref{asmp:main} and show that $\chi_{\mathrm{LOD}}''(\hat P)[H_0,H_0]\neq 0$. The minimax lower bound in Theorem \ref{thm:lod} therefore follows directly from Theorem \ref{thm:main}.

\subsection{Proof of Theorem \ref{thm:eqd}}
\label{sec:proof:eqd}

\begin{proposition}[Feasible perturbations under derivative constraints]\label{prop:eqd-feasible-perturbation}
	    Suppose $P$ has a density function $p$ (with respect to $\mu$) such that
	    \begin{equation}\label{eq:eqd-perturbation-slack}
	        \Delta_P
	        :=
	        \operatorname*{ess\,inf}_{(x,y)\in \gX\times[0,1]}
	        \min\left\{
	        \begin{array}{l}
	            p(x,y)-l_{\hat{P}}/2,\\
	            2u_{\hat{P}}-p(x,y),\\
	            2C_{X,1}-|\partial_{x_1}p(x,y)|,\\
	            2C_{Y,1}-|\partial_{y}p(x,y)|,\\
	            2C_{Y,2}-|\partial_{y}^2p(x,y)|
	        \end{array}
	        \right\}
	        >0.
	    \end{equation}
    Here the essential infimum and $\|\cdot\|_\infty$ are taken with respect to $\mu$, and derivatives are understood in the weak sense.

    Let $H$ be a signed measure with density $h=\dd H/\dd\mu$ such that $\int h\,\dd\mu=0$, $\|h\|_\infty<\infty$ and
    \[
        L_H
        :=
        \max\Bigl\{
            \|\partial_{x_1}h\|_\infty,\ 
            \|\partial_{y}h\|_\infty,\ 
            \|\partial_{y}^2h\|_\infty
        \Bigr\}
        <\infty.
    \]
    Then $H$ is a $\gM_1$-feasible perturbation of $P$. In particular, defining
    $M_H:=\max\{\|h\|_\infty,L_H\}$ and $r_H:=\Delta_P/M_H$, we have
    $P+tH\in \gM_1$ for all $|t|\le r_H$.

    Conversely, if $H$ is a $\gM_1$-feasible perturbation of $P$ with feasible radius $r>0$ (i.e.\ $P+tH\in\gM_1$ for all $|t|\le r$),
    then
    \[
        L_H
        \le
        4r^{-1}\max\{C_{X,1},C_{Y,1},C_{Y,2}\}.
    \]
\end{proposition}

\begin{proof}
    For the forward direction, fix $|t|\le r_H$. By \eqref{eq:eqd-perturbation-slack},
    $p(x,y)\ge l_{\hat{P}}/2+\Delta_P$ and $p(x,y)\le 2u_{\hat{P}}-\Delta_P$ for $\mu$-a.e.\ $(x,y)$. Hence
    \[
        p(x,y)+th(x,y)
        \ge
        l_{\hat{P}}/2+\Delta_P-|t|\|h\|_\infty
        \ge
        l_{\hat{P}}/2,
    \]
    and similarly $p(x,y)+th(x,y)\le 2u_{\hat{P}}$.
    Also, $\int (p+th)\,\dd\mu=1$ since $\int h\,\dd\mu=0$, so $P+tH$ is a probability measure.

    Moreover, for each of the constrained derivatives,
    \[
        |\partial_{x_1}(p+th)|
        \le
        |\partial_{x_1}p|+|t||\partial_{x_1}h|
        \le
        (2C_{X,1}-\Delta_P)+r_H L_H
        \le
        2C_{X,1},
    \]
    and the same argument applies to $|\partial_y(p+th)|$ and $|\partial_y^2(p+th)|$.
    Thus $P+tH\in\gM_1$ for all $|t|\le r_H$.

    For the converse direction, fix any $t\in(0,r]$. Since $P\pm tH\in\gM_1$, the weak derivatives $\partial_{x_1}(p\pm th)$ exist and satisfy $\|\partial_{x_1}(p\pm th)\|_\infty\le 2C_{X,1}$, and by linearity
    \[
        |\partial_{x_1}h|
        =
        \left|\frac{\partial_{x_1}(p+th)-\partial_{x_1}(p-th)}{2t}\right|
        \le
        \frac{|\partial_{x_1}(p+th)|+|\partial_{x_1}(p-th)|}{2t}
        \le
        \frac{4C_{X,1}}{t}.
    \]

    Taking $t=r$ yields $\|\partial_{x_1}h\|_\infty\le 4C_{X,1}/r$. The same argument gives
    $\|\partial_y h\|_\infty\le 4C_{Y,1}/r$ and $\|\partial_y^2 h\|_\infty\le 4C_{Y,2}/r$, proving the bound on $L_H$.
\end{proof}

We now verify the conditions needed to apply Theorem~\ref{thm:main} to the EQD functional in Theorem~\ref{thm:eqd} and then construct the perturbations required by Assumption~\ref{asmp:main}.

\paragraph{Verifying Assumption~\ref{asmp:density-bounded}.}
Take $r:=l_{\hat P}/4$. If $d_{\mu,\infty}(P,\hat P)\le r$ and $\dd P/\dd\mu=p$, then
$p(x,y)\ge \hat p(x,y)-r\ge 3l_{\hat P}/4$ and $p(x,y)\le \hat p(x,y)+r\le 2u_{\hat P}$ for $\mu$-a.e.\ $(x,y)$, so Assumption~\ref{asmp:density-bounded} holds.

\paragraph{Verifying Assumption~\ref{asmp:cond-prob-functional}.}
For EQD, the estimating equation uses
$\rho(o,\gamma)=\mathbbm{1}\{y\le \gamma(x)\}-q$ with $z=x$ and $w=y$.
Then $\E_P[\rho(O,\gamma)\mid X=x]=F_{Y\mid X=x}(\gamma(x))-q$, so we can take
\[
    \nu_\rho(x;P)=f_{Y\mid X=x}(\gamma(x;P))=\frac{p(x,\gamma(x;P))}{p_X(x)}
    \quad\text{and}\quad
    \upsilon_\rho(x;P)=\partial_y f_{Y\mid X=x}(y)\big|_{y=\gamma(x;P)}
    =
    \frac{\partial_y p(x,\gamma(x;P))}{p_X(x)},
\]
where $p_X(x):=\int_0^1 p(x,y)\dd y$ is the marginal density of $X$.
This is exactly the structure required by Assumption~\ref{asmp:cond-prob-functional}.

\textit{Bounds for $\gamma(\cdot;\hat P)$ and its $x_1$-derivative.}
We will repeatedly use that the conditional quantile stays away from the boundary and that it is differentiable in $x_1$ under Assumption~\ref{asmp:eqd-density}.

\begin{lemma}[Quantile stays away from the boundary]\label{lem:eqd-gamma-bound}
Under Assumption~\ref{asmp:eqd-density}, for all $x\in\gX$,
\[
    \frac{l_{\hat P} q}{2u_{\hat P}}
    \le
    \gamma(x;\hat P)
    \le
    1-\frac{l_{\hat P}(1-q)}{2u_{\hat P}}.
\]
\end{lemma}
\begin{proof}
Fix $x\in\gX$ and abbreviate $\gamma:=\gamma(x;\hat P)$ and $\hat p_X(x):=\int_0^1 \hat p(x,y)\dd y$.
Since $\gamma$ is a $q$-quantile of $Y\mid X=x$,
\[
    q
    =
    \frac{\int_0^\gamma \hat p(x,u)\dd u}{\hat p_X(x)}
    \le
    \frac{u_{\hat P}\gamma}{l_{\hat P}/2},
\]
which yields $\gamma\ge l_{\hat P}q/(2u_{\hat P})$. The upper bound is analogous, using
$1-q=\int_\gamma^1 \hat p(x,u)\dd u/\hat p_X(x)\le u_{\hat P}(1-\gamma)/(l_{\hat P}/2)$.
\end{proof}

\begin{lemma}[Derivative of the conditional quantile in $x_1$]\label{lem:eqd-gamma-x1-derivative}
Under Assumption~\ref{asmp:eqd-density}, the map $x\mapsto \gamma(x;\hat P)$ is differentiable in $x_1$, with
\[
    \begin{aligned}
    \partial_{x_1}\gamma(x;\hat P)
    &=
    -\frac{1}{\hat p(x,\gamma(x;\hat P))}
    \Bigg\{\int_0^{\gamma(x;\hat P)} \partial_{x_1}\hat p(x,u)\dd u
    - q\int_0^1 \partial_{x_1}\hat p(x,u)\dd u\Bigg\}.
    \end{aligned}
\]
In particular, $\|\partial_{x_1}\gamma(\cdot;\hat P)\|_\infty\le 2C_{X,1}/l_{\hat P}$.
\end{lemma}
\begin{proof}
For fixed $x$, define
$G(x,y):=\int_0^y \hat p(x,u)\dd u-q\int_0^1 \hat p(x,u)\dd u$.
Then $G(x,\gamma(x;\hat P))=0$ and $\partial_y G(x,y)=\hat p(x,y)\ge l_{\hat P}>0$.
By the implicit function theorem \citep[Chapter~1]{krantz2002implicit}, $\gamma(\cdot;\hat P)$ is differentiable in $x_1$ and
$\partial_{x_1}\gamma(x;\hat P)=-\partial_{x_1}G(x,\gamma)/\partial_y G(x,\gamma)$, yielding the displayed formula.
The bound follows from
\[
|\partial_{x_1}G(x,\gamma)|
\le \int_0^{\gamma}|\partial_{x_1}\hat p|\dd u + q\int_0^1|\partial_{x_1}\hat p|\dd u
\le 2C_{X,1}.
\]
\end{proof}

\paragraph{Verifying Assumption~\ref{asmp:regularity}.}
Fix $P$ in the $r$-neighborhood of $\hat P$ from Assumption~\ref{asmp:density-bounded} and write $p=\dd P/\dd\mu$.
Let $H,H'$ be any $P$-feasible perturbations of radius $r_P$ whose densities $h=\dd H/\dd\mu$ and $h'=\dd H'/\dd\mu$ satisfy
$\|h\|_\infty\vee\|h'\|_\infty\le C_P$.
By Proposition~\ref{prop:eqd-feasible-perturbation}, their first $x_1$-derivative and first/second $y$-derivatives are uniformly bounded by a constant depending on $r_P$.

We now derive (and bound) the first and mixed second directional derivatives of $\gamma(\cdot;P)$ and $\alpha(\cdot;P)$.
For each $x$, let
\[
    p_X(x):=\int_0^1 p(x,u)\dd u,\qquad
    h_X(x):=\int_0^1 h(x,u)\dd u,\qquad
    h'_X(x):=\int_0^1 h'(x,u)\dd u,
\]
and define
$A_x(y):=\int_0^y p(x,u)\dd u$ and $B_x(y):=\int_0^y h(x,u)\dd u$, $B_x'(y):=\int_0^y h'(x,u)\dd u$.

For fixed $x$, $\gamma(x;P+tH)$ is defined implicitly by
\[
    A_x(\gamma(x;P+tH))+tB_x(\gamma(x;P+tH)) = q\bigl(p_X(x)+t h_X(x)\bigr).
\]
Since $y\mapsto A_x(y)$ is continuously differentiable with derivative $p(x,y)$, and $p(x,\gamma(x;P))$ is bounded away from $0$
uniformly in $x$ (by Assumption~\ref{asmp:density-bounded}), the implicit function theorem gives twice differentiability of
$t\mapsto \gamma(x;P+tH)$ for $|t|\le r_P$ and yields the following standard formulas.

\begin{equation}\label{eq:eqd-gamma-first-derivative}
    \gamma_P'(x;P)[H]
    =
    -\frac{B_x(\gamma(x;P))-q h_X(x)}{p(x,\gamma(x;P))}.
\end{equation}

\begin{equation}\label{eq:eqd-gamma-second-derivative}
    \gamma_P''(x;P)[H,H']
    =
    -\frac{
        \partial_y p(x,\gamma)\,\gamma_P'(x;P)[H]\,\gamma_P'(x;P)[H']
        + h(x,\gamma)\,\gamma_P'(x;P)[H']
        + h'(x,\gamma)\,\gamma_P'(x;P)[H]
    }{p(x,\gamma)},
\end{equation}
where $\gamma=\gamma(x;P)$.

Since $|B_x(\gamma)-qh_X(x)|\le \int_0^\gamma |h|\dd u + q|h_X(x)|\le 2C_P$, we have
$\|\gamma_P'(\cdot;P)[H]\|_\infty\lesssim C_P$ uniformly over such $H$.
Similarly, using \eqref{eq:eqd-gamma-second-derivative}, the bounds on $\partial_y p$, and the already-derived bound on $\gamma_P'$,
we obtain $\|\gamma_P''(\cdot;P)[H,H']\|_\infty\lesssim C_P^2$ uniformly over such $(H,H')$.

Next, recall
\[
\alpha(x;P)=p(x,\gamma(x;P))^{-1}\partial_{x_1}\bigl(w(x)p_X(x)\bigr).
\]
Let $N(x;P):=\partial_{x_1}(w(x)p_X(x))$ and $D(x;P):=p(x,\gamma(x;P))$, so that $\alpha=N/D$.
Then for any perturbation direction $H$,
\[
    N_P'(x;P)[H] = \partial_{x_1}\bigl(w(x)h_X(x)\bigr),
    \qquad
    D_P'(x;P)[H]
    =
    h(x,\gamma)+\partial_y p(x,\gamma)\,\gamma_P'(x;P)[H].
\]
Therefore the first directional (G\^ateaux) derivative of $\alpha$ is
\begin{equation}\label{eq:eqd-alpha-first-derivative}
    \alpha_P'(x;P)[H]
    =
    \frac{\partial_{x_1}(w h_X)(x)}{p(x,\gamma)}
    -
    \alpha(x;P)\frac{h(x,\gamma)+\partial_y p(x,\gamma)\gamma_P'(x;P)[H]}{p(x,\gamma)}.
\end{equation}

For the second derivative, we additionally need the mixed second derivative of $D(x;P)$ along $(H,H')$.
A direct differentiation of $D(s,t)=p+s h+t h'$ evaluated at $\gamma(x;P+sH+tH')$ yields
\begin{equation}
    \notag
    \begin{aligned}
        D_P''(x;P)[H,H']
    & =
    \partial_y p(x,\gamma)\,\gamma_P''(x;P)[H,H']
    +
    \partial_y^2 p(x,\gamma)\,\gamma_P'(x;P)[H]\,\gamma_P'(x;P)[H'] \\
    &\quad
    +
    \partial_y h(x,\gamma)\,\gamma_P'(x;P)[H']
    +
    \partial_y h'(x,\gamma)\,\gamma_P'(x;P)[H].
    \end{aligned}
\end{equation}

Using the quotient rule for mixed derivatives of $N/D$ (and that $N$ is linear in $p_X$),
we obtain
\begin{equation}\label{eq:eqd-alpha-second-derivative}
\begin{aligned}
    \alpha_P''(x;P)[H,H']
    &=
    -\frac{N_P'(x;P)[H]\,D_P'(x;P)[H'] + N_P'(x;P)[H']\,D_P'(x;P)[H]}{p(x,\gamma)^2}\\
    &\quad
    -\alpha(x;P)\frac{D_P''(x;P)[H,H']}{p(x,\gamma)}
    +2\alpha(x;P)\frac{D_P'(x;P)[H]\,D_P'(x;P)[H']}{p(x,\gamma)^2}.
\end{aligned}
\end{equation}

Each term in \eqref{eq:eqd-alpha-first-derivative}--\eqref{eq:eqd-alpha-second-derivative} can be uniformly bounded
using: (i) the lower bound on $p(x,\gamma)$ from Assumption~\ref{asmp:density-bounded};
(ii) the bounds on $\partial_y p$ and $\partial_y^2 p$ from membership in $\gM_1$;
(iii) the bounds on $\gamma_P'$ and $\gamma_P''$ derived above; and
(iv) the bounds on the derivatives of $h,h'$ provided by Proposition~\ref{prop:eqd-feasible-perturbation}.
This verifies the derivative boundedness requirements in Assumption~\ref{asmp:regularity}(a)--(d).

Finally, for Assumption~\ref{asmp:regularity}(e), note that
$\upsilon_\rho(x;P)=\partial_y p(x,\gamma(x;P))/p_X(x)$.
Under $\gM_1$, $y\mapsto\partial_y p(x,y)$ is Lipschitz uniformly in $x$ (bounded $\partial_y^2 p$),
and $t\mapsto\gamma(x;\hat P+tH)$ is continuous uniformly in $x$ (bounded $\gamma_P'$).
Thus $\upsilon_\rho(x;\hat P+tH)\to \upsilon_\rho(x;\hat P)$ pointwise in $x$, and is dominated by an integrable constant,
so dominated convergence yields the required $L^1(\hat P)$ continuity.

\paragraph{Constructing $(G_0,G_1)$ with $\gamma(\cdot;\hat P+tG_0)=\gamma(\cdot;\hat P)$ and $\chi_{\mathrm{eqd}}''(\hat P)[G_0,G_1]\neq 0$.}
Define
\[
    r_\ell:=\frac{l_{\hat P}q}{4u_{\hat P}},
    \qquad
    r_u:=1-\frac{l_{\hat P}(1-q)}{4u_{\hat P}}.
\]
Then $\gamma(x;\hat P)\in[r_\ell,r_u]$ for all $x$ by Lemma~\ref{lem:eqd-gamma-bound}.

Define $g_0(x,y)$ (for $x\in\gX$) by
\[
    g_0(x,y)
    :=
    -\frac14\times
    \begin{cases}
        1-140\Bigl(\dfrac{y}{\gamma(x;\hat P)}\Bigr)^3\Bigl(1-\dfrac{y}{\gamma(x;\hat P)}\Bigr)^3, & 0\le y\le \gamma(x;\hat P),\\[1.25ex]
        1-140\Bigl(\dfrac{y-\gamma(x;\hat P)}{1-\gamma(x;\hat P)}\Bigr)^3\Bigl(1-\dfrac{y-\gamma(x;\hat P)}{1-\gamma(x;\hat P)}\Bigr)^3, & \gamma(x;\hat P)<y\le 1.
    \end{cases}
\]
A direct calculation (using $\int_0^1 v^3(1-v)^3\dd v=1/140$) shows that for each $x$:
\[
    \int_0^{\gamma(x;\hat P)} g_0(x,u)\dd u = 0,
    \qquad
    \int_{\gamma(x;\hat P)}^1 g_0(x,u)\dd u = 0,
    \qquad\text{and}\qquad
    g_0(x,\gamma(x;\hat P))=-\frac14.
\]
In particular, $g_0(x,\cdot)$ integrates to $0$ over $[0,1]$, so $G_0$ does not change the marginal distribution of $X$.
Let $G_0$ be the signed measure with density $g_0$ with respect to $\mu$.
Using Lemmas~\ref{lem:eqd-gamma-bound} and~\ref{lem:eqd-gamma-x1-derivative}, $g_0$ and its required derivatives
(first in $x_1$, first/second in $y$) are uniformly bounded, so $G_0$ is $\gM_1$-feasible by Proposition~\ref{prop:eqd-feasible-perturbation}.

Moreover, for any sufficiently small $t$, the conditional CDF of $Y\mid X=x$ under $\hat P+tG_0$ at $y=\gamma(x;\hat P)$ satisfies
\[
    F_{Y\mid X=x}^{\hat P+tG_0}(\gamma(x;\hat P))
    =
    \frac{\int_0^{\gamma(x;\hat P)}\bigl(\hat p(x,u)+t g_0(x,u)\bigr)\dd u}{\hat p_X(x)}
    =
    q+\frac{t}{\hat p_X(x)}\int_0^{\gamma(x;\hat P)} g_0(x,u)\dd u
    =
    q.
\]
Since the conditional density is bounded away from zero and $\hat P+tG_0$ remains $\gM_1$-feasible for all $|t|\le c_t$ for some $c_t>0$ (by Proposition~\ref{prop:eqd-feasible-perturbation}), the $q$-quantile is unique, hence
$\gamma(x;\hat P+tG_0)=\gamma(x;\hat P)$ for all $x$ and all sufficiently small $t$.

Next define
\[
    A(x):=\partial_{x_1}\bigl(w(x)\hat p_X(x)\bigr),
    \qquad
    \gamma(x):=\gamma(x;\hat P),
    \qquad
    I_2(x):=\frac{A(x)\,g_0(x,\gamma(x))}{\hat p(x,\gamma(x))^2}.
\]
By assumption, $\mu_X(\{x:\alpha(x;\hat P)\neq 0\})>0$, and $g_0(x,\gamma(x))=-1/4$ with $\hat p(x,\gamma(x))>0$, so $\mu_X(\{x:I_2(x)\neq 0\})>0$.

To construct a $\gM_1$-feasible $G_1$, we smooth $I_2$ in the $x_1$ coordinate. Let $K:\R\to\R$ be a fixed $C^\infty$ kernel supported on $[-1,1]$ with $\int K=1$, and write $K_\varepsilon(u):=\varepsilon^{-1}K(u/\varepsilon)$.
For $\varepsilon\in(0,1)$ define the (one-dimensional) convolution
\[
    I_{2,\varepsilon}(x_1,x_{-1})
    :=
    \int_0^1 I_2(u,x_{-1})K_\varepsilon(x_1-u)\dd u.
\]
Then $I_{2,\varepsilon}$ is $C^1$ in $x_1$, with
$\|I_{2,\varepsilon}\|_\infty\le \|I_2\|_\infty$ and
$\|\partial_{x_1}I_{2,\varepsilon}\|_\infty\lesssim \varepsilon^{-1}\|I_2\|_\infty$.
Moreover, $I_{2,\varepsilon}\to I_2$ in $L^2(\mu_X)$ as $\varepsilon\downarrow 0$.

Define, for $\gamma\in(r_\ell,r_u)$, the $C^2$ function
\[
    b_\gamma(y)
    :=
    \begin{cases}
        \dfrac{4}{\gamma^4}(\gamma-y)^3, & 0\le y<\gamma,\\[1.25ex]
        -\dfrac{4}{(1-\gamma)^4}(y-\gamma)^3, & \gamma\le y\le 1.
    \end{cases}
\]
Then $\int_0^\gamma b_\gamma(y)\dd y=1$ and $\int_0^1 b_\gamma(y)\dd y=0$.

For a constant $\kappa\neq 0$ (chosen small enough for feasibility), define
\[
    g_1(x,y)
    :=
    \kappa\, I_{2,\varepsilon}(x)\,b_{\gamma(x)}(y),
    \qquad\text{and let }G_1\text{ be the signed measure with density }g_1\text{ w.r.t.\ }\mu.
\]
By construction, $g_1(x,\cdot)$ integrates to $0$ over $[0,1]$, so $G_1$ preserves the $X$-marginal.
Also, since $b_{\gamma(x)}$ has uniformly bounded $y$-derivatives up to order $2$ (because $\gamma(x)\in[r_\ell,r_u]$),
and $x_1\mapsto I_{2,\varepsilon}(x)$ is $C^1$ with bounded derivative, it follows that $g_1$ has uniformly bounded
first $x_1$-derivative and first/second $y$-derivatives.
Thus $G_1$ is $\gM_1$-feasible by Proposition~\ref{prop:eqd-feasible-perturbation} (after taking $|\kappa|$ small enough).

Finally, we compute $\chi_{\mathrm{eqd}}''(\hat P)[G_0,G_1]$.
Consider the two-parameter path $P_{s,t}:=\hat P+sG_0+tG_1$.
Because both $G_0$ and $G_1$ preserve the $X$-marginal, the EQD Riesz representer
\[
\nu_m(x;P)= -p_X(x)^{-1}\partial_{x_1}\bigl(w(x)p_X(x)\bigr)
\]
is constant along $(s,t)$, and
\[
    \chi_{\mathrm{eqd}}(P_{s,t})
    =
    \int \nu_m(x;\hat P)\,\gamma(x;P_{s,t})\,p_X(x)\dd\mu_X(x)
    =
    -\int A(x)\,\gamma(x;P_{s,t})\,\dd\mu_X(x).
\]
Therefore,
\[
    \chi_{\mathrm{eqd}}''(\hat P)[G_0,G_1]
    =
    -\int A(x)\,\gamma_P''(x;\hat P)[G_0,G_1]\,\dd\mu_X(x).
\]
Since $\gamma_P'(x;\hat P)[G_0]=0$ (because $\int_0^{\gamma(x)}g_0(x,u)\dd u=0$) and $\gamma_P'(x;\hat P)[G_1]=-\int_0^{\gamma(x)}g_1(x,u)\dd u\,/\,\hat p(x,\gamma(x))$,
the mixed derivative formula \eqref{eq:eqd-gamma-second-derivative} gives
\[
    \gamma_P''(x;\hat P)[G_0,G_1]
    =
    -\frac{g_0(x,\gamma(x))\,\gamma_P'(x;\hat P)[G_1]}{\hat p(x,\gamma(x))}
    =
    \frac{g_0(x,\gamma(x))}{\hat p(x,\gamma(x))^2}\int_0^{\gamma(x)}g_1(x,u)\dd u.
\]
By the defining property of $b_{\gamma(x)}$,
\[
    \int_0^{\gamma(x)}g_1(x,u)\dd u
    =
    \kappa\,I_{2,\varepsilon}(x)\int_0^{\gamma(x)}b_{\gamma(x)}(u)\dd u
    =
    \kappa\,I_{2,\varepsilon}(x),
\]
and hence
\[
    \chi_{\mathrm{eqd}}''(\hat P)[G_0,G_1]
    =
    -\kappa\int I_2(x)\,I_{2,\varepsilon}(x)\,\dd\mu_X(x).
\]
Since $I_{2,\varepsilon}\to I_2$ in $L^2(\mu_X)$ and $\|I_2\|_{L^2(\mu_X)}^2>0$, the integral is nonzero for all sufficiently small $\varepsilon$.
Choosing such an $\varepsilon$ and any $\kappa\neq 0$ therefore yields $\chi_{\mathrm{eqd}}''(\hat P)[G_0,G_1]\neq 0$.

\textit{Constructing $H_0$ with $\alpha(\cdot;\hat P+tH_0)=\alpha(\cdot;\hat P)$ and $\chi_{\mathrm{eqd}}''(\hat P)[H_0]\neq 0$.}
We now construct a perturbation direction along which the ``regression'' $\alpha$ is invariant, but for which the second directional derivative of $\chi_{\mathrm{eqd}}$ is nonzero.

We first give a one-dimensional lemma that allows us to perturb a density while keeping the density \emph{at the moving quantile} unchanged.
(The smoothness assumptions ensure the resulting perturbations are $\gM_1$-feasible.)

\begin{lemma}[Perturbing a density while preserving the density at the moving quantile]\label{lem:alpha-invariant-perturbation}
Let $p$ be a twice continuously differentiable density on $[0,1]$ with $p(y)\ge l_p>0$ for all $y$, and let $y_q$ be its (unique) $q$-quantile.
Fix $0<r_\ell<y_q<r_u<1$ and define, for $y\in[r_\ell,r_u]\setminus\{y_q\}$,
\[
    \lambda(y)
    :=
    \frac{p(y)-p(y_q)}{\int_{y_q}^{y}p(z)\dd z},
    \qquad
    \lambda(y_q):=\frac{p'(y_q)}{p(y_q)}.
\]
Let $D(y):=\exp\bigl(\int_{y_q}^{y}\lambda(s)\dd s\bigr)$ and define $\delta(y):=D'(y)=\lambda(y)D(y)$ for $y\in[r_\ell,r_u]$.
Let $\bar\delta$ be any twice continuously differentiable extension of $\delta$ to $[0,1]$ such that
$\int_0^{r_\ell}\bar\delta(u)\dd u=D(r_\ell)$ and $\int_0^1 \bar\delta(u)\dd u=0$.
Let $y_{\eta,q}$ be the $q$-quantile of the density $p+\eta\bar\delta$.
If $y_{\eta,q}\in[r_\ell,r_u]$, then
\[
    p(y_{\eta,q})+\eta\bar\delta(y_{\eta,q})
    =
    p(y_q).
\]
\end{lemma}
\begin{proof}
For $y\in[r_\ell,r_u]$ we have $\int_0^{y}\bar\delta(u)\dd u=D(y)$: this holds at $y=r_\ell$ by assumption, and for $y>r_\ell$ by integrating $\delta=D'$.
The identity
\[
    \bigl(p(y_q)-p(y)\bigr)D(y)+\delta(y)\int_{y_q}^{y}p(z)\dd z=0
\]
follows immediately from $\delta(y)=\lambda(y)D(y)$ and the definition of $\lambda(y)$.
Now, by the definition of $y_{\eta,q}$,
\[
    \int_{y_q}^{y_{\eta,q}}p(u)\dd u+\eta\int_0^{y_{\eta,q}}\bar\delta(u)\dd u=0.
\]
If $y_{\eta,q}\in[r_\ell,r_u]$ then $\int_0^{y_{\eta,q}}\bar\delta(u)\dd u=D(y_{\eta,q})$, so
$\int_{y_q}^{y_{\eta,q}}p(u)\dd u=-\eta D(y_{\eta,q})$.
Plugging this into the identity above evaluated at $y=y_{\eta,q}$ yields
\[
    \bigl(p(y_q)-p(y_{\eta,q})-\eta\bar\delta(y_{\eta,q})\bigr)D(y_{\eta,q})=0.
\]
Since $D(y_{\eta,q})>0$, the claim follows.
\end{proof}

\begin{corollary}[Constructing an $\alpha$-invariant perturbation for EQD]\label{cor:eqd-alpha-invariant-perturbation}
Under Assumption~\ref{asmp:eqd-density}, there exists a $\gM_1$-feasible perturbation $\hat H_0$ with density $\hat h_0=\dd\hat H_0/\dd\mu$ and a constant $c_t>0$ such that:
\begin{enumerate}
    \item $\int_0^1 \hat h_0(x,u)\dd u=0$ for all $x$ (so $\hat H_0$ does not change the $X$-marginal);\label{it:eqd-h0-marginal}
    \item for all $|t|\le c_t$ and all $x$, writing $\gamma_t(x):=\gamma(x;\hat P+t\hat H_0)$,
    \[
        \frac{\dd(\hat P+t\hat H_0)}{\dd\mu}\bigl(x,\gamma_t(x)\bigr)=\hat p\bigl(x,\gamma(x;\hat P)\bigr);
    \]\label{it:eqd-h0-density-at-quantile}
    \item and $\int_0^{\gamma(x;\hat P)}\hat h_0(x,u)\dd u=\hat p_X(x)>0$ for all $x$.\label{it:eqd-h0-positive-mass}
\end{enumerate}
\end{corollary}
\begin{proof}
Fix $x\in\gX$ and consider the conditional density $p_x(y):=\hat p(x,y)/\hat p_X(x)$ on $[0,1]$, whose $q$-quantile is $y_q=\gamma(x;\hat P)$.
Define
\[
    r_\ell:=\frac{l_{\hat P}q}{4u_{\hat P}},
    \qquad
    r_u:=1-\frac{l_{\hat P}(1-q)}{4u_{\hat P}},
\]
so that Lemma~\ref{lem:eqd-gamma-bound} implies $y_q\in[2r_\ell,1-2(1-r_u)]\subset(r_\ell,r_u)$ uniformly in $x$.

Apply Lemma~\ref{lem:alpha-invariant-perturbation} to the density $p_x$, the quantile $y_q$, and the interval $[r_\ell,r_u]$.
This yields a function $\bar\delta_x$ on $[0,1]$ with $\int_0^1 \bar\delta_x=0$ and, on $[r_\ell,r_u]$, $\int_0^{y}\bar\delta_x(u)\dd u=D_x(y)$ where $D_x(y_q)=1$.
(Existence of a $C^2$ extension satisfying the two integral constraints is standard: start from any $C^2$ extension of $\delta_x$ to $[0,1]$ and then correct the two integrals by adding suitable $C^2$ bump functions supported in $[0,r_\ell]$ and $[r_u,1]$.)

Define
\[
    \hat h_0(x,y):=\hat p_X(x)\,\bar\delta_x(y),
    \qquad\text{and let }\hat H_0\text{ be the signed measure with density }\hat h_0\text{ w.r.t.\ }\mu.
\]
Then $\int_0^1 \hat h_0(x,u)\dd u=\hat p_X(x)\int_0^1 \bar\delta_x(u)\dd u=0$, proving~\ref{it:eqd-h0-marginal}.
Also,
\[
    \int_0^{\gamma(x;\hat P)}\hat h_0(x,u)\dd u
    =
    \hat p_X(x)\int_0^{y_q}\bar\delta_x(u)\dd u
    =
    \hat p_X(x)D_x(y_q)
    =
    \hat p_X(x),
\]
which gives~\ref{it:eqd-h0-positive-mass}.

Now let $\gamma_t(x)=\gamma(x;\hat P+t\hat H_0)$ be the $q$-quantile under the perturbed law.
Since the $X$-marginal is unchanged, the conditional density under $\hat P+t\hat H_0$ is $p_x+t\bar\delta_x$.
Lemma~\ref{lem:alpha-invariant-perturbation} therefore implies that if $\gamma_t(x)\in[r_\ell,r_u]$, then
\[
    p_x(\gamma_t(x))+t\bar\delta_x(\gamma_t(x))=p_x(y_q).
\]
Multiplying by $\hat p_X(x)$ yields~\ref{it:eqd-h0-density-at-quantile}.

Finally, $\gamma_t(x)$ stays in $[r_\ell,r_u]$ for all $|t|\le c_t$ uniformly in $x$ for some $c_t>0$.
Indeed, by \eqref{eq:eqd-gamma-first-derivative} applied at $P=\hat P$ and $H=\hat H_0$ and using that $\hat p(x,\gamma(x;\hat P))\ge l_{\hat P}$,
\[
    |\gamma_{\hat P}'(x;\hat P)[\hat H_0]|
    \le
    \frac{\int_0^1|\hat h_0(x,u)|\dd u + q|\hat h_{0,X}(x)|}{l_{\hat P}}
    \le
    \frac{\|\hat h_0\|_\infty}{l_{\hat P}},
\]
so $t\mapsto \gamma_t(x)$ is Lipschitz in $t$ uniformly in $x$.
Because $\gamma(x;\hat P)$ is uniformly at least distance $\min\{r_\ell,1-r_u\}>0$ from the boundary of $[r_\ell,r_u]$, choosing $c_t>0$ small enough yields $\gamma_t(x)\in[r_\ell,r_u]$ for all $x$ and $|t|\le c_t$.

The construction of $\bar\delta_x$ and the smoothness/boundedness assumptions on $\hat p$ imply that $\hat h_0$ has bounded first $x_1$-derivative and bounded first/second $y$-derivatives uniformly over $(x,y)$, so $\hat H_0$ is $\gM_1$-feasible by Proposition~\ref{prop:eqd-feasible-perturbation}.
\end{proof}

We now use $\hat H_0$ to build a perturbation $H_0$ that guarantees $\chi_{\mathrm{eqd}}''(\hat P)[H_0]\neq 0$.
Recall that for EQD,
\[
    \chi_{\mathrm{eqd}}''(\hat P)[H_0]
    =
    -\int \alpha(x;\hat P)\,\upsilon_\rho(x;\hat P)\,\bigl(\gamma_P'(x;\hat P)[H_0]\bigr)^2\,\dd\hat P(x,y)
\]
(see Proposition~\ref{prop:compute-2nd-order-differential} and that $\chi_{\mathrm{eqd}}'(\hat P)[H_0]=0$ by $\alpha$-invariance).
Define, with $\gamma(x)=\gamma(x;\hat P)$,
\[
    I_3(x):=\frac{\alpha(x;\hat P)\,\upsilon_\rho(x;\hat P)}{\hat p(x,\gamma(x))^2}.
\]
By assumption, $\mu_X(\{x:I_3(x)\neq 0\})>0$. Without loss of generality, suppose $\mu_X(\{x:I_3(x)>0\})>0$; otherwise the same argument applies on $\{I_3<0\}$.
Hence there exists $\delta_0>0$ with $\mu_X(\{x:I_3(x)\ge 4\delta_0\})>0$.

As in the construction of $I_{2,\varepsilon}$, let $I_{3,\varepsilon}$ be the $x_1$-mollification of $I_3$:
\[
    I_{3,\varepsilon}(x_1,x_{-1})
    :=
    \int_0^1 I_3(u,x_{-1})K_\varepsilon(x_1-u)\dd u.
\]
Let $S:=\{x:I_3(x)\ge 4\delta_0\}$, which has positive $\mu_X$-measure by construction. Choose $\varepsilon>0$ small enough that
$\|I_{3,\varepsilon}-I_3\|_{L^1(\mu_X)}<\delta_0\,\mu_X(S)$. Then
\[
    \mu_X\bigl(S\cap\{x:I_{3,\varepsilon}(x)\ge 3\delta_0\}\bigr)>0,
\]
since otherwise we would have $\int_S |I_{3,\varepsilon}-I_3|\dd\mu_X\ge \delta_0\,\mu_X(S)$.

Let $\varphi:\R\to[0,1]$ be a fixed $C^\infty$ cutoff such that $\varphi(t)=0$ for $t\le 2\delta_0$ and $\varphi(t)=1$ for $t\ge 3\delta_0$.
Define $s(x):=\varphi(I_{3,\varepsilon}(x))$ and set
\[
    h_0(x,y)
    :=
    \frac{s(x)\,\hat h_0(x,y)}{\int |s(x)\hat h_0(x,y)|\,\dd\mu(x,y)}.
\]
Let $H_0$ be the signed measure with density $h_0$ w.r.t.\ $\mu$.
Because $s$ is $C^1$ in $x_1$ (as a smooth function of $I_{3,\varepsilon}$) and bounded by $1$,
$h_0$ has bounded first $x_1$-derivative and bounded first/second $y$-derivatives, so $H_0$ is $\gM_1$-feasible by Proposition~\ref{prop:eqd-feasible-perturbation}.
Moreover, since $h_0(x,\cdot)$ is just a (possibly $x$-dependent) scalar multiple of $\hat h_0(x,\cdot)$, the density-at-quantile property in Corollary~\ref{cor:eqd-alpha-invariant-perturbation} implies that $\alpha(x;\hat P+tH_0)=\alpha(x;\hat P)$ for all $x$ and all $|t|\le c_t^{(H)}$ for some $c_t^{(H)}>0$ (for instance, one may take $c_t^{(H)}:=c_t\int |s\hat h_0|\,\dd\mu$, where $c_t$ is the constant from Corollary~\ref{cor:eqd-alpha-invariant-perturbation}).
Thus Assumption~\ref{asmp:main} holds with $H_1=H_0$.

Finally, we show $\chi_{\mathrm{eqd}}''(\hat P)[H_0]\neq 0$.
Because $\int_0^1 h_0(x,u)\dd u=0$, \eqref{eq:eqd-gamma-first-derivative} gives
\[
    \gamma_P'(x;\hat P)[H_0]
    =
    -\frac{\int_0^{\gamma(x)} h_0(x,u)\dd u}{\hat p(x,\gamma(x))}
    =
    -\frac{s(x)\,\hat p_X(x)}{\hat p(x,\gamma(x))\int |s\hat h_0|\,\dd\mu}.
\]
Therefore,
\[
    \chi_{\mathrm{eqd}}''(\hat P)[H_0]
    =
    -\frac{1}{\bigl(\int |s\hat h_0|\,\dd\mu\bigr)^2}
    \int_{\gX}
        I_3(x)\,s(x)^2\,\hat p_X(x)^3\,\dd\mu_X(x).
\]
Here we used $\int f(x)\dd\hat P(x,y)=\int f(x)\hat p_X(x)\dd\mu_X(x)$ for any integrable function $f$.
On the set $S\cap\{x:I_{3,\varepsilon}(x)\ge 3\delta_0\}$ we have $s(x)=1$ and $I_3(x)\ge 4\delta_0$, and $\hat p_X(x)\ge l_{\hat P}/2$, so the integral is strictly positive.
Hence $\chi_{\mathrm{eqd}}''(\hat P)[H_0]<0$, completing the construction.

Together with the construction of $(G_0,G_1)$ above, this verifies Assumption~\ref{asmp:main} and the non-degeneracy conditions
$\chi_{\mathrm{eqd}}''(\hat P)[G_0,G_1]\neq 0$ and $\chi_{\mathrm{eqd}}''(\hat P)[H_0]\neq 0$, so Theorem~\ref{thm:main} yields Theorem~\ref{thm:eqd}.

\section{Proof of Theorem \ref{thm:upper-bound-general}}
\label{sec:proof:upper-bound-general}

\
We write $\gamma_0:=\gamma(\cdot;P_0)$ and $\alpha_0(\cdot):=\alpha(\cdot;P_0,Q_0)$ for the true nuisance functions in the
covariate shift setting. Recall that the target parameter is
\[
\chi(P_0,Q_0)=\E_{Q_0}\big[m_1(Z,\gamma_0)\big].
\]
Since $\gamma_0$ satisfies the first-order optimality condition \eqref{eq:first-order-optimality-cond} under the training law
$P_0$, we have
\[
\E_{P_0}\big[\rho(O,\gamma_0(Z))\mid Z\big]=0
\qquad\text{almost surely,}
\]
hence
\begin{equation}
    \label{eq:upper-bound-general-1}
    \chi(P_0,Q_0)
    =
    \E_{Q_0}\big[m_1(Z,\gamma_0)\big]
    +
    \E_{P_0}\big[\alpha_0(Z)\,\rho(O,\gamma_0(Z))\big].
\end{equation}

Define the intermediate (population) quantity
\begin{equation}
    \label{eq:upper-bound-general-2}
    \tilde{\chi}
    =
    \E_{Q_0}\big[m_1(Z,\hat{\gamma})\big]
    +
    \E_{P_0}\big[\alpha_0(Z)\,\rho(O,\hat{\gamma}(Z))\big].
\end{equation}
Then
\begin{equation}
    \label{eq:upper-bound-general-3}
    \begin{aligned}
        |\tilde{\chi}-\chi(P_0,Q_0)|
        &= \Big|\E_{Q_0}\Big[m_1(Z,\hat{\gamma}) - m_1(Z,\gamma_0)\Big]
        + \E_{P_0}\Big[\alpha_0(Z)\big\{\rho(O,\hat{\gamma}(Z))-\rho(O,\gamma_0(Z))\big\}\Big]\Big|\\
        &= \Big|\E_{P_0}\Big[\big\{\hat{\gamma}(Z)-\gamma_0(Z)\big\}\nu_m(Z;P_0,Q_0)
        + \alpha_0(Z)\big\{\rho(O,\hat{\gamma}(Z))-\rho(O,\gamma_0(Z))\big\}\Big]\Big|\\
        &= \Big|\E_{P_0}\Big[\alpha_0(Z)\Big\{\rho(O,\hat{\gamma}(Z))-\rho(O,\gamma_0(Z))
        - \nu_\rho(Z;P_0)\big(\hat{\gamma}(Z)-\gamma_0(Z)\big)\Big\}\Big]\Big|\\
        &= \Big|\E_{P_0}\Big[\alpha_0(Z)\,\E_{P_0}\Big[\rho(O,\hat{\gamma}(Z))-\rho(O,\gamma_0(Z))
        - \nu_\rho(Z;P_0)\big(\hat{\gamma}(Z)-\gamma_0(Z)\big)\,\Big|\, Z\Big]\Big]\Big|\\
        &\le A\,\E_{P_0}\Big[\Big|\E_{P_0}\Big[\rho(O,\hat{\gamma}(Z))-\rho(O,\gamma_0(Z))
        - \nu_\rho(Z;P_0)\big(\hat{\gamma}(Z)-\gamma_0(Z)\big)\,\Big|\, Z\Big]\Big|\Big]\\
        &\le A\, C_{\rho,2}\,\|\hat{\gamma}(Z)-\gamma_0(Z)\|_{P_{0,Z},2}^2\\
        &\le A\, C_{\rho,2}\,\eps_{N,\gamma}^2.
    \end{aligned}
\end{equation}
Here the second equality uses the linearity of $\gamma\mapsto \E_{Q_0}[m_1(Z,\gamma)]$ together with the cross-population
Riesz representation \eqref{eq:riesz-representer}, and the third equality uses
$\nu_m(z;P_0,Q_0)=-\alpha_0(z)\nu_\rho(z;P_0)$ (by definition of $\alpha_0$ in \eqref{eq:weighted-riesz}).
The penultimate inequality is exactly the (conditional) second-order remainder bound
\eqref{eq:def-2nd-order-effect-rho-equiv} defining $C_{\rho,2}$.

\medskip

On the other hand, define the population analogue of the empirical estimator \eqref{eq:dml-estimator}:
\begin{equation}
    \label{eq:upper-bound-general-4}
    \chi'
    =
    \E_{Q_0}\big[m_1(Z,\hat{\gamma})\big]
    +
    \E_{P_0}\big[\hat{\alpha}(Z)\rho(O,\hat{\gamma}(Z))\big].
\end{equation}
Conditioning on the nuisance estimators (e.g., under sample-splitting/cross-fitting so that the evaluation samples are
independent of $\hat{\gamma}$ and $\hat{\alpha}$), the two empirical averages defining $\hat{\chi}$ are based on independent
i.i.d.\ summands with means matching the two terms in $\chi'$, and are uniformly bounded by $C_m$ and $A C_{\rho,0}$,
respectively. Therefore, by Chebyshev's inequality applied to each average and a union bound, we obtain that
\begin{equation}
    \label{eq:upper-bound-general-5}
    |\chi' - \hat{\chi}| \le C_\delta\big(C_m + A C_{\rho,0}\big)N^{-1/2}
\end{equation}
with probability at least $1-\delta$, where one may take $C_\delta=(2/\delta)^{1/2}$.

\medskip

Finally,
\begin{equation}
    \label{eq:upper-bound-general-6}
    \begin{aligned}
        |\chi' - \tilde{\chi}|
        &= \Big|\E_{P_0}\big[(\hat{\alpha}(Z) - \alpha_0(Z))\,\rho(O,\hat{\gamma}(Z))\big]\Big|\\
        &= \Big|\E_{P_0}\big[(\hat{\alpha}(Z) - \alpha_0(Z))\big\{\rho(O,\hat{\gamma}(Z))-\rho(O,\gamma_0(Z))\big\}\big]\Big|\\
        &\le C_{\rho,1}\,\E_{P_0}\big[|\hat{\alpha}(Z) - \alpha_0(Z)|\cdot|\hat{\gamma}(Z)-\gamma_0(Z)|\big]\\
        &\le C_{\rho,1}\,\|\hat{\alpha}(Z) - \alpha_0(Z)\|_{P_{0,Z},2}\,\|\hat{\gamma}(Z)-\gamma_0(Z)\|_{P_{0,Z},2}\\
        &\le C_{\rho,1}\,\eps_{N,\alpha}\eps_{N,\gamma}.
    \end{aligned}
\end{equation}
The second equality uses $\E_{P_0}[\rho(O,\gamma_0(Z))\mid Z]=0$ and the fact that $\hat{\alpha}(Z)-\alpha_0(Z)$ is
$Z$-measurable. The inequality on the third line uses the uniform Lipschitz property of $\rho(O,\cdot)$ with constant
$C_{\rho,1}$ (e.g., implied by a uniform bound on its derivative in $\gamma$).

Combining \eqref{eq:upper-bound-general-3}, \eqref{eq:upper-bound-general-5} and \eqref{eq:upper-bound-general-6} yields the
desired upper bound and concludes the proof.

\section{Proof of Proposition \ref{prop:main-asmp-suff-cond}}
\label{sec:proof:main-asmp-suff-cond}

For any $z\in\gZ$, define $\hat{\alpha}_z = \alpha(z,\hat{P})$. Under the given assumptions, for any fixed $z\in\gZ$,
\begin{equation}
    \label{eq:invariant-direction-equiv}
    \alpha(z;\hat{P}+tG_0)=\alpha(z;\hat{P}),\ \forall\,|t|\le c_t
    \iff
    \int \Big(F_0(z,w)-\hat{\alpha}_zF_1(z,w)\Big)\, g_0(w\mid z)\,\dd \mu_{\gW\mid\gZ}(w\mid z)=0,
\end{equation}
where we write $\tilde{F}_z(w):=F_0(z,w)-\hat{\alpha}_zF_1(z,w)$.

We show that there exists a nonzero perturbation that satisfies \eqref{eq:invariant-direction-equiv}. Fix $z\in\gZ$ and let
\[
    c := \int \tilde{F}_z(w)\,\dd \mu_{\gW\mid\gZ}(w\mid z).
\]
Then for any bounded and $\mu_{\gW\mid\gZ}(\cdot\mid z)$-measurable function $\tilde{g}_0(\cdot\mid z)$, define
\begin{equation}
    \label{eq:invariant-direction-gram-schmidt}
    \begin{aligned}
    g_0(w\mid z)
    &:=
    \tilde{g}_0(w\mid z)
    - \frac{\int \tilde{g}_0(w\mid z)\big(\tilde{F}_z(w)-c\big)\,\dd \mu_{\gW\mid\gZ}(w\mid z)}{\int\big(\tilde{F}_z(w)-c\big)^2\,\dd \mu_{\gW\mid\gZ}(w\mid z)}\big(\tilde{F}_z(w)-c\big)\\
    &\qquad\qquad - \int \tilde{g}_0(w\mid z)\,\dd \mu_{\gW\mid\gZ}(w\mid z).
    \end{aligned}
\end{equation}
This is exactly the Gram--Schmidt orthogonalization of $\tilde{g}_0(\cdot\mid z)$ against the two-dimensional subspace $\langle \tilde{F}_z(\cdot),1\rangle$ in $L^2(\mu_{\gW\mid\gZ}(\cdot\mid z))$, written out explicitly. By construction,
\[
    \int g_0(w\mid z)\,\dd \mu_{\gW\mid\gZ}(w\mid z)=0
    \quad\text{and}\quad
    \int \big(\tilde{F}_z(w)-c\big)\,g_0(w\mid z)\,\dd \mu_{\gW\mid\gZ}(w\mid z)=0.
\]
Hence $\int \tilde{F}_z(w)\,g_0(w\mid z)\,\dd \mu_{\gW\mid\gZ}(w\mid z)=0$ as well, proving that $g_0(\cdot\mid z)$ satisfies \eqref{eq:invariant-direction-equiv}. Note that \eqref{eq:invariant-direction-equiv} and \eqref{eq:invariant-direction-gram-schmidt} are well-defined because (26) guarantees that the denominator $\int(\tilde{F}_z(w)-c)^2\,\dd \mu_{\gW\mid\gZ}(w\mid z)$ is bounded away from $0$.

Since $\tilde{F}_z$, $\hat{\alpha}_z$ and $\tilde{g}_0(\cdot\mid z)$ are all bounded, the resulting $g_0(\cdot\mid z)$ is also bounded. If $g_0(\cdot\mid z)$ is not identically zero (for at least one $z$), then after multiplying by a normalizing constant we obtain the desired nonzero perturbation.

It remains to handle the case when $g_0(\cdot\mid z)$ is identically zero for every $z$, regardless of how we choose $\tilde{g}_0(\cdot\mid z)$. In that case, every bounded $\mu_{\gW\mid\gZ}(\cdot\mid z)$-measurable function lies in $\langle \tilde{F}_z(\cdot),1\rangle$, contradicting the assumption that $(\gW,\mu_{\gW\mid\gZ}(\cdot\mid z))$ is $3$-non-degenerate. This concludes the proof.

\end{document}